\pdfoutput=1
\documentclass{article}

\PassOptionsToPackage{numbers,compress,sort}{natbib}
\usepackage[final]{neurips_2020}

\usepackage[utf8]{inputenc} % allow utf-8 input
\usepackage[T1]{fontenc}    % use 8-bit T1 fonts
\usepackage[colorlinks = true,
            linkcolor = blue,
            urlcolor  = blue]{hyperref}

\usepackage{hyperref}       % hyperlinks
\usepackage{url}            % simple URL typesetting
\usepackage{booktabs}       % professional-quality tables
\usepackage{amsfonts}       % blackboard math symbols
\usepackage{nicefrac}       % compact symbols for 1/2, etc.
\usepackage{microtype}      % microtypography

\usepackage{stackengine, graphicx, wrapfig, lipsum}
\usepackage{amsmath}
\usepackage{xspace}
\usepackage{amsfonts}
\usepackage{float}
\usepackage{xcolor}
\usepackage{multirow}
\usepackage{appendix}
\usepackage{tablefootnote}

\usepackage{subcaption}
\usepackage[makeroom]{cancel}

\usepackage{pifont}% http://ctan.org/pkg/pifont
\newcommand{\cmark}{\ding{51}}%
\newcommand{\sh}[1]{}
\newcommand{\cutabstractup}{\vspace*{-0.2in}}

\newcommand{\cutsectiondown}{\vspace*{-0.12in}}

\newcommand{\cutsubsectiondown}{\vspace*{-0.07in}}
\newcommand{\cutparagraphup}{\vspace*{-0.1in}}

\usepackage[numbered,framed]{matlab-prettifier}

\title{Variational Interaction Information Maximization\\for Cross-domain Disentanglement}

\author{%
    HyeongJoo Hwang$^1$,
    Geon-Hyeong Kim$^2$,
    Seunghoon Hong$^2$,
    Kee-Eung Kim$^{1,2}$\\
    $^1$ Graduate School of AI, KAIST, Daejeon, Republic of Korea\\
    $^2$ School of Computing, KAIST, Daejeon, Republic of Korea\\
    \texttt{\{hjhwang, ghkim\}@ai.kaist.ac.kr, \{seunghoon.hong, kekim\}@kaist.ac.kr}
}
\begin{document}

\maketitle
\cutabstractup
\begin{abstract}
Cross-domain disentanglement is the problem of learning representations partitioned into domain-invariant and domain-specific representations,
which is a key to successful domain transfer or measuring semantic distance between two domains.
Grounded in information theory, we cast the simultaneous learning of domain-invariant and domain-specific representations as a joint objective of multiple information constraints, which does not require adversarial training or gradient reversal layers. We derive a tractable bound of the objective and propose a generative model named Interaction Information Auto-Encoder (IIAE). Our approach reveals insights on the desirable representation for cross-domain disentanglement and its connection to Variational Auto-Encoder (VAE).
We demonstrate the validity of our model in % learning disentangled representations with 
the image-to-image translation and the cross-domain retrieval tasks.
We further show that our model achieves the state-of-the-art performance
in the zero-shot sketch based image retrieval task, even without external knowledge. Our implementation is publicly available at: \url{https://github.com/gr8joo/IIAE}
%Specifically, we propose to enforce domain invariant representation to capture correlations between two data domains by maximizing interaction information between data domains and shared representation.  while minimize mutual information between s 
\end{abstract}

% !TEX root = neurips_2020.tex
\section{Introduction}
\cutsectiondown
There have been great interests in learning disentangled representation for various purposes, such as identifying sources of variation \citep{chen2016infogan, higgins2017beta, pmlr-v80-kim18b, chen2018isolating, jeongICML19} for interpretability, obtaining representation invariant to nuisance factors \citep{tishby2000information, DBLP:journals/corr/LouizosSLWZ15, 45903, song2018learning, moyer2018invariant, federici2020}, and domain transfer \citep{zhu2017toward, gonzalez-garcia2018NeurIPS, NIPS2018_7525, lee2018diverse, press2018emerging, yu2019multi}.
In particular, the cross-domain disentanglement problem \citep{gonzalez-garcia2018NeurIPS} assumes the dataset composed of paired samples $(x \in X,  y \in Y)$ where every sample has some shared information.
The problem requires a model to learn a representation explicitly separated into three parts: domain-invariant representation shared across two data domains and domain-specific representations exclusive to each domain.
This task is challenging since those representations must be (1) disentangled so that they are independent to one another, while (2) informative in such a way that every factor of variation is captured in the right part of the representation.

%%% Tackle heuristic, ad-hoc, non-principled, and no guranteed approaches of previous works.
In recent studies, many models have been proposed to tackle important tasks related to cross-domain disentanglement, such as image-to-image translation \citep{gonzalez-garcia2018NeurIPS, NIPS2018_7525, lee2018diverse, press2018emerging, yu2019multi} and Zero-Shot Sketch Based Image Retrieval (ZS-SBIR) \citep{kodirov2017semantic, felix2018multi, 'shen2018cvpr', Yelamarthi_2018_ECCV, Dutta2019SEMPCYC, linlearning}.
Although those models perform reasonably well with realistic datasets, %most of them take heuristic approach to regularize the latent space, such as cycle consistency loss \citep{CycleGAN2017}, cross-reconstruction loss \citep{linlearning}, adversarial training \citep{10.5555/2969033.2969125, DBLP:conf/nips/XieDDHN17}, and Gradient Reversal Layer (GRL) \citep{pmlr-v37-ganin15}.
most of them take a heuristic combination of techniques that regularize the latent space, such as cycle consistency loss \citep{CycleGAN2017}, cross-reconstruction loss \citep{linlearning}, adversarial training \citep{10.5555/2969033.2969125, DBLP:conf/nips/XieDDHN17}, and Gradient Reversal Layer (GRL) \citep{pmlr-v37-ganin15}.
Consequently, it is not obvious to interpret each module or identify key factors that contribute to disentanglement in their models.

In this paper, we address the cross-domain disentanglement problem with a novel principle based on information theory. 
Specifically, we train a generative model named \emph{Interaction Information Auto-Encoder} (IIAE) whose representations are enforced to be informative but disentangled by information regularization terms that we will describe shortly. 
Leveraging representations learned by IIAE, we show that image manipulation tasks such as image translation and synthesis can be done in fine details. Furthermore, we demonstrate that IIAE outperforms Generative Adversarial Network (GAN) \citep{NIPS2014_5423} based models in the cross-domain retrieval task.
Lastly, we empirically show that our model outperforms the state-of-the-art models for ZS-SBIR which strongly depend on external knowledge such as word embedding of class labels.
Our contributions are three-fold:
\begin{enumerate}
    \item We propose a novel information-theoretic framework to learn and disentangle shared and exclusive representations and derive a tractable lower bound of the optimization objective.
    \item By bridging the lower bound of the objective and the Evidence Lower Bound (ELBO), we introduce IIAE, a simple and interpretable generative model trained by maximizing the lower bound.
    \item The performance of IIAE are demonstrated on an extensive set of tasks, such as cross-domain image translation, cross-domain retrieval, and ZS-SBIR.
\end{enumerate}
% \section{Interaction Information}
% \label{gen_inst}
% Interaction information was first introduced by \citep{1057469} as a generalization of mutual information among three or more random variables. It quantifies the shared information among any subset of those random variables. The interaction information is also known as the amount of information \citep{ting1962amount} or co-information \citep{bell2003co}. Since the sign convention is not unified, we use the following definition to refer to the interaction information among three random variables $X,Y,Z$:
% \begin{equation}
% \begin{aligned}
%     I\left(X; Y; Z\right) &= H(X) + H(Y) + H(Z) - H(X,Y) - H(Y,Z) - H(X,Z) + H(X,Y,Z) \\
% \end{aligned}
% \end{equation}
% From the definition, the interaction information exhibits symmetry:
% \begin{equation}
% \begin{aligned}
%     I\left(X; Y; Z\right) = I\left(X; Y\right) - I\left(X; Y | Z\right) = I\left(X; Z\right) - I\left(X; Z | Y\right) = I\left(Y; Z\right) - I\left(Y; Z | X\right) \\
% \end{aligned}
% \end{equation}
% This quantity is positive when one random variable reasons out the correlation between the other two random variables by decreasing their conditional dependency. It can be negative in the opposite case, while mutual information is always non-negative.
% !TEX root = neurips_2020.tex
\section{Method}
\label{headings}
\cutsectiondown
Consider a set of paired data sampled from an unknown joint distribution $(x,y)\sim p_{D}(x,y)$, where each element of a pair $x\in X$ and $y\in Y$ is extracted from different domains $X$ and $Y$, respectively.
We assume that two domains exhibit domain-specific factors of variations while sharing some common factors of variations.
For instance, $x$ and $y$ can be images in different styles (\emph{e.g.}, sketch and photo) sharing the same semantic content, or images of different content (\emph{e.g.}, different types of car)  sharing the same factors of variation (\emph{e.g.}, rotation and scale).

Given this data, the goal of cross-domain disentanglement is to find the structured representation that can be factorized into three parts: domain-specific representations $Z^X$ and $Z^Y$ that capture the distinctive and exclusive characteristics of each domain $X$ and $Y$, respectively, and the shared representation $Z^S$ that captures common factors shared across the domains.
Figure~\ref{subfig:generative} describes our graphical model encoding this structure.

% Formally, we define the joint data distribution as follows:
A typical way to learn a latent variable model is maximizing the marginal likelihood~\citep{DBLP:journals/corr/KingmaW13}. 
In our problem, we maximize the marginal likelihood of the joint distribution of $X$ and $Y$:
\begin{equation}
 p_{\theta}(x,y)=\int d z^{x} d z^{s} d z^{y} p_{\theta_X}(x|z^{x},z^{s}) p_{\theta_Y}(y|z^{y},z^{s})p(z^{x})p(z^{s})p(z^{y}),
 \label{eqn:joint_prob_opt}
\end{equation}
where $\theta=\{\theta_X,\theta_Y\}$ denotes the parameter modeling the conditional distributions. 
Our objective is then training the generative model $p_{\theta}(x,y)$ that not only maximizes the joint distribution $p_{D}(x,y)$ by optimizing $\theta$, but also disentangles the exclusive representations $Z^{X}$ and $Z^{Y}$ from the shared representation $Z^{S}$.
Below, we describe our approach to optimize the Eq.~\eqref{eqn:joint_prob_opt} while enforcing the disentanglement constraints on the latent representations. 

\iffalse
\sh{Move the following to the end of the method section?}
Once we learn the model, we can generate a paired sample by (1) sampling three latent factors $z^x$, $z^y$ and $z^s$ from the prior distributions $p(z^x)$, $p(z^y)$ and $p(z^s)$, and (2) generating data $x$ and $y$ independently from conditional distributions $p_{\theta^{*}}(x|z^{x},z^{s})$ and $p_{\theta^{*}}(y|z^{y},z^{s})$. 
Also, considering the shared latent variable $z^s$ as a domain-invariant representation, we can solve other downstream tasks such as cross-domain retrieval (Section~\ref{}).
\fi

\subsection{Generative model for the joint distribution $p_{D}(x,y)$}
\begin{figure}[!t]
    \centering
    \begin{subfigure}[b]{0.33\textwidth}
        \centering
        \includegraphics[width=0.7\textwidth]{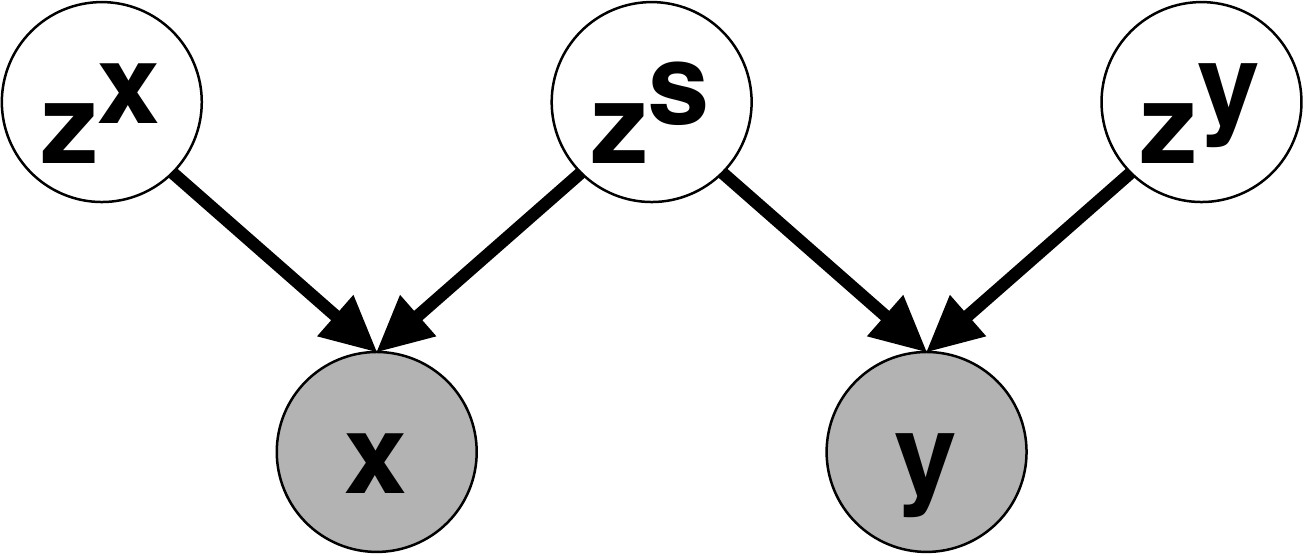}
        \caption{Generative model $p_{\theta}$}
        \label{subfig:generative}
    \end{subfigure}
    \hspace{20mm}
    \begin{subfigure}[b]{0.33\textwidth}
        \centering
        \includegraphics[width=0.7\textwidth]{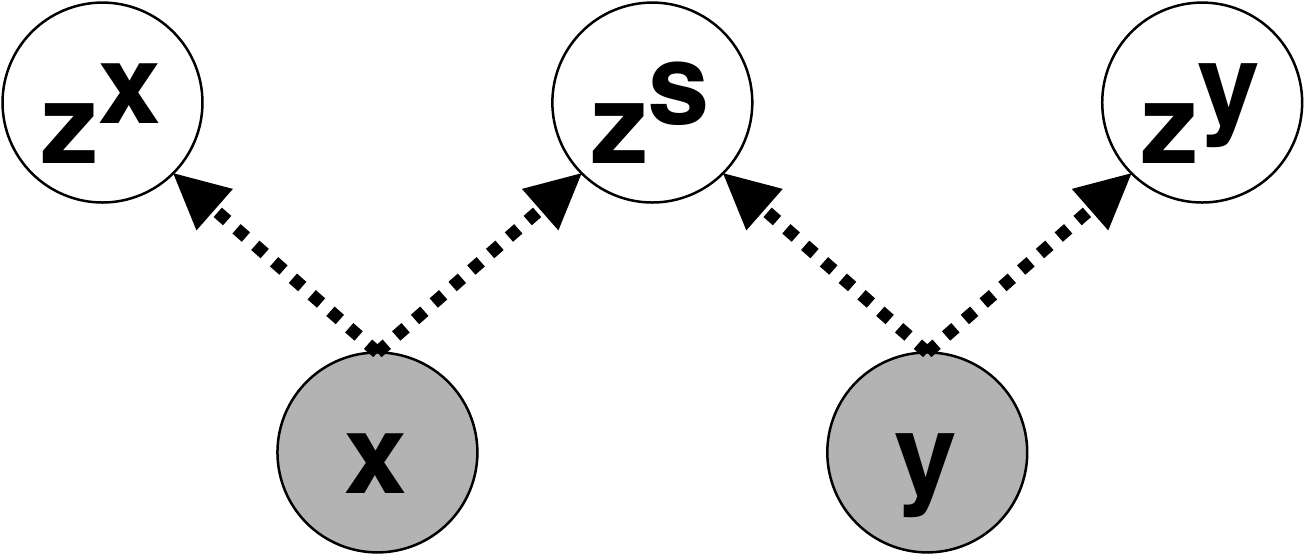}
        \caption{Approximate inference model $q_{\phi}$}
        \label{subfig:inference}
    \end{subfigure}
    \caption{Graphical models for cross-domain disentanglement.}
    \label{fig:graphical}
    \vspace{-4mm}
\end{figure}

\cutsubsectiondown
Since the direct optimization of Eq.~\eqref{eqn:joint_prob_opt} is intractable, we employ variational inference based on Variational Auto-Encoder (VAE)~\citep{DBLP:journals/corr/KingmaW13}.
Specifically, we approximate the true posterior distribution $p_{\theta}(z^{x},z^{s},z^{y}|x,y)$ using the approximated posterior $q_\phi(z^{x},z^{s},z^{y}|x,y)$, which is  factorized according to the graphical model in Figure~\ref{subfig:inference} as follows: 
% Our goal is to train a generative model $p_{\theta}(x,y)$ that not only learns the joint distribution $p_{D}(x,y)$ by optimizing $\theta \approx \theta^*$,  but also disentangles the exclusive representations $Z^{X}$ and $Z^{Y}$ from the shared representation $Z^{S}$. 
% A typical way to learn the latent representation is maximizing the marginal likelihood \citep{DBLP:journals/corr/KingmaW13}. In this case, we maximize the marginal likelihood of the joint distribution of $X,Y$.
% We model the joint data distribution with the following VAE model that naturally reflects the assumption of independent shared / exclusive representations.
% Since the true posterior $p_{\theta}(z^{x}, z^{s}, z^{y}|x,y)=\frac{p_{\theta}(x,y,z^{x},z^{s},z^{y})}{p_{\theta}(x,y)}$ involves an intractable integration,
% we would like to fit the true posterior  with an approximate posterior $q_{\phi}$, which is factored into three terms:
\begin{equation}
    q_{\phi}(z^{x}, z^{s}, z^{y}|x,y) = q_{\phi_X}(z^{x} | x) q_{\phi_S}(z^{s} | x, y) q_{\phi_Y}(z^{y} | y)\label{eq:JointQ},
\end{equation}
where $q_{\phi_X}$ and $q_{\phi_Y}$ are encoders for domain-specific latent variable $Z^X$ and $Z^Y$, respectively, $q_{\phi_S}$ is the encoder for the shared latent variable $Z^S$, and $\phi=\{\phi_X,\phi_S,\phi_Y\}$ is the encoder parameter.
In the following, we omit subscripts $\theta$ and $\phi$ for brevity. 
Using the Eq.~\eqref{eq:JointQ}, we can derive the ELBO of Eq.~\eqref{eqn:joint_prob_opt} as follow (see \ref{appendix:ELBO} in the supplementary material for the derivation):
\begin{align}
\log p(x, y) &\geq \mathbb{E}_{q(z^{x}, z^{s}, z^{y}|x,y)}\left[ \log \frac{p(x,y,z^{x},z^{s},z^{y})}{q(z^{x}, z^{s}, z^{y}|x,y)} \right] \label{eq:ELBO}\\
&= \mathbb{E}_{q(z^{x} | x) q(z^{s} | x, y)} \left[ \log p(x | z^{x},z^{s}) \right] + \mathbb{E}_{q(z^{y} | y) q(z^{s} | x, y)} \left[ \log p(y | z^{y},z^{s}) \right] \nonumber \\
&\quad - D_{KL}\left[q(z^{x} | x) \| p(z^{x})\right] - D_{KL}\left[q(z^{y} | y) \| p(z^{y})\right]\nonumber \\
&\quad - D_{KL}\left[q(z^{s} | x, y) \| p(z^{s})\right]. \label{eq:ELBO_VIB}
\end{align}

% Unfortunately, maximizing the ELBO does not necessarily encourage that the representations learned by $q(z^{x} | x)$,  $q(z^{s} | x, y)$, and $q(z^{y} | y)$ are properly disentangled. 
% Unfortunately, maximizing the ELBO does not necessarily encourage the disentanglement of latent representations learned by $q(z^{x} | x)$,  $q(z^{s} | x, y)$, and $q(z^{y} | y)$. 
% Specifically, the following desiderata are not reflected in the objective:
Unfortunately, maximizing the ELBO does not necessarily encourage the structured representations. 
This is mainly because we have no control over the assignment of the generative factors to representations learned by three different encoders $q(z^{x} | x)$, $q(z^{s} | x, y)$, and $q(z^{y} | y)$. 
Specifically, the following desiderata of the cross-domain disentanglement should be reflected in the objective:
\begin{enumerate}
    % \item The shared representation $Z^{S}$ should not be ignored. (All representations should be learned informative.)
    \item \textbf{Disentanglement of $Z^X, Z^Y$ and $Z^S$}: 
    % one degenerate solution of Eq.~\eqref{eq:ELBO} is the case of $Z^S$ being entangled with $Z^X$ and $Z^Y$, which leads to dependency between shared and exclusive representation. To prevent this, we have to ensure that $Z^X$ and $Z^Y$ is independent of $Z^S$, and vice versa.
    % generative factors encoded in domain-specific representations $Z^X$ and $Z^Y$ should be mutually exclusive to the shared representation $Z^S$.
    the generative factors learned by $Z^X$, $Z^Y$ and $Z^S$ should be mutually exclusive to each other to avoid encoding redundant information.
    \item \textbf{Decomposition of domain-specific and shared representations}: 
    the generative factors exclusively presented in each domain should be captured by $Z^X$ and $Z^Y$, while the rest of factors shared across the domains should be encoded in $Z^S$.
    % \item \textbf{Correct mapping between representations and factors of variation}: another degenerate solution is that $Z^{S}$ / $Z^{X}$ learns some factors of domain-specific / domain-invariant variation. To prevent this, we have to introduce some explicit guidance over the assignment of factors of variation to right representation.
    % \item The exclusive representations $Z^{X}$ and $Z^{Y}$ should be disentangled from $Z^{S}$.
\end{enumerate}
To guide the model to learn desirable latent representations that satisfy the above properties, we propose to introduce regularizations on $q$ motivated by information theory, which are described below.
% This is mainly because we have no control over the assignment of the factors of variation to representations learned by three different encoders $q(z^{x} | x)$, $q(z^{s} | x, y)$, and $q(z^{y} | y)$. Thus, we need to devise a proper information regularization on those approximate posterior encoders for disentangled yet informative shared and exclusive representations.

\subsection{Information regularization on $q$ for cross-domain disentanglement}
\cutsubsectiondown
%However, this may lead to a degenerate solution that $Z^S$ donimates the $Z^X$ and $Z^Y$, encoding all factors of variations in joint distribution. To prevent this and enforce $Z^X$ and $Z^Y$ to encode domain-specific information of $\mathcal{X}$ and $Y$,
\paragraph{Enforcing disentanglement}
% A desirable exclusive representation for any domain should be disentangled from the shared representation. 
Desirable shared and exclusive representations must be disentangled so that none of factors of variation is shared across any representations.
Thus, we introduce regularizations that minimize the mutual information $I(Z^{X};Z^{S})$ and $I(Z^{Y};Z^{S})$ so that exclusive representations are statistically independent to shared representation, and vice versa. 
Here we only present our formulation for domain $X$, as the one for domain $Y$ is analogous.

To gain better insights on how minimizing the mutual information impacts the disentanglement, we rewrite $I(Z^{X};Z^{S})$ as follows (see \ref{appendix:MI} in the supplementary for details):
%(the same analysis can be applied to $I_{q}(Z^{Y};Z^{S})$):
% \sh{TODO: move the original derivation to supp}
%
\begin{align}
I(Z^{X}; Z^{S}) &= - I(X; Z^{X}, Z^{S}) + I(X; Z^{X}) + I(X; Z^{S}). \label{eq:MI}
\end{align}
%
% Surprisingly, minimization of $I(Z^{X}; Z^{S})$ leads $Z^{X}$ and $Z^{S}$ jointly informative to $X$ under the information constraints $I(Z^{X}; X)$ and $I(Z^{S}; X)$. Since those two information constraints simply penalizes the total amount of information that $Z^{X}$ and $Z^{S}$ encoded, they cannot provide any guidance on what factors of variation to be learned in specific representation. This issue can be resolved by following additional information constraint.
% Eq.~\eqref{eq:MI} shows that minimizing the mutual information  
% It implies that maximizing Eq.~\eqref{eq:MI} encourages $Z^{X}$ and $Z^{Y}$ to be jointly informative to $X$ (the first term), while minimizing the amount of information encoded in $Z^X$ and $Z^S$ (the second and third terms).
Surprisingly, Eq.~\eqref{eq:MI} implies that minimizing the mutual information of $Z^X$ and $Z^S$ encourages them to be \emph{jointly informative} to domain $X$ (the first term in RHS).
Since the last two terms will penalize the total amount of information in $Z^X$ and $Z^S$, minimizing Eq.~\eqref{eq:MI} will naturally encourage $Z^S$ and $Z^X$ to encode the \emph{mutually exclusive} information of domain $X$.
% This naturally encourages $Z^S$ and $Z^X$ to encode mutually exclusive information of domain X, as any redundant information between them will be penalized by the last two terms.
% This naturally encourages the information of domain X to be encoded exclusively to $Z^S$ and $Z^X$, as encoding any redundant information will be penalized by the last two terms.

% The third term in Eq.~\eqref{eq:MI} encourages $Z^{X}$ and $Z^{S}$ to be jointly informative to $X$, while the first and second term constrain the amount of information encoded in $Z^X$ and $Z^S$.
% {\color{red}It prevents encoding redundant information of domain X in $Z^S$ and $Z^X$, which encourages them to be mutually exclusive}. This is because the redundant information that overlaps does not give any extra informativeness to $X$ required by the third term in Eq.~\eqref{eq:MI} while $Z^X$ and $Z^S$ are constrained to have minimal sufficient encoding \citep{tishby2000information, 45903} by the first and second term in Eq.\eqref{eq:MI}.
% \sh{perhapse more emphasizing the third term..?}

However, we also notice that minimization of Eq.~\eqref{eq:MI} does not enforce any constraints on separation of domain-specific and domain-invariant representation to $Z^X$ and $Z^S$; 
any arbitrary mutually exclusive factorization will be equally preferred, even those with
no information captured in $Z^S$.
% This is mainly because $Z^{X}$ and $Z^{Y}$ are under the same constraints on the total amount of information that $Z^{X}$ and $Z^{S}$, which do not embed any preference on learning specific types of factors of variation in those representations. 
It motivates us to introduce additional regularization to enforce a proper disentanglement on domain-specific and shared information.

% Since those two information constraints simply penalizes the total amount of information that $Z^{X}$ and $Z^{S}$ encoded, they cannot provide any guidance on what factors of variation to be learned in specific representation. This issue can be resolved by following additional information constraint.
\cutparagraphup
\paragraph{Enforcing decomposition}
To encourage decomposition of domain-specific and shared representation, we introduce a regularization on the shared latent variable $Z^S$. 
Specifically, we encourage  $Z^S$ to capture the \emph{shared} information across domains, which is enforced based on \emph{interaction information}~\citep{1057469} (also known as co-information~\citep{bell2003co}).

Interaction information
% \footnote{Interaction Information is also known as co-information \citep{bell2003co}. Since its sign convention is not unified, we define it as Eq.~\eqref{eq:InterInfo}.}
 is a generalization of mutual information among three or more random variables, and quantifies the amount of shared information among them. 
Specifically, we define the interaction information among two domains $X$, $Y$ and the shared representation $Z^S$ as follows:
% \sh{only demonstrated for domain X. need to mention it.}
%
\begin{align}
    % I\left(X; Y; Z\right) = I\left(X; Y\right) - I\left(X; Y | Z\right) = I\left(X; Z\right) - I\left(X; Z | Y\right) = I\left(Y; Z\right) - I\left(Y; Z | X\right) 
    I(X;Y;Z^{S}) &= I\left(X; Z^{S}\right) - I\left(X; Z^{S} | Y\right) \label{eq:InterInfo}\\
    &= I\left(Y; Z^{S}\right) - I\left(Y; Z^{S} | X\right), \label{eq:InterInfoSym}
\end{align}
where the equality in Eq.~\eqref{eq:InterInfoSym} holds due to symmetry.
The above equations show how maximizing interaction information encourages $Z^S$ to encode the shared information. 
For instance, in Eq.~\eqref{eq:InterInfo}, the first term in RHS is maximized when $Z^{S}$ becomes informative to $X$, while the second term will be minimized if such information in $Z^S$ can be \emph{also inferred} from $Y$;
the combination of both terms will naturally make $Z^S$ to encode information shared between $X$ and $Y$. %the informativeness of $Z^{S}$ to $X$ when $Y$ is observed. 

\cutparagraphup
\paragraph{Joint regularization} 
% Individual optimization of Eq.~\eqref{eq:MI} or Eq.~\eqref{eq:InterInfo} leads either of failure to decompose domain-specific and shared representation or stopping shared representation from being too informative. 
% Fortunately, simultaneous maximization of $I_{q}(X;Y;Z^{S})$ and minimization of $I(Z^{X}; Z^{S})$ resolves the issue of optimizing any one of them.
Our final regularization on cross-domain disentanglement is obtained by combining regularizations on disentanglement and decomposition.
To make analysis easier, we first present the objective with respect to domain $X$ and show the complete one on both domains later.

Combining Eq.~\eqref{eq:MI} and \eqref{eq:InterInfo}, 
our preference for $q$ on domain $X$ (the negative of regularization) 
becomes
\begin{align}
\max_q \, &I(X;Y;Z^{S}) - I(Z^{X}; Z^{S})\nonumber\\
&= \underbrace{\cancel{I(X;Z^{S})} - I(X;Z^{S}|Y)}_{I(X;Y;Z^{S})} + \underbrace{I(X ; Z^{X}, Z^{S}) - I(X;Z^{X}) - \cancel{I(X;Z^{S})}}_{- I(Z^{X}; Z^{S})} \nonumber\\
&= I(X; Z^{X}, Z^{S}) - I(X;Z^{X}) - I(X;Z^{S}|Y). \label{eqn:joint_inforeg}
% &\geq H(X) + \mathbb{E}_{p_{D}(x, y) q_{s}(z^{s} | x,y) q_{x}(z^{x} | x)}\left[\log p(x | z^{x}, z^{s})\right] - \mathbb{E}_{p_{D}(x,y)}\left[D_{KL}\left[ q_{x}(z^{x} | x) \| r_{x}(z^{x})\right]\right]\\
% &\quad \quad \quad \quad - \mathbb{E}_{p_{D}(x,y)}\left[D_{KL}\left[ q_{s}(z^{s} | x, y) \| r_{s}(z^{s} | y)\right]\right] \\
\end{align}

\paragraph{Optimization}
Direct optimization of Eq.~\eqref{eqn:joint_inforeg} is intractable since each term involves several intractable integrals.
The details are in~\ref{appendix:InfoConst} in the supplementary material.

The first term $I(X; Z^X, Z^S)$ in Eq.~\eqref{eqn:joint_inforeg} is intractable since $q(x|z^{x}, z^{s}) = \frac {q(z^{x}, z^{s} | x) p_{D}(x)}{\int p_{D}(x,y) \hspace*{0.1cm} q(z^{x}, z^{s} | x, y) \hspace*{0.1cm} dx dy}$ involves intractable integral (unknown $p_D(x,y)$ and $p_D(x)$).
Thus, we derive its lower bound with the generative distribution $p(x|z^{x}, z^{s})$ as follows:
\begin{align}
&I(X; Z^{X}, Z^{S}) = \mathbb{E}_{q(z^{x}, z^{s} | x) p_{D}(x)}\left[\log \frac{q(x | z^{x}, z^{s})}{p_{D}(x)}\right] \nonumber\\
&\quad \quad \quad = H(X) + \mathbb{E}_{q(z^{x}, z^{s} | x) p_{D}(x)}\left[\log p(x | z^{x}, z^{s}) \right] + \mathbb{E}_{q(z^{x}, z^{s})}\left[ D_{KL} \left[ q(x | z^{x}, z^{s}) \| p(x | z^{x}, z^{s}) \right] \right] \nonumber\\
&\quad \quad \quad \geq H(X) + \mathbb{E}_{q(z^{x}, z^{s} | x) p_{D}(x)}\left[\log p(x | z^{x}, z^{s}) \right] \nonumber\\
&\quad \quad \quad = H(X) + \mathbb{E}_{p_{D}(x,y) \hspace*{0.1cm} q(z^{x} | x) \hspace*{0.1cm} q(z^{s} | x, y)}\left[\log p(x | z^{x}, z^{s}) \right] \label{eqn:opt1}.
\end{align}
Note that maximization of Eq.~\eqref{eqn:opt1} not only maximizes $I(X; Z^{X}, Z^{S})$ but also fits $p(x | z^{x}, z^{s})$ to $q(x | z^{x}, z^{s})$ so that we can utilize it as a decoder.

The second term $-I(X;Z^X)$ is intractable since $q(z^{x}) = \int p_{D}(x) q(z^{s} | x) \hspace*{0.1cm} dx$ is intractable (unknown distribution $p_D(x)$). 
We use $-\mathbb{E}_{p_D(x)} \left[D_{KL}\left[q(z^{x} | x) \| p(z^{x})\right] \right]$ as its lower bound with the generative distribution $p(z^{x})$ defined as
the standard Gaussian, which is also known as the Variational Information Bottleneck (VIB)~\citep{45903}.

The last term is also intractable because $q(z^{s} | y) = \int p_{D}(x|y) q(z^{s} | x, y) dx$ is intractable (unknown $p_D(x|y)$). Similar to VIB, we use variational distribution $r^{y}(z^{s} | y)$ to maximize its lower bound:
\begin{align}
-I(X;Z^{S}|Y)
&= -\mathbb{E}_{p_{D}(x,y) q(z^{s} | x, y)}\left[ \log \frac{q(z^{s} | x, y)}{q(z^{s} | y)} \right] \nonumber\\%\hspace{-0.5ex}
&= -\mathbb{E}_{p_{D}(x,y) q(z^{s} | x, y)}\left[ \log \frac{q(z^{s} | x, y) r^{y}(z^{s} | y)}{r^{y}(z^{s} | y)q(z^{s} | y)} \right] \nonumber \\
&= -\mathbb{E}_{p_{D}(x,y)}\left[D_{KL}\left[ q(z^{s} | x, y) \| r^{y}(z^{s} | y)\right]\right] + \mathbb{E}_{p_{D}(y)}\left[D_{KL}\left[ q(z^{s} | y) \| r^{y}(z^{s} | y)\right]\right] \nonumber \\
&\geq -\mathbb{E}_{p_{D}(x,y)}\left[D_{KL}\left[ q(z^{s} | x, y) \| r^{y}(z^{s} | y)\right]\right].\label{eqn:opt3}
\end{align}
Thus, the maximization of  Eq.~\eqref{eqn:opt3} not only minimizes $I(X;Z^{S}|Y)$ but also fits $r^{y}(z^{s} | y)$ to $q(z^{s} | y)$.
% Combining lower bounds of those three terms, we achieve the lower bound on Eq.~\eqref{eqn:joint_inforeg} as well as variational distributions $p(x|z^{x}, z^{s})$ and $r^{y}(z^{s} | y)$.
Putting together, we are ready to derive the lower bound of
the preference for $q$ on domain $X$ and $Y$: %Eq.~\eqref{eqn:final_inforeg}.
\begin{align}
&(I(X;Y;Z^{S}) - I(Z^{X};Z^{S})) + (I(X;Y;Z^{S}) - I(Z^{Y};Z^{S}))\nonumber \\
&=2 \cdot I(X;Y;Z^{S}) - I(Z^{X};Z^{S}) - I(Z^{Y};Z^{S}) \nonumber \\
&= I(X; Z^{X}, Z^{S}) + I(Y; Z^{Y}, Z^{S}) - I(X;Z^{X}) - I(Y;Z^{Y}) - I(X;Z^{S}|Y) - I(Y;Z^{S}|X) \nonumber\\
&\geq \mathbb{E}_{p_D(x,y)} \left[\hspace*{0.1cm} \mathbb{E}_{q(z^{s} | x,y) q(z^{x} | x)}\left[\log p(x | z^{x}, z^{s}) \right] + \mathbb{E}_{q(z^{s} | x,y) q(z^{y} | y)}\left[ \log p(y | z^{y}, z^{s})\right] \hspace*{0.1cm} \right] \nonumber\\
&\quad - \mathbb{E}_{p_{D}(x,y)}\left[ \hspace*{0.1cm} D_{KL}\left[ q(z^{x} | x) \| p(z^{x}) \right] + D_{KL}\left[q(z^{y} | y) \| p(z^{y})\right] \hspace*{0.1cm} \right] \nonumber\\
&\quad - \mathbb{E}_{p_{D}(x,y)}\left[ \hspace*{0.1cm} D_{KL}\left[ q(z^{s} | x, y) \| r^{y}(z^{s} | y) \right] + D_{KL}\left[ q(z^{s} | x, y) \| r^{x}(z^{s} | x)\right] \hspace*{0.1cm}\right] \nonumber\\
&\quad + H(X) + H(Y).\label{eqn:final_inforeg}
\end{align}
Surprisingly, many of the terms are also present in the ELBO. Thus, when we
add the above lower bound to the ELBO objective to perform joint
optimization, many of the terms above are obtained with 
very little additional cost by sharing parameters and computations, which we describe below.

\subsection{Interaction Information Auto-Encoder}
\cutsubsectiondown
\begin{figure}[!t]
 \centering
 \includegraphics[width=.9\textwidth]{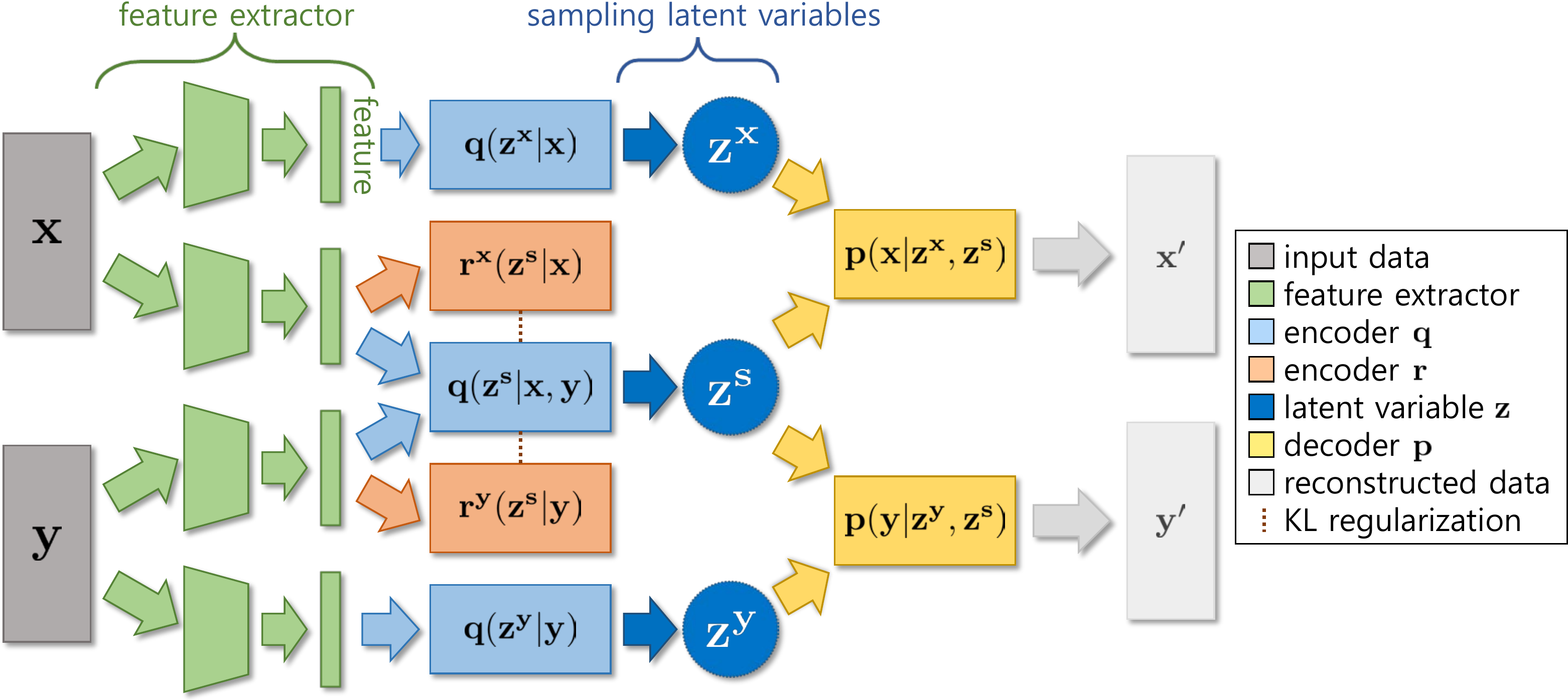}
 \caption{The architecture of Interaction Information Auto-Encoder.}
 \label{fig:IIAE}
 \vspace{-0.1cm}
\end{figure}
% Our goal is to train a generative model with the ELBO in Eq.~\eqref{eq:ELBO} under the information constraints for cross-domain disentanglement defined in Eq.~\eqref{eqn:final_inforeg}. 
% However, the constrained optimization is relatively difficult to optimize. 
% Therefore, we relax the information constraint by allowing the model to maximize it jointly with the ELBO as done in a similar way in \citep{moyer2018invariant}. The final objective function is defined as follow (see \ref{appendix:InfoConst} in the supplementary file for full proofs).
Our goal is to learn a latent variable model with maximum likelihood objective (ELBO in Eq.~\eqref{eq:ELBO}) under the the information regularization for cross-domain disentanglement (Eq.~\eqref{eqn:final_inforeg}).
Due to the difficulties in the constrained optimization, we relax this problem as a joint maximization problem similar to \citep{moyer2018invariant},
which we name 
Interaction Information Auto-Encoder (IIAE) shown in figure \ref{fig:IIAE},
as follows (see \ref{appendix:InfoConst} in the supplementary material for details):
\begin{align}
%&\max \mathbb{E}_{p_{D}(x, y)}\left[\log p_{\theta}(x,y)\right]
&\max_{p,q} \mathbb{E}_{q(z^{x}, z^{s}, z^{y}, x,y)}\left[ \log \frac{p(x,y,z^{x},z^{s},z^{y})}{q(z^{x}, z^{s}, z^{y}|x,y)} \right] + \lambda \left(2\cdot I(X;Y;Z^{S}) - I(Z^{X};Z^{S}) - I(Z^{Y};Z^{S}) \right) \nonumber\\
&\geq \max_{p,q,r} (1+\lambda) \cdot \mathbb{E}_{p_{D}(x,y)} \left[ \hspace*{0.1cm} ELBO(p,q) \hspace*{0.1cm} \right]\nonumber\\
&\quad \quad \quad \quad + \lambda \cdot \mathbb{E}_{p_{D}(x,y)} \left[ \hspace*{0.1cm} D_{KL}\left[ q(z^{s} | x, y) \| p(z^{s})\right] \hspace*{0.1cm} \right] \label{eqn:final2}\\
&\quad \quad \quad \quad - \lambda \cdot \mathbb{E}_{p_{D}(x,y)} \left[ \hspace*{0.1cm} D_{KL}\left[ q(z^{s} | x, y) \| r^{y}(z^{s} | y) \right] + D_{KL}\left[ q(z^{s} | x, y) \| r^{x}(z^{s} | x) \right] \hspace*{0.1cm} \right].\label{eqn:final3}
\end{align}
% Ignoring constant entropy terms $H(X)$ and $H(Y)$, we proved the inequality of \eqref{eq:final_inequal} in \ref{appendix:InfoConst} in the supplementary file.

% We indicate that sources of Eq.~\eqref{eq:final_inequal} and Eq.~\eqref{eq:final_ex_VIB} (with blue underline) are from both of the ELBO and the information constraints, Eq.~\eqref{eq:final_sh_VIB} (green) from only the ELBO, and Eq.~\eqref{eq:final_sh_invariant} (red) from only the information constraints.
This objective is essentially augmenting the ELBO with Eq.~\eqref{eqn:final2} and Eq.~\eqref{eqn:final3}, which trades off the overall amount of information captured by the shared representation with that from the domain-specific information, by factor $\lambda$.
This augmented term encourages the shared representation to exclude domain-specific factors of variation.
Finally, note that Eq.~\eqref{eqn:final3} yields variational encoders $r^{x}(z^{s} | x)$ and $r^{y}(z^{s} | y)$ as byproducts of optimization, which is useful for many tasks such as image translation and retrieval where we need to extract the shared representation $z^s$ only from $x$ or $y$.
\section{Related Work}
% \cutsectiondown
\label{others}
% \paragraph{Information bottleneck}
\cutparagraphup
\paragraph{Invariant representation}
Representation learning \citep{lecun2015deep} focuses on feature extraction from the data that is 
informative to the given task. Information bottleneck (IB) \citep{tishby2000information} was
introduced as an information theoretic regularization method to achieve minimal sufficient encoding 
by constraining the amount of information that latent variable encodes observed variable. IB  
enables the encoder to filter out nuisance factors and thus to generalize well. 
IB is later extended to deep VIB \citep{45903}, which parameterizes IB with a neural network 
and optimizes the variational lower bound of the IB objective. VIB showed a close relationship to 
VAEs \citep{DBLP:journals/corr/KingmaW13} and $\beta$-VAEs \citep{higgins2017beta} by extending 
their models to unsupervised learning. Based on VIB, several  methods were developed  
\citep{song2018learning, moyer2018invariant} to learn encoders that capture only the factors of 
variation invariant to the given attribute. Similarly, a variant of VIB was proposed by 
\citep{federici2020} to learn a domain invariant representation by discarding domain specific 
variations. GRL \citep{pmlr-v37-ganin15} is another approach to achieve an invariant approach, which 
has been widely adopted to the tasks such as unlearning the bias in the input data 
\citep{kim2019learning}, domain adaptation \citep{pmlr-v37-ganin15, gonzalez-garcia2018NeurIPS}, 
and zero-shot image retrieval \citep{Dey_2019_CVPR}. The idea of learning invariant representations 
in zero-shot learning has been explored as well \citep{kodirov2017semantic, felix2018multi, 
'shen2018cvpr', Yelamarthi_2018_ECCV, Dutta2019SEMPCYC, linlearning}, aiming to achieve 
domain-invariant representation by regularizing the model with multiple tasks or objectives.

\cutparagraphup
\paragraph{Disentangled representation}
Based on $\beta$-VAEs \citep{higgins2017beta}, there has been extensive research on disentangled 
representation. Total correlation \citep{watanabe1960information} is quantified as a measure of
statistical dependency among all dimensions of the latent variable, which was the basis of the work
by  
\cite{pmlr-v80-kim18b, chen2018isolating, pmlr-v89-gao19a, pmlr-v89-esmaeili19a, jeongICML19}. 
Modeling hierarchical structure in the latent space was also introduced by \citep{pmlr-v70-zhao17c, 
hsu2017unsupervised}, expecting that representations learned in each level is disentangled from 
other levels in the hierarchy. Extending the conditional generative models \citep{pix2pix2017, 
zhu2017toward}, Cross-domain Disentanglement Networks (CdDN) \citep{gonzalez-garcia2018NeurIPS} 
introduced the concept of cross-domain disentanglement for image-to-image translation task, which 
is about disentangling domain-specific representation from the shared representation. As 
cross-domain disentanglement problem assumes paired dataset, there have been several follow-up studies 
\citep{NIPS2018_7525, lee2018diverse, press2018emerging, yu2019multi} that extend cross-domain 
disentanglement to the case only unpaired data is available.

% \subsection{Image translation}
% \subsection{Zero-shot sketch based image retrieval}

% !TEX root = neurips_2020.tex
% \cutsectionup
\section{Experiments}
\label{sec:experiment}
\cutsectiondown
We employ experiments on image-to-image translation and image retrieval tasks to evaluate the quality of cross-domain 
disentanglement.
% In order to evaluate the quality of cross-domain disentanglement, \update{we employ image-to-image translation and image retrieval tasks.}
% we employ the two tasks. 
% Image-to-image translation for qualitative analysis, and cross-domain image retrieval for quantitative evaluation.
In both tasks, the main objective is to evaluate how our method encodes the 
domain-specific and the shared information into different representations ($Z^X$, 
$Z^Y$, and $Z^S$).

\subsection{Cross-domain Image Translation}
% \cutsubsectiondown
% For qualitative assessment of cross-domain disentanglement, we apply our method to cross-domain image-to-image translation.

\cutparagraphup
\paragraph{Datasets}
We evaluate our method on two datasets: MNIST-CDCB \citep{gonzalez-garcia2018NeurIPS} and Cars \citep{reed2015deep} datasets.
In MNIST-CDCB \citep{gonzalez-garcia2018NeurIPS} dataset, each pair $(x,y)$ consists of two images of the same digit but in different color patterns. 
Specifically, images in domain $X$ have color variations in the background, while the ones in domain $Y$ have variations in the foreground.%font face. 
We use 50,000 / 10,000 pairs of train/test samples following \citep{MNIST}. 
Cars \citep{reed2015deep} is a dataset of car CAD images with equally spaced variations in orientation, 4 different angles in pitch and 24 in yaw. 
We employ 92 pairs of $(x,y)$ per a car, where $x$ is fixed as a frontal view of every 
pitch, and $y$ is rotated view of rest 23 different angles in yaw. 
Out of those 16,836 pairs of 183 cars, we assigned 16,192 pairs of 176 cars to train 
set and 644 pairs of 7 cars to test set.

\cutparagraphup
\paragraph{Method}
Translating an image across domains ($X\to Y$ or $Y\to X$) can be done naturally by our method.
% For instance, we can translate an image $x\inX$ to $y'\inY$ by (1) extracting its shared representation using the encoder $\mu^s=r_x(z^s|x)$, (2) sampling the domain-specific representation either from the prior $z^y\sim p(z^y)$ or using the reference image $z^y=\mu^y=q(z^y|y)$, and (3) generating the image by $y'=p(y|\mu^s,z^y)$.
Specifically, we translate image $x$ in domain $X$ to domain $Y$ by (1) extracting its shared representation using the mean $\mu_x^s$ of $r^x(z^s|x)$, (2) sampling the domain-specific representation from the prior $z^y\sim p(z^y)$, and (3) generating the image by the mean $y'$ of $p(y|\mu^s_x,z^y)$.
When we have the reference image $y$ in another domain, we can also conduct a guided translation by replacing the second step to extract domain-specific representation using the $z^y=\mu^y$, the mean of $q(z^y|y)$.
Note that translation in the opposite direction can be done similarly. 
For network architecture, we employ the settings used in \citep{gonzalez-garcia2018NeurIPS} with some minor modifications.
We leave all the implementation details and hyperparameter settings in \ref{appendix:implementation} in the supplementary material.

\cutparagraphup
\paragraph{Results}
Table \ref{tab:I2I} shows the result of image translation with IIAE. 
For each row, we show the ground-truth pair $(x,y)$ (the first and sixth column), and the translated images between the domains.
We present two types of translation results obtained by (1) sampling domain-specific representation from the prior $z\sim p(z)$ (columns 2$\sim$4 and 7$\sim$9) and (2) using the one extracted from the ground-truth pair $\mu\sim r(\cdot)$ (columns 5 and 10). More  results can be found in \ref{appendix:visual} in the supplementary material.

% Those samples are achieved by random sampling from the prior (columns 2$\sim$4) or the mean of domain specific posterior (column 5) extracted from the given auxiliary input image from opposite domain. We evaluate cross reconstruction as well, a task of recovering an image using the exclusive representation extracted from the original image and the shared representation from an auxiliary image in opposite domain. We set the auxiliary input as the image with the same shape (MNIST-CDCB) or the same content (Cars) which are the input of the opposite mapping.

% discussions on MNIST
In MNIST-CDCB \citep{gonzalez-garcia2018NeurIPS} dataset (upper half), we observe that shared and exclusive representations learned by IIAE are disentangled in a way that shared representation encoders only preserve the shape information and domain-specific encoders capture only the color information. 
We also observe multi-modal outputs in the translation results, which implies that various generative factors exclusively presented in each domain are captured by domain-specific representation. 
% Those generated images in any domain show variation in colors, which implies that the learned domain-specific representations are able to express multimodal distributions of colors. 
Furthermore, the images of the first and the last columns look alike as well as fifth column and sixth column do, which tells us that our shared representation encoders $r^{x}(z^{s} | x)$ and $r^{y}(z^{s} | y)$ provide nicely aligned representation.

% discussions on Car dataset
In Cars~\citep{reed2015deep} dataset (bottom half), cross-domain disentanglement is much more challenging since the object in each training pair $(x,y)$ can have different geometric configurations. 
From these data, the model should learn that the shared representation is the car identity, and the domain-specific variations are about the types of geometric transformations (fixed to front-view in domain $X$ and different rotation angles in domains $Y$'s). 
%  Furthermore, the model is prone to overfitting due to the small number of cars. 
Under those challenges, the results show that IIAE can successfully learn disentangled representations.
When translating an image from $X$ to $Y$ domain, it generates various orientations 
while keeping the car identity (second to fourth columns), whereas producing the 
consistent front-view images when translated in reverse direction (seventh to tenth 
columns). Quantitative evaluation on the sample generation is in the supplementary material \ref{appendix:sample_quality}.
%  According to the different values of domain-specific representation from the prior of Y domain, cars in the second to  columns are varying only the orientation while keeping the identity of the car consistent. On the other hand, cars in seventh to tenth columns are stationary.
%  This is a desirable phenomenon since there is no factors of variation in $X$ domain since orientation is fixed and car identity is shared with $Y$ domain.
% of IIAE on learning disentangled representations by translating images from one domain to the opposite domain. .we used MNIST-CDCB \citep{gonzalez-garcia2018NeurIPS} and Cars datasets \citep{reed2015deep}. MNIST-CDCB is composed of two domains of images, colorized digit (CD) and colorized background (CB). Every sample in the dataset is a pair of $x,y$ whose shapes are aligned while colors in each domain are independently chosen. \ref{tab:I2I} shows the result of image translation with different sources of the exclusive representations. 
\begin{table}
    \begin{center}
    \begin{tabular}{ cccccccccccc }
    \toprule
    \multicolumn{5}{c}{\textbf{ $X \rightarrow Y$ }} & \multicolumn{5}{c}{\textbf{ $Y \rightarrow X$ }} \\
    \cmidrule(lr){1-5}\cmidrule(lr){6-10}
    \textbf{Input} & \multicolumn{4}{c}{\textbf{ Outputs w/ different $z^{y}$ }} & \textbf{Input} &  \multicolumn{4}{c}{\textbf{ Outputs w/ different $z^{x}$ }} \\
    \cmidrule(lr){1-1}\cmidrule(lr){2-5}\cmidrule(lr){6-6}\cmidrule(lr){7-10}
    x & \multicolumn{3}{c}{$z^{y}_{1},z^{y}_{2},z^{y}_{3} \sim p(z^{y})$} & $\mu^{y}$ & y & \multicolumn{3}{c}{$z^{x}_{1},z^{x}_{2},z^{x}_{3} \sim p(z^{x})$}  & $\mu^{x}$\\
    \cmidrule(lr){2-4}\cmidrule(lr){7-9}
    \includegraphics[width=0.06\textwidth]{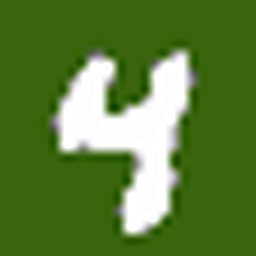} &
    \includegraphics[width=0.06\textwidth]{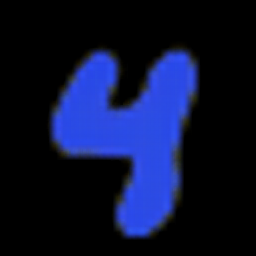} &
    \includegraphics[width=0.06\textwidth]{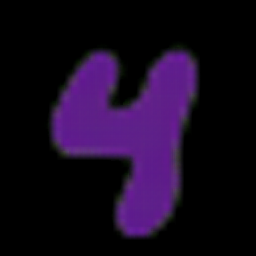} &
    \includegraphics[width=0.06\textwidth]{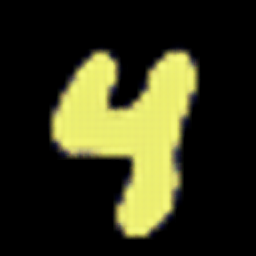} &
    \includegraphics[width=0.06\textwidth]{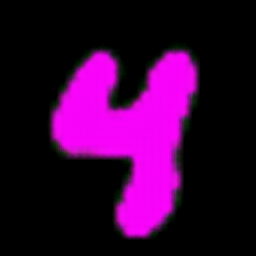} &
    \includegraphics[width=0.06\textwidth]{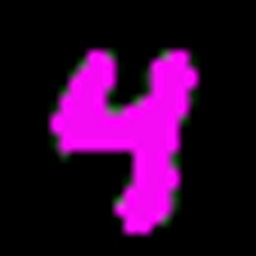} &
    \includegraphics[width=0.06\textwidth]{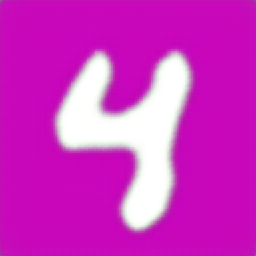} &
    \includegraphics[width=0.06\textwidth]{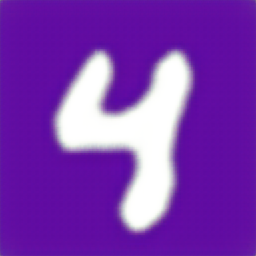} &
    \includegraphics[width=0.06\textwidth]{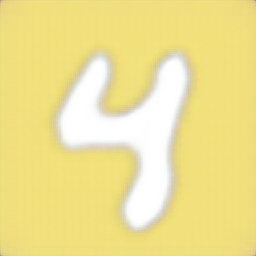} &
    \includegraphics[width=0.06\textwidth]{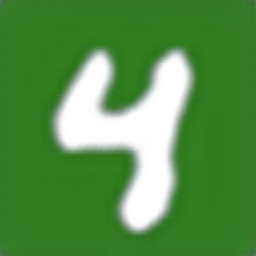} \\
    \includegraphics[width=0.06\textwidth]{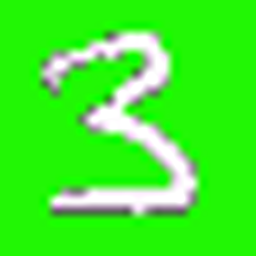} &
    \includegraphics[width=0.06\textwidth]{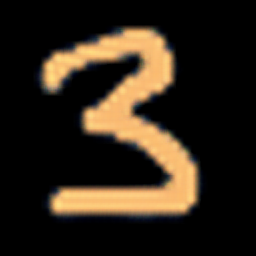} &
    \includegraphics[width=0.06\textwidth]{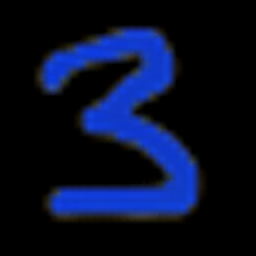} &
    \includegraphics[width=0.06\textwidth]{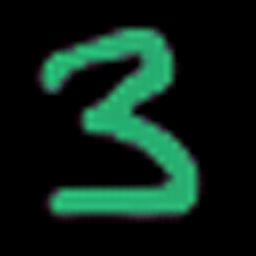} &
    \includegraphics[width=0.06\textwidth]{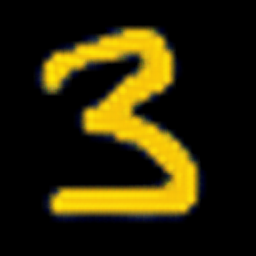} &
    \includegraphics[width=0.06\textwidth]{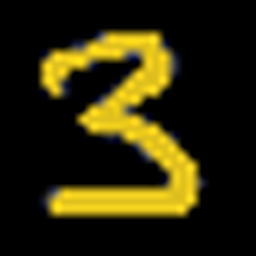} &
    \includegraphics[width=0.06\textwidth]{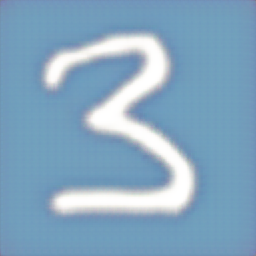} &
    \includegraphics[width=0.06\textwidth]{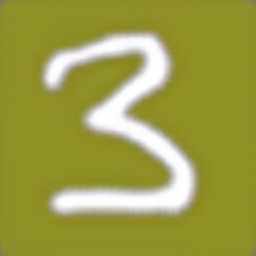} &
    \includegraphics[width=0.06\textwidth]{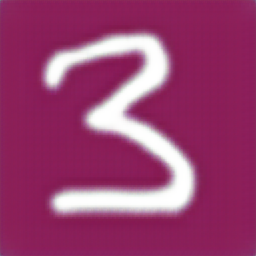} &
    \includegraphics[width=0.06\textwidth]{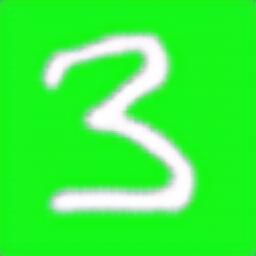} \\
    \includegraphics[width=0.06\textwidth]{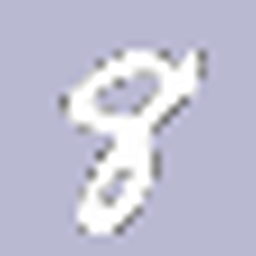} &
    \includegraphics[width=0.06\textwidth]{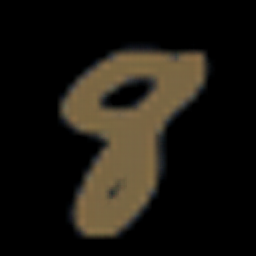} &
    \includegraphics[width=0.06\textwidth]{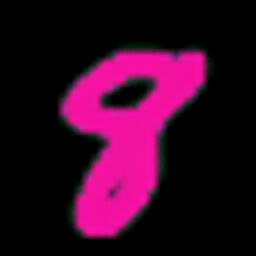} &
    \includegraphics[width=0.06\textwidth]{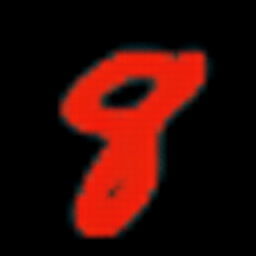} &
    \includegraphics[width=0.06\textwidth]{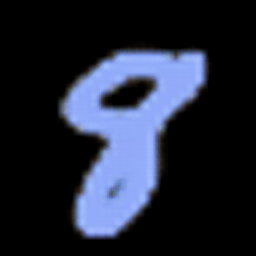} &
    \includegraphics[width=0.06\textwidth]{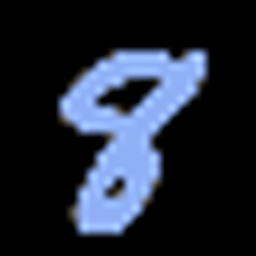} &
    \includegraphics[width=0.06\textwidth]{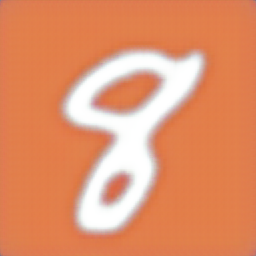} &
    \includegraphics[width=0.06\textwidth]{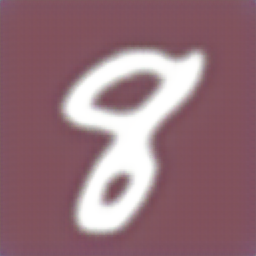} &
    \includegraphics[width=0.06\textwidth]{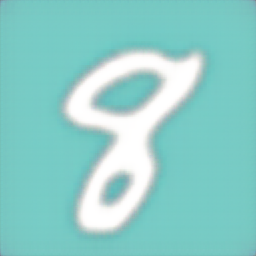} &
    \includegraphics[width=0.06\textwidth]{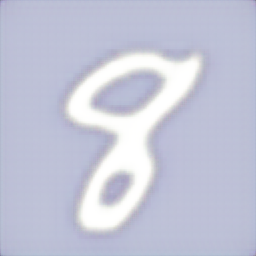} \\
    \includegraphics[width=0.06\textwidth]{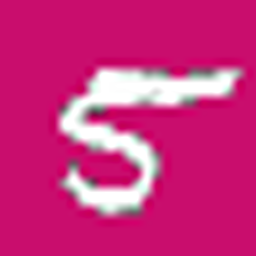} &
    \includegraphics[width=0.06\textwidth]{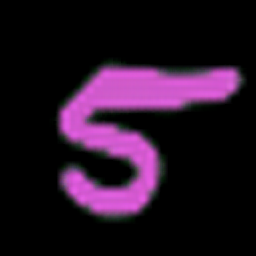} &
    \includegraphics[width=0.06\textwidth]{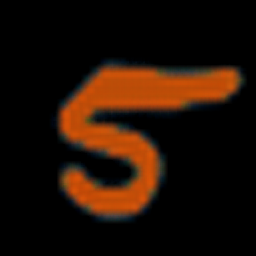} &
    \includegraphics[width=0.06\textwidth]{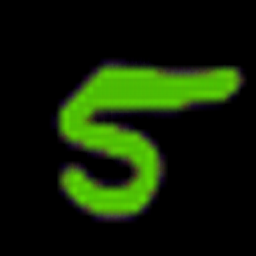} &
    \includegraphics[width=0.06\textwidth]{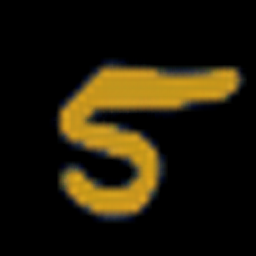} &
    \includegraphics[width=0.06\textwidth]{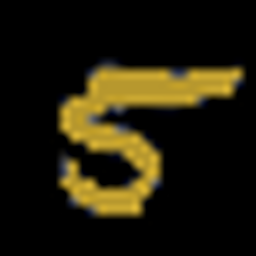} &
    \includegraphics[width=0.06\textwidth]{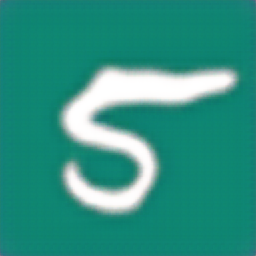} &
    \includegraphics[width=0.06\textwidth]{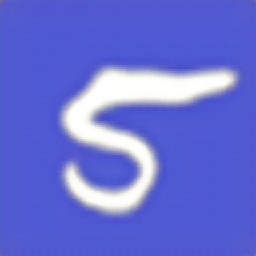} &
    \includegraphics[width=0.06\textwidth]{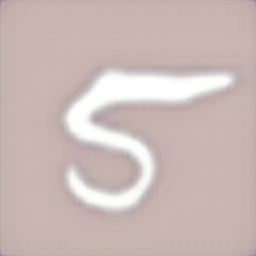} &
    \includegraphics[width=0.06\textwidth]{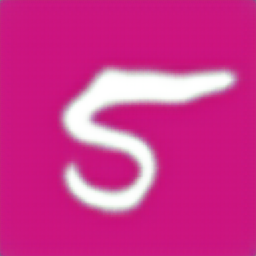} \\
    \includegraphics[width=0.06\textwidth]{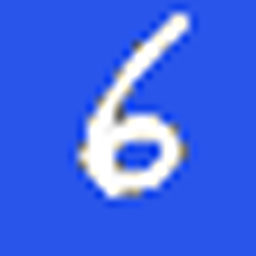} &
    \includegraphics[width=0.06\textwidth]{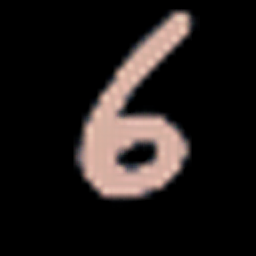} &
    \includegraphics[width=0.06\textwidth]{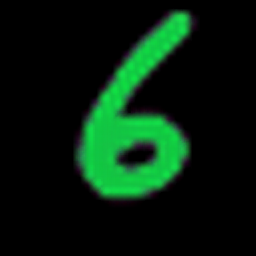} &
    \includegraphics[width=0.06\textwidth]{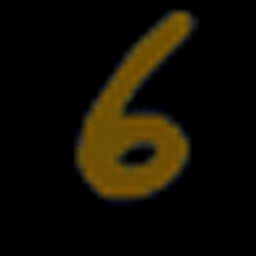} &
    \includegraphics[width=0.06\textwidth]{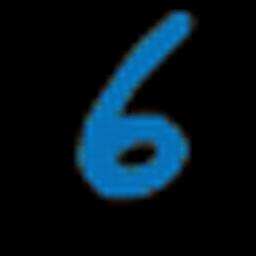} &
    \includegraphics[width=0.06\textwidth]{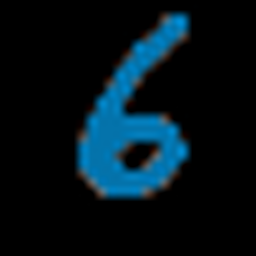} &
    \includegraphics[width=0.06\textwidth]{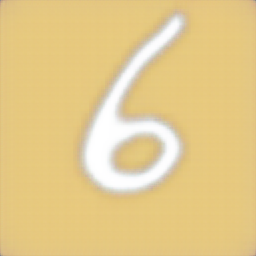} &
    \includegraphics[width=0.06\textwidth]{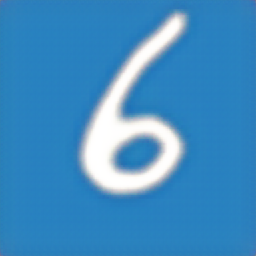} &
    \includegraphics[width=0.06\textwidth]{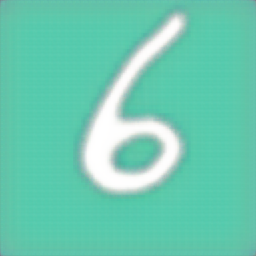} &
    \includegraphics[width=0.06\textwidth]{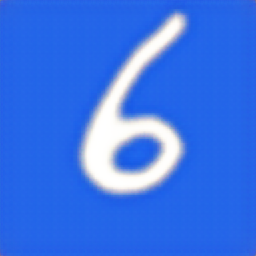} \\
    \midrule
    \includegraphics*[width=0.07\textwidth, viewport=48 64 216 196]{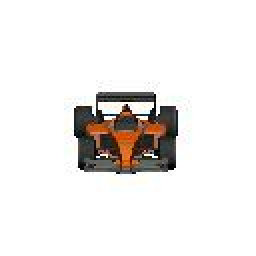} &
    \includegraphics*[width=0.07\textwidth, viewport=48 64 216 196]{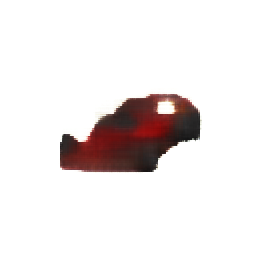} &
    \includegraphics*[width=0.07\textwidth, viewport=20 64 236 196]{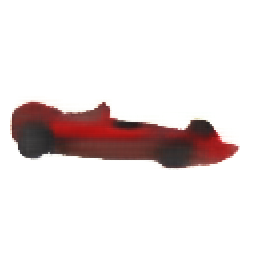} &
    \includegraphics*[width=0.07\textwidth, viewport=20 64 236 196]{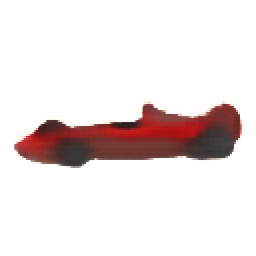} &
    \includegraphics*[width=0.07\textwidth, viewport=20 64 236 196]{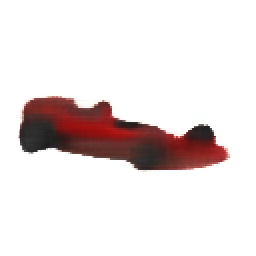} &
    \includegraphics*[width=0.07\textwidth, viewport=20 64 236 196]{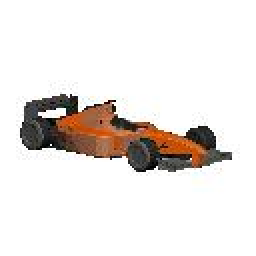} &
    \includegraphics*[width=0.07\textwidth, viewport=48 64 216 196]{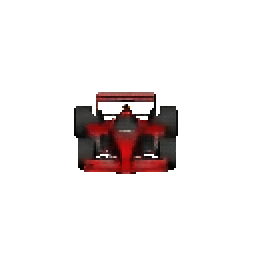} &
    \includegraphics*[width=0.07\textwidth, viewport=48 64 216 196]{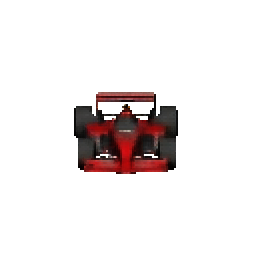} &
    \includegraphics*[width=0.07\textwidth, viewport=48 64 216 196]{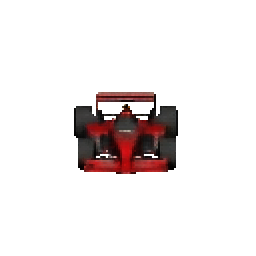} &
    \includegraphics*[width=0.07\textwidth, viewport=48 64 216 196]{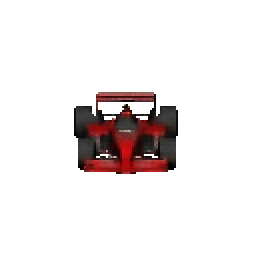} \\
    \includegraphics*[width=0.07\textwidth, viewport=48 64 216 196]{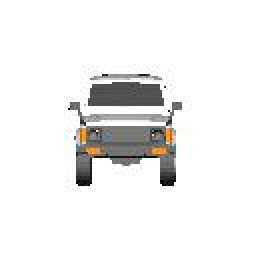} &
    \includegraphics*[width=0.07\textwidth, viewport=20 64 236 196]{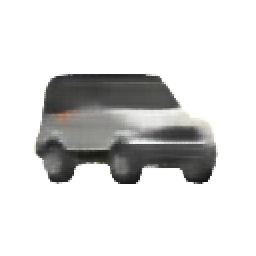} &
    \includegraphics*[width=0.07\textwidth, viewport=32 64 224 196]{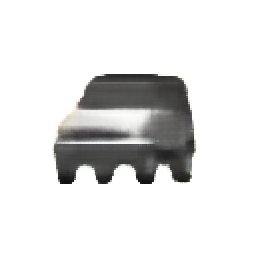} &
    \includegraphics*[width=0.07\textwidth, viewport=32 64 224 196]{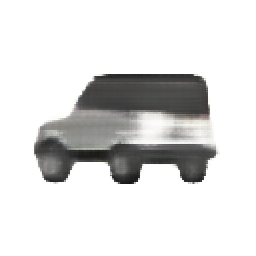} &
    \includegraphics*[width=0.07\textwidth, viewport=32 64 224 196]{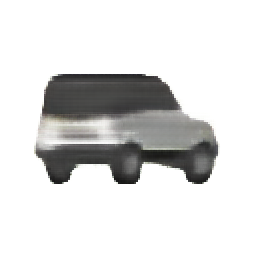} &
    \includegraphics*[width=0.07\textwidth, viewport=32 64 224 196]{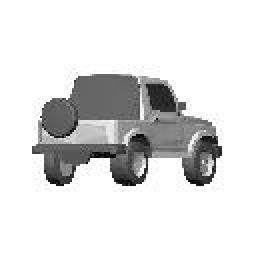} &
    \includegraphics*[width=0.07\textwidth, viewport=48 64 216 196]{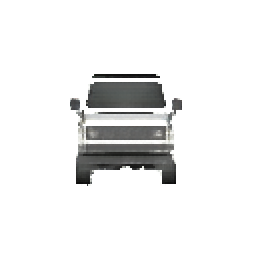} &
    \includegraphics*[width=0.07\textwidth, viewport=48 64 216 196]{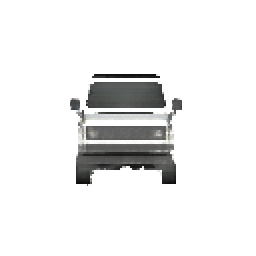} &
    \includegraphics*[width=0.07\textwidth, viewport=48 64 216 196]{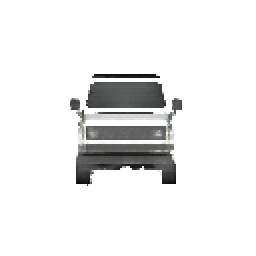} &
    \includegraphics*[width=0.07\textwidth, viewport=48 64 216 196]{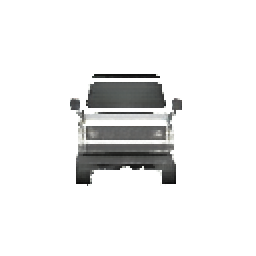} \\
    \includegraphics*[width=0.07\textwidth, viewport=56 64 200 192]{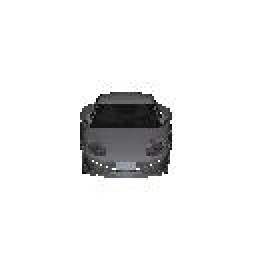} &
    \includegraphics*[width=0.07\textwidth, viewport=16 64 240 192]{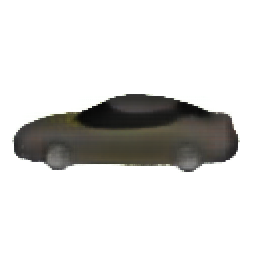} &
    \includegraphics*[width=0.07\textwidth, viewport=16 64 240 192]{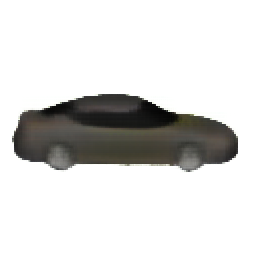} &
    \includegraphics*[width=0.07\textwidth, viewport=32 64 224 192]{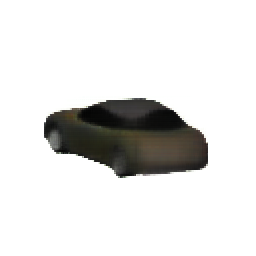} &
    \includegraphics*[width=0.07\textwidth, viewport=48 64 218 192]{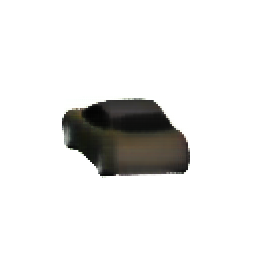} &
    \includegraphics*[width=0.07\textwidth, viewport=48 64 218 192]{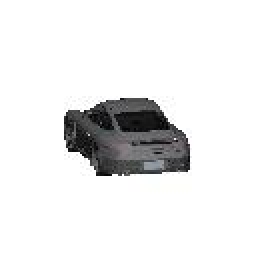} &
    \includegraphics*[width=0.07\textwidth, viewport=56 64 200 192]{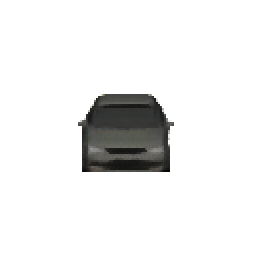} &
    \includegraphics*[width=0.07\textwidth, viewport=56 64 200 192]{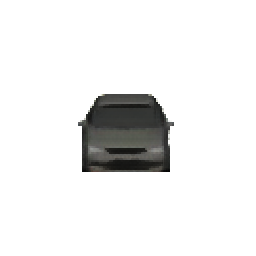} &
    \includegraphics*[width=0.07\textwidth, viewport=56 64 200 192]{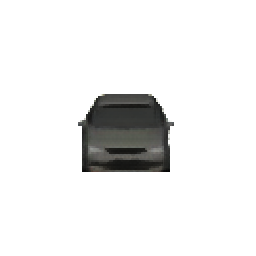} &
    \includegraphics*[width=0.07\textwidth, viewport=56 64 200 192]{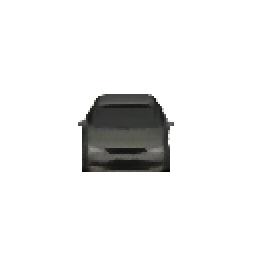} \\
    \bottomrule
    \end{tabular}
    \caption{Cross-domain translation results in MNIST-CDCB \citep{gonzalez-garcia2018NeurIPS} (top half) and Cars \citep{reed2015deep} (bottom half) generated by IIAE. In MNIST-CDCB, domain-specific factors are color variation in the background ($X$) and the foreground ($Y$) while the common factor is the digit identity. 
    In Cars, domain-specific factors only exists in $Y$, views in 23 different yaw angles, while the ones in $X$ is fixed to the front-view. The shared factor is the car identity.}
    \label{tab:I2I}
    \end{center}
    \vspace{-0.6cm}
\end{table}

% \cutsubsectionup
\subsection{Image Retrieval}
\cutsubsectiondown
% For quantative assessment of cross-domain disentanglement, we apply our method to cross-domain retrieval and zero-shot sketch based image retrieval task.
For quantitative evaluation of cross-domain disentanglement, we apply our method to the task of image retrieval. 
Given a query image, the objective is to find its nearest neighbors from the database images, where the query and images in a database are from different domains exhibiting some exclusive characteristics. 
% Similar to naive image retrieval, this task aims to find the nearest neighbors of the query image in a database, but the query and database images are in different domains thus exhibit exclusive characteristics (\emph{e.g.}, label map and photo).
The main challenge in this task is to learn image representation \emph{invariant} to domain-specific characteristics, such that the distance between the query and database image in a representation space is aligned with their semantic similarity.

% In this task, the model takes a query image, and finds its nearest neighbors in a database, where the query and database images belong to the different domains thus have different characteristics.
We address this task by exploiting the \emph{shared} representation learned by our method. 
Given a query image $x\in X$ and the one from a database $y\in Y$, we compute their similarity by (1) extracting the shared representations independently by the mean $\mu^s_x$ of $r^x(z^s|x)$ and the mean $\mu^s_y$ of $r^y(z^s|y)$ and (2) computing their distance by $d(\mu^s_x, \mu^s_y)$ with a distance metric $d$ (\emph{e.g.}, Euclidean distance, cosine distance, \emph{etc.}).
Then the retrieval is performed by extracting the $K$-nearest neighbors in the database.
%\sh{If there are some algorithms that require a pair to extract their semantic similarity, we can talk about our advantage in terms of computational efficiency (linear instead of combinatorial)}
% Based on those two simple procedures, top-k images from domain $Y$ closest to $x$ can be retrieved for every $x$ in $X$. 
% To see if any correlation exist between exclusive representations between two domains, we can also measure the distance using exclusive representations by replacing the first step with extracting domain-specific representation using the encoder by $z^x=\mu^x=q(z^x|x)$ and $z^y=\mu^y=q(z^y|y)$.
%\cutsubsectionup
\subsubsection{Cross-domain retrieval}
% \cutsubsectiondown
% Cross-domain retrieval \citep{gonzalez-garcia2018NeurIPS} supports bi-directional retrieval, where the given two different data domains can be in any side of query or database as long as they are opposite. 
\cutparagraphup
\paragraph{Datasets}
We tested our model with MNIST-CDCB \citep{gonzalez-garcia2018NeurIPS}, Facades \citep{10.1007/978-3-642-40602-7_39}, and Maps \citep{pix2pix2017} datasets.
In Facades \citep{10.1007/978-3-642-40602-7_39} dataset, each pair $(x,y)$ is made up of an image of semantic label map and photo of the same building. 
% To be specific, images in skeleton domain have color variation, while the ones in facade domain have various style of the building. 
We use 400 / 100 / 106 pairs of train/valid/test samples following \citep{10.1007/978-3-642-40602-7_39}.
In Maps \citep{pix2pix2017} dataset, each pair $(x,y)$ is composed of an image of map and a satellite image of the same area.
% To be specific, images in map domain have color variation, while the ones in map domain have various geological style of the area. 
We use 1096 / 1098 pairs of train/test samples following \citep{pix2pix2017}. 
% Each pair of images in those three datasets are geometrically aligned.

\begin{figure}
    \centering
    \hrule
    \vspace{0.02cm}
    \begin{subfigure}{0.49\textwidth}
        \captionsetup{justification=raggedright,singlelinecheck=false}
        \caption*{Query(S) \hspace*{0.15cm} GT(M) \hspace*{1.88cm} S $\rightarrow$ M}
        \centering
        \vspace{-0.2cm}
        \includegraphics*[width=0.18\textwidth, viewport=0 0 600 600]{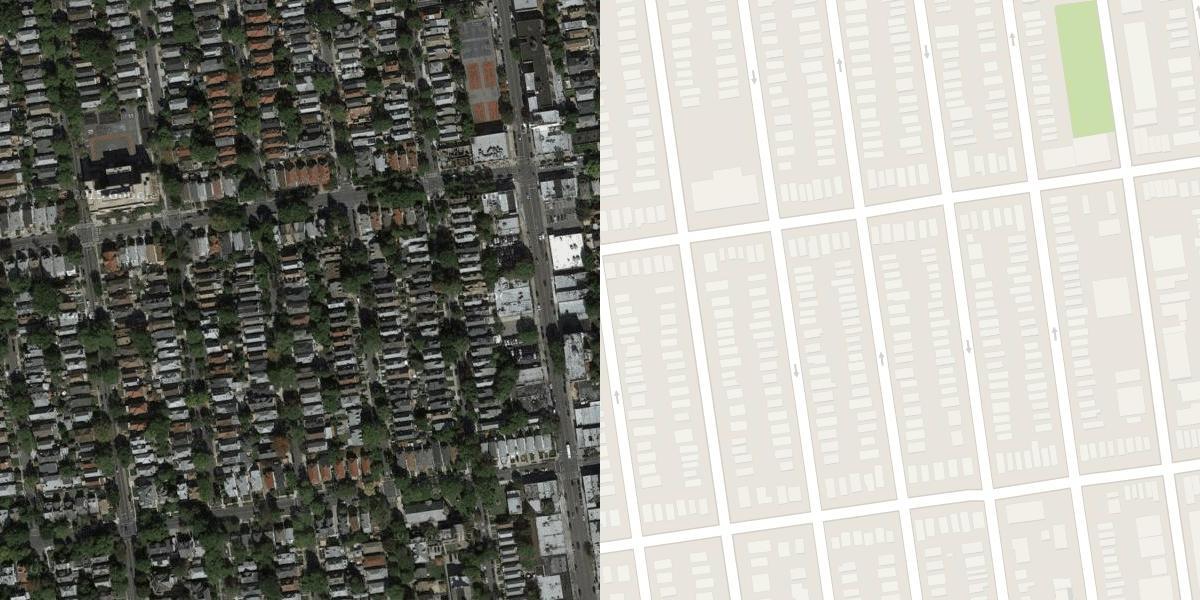}%
        \hspace*{0.02cm}
        \includegraphics*[width=0.18\textwidth, viewport=602 0 1200 600]{figs/Maps/388.jpg}
        \hfill
        \includegraphics*[width=0.18\textwidth, viewport=602 0 1200 600]{figs/Maps/388.jpg}
        \includegraphics*[width=0.18\textwidth, viewport=602 0 1200 600]{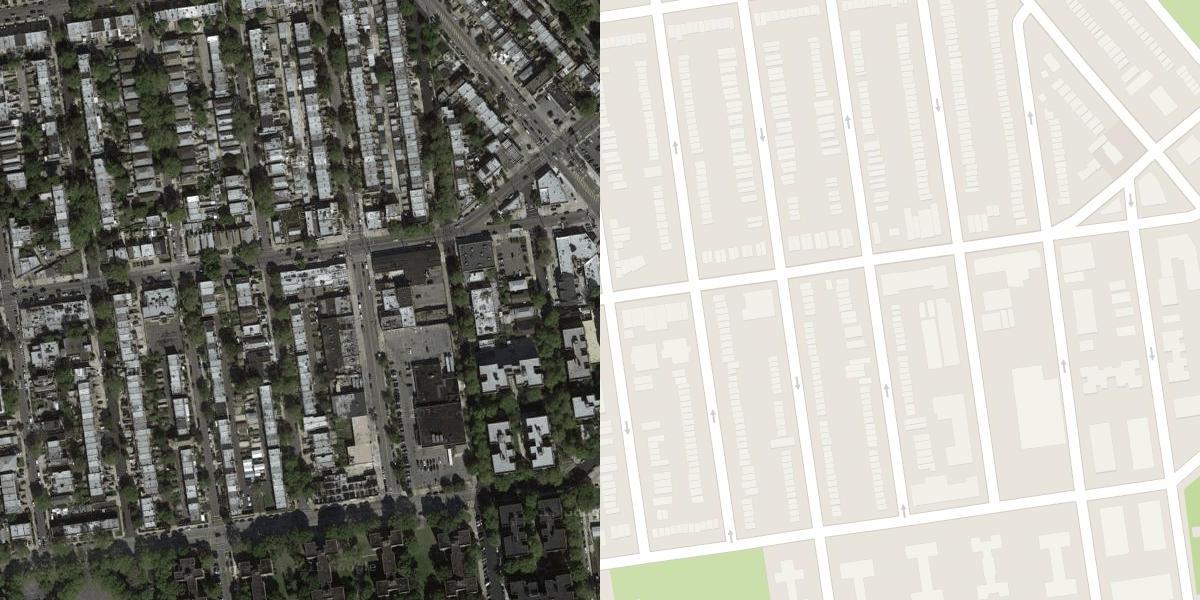}
        \includegraphics*[width=0.18\textwidth, viewport=602 0 1200 600]{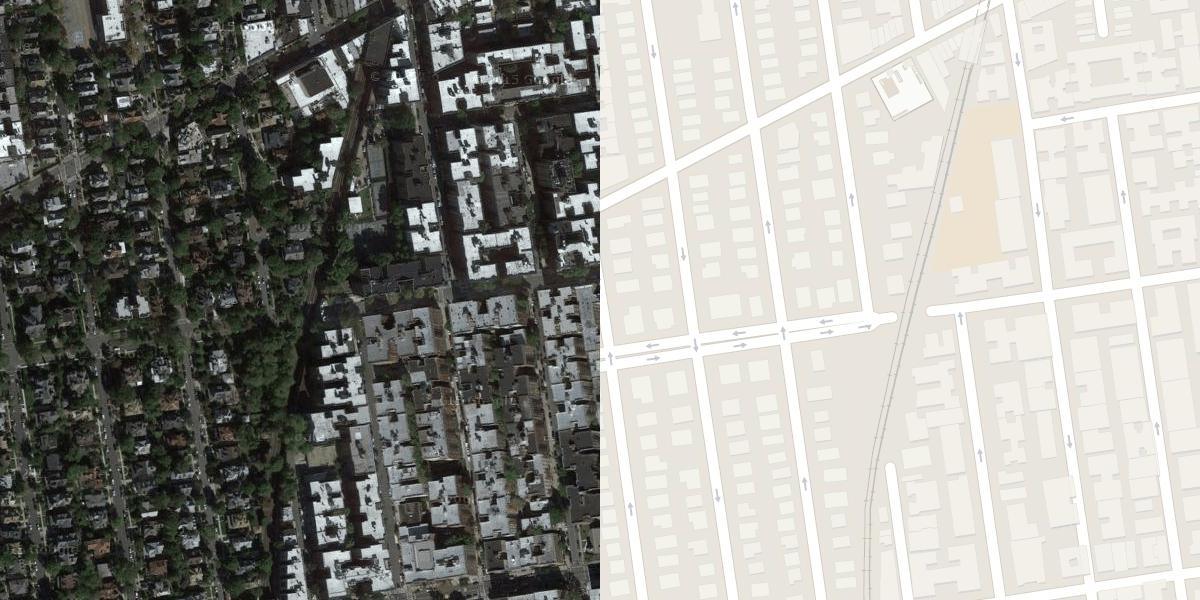}
        \\
        \includegraphics*[width=0.18\textwidth, viewport=0 0 600 600]{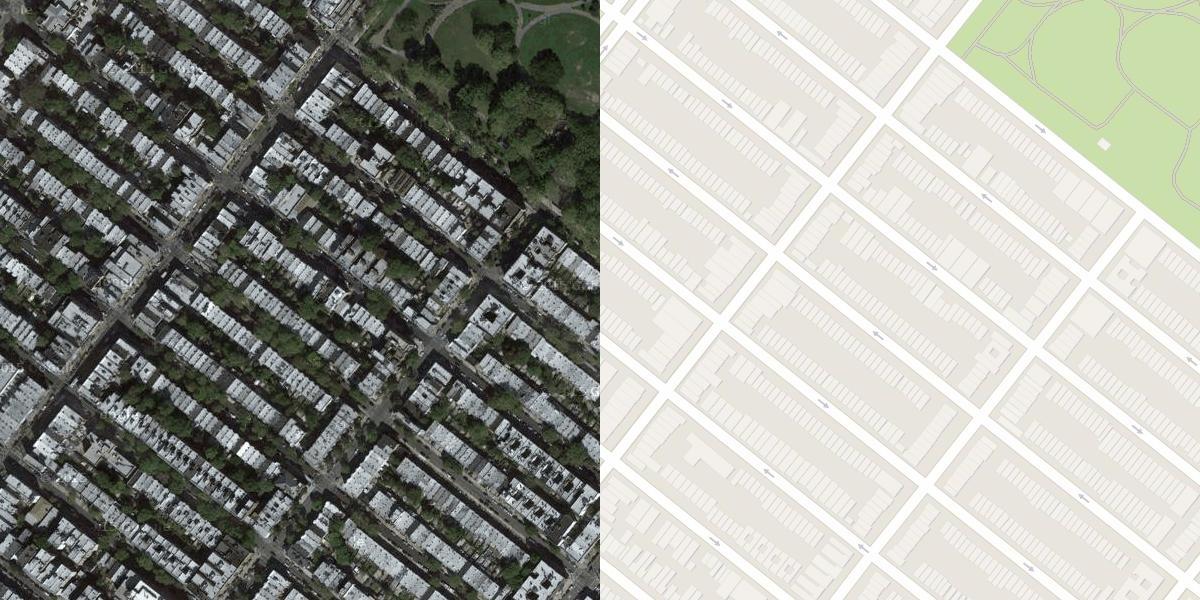}%
        \hspace*{0.02cm}
        \includegraphics*[width=0.18\textwidth, viewport=602 0 1200 600]{figs/Maps/391.jpg}
        \hfill
        \includegraphics*[width=0.18\textwidth, viewport=602 0 1200 600]{figs/Maps/391.jpg}
        \includegraphics*[width=0.18\textwidth, viewport=602 0 1200 600]{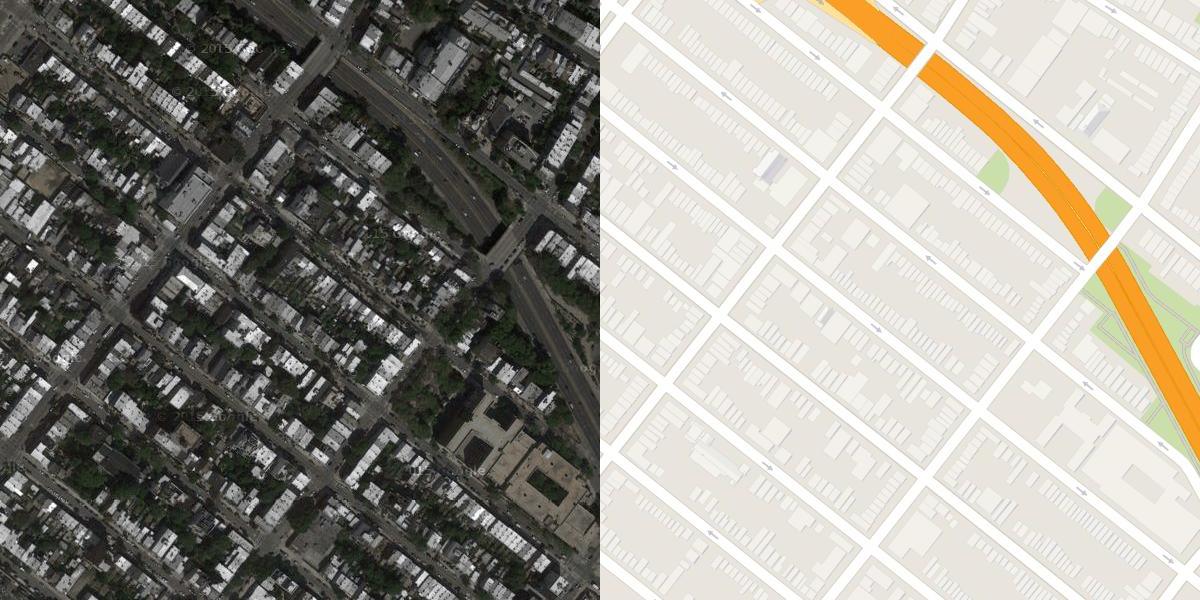}
        \includegraphics*[width=0.18\textwidth, viewport=602 0 1200 600]{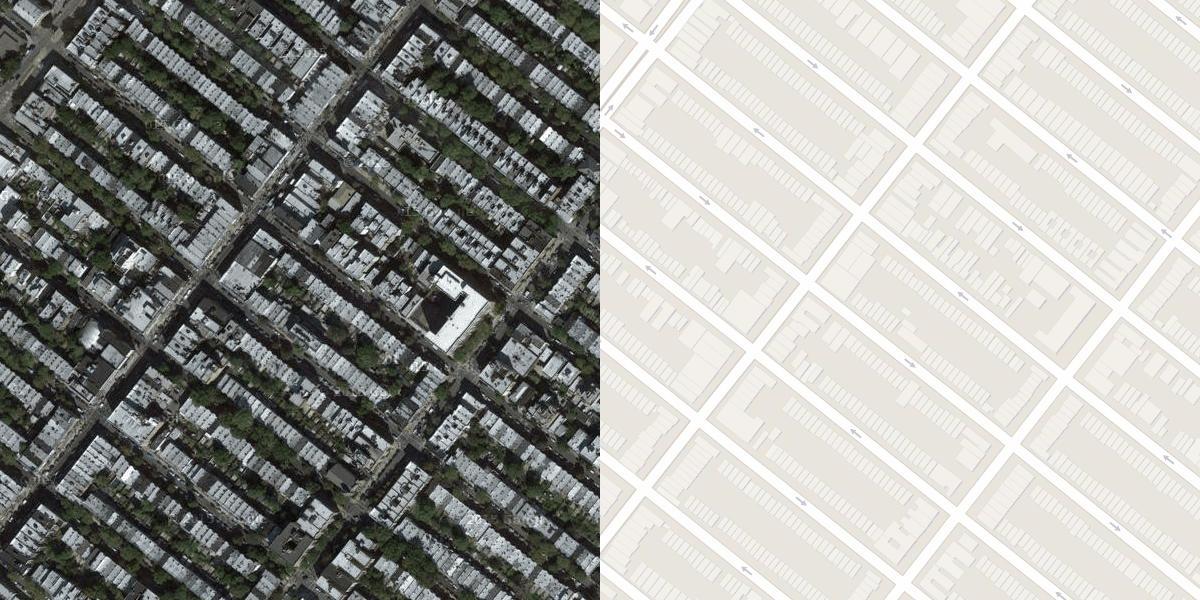}%
        % \label{subfig:FL}
    \end{subfigure}
    \vrule
    \hspace*{0.05cm}
    \begin{subfigure}{0.49\textwidth}
        \captionsetup{justification=raggedright,singlelinecheck=false}
        \centering
        \caption*{Query(M) \hspace*{0.15cm} GT(S) \hspace*{1.88cm} M $\rightarrow$ S}
        \vspace{-0.2cm}
        \includegraphics*[width=0.18\textwidth, viewport=602 0 1200 600]{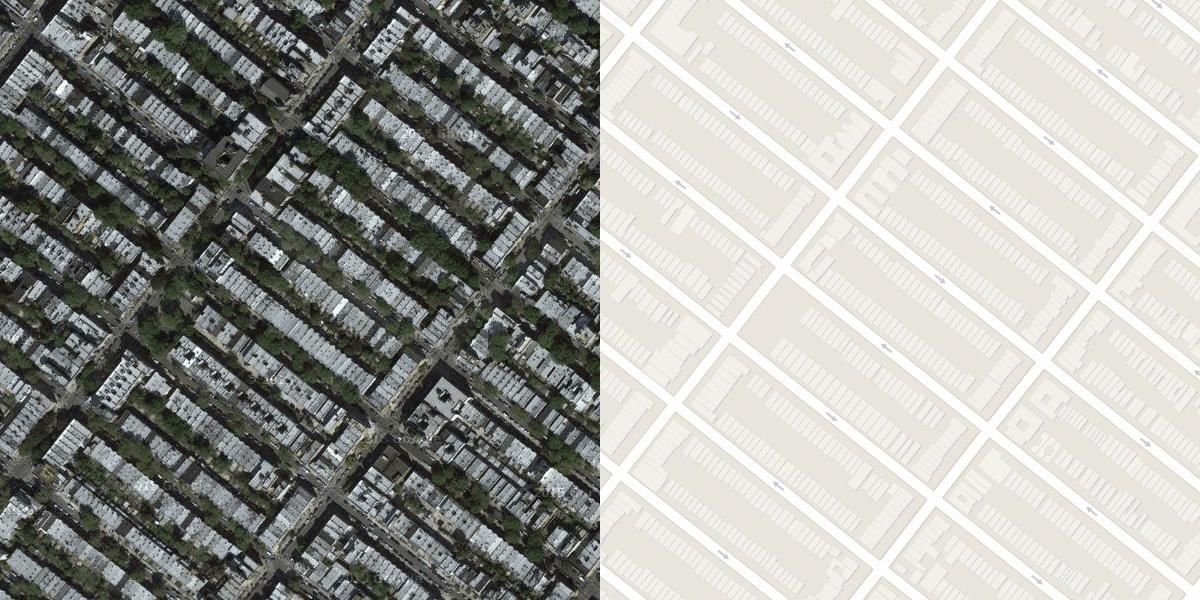}
        \includegraphics*[width=0.18\textwidth, viewport=0 0 600 600]{figs/Maps/389.jpg}%
        \hfill
        \includegraphics*[width=0.18\textwidth, viewport=0 0 600 600]{figs/Maps/389.jpg}
        \includegraphics*[width=0.18\textwidth, viewport=0 0 600 600]{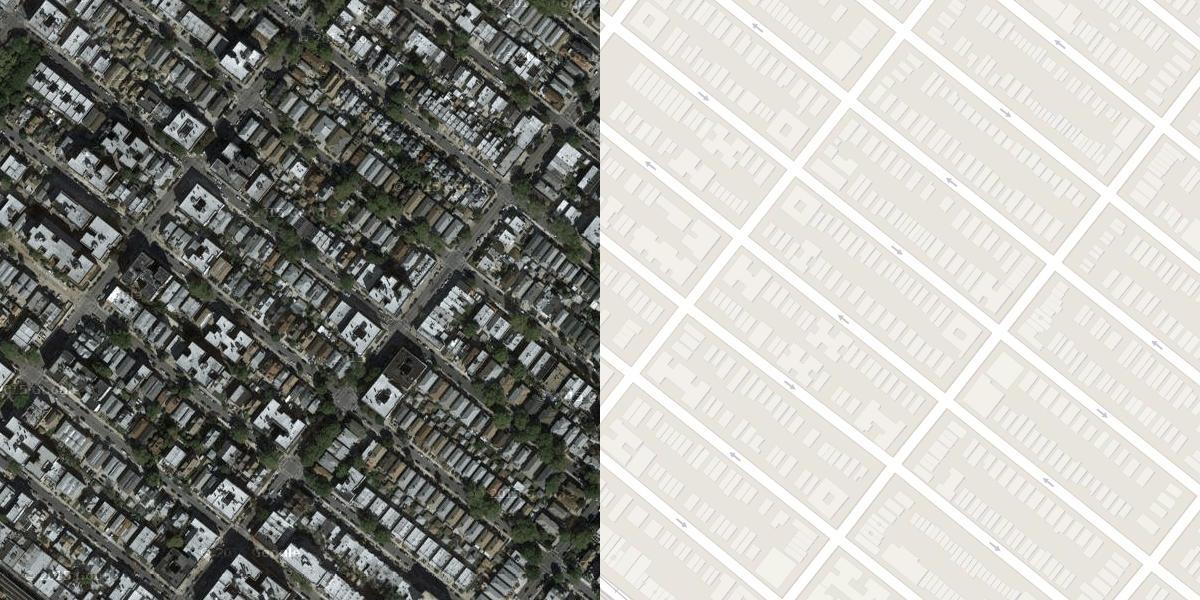}
        \includegraphics*[width=0.18\textwidth, viewport=0 0 600 600]{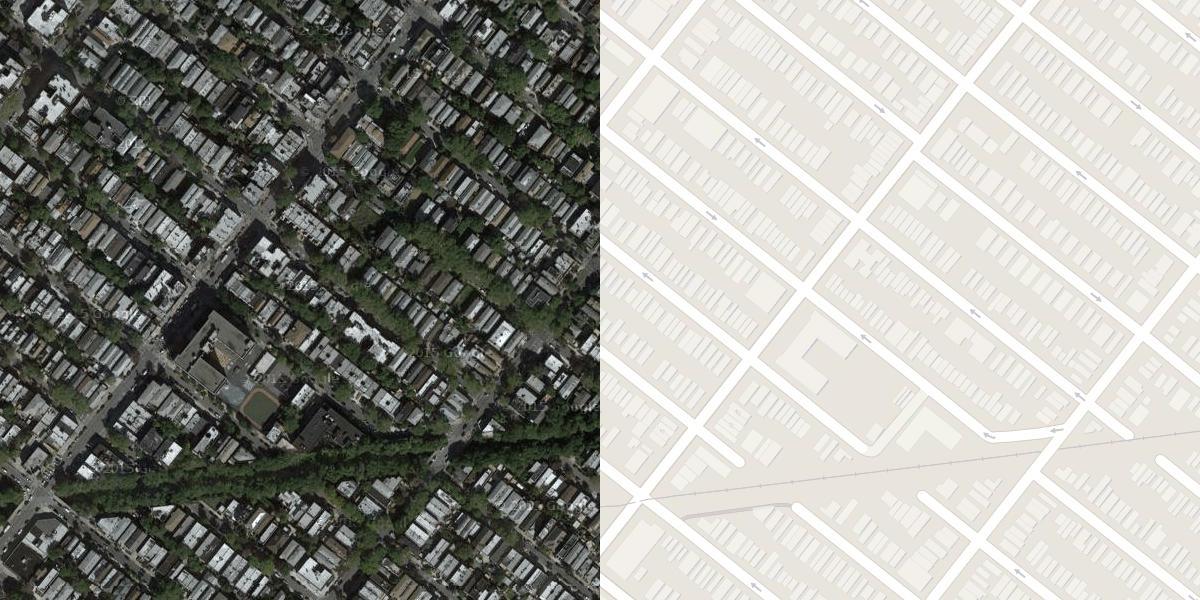}%
        \\
        \includegraphics*[width=0.18\textwidth, viewport=602 0 1200 600]{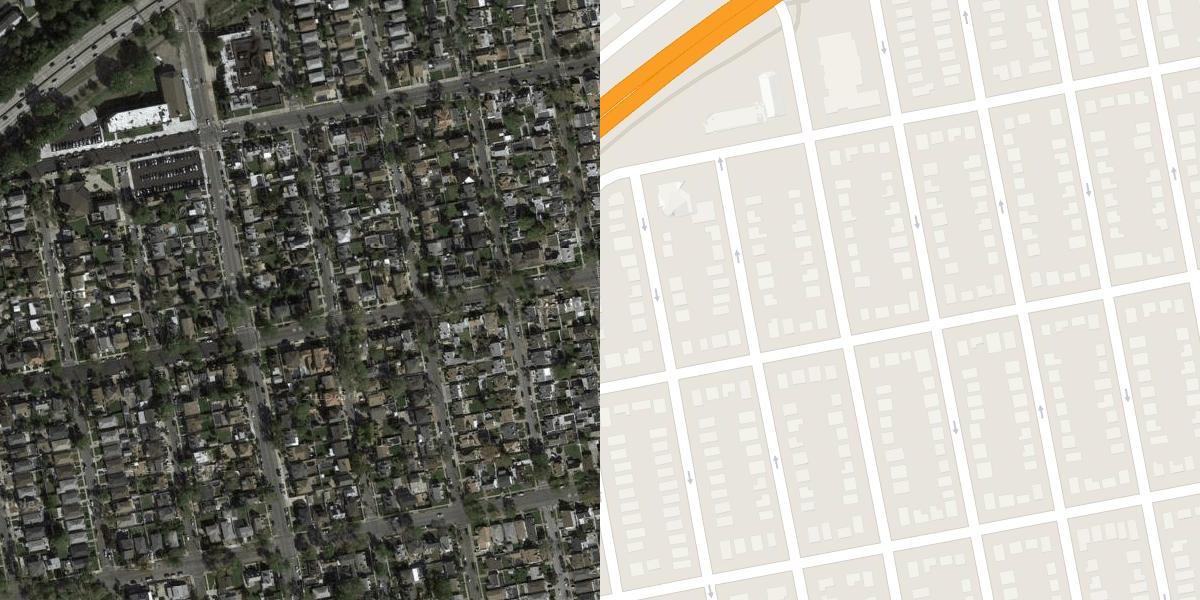}
        \includegraphics*[width=0.18\textwidth, viewport=0 0 600 600]{figs/Maps/390.jpg}%
        \hfill
        \includegraphics*[width=0.18\textwidth, viewport=0 0 600 600]{figs/Maps/390.jpg}
        \includegraphics*[width=0.18\textwidth, viewport=0 0 600 600]{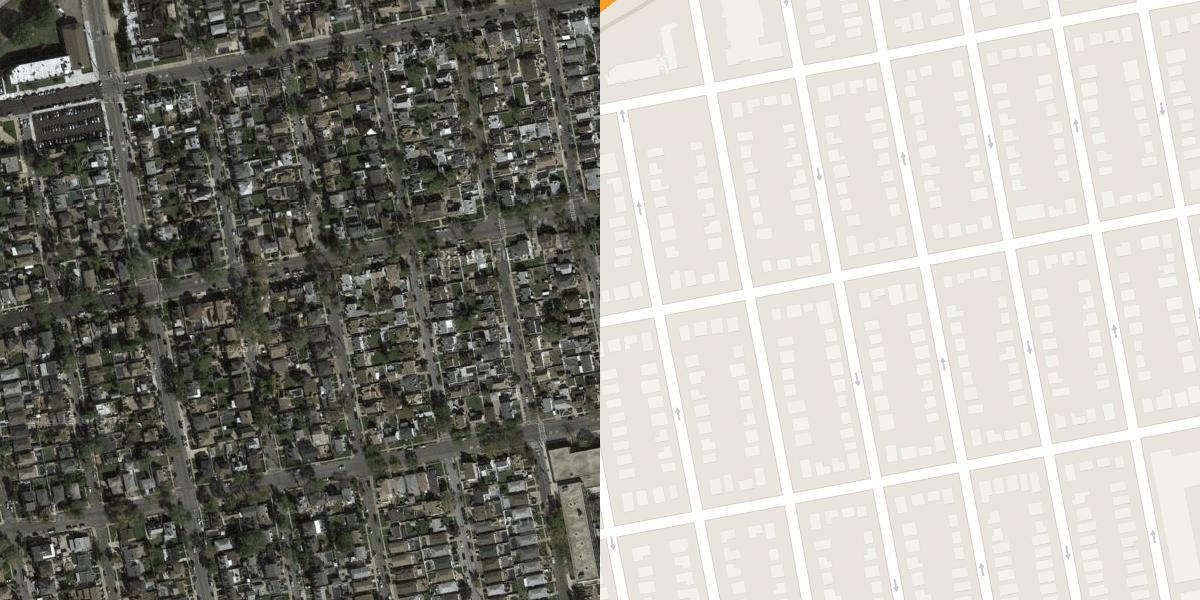}
        \includegraphics*[width=0.18\textwidth, viewport=0 0 600 600]{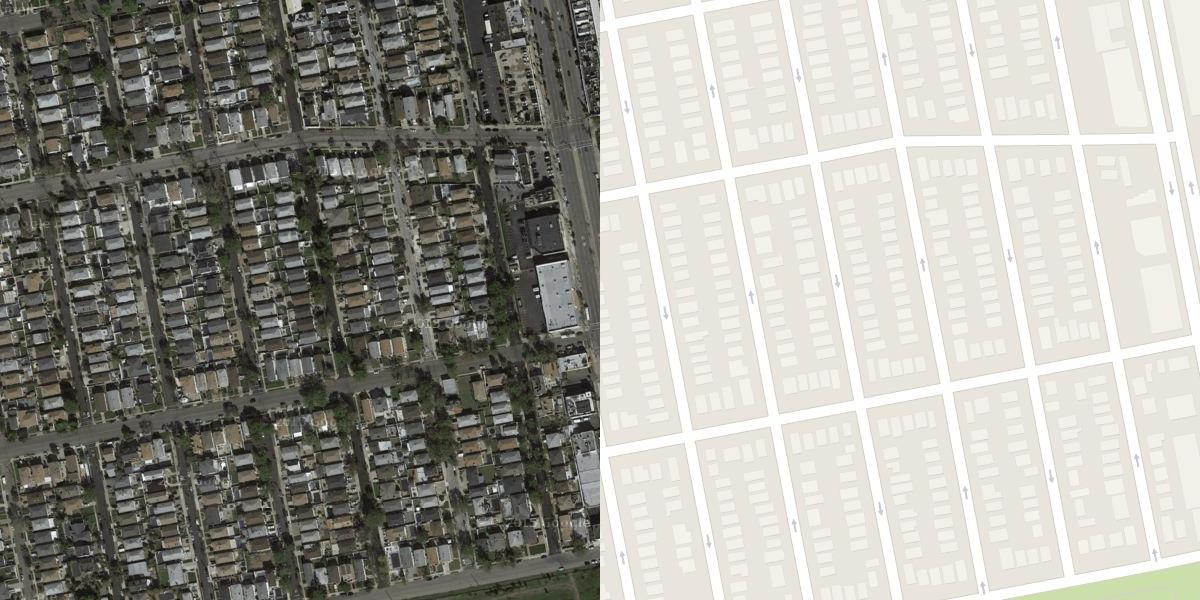}%
        % \label{subfig:LF}
    \end{subfigure}
    \vspace{0.1cm}
    \hrule
    \begin{subfigure}{0.49\textwidth}
        \captionsetup{justification=raggedright,singlelinecheck=false}
        \caption*{Query(F) \hspace*{0.16cm} GT(L) \hspace*{2cm} F $\rightarrow$ L}
        \vspace{-0.2cm}
        \centering
        \includegraphics*[width=0.18\textwidth, viewport=0 0 256 256]{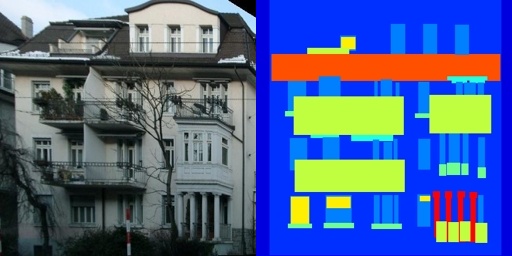}%
        \hspace*{0.02cm}
        \includegraphics*[width=0.18\textwidth, viewport=256 0 512 256]{figs/Facades/5.jpg}
        \hfill
        \includegraphics*[width=0.18\textwidth, viewport=256 0 512 256]{figs/Facades/5.jpg}
        \includegraphics*[width=0.18\textwidth, viewport=256 0 512 256]{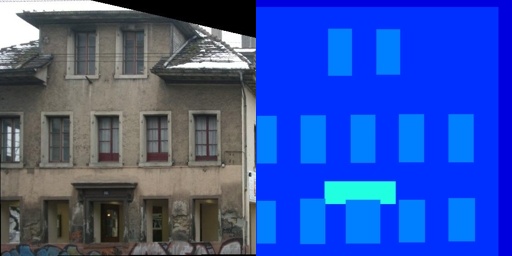}
        \includegraphics*[width=0.18\textwidth, viewport=256 0 512 256]{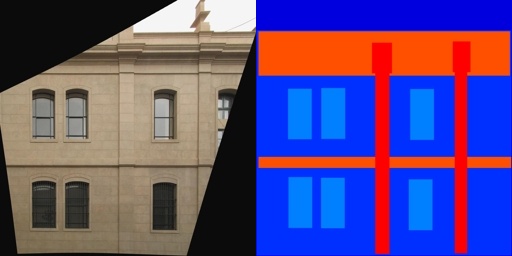}%
        \\
        \includegraphics*[width=0.18\textwidth, viewport=0 0 256 256]{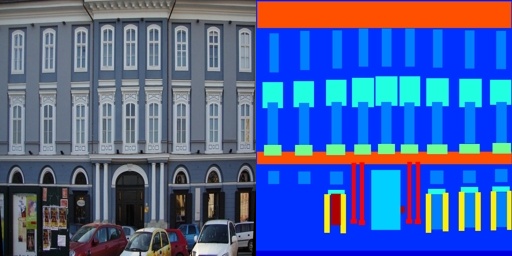}%
        \hspace*{0.02cm}
        \includegraphics*[width=0.18\textwidth, viewport=256 0 512 256]{figs/Facades/8.jpg}
        \hfill
        \includegraphics*[width=0.18\textwidth, viewport=256 0 512 256]{figs/Facades/8.jpg}
        \includegraphics*[width=0.18\textwidth, viewport=256 0 512 256]{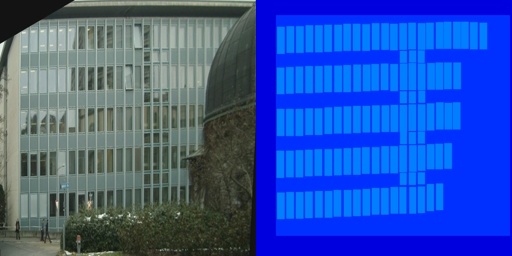}
        \includegraphics*[width=0.18\textwidth, viewport=256 0 512 256]{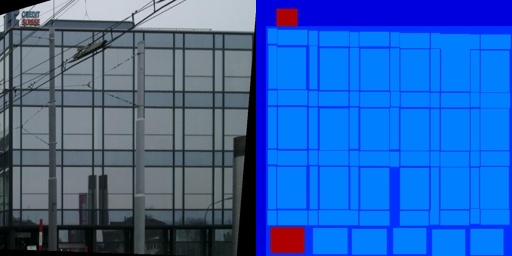}%
        % \label{subfig:LF}
    \end{subfigure}
    %\hfill
    \vrule
    \hspace*{0.03cm}
    % \hspace{10mm}
    \begin{subfigure}{0.49\textwidth}
        \captionsetup{justification=raggedright,singlelinecheck=false}
        \caption*{Query(L) \hspace*{0.16cm} GT(F) \hspace*{2cm} L $\rightarrow$ F}
        \vspace{-0.2cm}
        \centering
        \includegraphics*[width=0.18\textwidth, viewport=256 0 512 256]{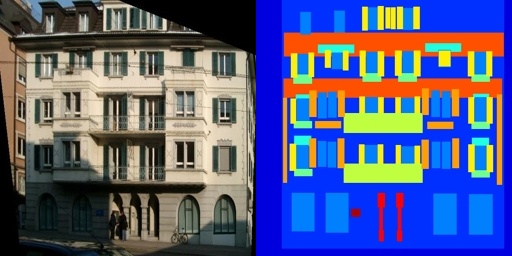}
        \includegraphics*[width=0.18\textwidth, viewport=0 0 256 256]{figs/Facades/3.jpg}%
        \hfill
        \includegraphics*[width=0.18\textwidth, viewport=0 0 256 256]{figs/Facades/3.jpg}
        \includegraphics*[width=0.18\textwidth, viewport=0 0 256 256]{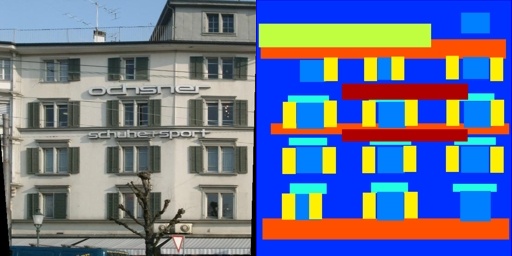}
        \includegraphics*[width=0.18\textwidth, viewport=0 0 256 256]{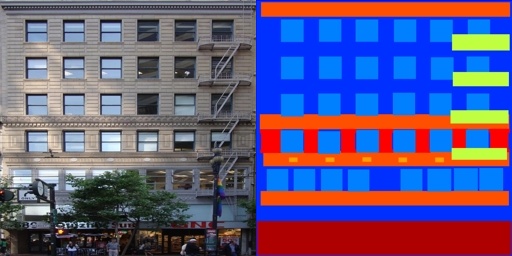}%
        \\
        \includegraphics*[width=0.18\textwidth, viewport=256 0 512 256]{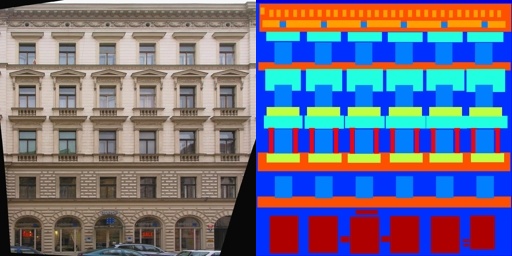}
        \includegraphics*[width=0.18\textwidth, viewport=0 0 256 256]{figs/Facades/10.jpg}%
        \hfill
        \includegraphics*[width=0.18\textwidth, viewport=0 0 256 256]{figs/Facades/10.jpg}
        \includegraphics*[width=0.18\textwidth, viewport=0 0 256 256]{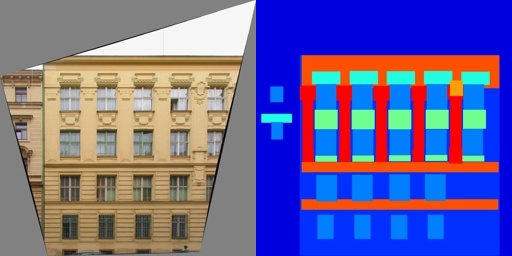}
        \includegraphics*[width=0.18\textwidth, viewport=0 0 256 256]{figs/Facades/45.jpg}%
        % \vspace{-0.2cm}
        % \label{subfig:FL}
    \end{subfigure}
    \vspace{0.1cm}
    \hrule
    \vspace{0.1cm}
    \caption{Qualitative examples of cross-domain retrieval (Top-1,2,3 from the left to the right) in Facades and Maps using IIAE. GT stands for ground truth.}
    \label{fig:SuccessCdMini}
    \vspace{-0.5cm}
\end{figure}

\cutparagraphup
\paragraph{Results}% \fbox{please fix this paragraph}
Following \citep{gonzalez-garcia2018NeurIPS}, we compute the nearest neighbor using the Euclidean distance and evaluate the performance by the Recall$@$1 metric.\footnote{In the MNIST-CDCB\citep{gonzalez-garcia2018NeurIPS} dataset, we only count the ground-truth pair of the query as a hit, whereas in \citep{gonzalez-garcia2018NeurIPS} any 
retrieved image containing the same digit as a hit, which is why the scores are lower
than originally reported.
% any retrieved image with the same digit identity is also counted in % \citep{gonzalez-garcia2018NeurIPS}. 
In the Facades \citep{10.1007/978-3-642-40602-7_39} dataset, we present the results on the \emph{test set}, while the results on the validation set is reported in \citep{gonzalez-garcia2018NeurIPS}. We also report the result on the validation set in the supplementary material \ref{appendix:additional_note}.}
We compare our method with two baselines, CdDN \citep{gonzalez-garcia2018NeurIPS} and DRIT \citep{lee2018diverse}, each of which is one of the most representative image to image translation models that encourage the cross-domain disentanglement in the representation with paired and unpaired dataset respectively.
In order to make a fair comparison, we re-trained DRIT using the paired data via minor modification to the author's code to take advantage of the paired data.
Table \ref{tab:MNIST-MAPS} summarizes the result of cross-domain retrieval with MNIST-CDCB, Maps, and Facades datasets.
Evaluation of DRIT on MNIST-CDCB was intractable because 
the dimensionality of the shared representation as well as the size of the test set
were too large.
% due to the high dimensionality of shared representation and the size of the test set.}
In MNIST-CDCB, both IIAE and CdDN both perform almost perfectly. 
This might be because the ground truth factors of variation inherent in the dataset is simple. 
However, in Maps \citep{pix2pix2017} and Facades \citep{10.1007/978-3-642-40602-7_39} datasets, we observe that IIAE outperforms all the baselines in any direction of the retrieval exhibiting well balanced performance in two directions.
On the other hand, CdDN shows relatively poor performance on satellite$\to$map in Maps and facade$\to$label in Facades, and DRIT shows the worst performance in Facades and Maps, implying that the learned latent representations of two data domains are not aligned well.
This shows that IIAE is more successful in capturing the complex factors of variation that are present in more realistic datasets such as Maps and Facades.
Figure \ref{fig:SuccessCdMini} presents the examples of top-3 images retrieved by IIAE in Maps (top two rows) and Facades (bottom two rows).
All of top-1 images in figure \ref{fig:SuccessCdMini} are the ground truth of the query.
Furthermore, it is remarkable that most of images retrieved as second or third closest ones also have geometrical structure similar to the query image.
Additional qualitative results of the retrieval can be found in supplementary material \ref{appendix:retrieval}.
\cutparagraphup
\paragraph{Ablation study}
We also conducted cross-domain retrieval
with \textit{domain-specific representations} as an ablation study.
The results are summarized in Table~\ref{tab:MNIST-MAPS} with parenthesized numbers.
We observe that the retrieval accuracy approaches near zero,
% \update{We observe that the retrieval accuracy is close to a random guess (100/N\%)},
which indicates that the learned domain-specific representations encode information only presented in each domain, as desired.
%\update{In contrast, the results from CdDN are noticeably high or low, suggesting that it was relatively unsuccessful in disentangling the representations.}
\begin{table}
    \centering
    \caption{Shared (exclusive) representation based retrieval on MNIST-CDCB \citep{gonzalez-garcia2018NeurIPS}, Maps \citep{pix2pix2017}, and Facades \citep{10.1007/978-3-642-40602-7_39} dataset. CD/CB stand for colored digit/background, S/M stand for satellite/map, and F/L stand for facade/label respectively.}
    \begin{tabular}{ccccccc}
        \toprule
        \textbf{Dataset} & \multicolumn{2}{c}{\textbf{MNIST-CDCB}} & \multicolumn{2}{c}{\textbf{Maps}} & \multicolumn{2}{c}{\textbf{Facades}} \\
        \cmidrule(lr){2-3}
        \cmidrule(lr){4-5}
        \cmidrule(lr){6-7}
        \textbf{Models} & CD $\rightarrow$ CB & CB $\rightarrow$ CD & S $\rightarrow$ M & M $\rightarrow$ S & F $\rightarrow$ L & L $\rightarrow$ F \\
        \midrule
        DRIT \citep{lee2018diverse}             & -  & -  & 33.8 (0.09)  & 37.3 (0.09) &  31.1 (0.94)  &  44.3 (0.94) \\
        CdDN \citep{gonzalez-garcia2018NeurIPS} & 99.6 (0.0)   & 99.6 (0.0)   & 91.4 (0.18)  & 96.9 (0.09) & 84.9 (0.94)  & 89.6 (0.0)\\
        \midrule
        IIAE                                    & \textbf{99.7} (0.01)  & \textbf{99.7} (0.01)  & \textbf{96.6} (0.09)  & \textbf{97.3} (0.0) &  \textbf{96.2} (0.94)  &  \textbf{99.1} (0.94) \\
        \bottomrule             
    \end{tabular}
    \label{tab:MNIST-MAPS}
    \vspace{-0.1cm}
\end{table}

% \cutsubsectionup
\subsubsection{Zero-shot sketch based image retrieval (ZS-SBIR)}
\label{subsubsec:ZS-SBIR}
\begin{figure}
    \centering
    \includegraphics*[width=0.078\linewidth, height=0.078\linewidth]{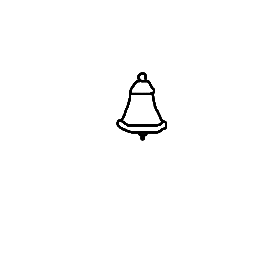}
    \includegraphics*[width=0.078\linewidth, height=0.078\linewidth]{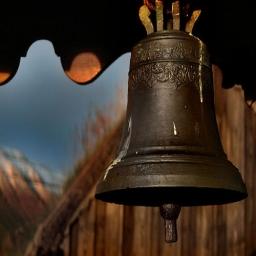}
    \includegraphics*[width=0.078\linewidth, height=0.078\linewidth]{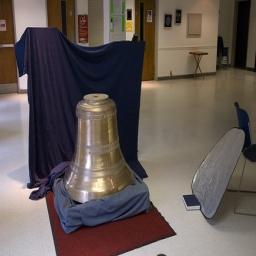}
    \includegraphics*[width=0.078\linewidth, height=0.078\linewidth]{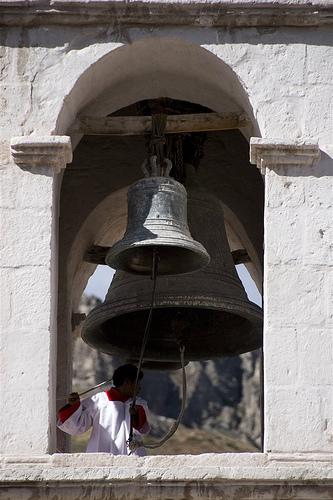}
    \includegraphics*[width=0.078\linewidth, height=0.078\linewidth]{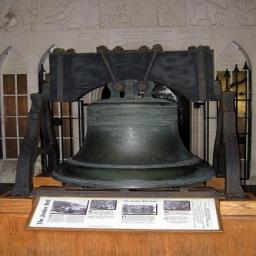}
    \includegraphics*[width=0.078\linewidth, height=0.078\linewidth]{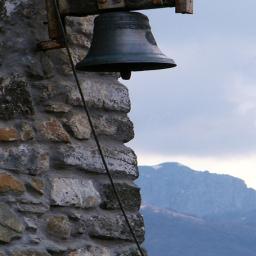}
    \hfill
    \includegraphics*[width=0.078\linewidth, height=0.078\linewidth]{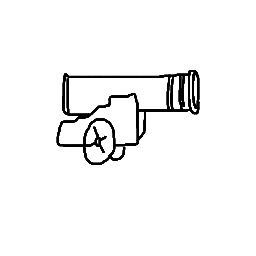}
    \includegraphics*[width=0.078\linewidth, height=0.078\linewidth]{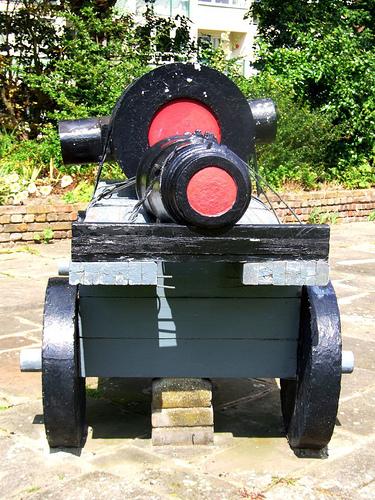}
    \includegraphics*[width=0.078\linewidth, height=0.078\linewidth]{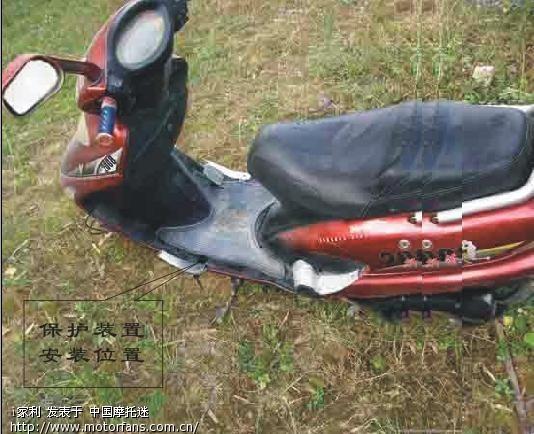}
    \includegraphics*[width=0.078\linewidth, height=0.078\linewidth]{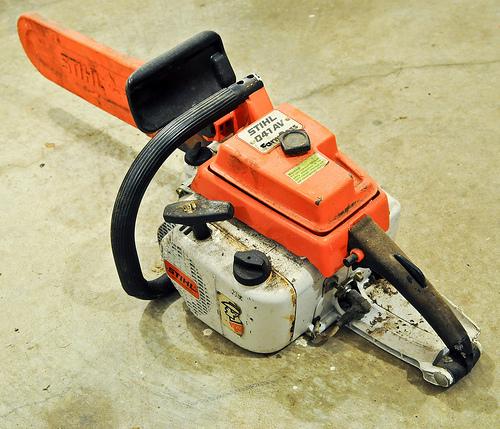}
    \includegraphics*[width=0.078\linewidth, height=0.078\linewidth]{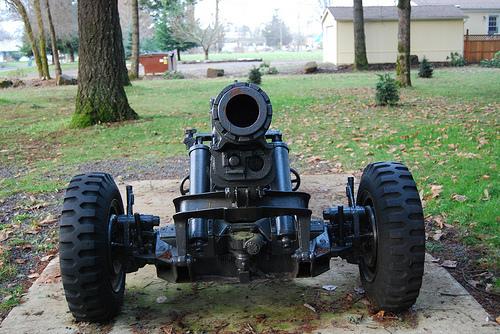}
    \includegraphics*[width=0.078\linewidth, height=0.078\linewidth]{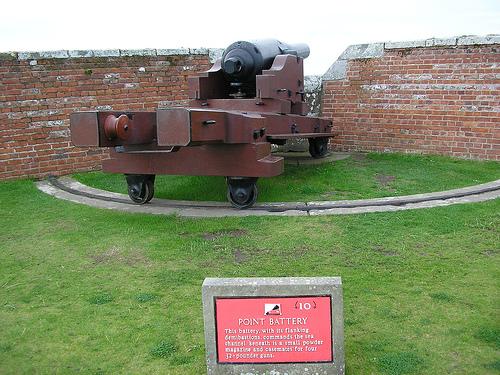}\\
    \vspace{-0.3em}
    \hspace*{0.078\linewidth}
    \includegraphics*[width=0.078\linewidth, height=\fontcharht\font`\B]{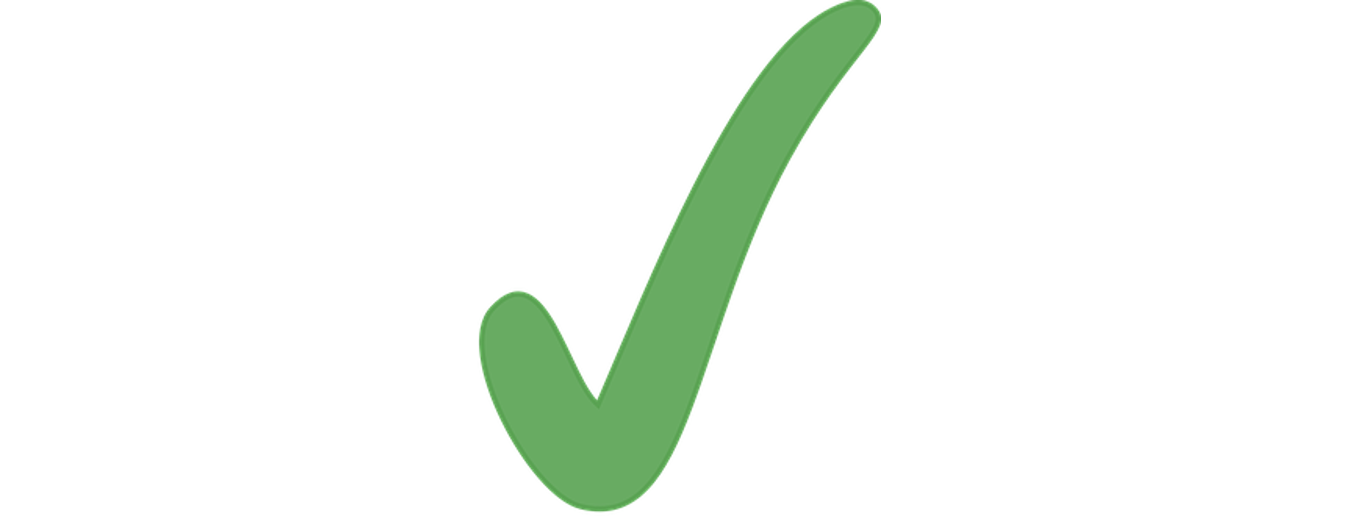}
    \includegraphics*[width=0.078\linewidth, height=\fontcharht\font`\B]{figs/gc}
    \includegraphics*[width=0.078\linewidth, height=\fontcharht\font`\B]{figs/gc}
    \includegraphics*[width=0.078\linewidth, height=\fontcharht\font`\B]{figs/gc}
    \includegraphics*[width=0.078\linewidth, height=\fontcharht\font`\B]{figs/gc}
    \hfill
    \hspace*{0.078\linewidth}
    \includegraphics*[width=0.078\linewidth, height=\fontcharht\font`\B]{figs/gc}
    \includegraphics*[width=0.078\linewidth, height=\fontcharht\font`\B]{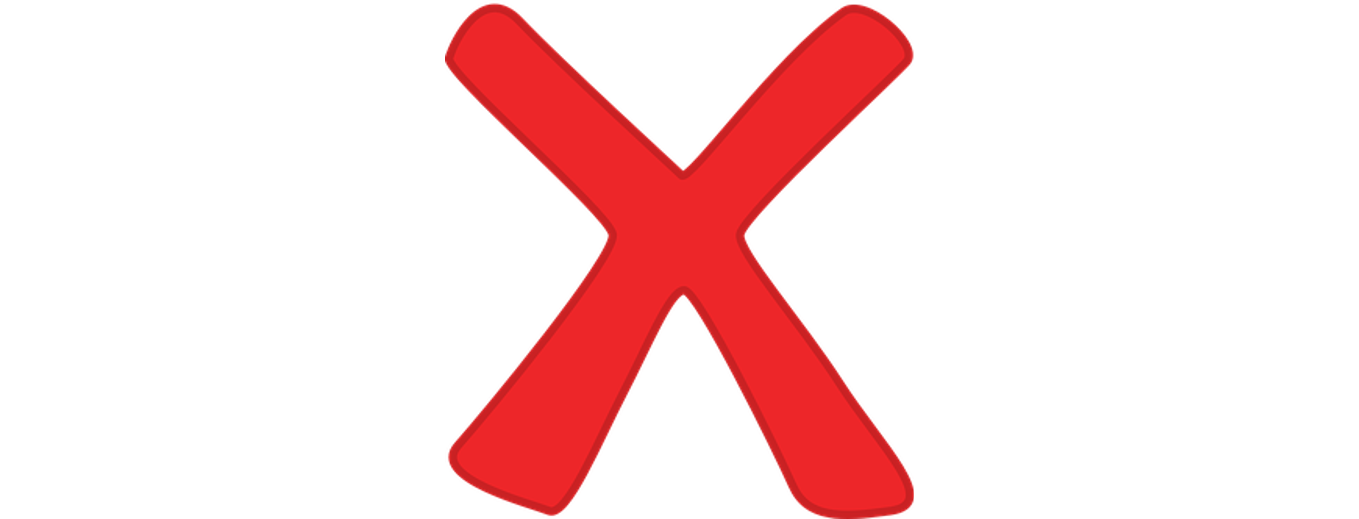}
    \includegraphics*[width=0.078\linewidth, height=\fontcharht\font`\B]{figs/rc}
    \includegraphics*[width=0.078\linewidth, height=\fontcharht\font`\B]{figs/gc}
    \includegraphics*[width=0.078\linewidth, height=\fontcharht\font`\B]{figs/gc}\\
    \vspace{0.3em}
    \includegraphics*[width=0.078\linewidth, height=0.078\linewidth]{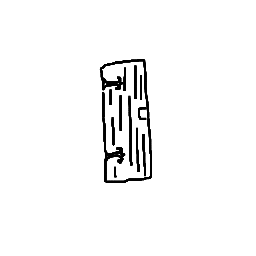}
    \includegraphics*[width=0.078\linewidth, height=0.078\linewidth]{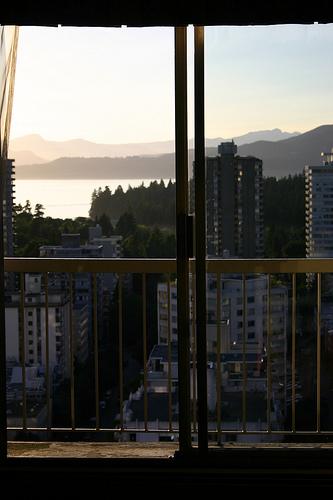}
    \includegraphics*[width=0.078\linewidth, height=0.078\linewidth]{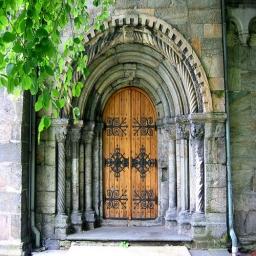}
    \includegraphics*[width=0.078\linewidth, height=0.078\linewidth]{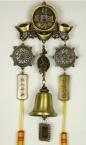}
    \includegraphics*[width=0.078\linewidth, height=0.078\linewidth]{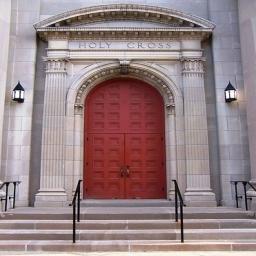}
    \includegraphics*[width=0.078\linewidth, height=0.078\linewidth]{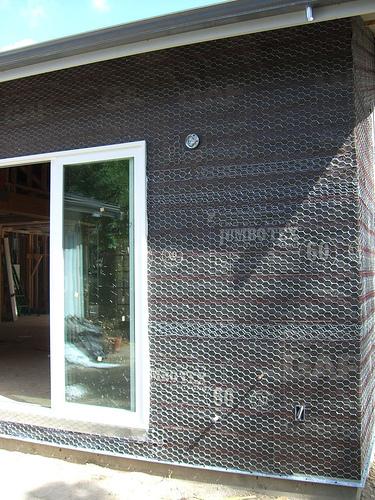}
    \hfill
    \includegraphics*[width=0.078\linewidth, height=0.078\linewidth]{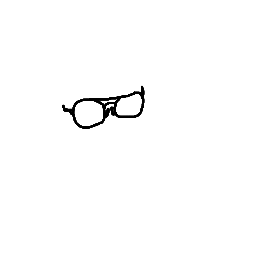}
    \includegraphics*[width=0.078\linewidth, height=0.078\linewidth]{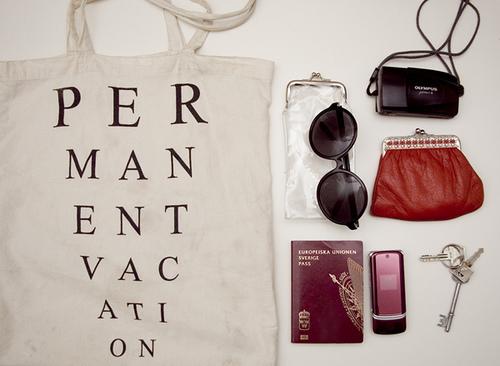}
    \includegraphics*[width=0.078\linewidth, height=0.078\linewidth]{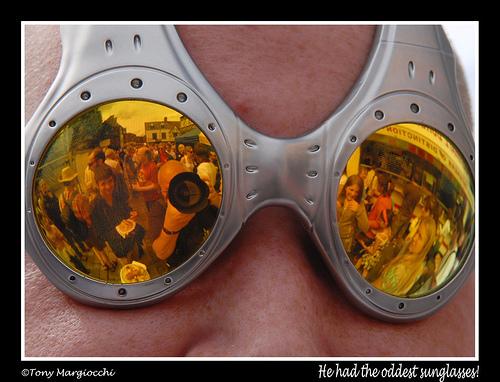}
    \includegraphics*[width=0.078\linewidth, height=0.078\linewidth]{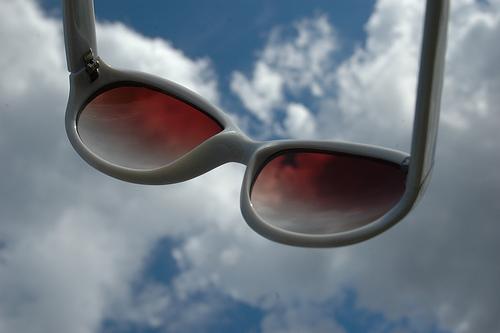}
    \includegraphics*[width=0.078\linewidth, height=0.078\linewidth]{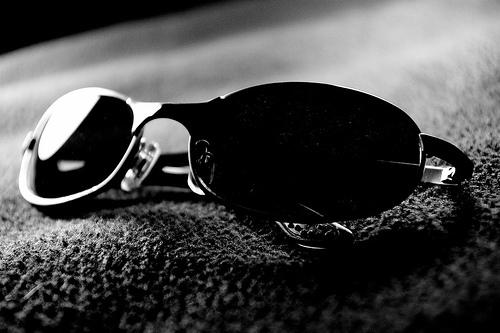}
    \includegraphics*[width=0.078\linewidth, height=0.078\linewidth]{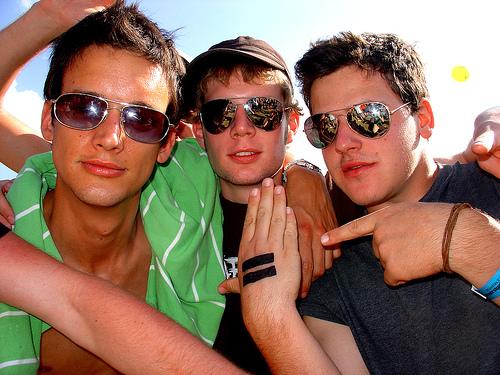}\\
    \vspace{-0.3em}
    \hspace*{0.078\linewidth}
    \includegraphics*[width=0.078\linewidth, height=\fontcharht\font`\B]{figs/gc}
    \includegraphics*[width=0.078\linewidth, height=\fontcharht\font`\B]{figs/gc}
    \includegraphics*[width=0.078\linewidth, height=\fontcharht\font`\B]{figs/rc}
    \includegraphics*[width=0.078\linewidth, height=\fontcharht\font`\B]{figs/gc}
    \includegraphics*[width=0.078\linewidth, height=\fontcharht\font`\B]{figs/gc}
    \hfill
    \hspace*{0.078\linewidth}
    \includegraphics*[width=0.078\linewidth, height=\fontcharht\font`\B]{figs/gc}
    \includegraphics*[width=0.078\linewidth, height=\fontcharht\font`\B]{figs/gc}
    \includegraphics*[width=0.078\linewidth, height=\fontcharht\font`\B]{figs/gc}
    \includegraphics*[width=0.078\linewidth, height=\fontcharht\font`\B]{figs/gc}
    \includegraphics*[width=0.078\linewidth, height=\fontcharht\font`\B]{figs/gc}\\
    %\vspace{0.3em}
    \caption{Top-5 ZS-SBIR samples from IIAE on the Sketchy Extended dataset. Sketches in the first and seventh columns are queries and rest are retrieved candidates (Top-1 to 5 from the left to the right). Green checkmark indicates correct retrieval, whereas red crossmark indicates wrong retrieval.}
    \label{fig:ZS-SBIR-MINI}
    \vspace{-0.3cm}
\end{figure}

\cutsubsectiondown
\paragraph{Dataset}
\sh{there is no explanations that this is the zero shot learning setting...}
ZS-SBIR \citep{Yelamarthi_2018_ECCV} is an extension of sketch based image retrieval task where none of the classes in the test set is exposed when training a retrieval model. We evaluate our model on Sketchy (Extended) \citep{sangkloy2016sketchy, liu2017deep}, one of the most widely used datasets of sketch and photo images in sketch-based image retrieval (SBIR) task.
% The original Sketchy dataset \citep{sangkloy2016sketchy} has 75,471 sketches and 12,500 photos of 125 classes which are paired up according to the geometrical correspondence within in every class.
% Sketchy is extended by \citep{liu2017deep} with extra 60,502 photos that do not have any geometrical correspondence to sketch, which results in 73,002 images.
% We used the extended one, which we call Sketchy Extended, while ignoring all the aligned pairs.
% Instead, we randomly sampled one sketch and one photo per category to pair up one training sample. The factors of variation shared across two domains is the class of the object, while the exclusive ones are scale, translation, orientation, and style in both domains.
We employ the extended version of Sketcy dataset (Sketcy Extended) \citep{liu2017deep}, which is composed of \sh{add number of images and class} \emph{un-aligned} images of 73,002 photos and 75,479 sketches distributed in 125 different classes.
To learn our model without ground-truth pairs, we randomly sample one sketch and one photo per category to pair up one training sample. The factors of variation shared across two domains are the class of the object, while the exclusive ones are scale, translation, orientation, and style in both domains.
We used train / test splits (100/25 categories) same as \citep{Dutta2019SEMPCYC, linlearning} and extracted features of images from VGG16 % \citep{DBLP:journals/corr/SimonyanZ14a} pretrained with ImageNet \citep{deng2009imagenet} dataset 
and finetuned with the train set of Sketchy Extended. Those extracted features are used as input to IIAE.
\cutparagraphup
\paragraph{Results}
We conducted the retrieval using cosine similarities, as used by \citep{Yelamarthi_2018_ECCV}, between shared representations extracted from IIAE.
We compare IIAE with various baselines, SAE \citep{kodirov2017semantic}, FRWGAN \citep{felix2018multi}, ZSIH \citep{'shen2018cvpr'}, CAAE \citep{Yelamarthi_2018_ECCV}, SEM-PCYC \citep{Dutta2019SEMPCYC}, and LCALE \citep{linlearning}, which are designed for ZS-SBIR or general zero shot learning.
Following the previous works \citep{Dutta2019SEMPCYC, 'shen2018cvpr'}, we chose mean average precision (mAP) and Precision$@$100 (P$@$100) as evaluation metric.
Table \ref{tab:ZS-SBIR} summarizes the result.
It shows that IIAE outperforms all competitive methods, although some of them are specialized to this task and exploit side information such as attribute information of image, word embedding\citep{mikolov2013distributed}, or WordNet \citep{miller1995wordnet}.
% class name or attribute information on the given image or sketch.
The result implies that IIAE successfully learns to associate semantic structure of sketches and images while being generalized well to unseen classes, which can be explained by two different information constraints on the shared representation;
Eq.~\eqref{eqn:final3} enforces $Z^{S}$ to discard domain specific information while Eq.~\eqref{eq:ELBO_VIB} encourages $Z^{S}$ to be a minimal sufficient statistic so that it generalizes well to unseen classes. 
Note that we can control the balance between being invariant and being compressive with $\lambda$.
We also evaluated the effect of terms in the IIAE objective as an ablation study in the supplementary material \ref{appendix:ablation}.
Figure \ref{fig:ZS-SBIR-MINI} shows the qualitative result of ZS-SBIR.
It is notable that even the incorrectly retrieved images in figure \ref{fig:ZS-SBIR-MINI} have visual or semantic correspondence to their query images. For instance, given a sketch of cannon as a query, a motorcycle and a saw are wrongly retrieved by IIAE, but the motorcycle is semantically relevant to the cannon due to its wheels whereas the saw is visually close to the motorcycle.
Similarly, an image of bells is falsely retrieved by a sketch of door due to their visual similarity.
Additional visualization of the ZS-SBIR results is in the supplementary material 
\ref{appendix:ZS-SBIR}.

\begin{table}
    \centering
    \caption{Evaluation on the Sketchy Extended dataset \citep{sangkloy2016sketchy, liu2017deep}. WordEmb stands for word embedding.}
    \label{tab:ZS-SBIR}
    \begin{tabular}{ccccccc}
        \toprule
        & Feature & \multicolumn{2}{c}{\textbf{Evaluation metric}} & \multicolumn{3}{c}{\textbf{External knowledge}} \\
        \cmidrule(lr){3-4}
        \cmidrule(lr){5-7}
        \textbf{Models} & Dimension  & mAP & P$@$100 & Attribute & WordEmb. & WordNet \citep{miller1995wordnet}\\
        \midrule
        SAE \citep{kodirov2017semantic}     & 300   & 0.216         & 0.293         & \cmark    & \cmark    & - \\
        FRWGAN \citep{felix2018multi}       & 512   & 0.127         & 0.169         & \cmark    & -    & - \\
        ZSIH \citep{'shen2018cvpr'}         & 64    & 0.258         & 0.342         & -    & \cmark    & - \\
        CAAE \citep{Yelamarthi_2018_ECCV}   & 4096  & 0.196         & 0.284         & -    & -    & - \\
        SEM-PCYC \citep{Dutta2019SEMPCYC}   & 64    & 0.349         & 0.463         & -    & \cmark    & \cmark \\
        LCALE \citep{linlearning}           & 64    & 0.476         & 0.583         & -    & \cmark    & - \\
        \midrule
        IIAE                                & 64    & \textbf{0.573}& \textbf{0.659}& -    & -    & - \\
        \bottomrule             
    \end{tabular}
    %\vspace{-0.2cm}
\end{table}

% \cutparagraphup

% \cutsectionup
\section{Conclusion}
\cutsectiondown
\label{sec:conclusion}
In this paper, we investigate an approach for cross-domain disentanglement.
The proposed approach, coined Interaction Information Auto-Encoder, extends the VAE with a novel regularization inspired by information theory, which are principled, interpretable, and nicely integrated into ELBO objective to encourage disentanglement of domain-specific and shared representations.
The effectiveness of the proposed method is demonstrated on multiple applications, such as image-to-image translation and image retrieval.
% Compared to the existing approaches, the proposed approach is built on

% \bibliographystyle{ieee}
% \bibliographystyle{plainnat}

\section*{Broader Impact}

Our method provides an information theoretic perspective on  representation
learning, and is likely to accelerate research in areas that involve datasets with two data domains with some common factors of variation. One of such areas is image to image translation we tackled in this paper. Beyond the image translation task, our method could be potentially 
applied to NLP tasks, such as language translation or text summarization where the source and the target data domains share semantics while they also have domain specific factors of variation in syntax. 
% (language dynamics for language translation and writing style for text summarization). 
Leveraging IIAE, one could transform a sample from one domain to the other and measure the semantic similarity between languages from two different domains. However, one may exploit disentangled representations for 
wrongful purposes. For example, our approach could be adopted for
Deepfake to generate more diverse fake images. 
% finer control over the representations learned by our method can be abused to generate illegal images similar to Deepfake.
Lastly, we do not see any serious consequences of system failure.

% \section*{Acknowledgements}
% This work was supported by the National Research Foundation (NRF) of Korea (NRF-
% 2019R1A2C1087634 and NRF-2019M3F2A1072238), the Ministry of Science and Information
% communication Technology (MSIT) of Korea (IITP No. 2020-0-00940, IITP No. 2019-0-00075, IITP
% No. 2017-0-01779, IITP No. 2020-0-00153, and IITP No. 2016-0-00464), the ETRI (Contract No. 20ZS1100), and Samsung Electronics.

\begin{ack}
This work was supported by the National Research Foundation (NRF) of Korea (NRF-
2019R1A2C1087634 and NRF-2019M3F2A1072238), the Ministry of Science and Information
communication Technology (MSIT) of Korea (IITP No. 2020-0-00940, IITP No. 2019-0-00075, IITP
No. 2017-0-01779, IITP No. 2020-0-00153, and IITP No. 2016-0-00464), the ETRI (Contract No. 20ZS1100), and Samsung Electronics.
\end{ack}

\bibliographystyle{abbrvnat}
\bibliography{bibtex}
\clearpage
% \title{Supplementary Materials: Learning Hierarchical Semantic Image Manipulation through Structured Representations}
% \maketitle
% !TEX root = neurips_2020.tex
\section*{Appendices}
\appendix

\section{Proofs}
\subsection{Evidence Lower Bound on $p_{\theta}\left(x,y\right)$}
\label{appendix:ELBO}

\begin{align*}
\log p(x, y)
&= \log \int p(x | z^{x},z^{s})p(y | z^{y},z^{s}) \hspace*{0.1cm} p(z^{x})p(z^{s})p(z^{y}) \hspace{0.3cm} d z^{x} d z^{s} d z^{y}\\
&= \log \int \frac{p(x | z^{x},z^{s})p(y | z^{y},z^{s}) \hspace*{0.1cm} p(z^{x})p(z^{s})p(z^{y})}{q(z^{x} | x) q(z^{s} | x, y) q(z^{y} | y)} \hspace{0.1cm}q(z^{x} | x) q(z^{s} | x, y) q(z^{y} | y) \hspace{0.3cm} d z^{x} d z^{s} d z^{y}\\
&= \log \mathbb{E}_{q(z^{x} | x) q(z^{s} | x, y) q(z^{y} | y)}\left[\frac{p(x | z^{x},z^{s})p(y | z^{y},z^{s}) \hspace*{0.1cm} p(z^{x})p(z^{s})p(z^{y})}{q(z^{x} | x) q(z^{s} | x, y) q(z^{y} | y)}\right] \\
&\geq \mathbb{E}_{q(z^{x} | x) q(z^{s} | x, y) q(z^{y} | y)} \left[ \log \frac{p(x | z^{x},z^{s})p(y | z^{y},z^{s}) \hspace*{0.1cm} p(z^{x})p(z^{s})p(z^{y})}{q(z^{x} | x) q(z^{s} | x, y) q(z^{y} | y)}\right]\\
&= \mathbb{E}_{q(z^{x} | x) q(z^{s} | x, y)} \left[ \log p(x | z^{x},z^{s}) \right] + \mathbb{E}_{(z^{s} | x, y) q(z^{y} | y)} \left[ \log p(y | z^{y},z^{s}) \right]\\
&\quad - D_{KL}\left[q(z^{x} | x) \| p(z^{x})\right] - D_{KL}\left[q(z^{s} | x, y) \| p(z^{s})\right] - D_{KL}\left[q(z^{y} | y) \| p(z^{y})\right].
\end{align*}

\subsection{$I(Z^{X}; Z^{S})=I(X;Z^{X}) + I(X;Z^{S}) - I(X; Z^{X}, Z^{S}).$}
\label{appendix:MI}
Interaction information \citep{1057469} between three random variables is defined as follows.
\begin{equation}
% \label{eq:InterInfoSym}
\begin{aligned}
    I\left(X; Y; Z\right) = I\left(X; Y\right) - I\left(X; Y | Z\right) = I\left(X; Z\right) - I\left(X; Z | Y\right) = I\left(Y; Z\right) - I\left(Y; Z | X\right). \\
\end{aligned}
\end{equation}
Using the last equality, we obtain the following expression of mutual information between $Z^X$ and $Z^Y$:
\begin{align}
I(Z^{X}; Z^{S}) = I(Z^{X}; X) - I(Z^{X}; X | Z^{S}) + I(Z^{X}; Z^{S} | X).
\end{align}
% Below we omit subscript $\phi$ for all entropic terms that involve with any of latent variables $z^{s}$, $z^{x}$, or $z^{y}$ for simplicity. Furthermore, 
Due to the structural assumption on $q$, $q(z^{x}|x)=q(z^{x}|x, z^{s})$ holds. Thus, the last term in the above equation disappears:
\begin{align}
I(Z^{X}; Z^{S} | X) &= H(Z^{X} | X) - H(Z^{X} | X, Z^{S}) = H(Z^{X} | X) - H(Z^{X} | X) = 0,\nonumber
\end{align}
which yields
\begin{align}
I(Z^{X}; Z^{S}) &= I(X;Z^{X}) - I(X; Z^{X} | Z^{S})\nonumber\\
&= I(X;Z^{X}) + I(X;Z^{S}) - I(X; Z^{X}, Z^{S}).
\end{align}

% \subsection{$q(z^{x} | z^{s}, x) = q(z^{x} | x)$ and $q(z^{y} | z^{s}, y) = q(z^{y} | y)$}
% Given: $q(z^{x}, z^{s}, z^{y} | x, y) = q(z^{x} | x) q(z^{s} | x, y) q(z^{y} | y)$
% \label{appendix:Integration}
% \begin{equation}
% \begin{aligned}
% &q(z^{x}, z^{s} | x, y) = \int q(z^{x}, z^{s}, z^{y} | x, y) d z^{y} = q(z^{x} | x) q(z^{s} | x, y) \int q(z^{y} | y) d z^{y} = q(z^{x} | x) q(z^{s} | x, y)\\
% &q(z^{y}, z^{s} | x, y) = \int q(z^{x}, z^{s}, z^{y} | x, y) d z^{x} = q(z^{y} | y) q(z^{s} | x, y) \int q(z^{x} | x) d z^{x} = q(z^{y} | y) q(z^{s} | x, y)\\
% \end{aligned}
% \end{equation}

% subsequently,
% \begin{equation}
% \begin{aligned}
% q(z^{x}, z^{s} | x) &= \int q(z^{x}, z^{s}, y | x) d y = \int q(z^{x}, z^{s} | y, x) p_{D}(y|x) d y\\
% &= q(z^{x}|x) \int q(z^{s} | y, x) p_{D}(y|x) d y = q(z^{x}|x) q(z^{s}|x)\\
% q(z^{y}, z^{s} | y) &= \int q(z^{y}, z^{s}, x | y) d x = \int q(z^{y}, z^{s} | x, y) p_{D}(x|y) d x\\
% &= q(z^{y}|y) \int q(z^{s} | x, y) p_{D}(x|y) d x = q(z^{y}|y) q(z^{s}|y)\\
% \end{aligned}
% \end{equation}

% Thus,
% \begin{equation}
% \begin{aligned}
% &q(z^{x} | z^{s}, x) = \frac{q(z^{x}, z^{s} | x)}{q(z^{s} | x)} = \frac{q(z^{x} | x) q(z^{s} | x)}{q(z^{s} | x)} = q(z^{x} | x)\\
% &q(z^{y} | z^{s}, y) = \frac{q(z^{y}, z^{s} | y)}{q(z^{s} | y)} = \frac{q(z^{y} | y) q(z^{s} | y)}{q(z^{s} | y)} = q(z^{y} | y)\\
% \end{aligned}
% \end{equation}
% Finally, $H(Z^{X} | X, Z^{S}) = H(Z^{X} | X)$ and $H(Z^{Y} | Y, Z^{S}) = H(Z^{Y} | Y)$.

% \subsection{Lower bound on $2\cdot I(X;Y;Z^{S}) - I(Z^{X}; Z^{S}) - I(Z^{Y}; Z^{S})$ (To be repaired)}
\subsection{Derivation of full objective}
\label{appendix:InfoConst}
\subsubsection{Lower bound on $(I(X;Y;Z^{S}) - I(Z^{X};Z^{S})) + (I(X;Y;Z^{S}) - I(Z^{Y};Z^{S}))$}
Here we derive the lower bound on $I(X;Y;Z^{S}) - I(Z^{X}; Z^{S})$ since the one on $I(X;Y;Z^{S}) - I(Z^{Y}; Z^{S})$ is analogous.

\begin{align}
& I(X;Y;Z^{S}) - I(Z^{X}; Z^{S}) \nonumber\\
&=\left(\cancel{I(X;Z^S)}-I(X;Z^{S}|Y)\right) - \left(I(X; Z^X) + \cancel{I(X; Z^S)} - I(X; Z^X,Z^S)\right) \nonumber\\
&= I(X; Z^{X}, Z^{S}) - I(X; Z^{X}) - I(X; Z^{S}|Y) \label{eq:DecomposedInfoConst}\\
&= H(X) + \mathbb{E}_{p_{D}(x)q(z^{s}, z^{x} | x)}\left[\log q(x | z^{x}, z^{s})\right] - \mathbb{E}_{p_{D}(x)}\left[D_{KL}\left[ q(z^{x} | x) \| q(z^{x})\right]\right] \nonumber\\
&\quad \quad \quad \quad - \mathbb{E}_{p_{D}(x,y)}\left[D_{KL}\left[ q(z^{s} | x, y) \| q(z^{s} | y)\right]\right] \nonumber
% &= H(X) + \mathbb{E}_{p_{D}(x, y) q(z^{s} | x,y) q(z^{x} | x)}\left[\log q(x | z^{x}, z^{s})\right] - \mathbb{E}_{p_{D}(x,y)}\left[D_{KL}\left[ q(z^{x} | x) \| q(z^{x})\right]\right] \nonumber\\
% &\quad \quad \quad \quad - \mathbb{E}_{p_{D}(x,y)}\left[D_{KL}\left[ q(z^{s} | x, y) \| q(z^{s} | y)\right]\right] \nonumber
\end{align}
where $q(x|z^{x}, z^{s}) = \frac {q(z^{x}, z^{s} | x) p_{D}(x)}{\int p_{D}(x,y) \hspace*{0.1cm} q(z^{x}, z^{s} | x, y) \hspace*{0.1cm} dx dy}$, $q(z^{x})=\int q(z^{x}|x)p_{D}(x) dx$,\\
and $q(z^{y})=\int q(z^{y}|y)p_{D}(y) dy$ require intractable integrals.
Thus, we need to derive the lower bound on Eq.~\eqref{eq:DecomposedInfoConst}. 

\paragraph{Variational lower bound on $I(X; Z^{X}, Z^{S})$}:\\
Note that $q(z^{x}, z^{s})$ is intractable due to the unknown density of $p_{D}(x,y)$: $q(z^{x}, z^{s}) = \int p_{D}(x,y) \hspace*{0.1cm} q(z^{x}, z^{s} | x, y) \hspace*{0.2cm} dx \hspace*{0.1cm}dy$
% \begin{equation}
% \begin{aligned}
% q(z^{x}, z^{s}) &= \int p_{D}(x,y) \hspace*{0.1cm} q(z^{x}, z^{s} | x, y) \hspace*{0.2cm} dx \hspace*{0.1cm}dy\\
% &= \int p_{D}(x,y) \hspace*{0.1cm} q(z^{x}| x) \hspace*{0.1cm} q(z^{s} | x, y) \hspace*{0.2cm}dx \hspace*{0.1cm}dy\\
% &= \int p_{D}(x) \hspace*{0.1cm} q(z^{x}| x) \left( \int p_{D}(y|x) \hspace*{0.1cm} q(z^{s} | x, y)  \hspace*{0.1cm} dy \right) dx    
% \end{aligned}
% \end{equation}

Consequently, $q(x|z^{x}, z^{s}) = \frac {q(z^{x}, z^{s} | x) p_{D}(x)}{q(z^{x}, z^{s})}$ is intractable. Thus, we would like to bring the generative distribution $p(x | z^{x}, z^{s})$  to derive a lower bound such that:
\begin{align*}
I(X; Z^{X}, Z^{S})
&= \mathbb{E}_{q(z^{x}, z^{s} | x) p_{D}(x)}\left[\log \frac{q(x | z^{x}, z^{s})}{p_{D}(x)}\right]\\
&= H(X) + \mathbb{E}_{q(z^{x}, z^{s} | x) p_{D}(x)}\left[\log q(x | z^{x}, z^{s}) - \log p(x | z^{x}, z^{s}) + \log p(x | z^{x}, z^{s})\right]\\
&= H(X) + \mathbb{E}_{q(z^{x}, z^{s} | x) p_{D}(x)}\left[\log p(x | z^{x}, z^{s}) \right] + \mathbb{E}_{q(z^{x}, z^{s})}\left[ D_{KL} \left[ q(x | z^{x}, z^{s}) \| p(x | z^{x}, z^{s}) \right] \right]\\
&\geq H(X) + \mathbb{E}_{q(z^{x}, z^{s} | x) p_{D}(x)}\left[\log p(x | z^{x}, z^{s}) \right]\\
&= H(X) + \int q(z^{x}, z^{s} | x) p_{D}(x) \hspace*{0.1cm} \log p(x | z^{x}, z^{s}) \hspace*{0.2cm}d x \hspace*{0.1cm}d z^{x} \hspace*{0.1cm}d z^{s}\\
&= H(X) + \int p_{D}(x) \left( \int q(z^{x}, z^{s} | x, y) \hspace*{0.1cm} p_{D}(y|x) \hspace*{0.1cm}d y\right) \log p(x | z^{x}, z^{s}) \hspace*{0.2cm}d x \hspace*{0.1cm}d z^{x} \hspace*{0.1cm}d z^{s}\\
&= H(X) + \int p_{D}(x) \hspace{0.1cm} q(z^{x} | x)\left( \int q(z^{s} | x, y) \hspace*{0.1cm} p_{D}(y|x) \hspace*{0.1cm}d y\right) \log p(x | z^{x}, z^{s}) \hspace*{0.2cm}d x \hspace*{0.1cm}d z^{x} \hspace*{0.1cm}d z^{s}\\
&= H(X) + \int p_{D}(x,y) \hspace*{0.1cm} q(z^{x} | x) \hspace*{0.1cm} q(z^{s} | x, y) \log p(x | z^{x}, z^{s}) \hspace*{0.2cm}d x \hspace*{0.1cm}d y \hspace*{0.1cm}d z^{x} \hspace*{0.1cm}d z^{s}\\
&= H(X) + \mathbb{E}_{p_{D}(x,y) \hspace*{0.1cm} q(z^{x} | x) \hspace*{0.1cm} q(z^{s} | x, y)}\left[\log p(x | z^{x}, z^{s}) \right]\\
\end{align*}
Thus, maximization of $\mathbb{E}_{p_{D}(x,y) \hspace*{0.1cm} q(z^{x} | x) \hspace*{0.1cm} q(z^{s} | x, y)}\left[\log p(x | z^{x}, z^{s}) \right]$ not only maximizes $I(X; Z^{X}, Z^{S})$,\\
but also fits $p(x | z^{x}, z^{s})$ to $q(x | z^{x}, z^{s})$, so that we can utilize it as a decoder.

\paragraph{Variational upper bound on $I(X;Z^{S}|Y)$}:\\
Note that $q(z^{s} | y) = \int p_{D}(x|y) \hspace*{0.1cm} q(z^{s} | x, y) \hspace*{0.2cm} dx$ is intractable. Thus, 
\begin{align}
I(X;Z^{S}|Y)
&= \mathbb{E}_{p_{D}(x,y) q(z^{s} | x, y)}\left[ \log \frac{q(z^{s} | x, y)}{q(z^{s} | y)} \right] \hspace*{1.2cm}\left( = \mathbb{E}_{p_{D}(x,y)}\left[D_{KL}\left[ q(z^{s} | x, y) \| q(z^{s} | y)\right]\right] \right)\nonumber\\
&= \mathbb{E}_{p_{D}(x,y) q(z^{s} | x, y)}\left[ \log \frac{q(z^{s} | x, y) r^{y}(z^{s} | y)}{r^{y}(z^{s} | y)q(z^{s} | y)} \right]\nonumber\\
&= \mathbb{E}_{p_{D}(x,y)}\left[D_{KL}\left[ q(z^{s} | x, y) \| r^{y}(z^{s} | y)\right]\right] - \mathbb{E}_{p_{D}(y)}\left[D_{KL}\left[ q(z^{s} | y) \| r^{y}(z^{s} | y)\right]\right]\nonumber\\
&\leq \mathbb{E}_{p_{D}(x,y)}\left[D_{KL}\left[ q(z^{s} | x, y) \| r^{y}(z^{s} | y)\right]\right]\label{eq:IneqUpperbound}
\end{align}
Thus, minimization of $\mathbb{E}_{p_{D}(x,y)}\left[D_{KL}\left[ q(z^{s} | x, y) \| r^{y}(z^{s} | y)\right]\right]$ not only minimizes $I(X;Z^{S}|Y)$,\\
\hspace*{8.5cm} but also fits $r^{y}(z^{s} | y)$ to $q(z^{s} | y)$.
\paragraph{Variational upper bound on $I(X;Z^{X})$}:\\
Similar to Eq.~\eqref{eq:IneqUpperbound}, $I(X;Z^{X}) = \mathbb{E}_{p_{D}(x)}\left[D_{KL}\left[ q(z^{x} | x) \| q(z^{x})\right]\right] \leq \mathbb{E}_{p_{D}(x)}\left[D_{KL}\left[ q(z^{x} | x) \| p(z^{x})\right]\right]$.

\paragraph{Overall information preference}:\\
Putting together, we can derive the lower bound of
the preference for $q$ on domain $X$ and $Y$: %Eq.~\eqref{eqn:final_inforeg}.
\begin{align}
&(I(X;Y;Z^{S}) - I(Z^{X};Z^{S})) + (I(X;Y;Z^{S}) - I(Z^{Y};Z^{S}))\nonumber \\
&=2 \cdot I(X;Y;Z^{S}) - I(Z^{X};Z^{S}) - I(Z^{Y};Z^{S}) \nonumber \\
&= I(X; Z^{X}, Z^{S}) + I(Y; Z^{Y}, Z^{S}) - I(X;Z^{X}) - I(Y;Z^{Y}) - I(X;Z^{S}|Y) - I(Y;Z^{S}|X) \nonumber \\
&\geq \mathbb{E}_{p_D(x,y)} \left[\hspace*{0.1cm} \mathbb{E}_{q(z^{s} | x,y) q(z^{x} | x)}\left[\log p(x | z^{x}, z^{s}) \right] + \mathbb{E}_{q(z^{s} | x,y) q(z^{y} | y)}\left[ \log p(y | z^{y}, z^{s})\right] \hspace*{0.1cm} \right] \label{eq:rec}\\
&\quad - \mathbb{E}_{p_{D}(x,y)}\left[ \hspace*{0.1cm} D_{KL}\left[ q(z^{x} | x) \| p(z^{x}) \right] + D_{KL}\left[q(z^{y} | y) \| p(z^{y})\right] \hspace*{0.1cm} \right] \label{eq:kl_ex}\\
&\quad - \mathbb{E}_{p_{D}(x,y)}\left[ \hspace*{0.1cm} D_{KL}\left[ q(z^{s} | x, y) \| r^{y}(z^{s} | y) \right] + D_{KL}\left[ q(z^{s} | x, y) \| r^{x}(z^{s} | x)\right] \hspace*{0.1cm}\right] \nonumber\\
&\quad + H(X) + H(Y).\nonumber
\end{align}

\subsubsection{Merging ELBO and information preference}
Note that Eq.~\eqref{eq:rec} and Eq.~\eqref{eq:kl_ex} coexist in ELBO as reconstruction and KL regularization terms (on exclusive representation) respectively. Thus, the final objective is as follows:
\begin{align}
&\max_{p,q} \mathbb{E}_{q(z^{x}, z^{s}, z^{y}, x,y)}\left[ \log \frac{p(x,y,z^{x},z^{s},z^{y})}{q(z^{x}, z^{s}, z^{y}|x,y)} \right] + \lambda \left(2\cdot I(X;Y;Z^{S}) - I(Z^{X};Z^{S}) - I(Z^{Y};Z^{S}) \right) \nonumber \\
&\geq \max_{p,q,r} \left(1+\lambda \right) \cdot \mathbb{E}_{p_{D}(x,y)} \left[\hspace*{0.1cm} \mathbb{E}_{q(z^{x} | x) q(z^{s} | x,y)} \left[ \log p(x | z^{x},z^{s}) \right] + \mathbb{E}_{q(z^{y} | y) q(z^{s} | x,y)} \left[ \log p(y | z^{y},z^{s}) \right] \hspace*{0.1cm}\right] \nonumber \\%\label{eq:final_inequal}\\
&\quad \quad \quad - \left(1+\lambda \right) \cdot \mathbb{E}_{p_{D}(x,y)} \left[\hspace*{0.1cm} D_{KL}\left[q(z^{x} | x) \| p(z^{x})\right] + D_{KL}\left[ q(z^{y} | y) | p(z^{y}) \right] \hspace*{0.1cm} \right] \nonumber \\%\label{eq:final_ex_VIB}\\
&\quad \quad \quad - \mathbb{E}_{p_{D}(x,y)} \left[ \hspace*{0.1cm} D_{KL}\left[ q(z^{s} | x, y) \| p(z^{s})\right] \hspace*{0.1cm} \right] \nonumber\\%\label{eq:final_sh_VIB}\\
&\quad \quad \quad - \lambda \cdot \mathbb{E}_{p_{D}(x,y)} \left[ \hspace*{0.1cm} D_{KL}\left[ q(z^{s} | x, y) \| r^{y}(z^{s} | y) \right] + D_{KL}\left[ q(z^{s} | x, y) \| r^{x}(z^{s} | x) \right] \hspace*{0.1cm} \right] \nonumber \\
&= \max_{p,q,r} (1+\lambda) \cdot \mathbb{E}_{p_{D}(x,y)} \left[ \hspace*{0.1cm} ELBO(p,q) \hspace*{0.1cm} \right]\nonumber\\
&\quad \quad \quad + \lambda \cdot \mathbb{E}_{p_{D}(x,y)} \left[ \hspace*{0.1cm} D_{KL}\left[ q(z^{s} | x, y) \| p(z^{s})\right] \hspace*{0.1cm} \right] \nonumber\\
&\quad \quad \quad - \lambda \cdot \mathbb{E}_{p_{D}(x,y)} \left[ \hspace*{0.1cm} D_{KL}\left[ q(z^{s} | x, y) \| r^{y}(z^{s} | y) \right] + D_{KL}\left[ q(z^{s} | x, y) \| r^{x}(z^{s} | x) \right] \hspace*{0.1cm} \right].\nonumber
\end{align}
\section{Additional quantitative results}
\label{appendix:quant}
\subsection{Sample quality evaluation}%table here (The lower, the better)
\label{appendix:sample_quality}
\begin{table}[H]
    \centering
    \begin{tabular}{cccc}
        \toprule
        % \cmidrule(lr){2-3}
        \textbf{Translation} & pix2pix [23] & CdDN [12] & IIAE\\
        \midrule
        $X \rightarrow Y$ & 0.24987 $\pm$ 0.00780 & 0.23517 $\pm$ 0.00799 & \textbf{0.21478} $\pm$ 0.00844 \\
        % \midrule
        $Y \rightarrow X$ & 0.21524 $\pm$ 0.00704 & 0.19295 $\pm$ 0.00687 & \textbf{0.15277} $\pm$ 0.00774 \\
        \bottomrule
    \end{tabular}
    \caption{Sample quality evaluation on the Cars (bimodal) dataset.}
    \vspace{-0.6cm}
    \label{tab:sample_qual}
\end{table}
We report quantitative evaluation on the quality of samples in table \ref{tab:sample_qual}.
We followed the exact experimental setting for the Cars dataset as in \citep{gonzalez-garcia2018NeurIPS}, except we use freshly generated training data (the data from \citep{gonzalez-garcia2018NeurIPS} was unavailable) and the updated version of the evaluation metric LPIPS.
Thus, the numbers here do not exactly match those in \citep{gonzalez-garcia2018NeurIPS}.
The results show that the sample quality of IIAE clearly exceeds the quality of GAN-based methods.

\subsection{Additional notes on table \ref{tab:MNIST-MAPS} } 
\label{appendix:additional_note}
\begin{wraptable}{r}{9cm}% automatically uses minimum width
% \raisebox{\baselineskip}[20pt][\dimexpr\depth-2\baselineskip]{
\vspace{-2ex}
    \begin{tabular}{cccc}
        \toprule
        Facades(Val) & BicycleGAN \citep{zhu2017toward} & CdDN \citep{gonzalez-garcia2018NeurIPS} & IIAE \\
        \midrule
        F $\rightarrow$ L (\%) & - & 95.0 (1.0) & \textbf{100.0} (1.0) \\
        L $\rightarrow$ F (\%) & 45.0 & 97.0 (1.0) & \textbf{100.0} (0.0) \\
        \bottomrule
    \end{tabular}
    \vspace{-0.2cm}
    \caption{Shared (exclusive) representation based retrieval on validation set in the Facades \citep{10.1007/978-3-642-40602-7_39} dataset.}
    \vspace{-0.1cm}
    \label{tab:Facades_val}
\vspace{-2ex}
\end{wraptable}
Regarding the numbers from the Facades dataset in table \ref{tab:MNIST-MAPS}, they are different from \citep{gonzalez-garcia2018NeurIPS} since we used test set rather than the validation set (stated in the footnote).
The table \ref{tab:Facades_val} shows the result on the validation set, which matches the numbers in \citep{gonzalez-garcia2018NeurIPS}.

\subsection{Ablation study}
\label{appendix:ablation}
\begin{table}[H]
    \centering
    \begin{tabular}{ccccc}
        \toprule
        % \cmidrule(lr){2-3}
        \textbf{Metric} & II & II-MI &  ELBO+$\lambda$II & ELBO+$\lambda$(II-MI) \\
        \midrule
        mAP & 0.517 & 0.534 & 0.516 & \textbf{0.573} \\
        % \midrule
        P$@$100 & 0.605 & 0.616 & 0.595 & \textbf{0.659} \\
        \bottomrule             
    \end{tabular}
    \caption{Ablation study on ZS-SBIR.}
    \label{tab:Ablation}
    \vspace{-0.6cm}
\end{table}
We evaluated the effect of terms in the IIAE objective using the ZS-SBIR dataset. Table \ref{tab:Ablation} summarizes the result.
II represents maximizing only the interaction information among $X$,$Y$, and $Z^S$ (Eq.~\eqref{eqn:II}), whose lower bound is as follows: 
\begin{align}
2 \cdot I(X;Y;Z^{S}) &= I(X; Z^{S}) + I(Y; Z^{S}) - I(X;Z^{S}|Y) - I(Y;Z^{S}|X) \label{eqn:II}\\
&\geq \mathbb{E}_{p_D(x,y)} \left[\hspace*{0.1cm} \mathbb{E}_{q(z^{s} | x,y)}\left[\log p(x | z^{s}) \right] + \mathbb{E}_{q(z^{s} | x,y)}\left[ \log p(y | z^{s})\right] \hspace*{0.1cm} \right] \nonumber\\
&\quad - \mathbb{E}_{p_{D}(x,y)}\left[ \hspace*{0.1cm} D_{KL}\left[ q(z^{s} | x, y) \| r^{y}(z^{s} | y) \right] + D_{KL}\left[ q(z^{s} | x, y) \| r^{x}(z^{s} | x)\right] \hspace*{0.1cm}\right] \nonumber\\
&\quad + H(X) + H(Y). \label{eqn:LowerII}
\end{align}
II-MI is the joint information preference of maximizing the interaction information and minimizing mutual information between shared and domain-specific representations simultaneously (Eq.~\eqref{eqn:II-MI}), whose lower bound is Eq.~\eqref{eqn:LowerII-MI}.
\begin{align}
% &(I(X;Y;Z^{S}) - I(Z^{X};Z^{S})) + (I(X;Y;Z^{S}) - I(Z^{Y};Z^{S}))\nonumber \\
&2 \cdot I(X;Y;Z^{S}) - I(Z^{X};Z^{S}) - I(Z^{Y};Z^{S}) \label{eqn:II-MI} \\
&= I(X; Z^{X}, Z^{S}) + I(Y; Z^{Y}, Z^{S}) - I(X;Z^{X}) - I(Y;Z^{Y}) - I(X;Z^{S}|Y) - I(Y;Z^{S}|X) \label{eqn:II-MI-2} \\
&\geq \mathbb{E}_{p_D(x,y)} \left[\hspace*{0.1cm} \mathbb{E}_{q(z^{s} | x,y) q(z^{x} | x)}\left[\log p(x | z^{x}, z^{s}) \right] + \mathbb{E}_{q(z^{s} | x,y) q(z^{y} | y)}\left[ \log p(y | z^{y}, z^{s})\right] \hspace*{0.1cm} \right] \nonumber\\
&\quad - \mathbb{E}_{p_{D}(x,y)}\left[ \hspace*{0.1cm} D_{KL}\left[ q(z^{x} | x) \| p(z^{x}) \right] + D_{KL}\left[q(z^{y} | y) \| p(z^{y})\right] \hspace*{0.1cm} \right] \nonumber\\
&\quad - \mathbb{E}_{p_{D}(x,y)}\left[ \hspace*{0.1cm} D_{KL}\left[ q(z^{s} | x, y) \| r^{y}(z^{s} | y) \right] + D_{KL}\left[ q(z^{s} | x, y) \| r^{x}(z^{s} | x)\right] \hspace*{0.1cm}\right] \nonumber\\
&\quad + H(X) + H(Y). \label{eqn:LowerII-MI}
\end{align}

Last two columns in table \ref{tab:Ablation} represent taking weighted sum with the ELBO, treating $\lambda=2$ as the hyperparameter.
The final column is the objective of IIAE.
%Comparing to table \ref{tab:ZS-SBIR}, all settings significantly outperform SOTA, and shows that subtracting MI from II always help.

The first two columns imply that augmenting the minimization of the mutual information to the maximization of the interaction information is beneficial.
This is because the optimization of Eq.~\eqref{eqn:II} gives $Z^S$ an implicit trade-off between capturing domain-specific information to maximize the first and second terms and emptying domain-specific information to minimize the third and fourth terms in Eq.~\eqref{eqn:II}.
Thus, encoding the domain-specific information in addition to the shared information can be one of optimal solutions for $Z^S$.
On the other hand, optimizing Eq.~\eqref{eqn:II-MI} (or Eq.~\eqref{eqn:II-MI-2}) eliminates the trade-off since the first and second terms in Eq.~\eqref{eqn:II-MI-2} allow $Z^{S}$ to share with $Z^{X}$ and $Z^{Y}$ the burden of being informative to $X$ and $Y$.
Consequently, the optimal solution of Eq.~\eqref{eqn:II-MI} is that $Z^S$ encodes only the information shared across $X$ and $Y$ while $Z^X$ and $Z^Y$ encode only the domain-specific information.

Finally, the last two columns show that the joint information preference is better suited to ELBO than maximization of the interaction information only and gains further performance improvement.

\section{Visualization}
% Additional generation of IIAE trained with Cars dataset
% Some retrieval result of ZS-SBIR
\subsection{Additional samples of cross-domain image translation}
\label{appendix:visual}
\subsubsection{MNIST-CDCB \citep{gonzalez-garcia2018NeurIPS}}
We present additional samples of image translation with IIAE in table \ref{tab:I2I_MNIST}. 
Furthermore, we generate visual analogies using IIAE which are presented in table \ref{tab:VisAnalogy_MNIST}. For each row, we show the queries (the first and fourth columns), references (the second and fifth columns), and the synthesized images (the third and sixth columns). Queries are sources of shared representation, which is digit identity, whereas references are sources of exclusive representations, which are color variations. Tables \ref{tab:I2I_MNIST} and \ref{tab:VisAnalogy_MNIST} shows that IIAE extracts and preserves both of domain specific and shared representations properly.

\begin{table}[H]
    \begin{center}
    \begin{tabular}{ cccccccccccc }
    \toprule
    \multicolumn{5}{c}{\textbf{ $X \rightarrow Y$ }} & \multicolumn{5}{c}{\textbf{ $Y \rightarrow X$ }} \\
    \cmidrule(lr){1-5}\cmidrule(lr){6-10}
    \textbf{Input} & \multicolumn{4}{c}{\textbf{ Outputs w/ different $z^{y}$ }} & \textbf{Input} &  \multicolumn{4}{c}{\textbf{ Outputs w/ different $z^{x}$ }} \\
    \cmidrule(lr){1-1}\cmidrule(lr){2-5}\cmidrule(lr){6-6}\cmidrule(lr){7-10}
    x & \multicolumn{3}{c}{$z^{y}_{1},z^{y}_{2},z^{y}_{3} \sim p(z^{y})$} & $\mu^{y}$ & y & \multicolumn{3}{c}{$z^{x}_{1},z^{x}_{2},z^{x}_{3} \sim p(z^{x})$}  & $\mu^{x}$\\
    \cmidrule(lr){2-4}\cmidrule(lr){7-9}
    \includegraphics[width=0.07\textwidth]{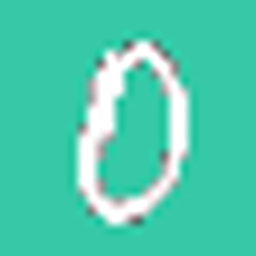} &
    \includegraphics[width=0.07\textwidth]{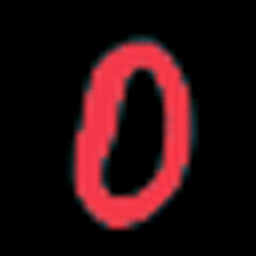} &
    \includegraphics[width=0.07\textwidth]{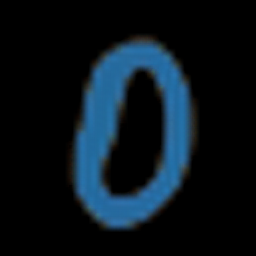} &
    \includegraphics[width=0.07\textwidth]{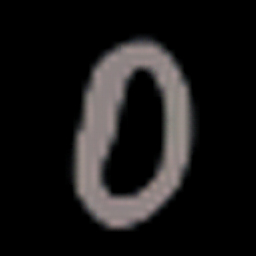} &
    \includegraphics[width=0.07\textwidth]{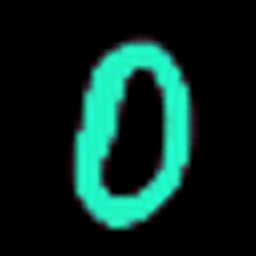} &
    \includegraphics[width=0.07\textwidth]{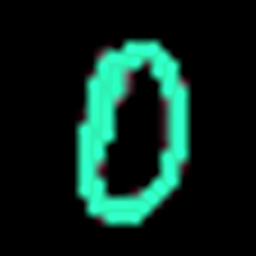} &
    \includegraphics[width=0.07\textwidth]{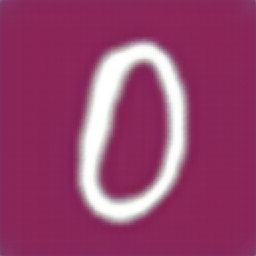} &
    \includegraphics[width=0.07\textwidth]{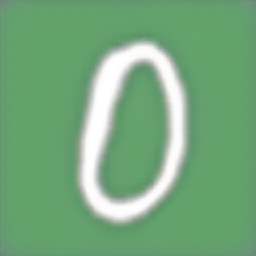} &
    \includegraphics[width=0.07\textwidth]{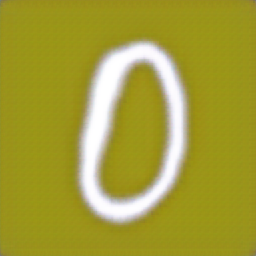} &
    \includegraphics[width=0.07\textwidth]{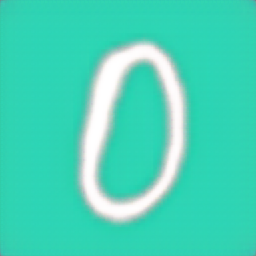} \\
    \includegraphics[width=0.07\textwidth]{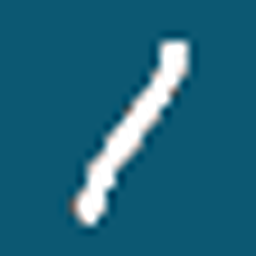} &
    \includegraphics[width=0.07\textwidth]{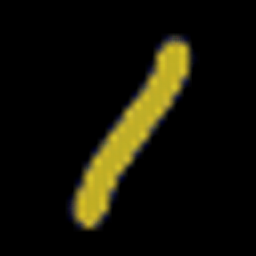} &
    \includegraphics[width=0.07\textwidth]{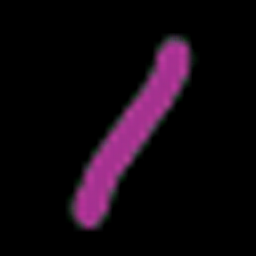} &
    \includegraphics[width=0.07\textwidth]{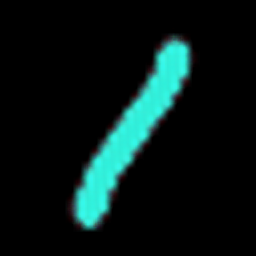} &
    \includegraphics[width=0.07\textwidth]{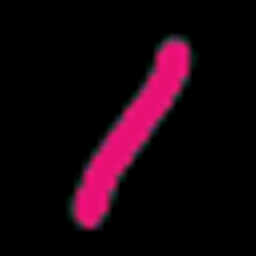} &
    \includegraphics[width=0.07\textwidth]{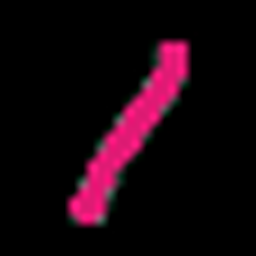} &
    \includegraphics[width=0.07\textwidth]{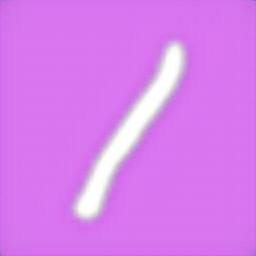} &
    \includegraphics[width=0.07\textwidth]{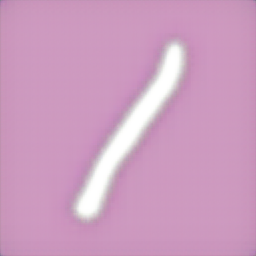} &
    \includegraphics[width=0.07\textwidth]{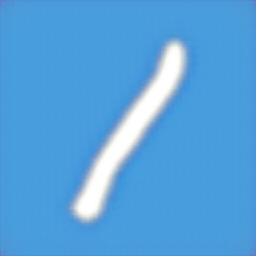} &
    \includegraphics[width=0.07\textwidth]{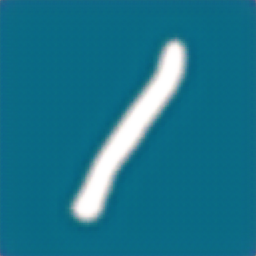} \\
    \includegraphics[width=0.07\textwidth]{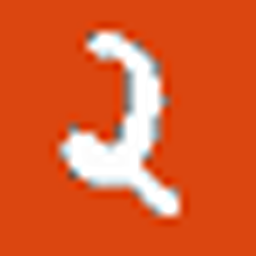} &
    \includegraphics[width=0.07\textwidth]{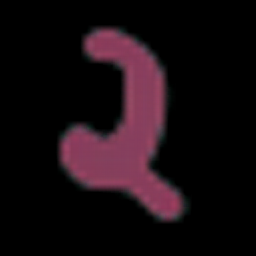} &
    \includegraphics[width=0.07\textwidth]{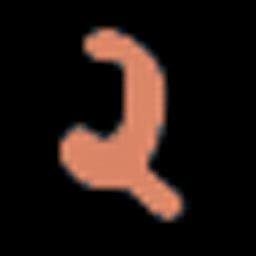} &
    \includegraphics[width=0.07\textwidth]{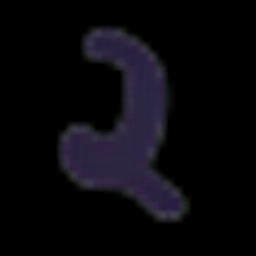} &
    \includegraphics[width=0.07\textwidth]{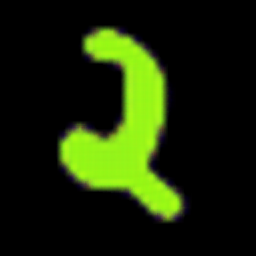} &
    \includegraphics[width=0.07\textwidth]{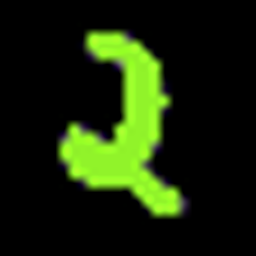} &
    \includegraphics[width=0.07\textwidth]{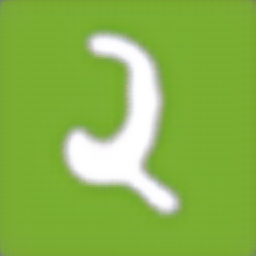} &
    \includegraphics[width=0.07\textwidth]{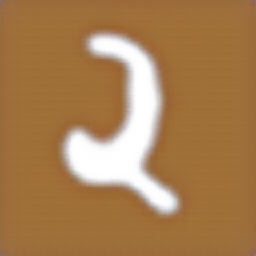} &
    \includegraphics[width=0.07\textwidth]{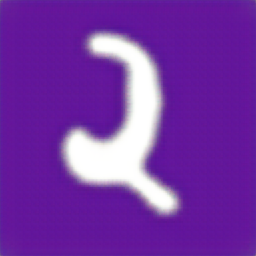} &
    \includegraphics[width=0.07\textwidth]{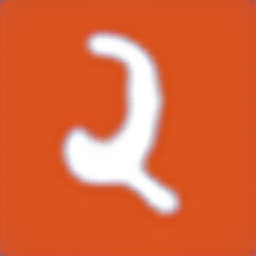} \\
    \includegraphics[width=0.07\textwidth]{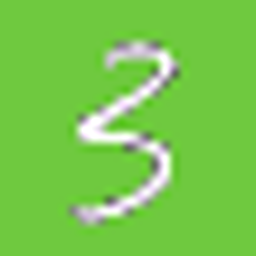} &
    \includegraphics[width=0.07\textwidth]{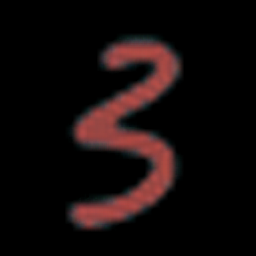} &
    \includegraphics[width=0.07\textwidth]{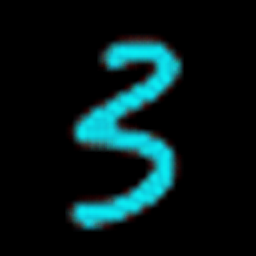} &
    \includegraphics[width=0.07\textwidth]{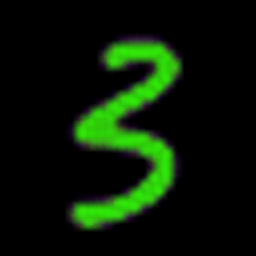} &
    \includegraphics[width=0.07\textwidth]{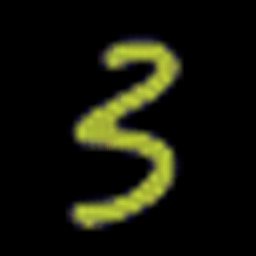} &
    \includegraphics[width=0.07\textwidth]{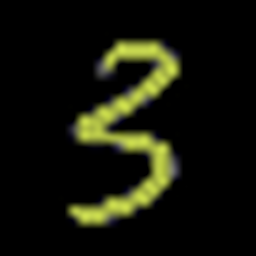} &
    \includegraphics[width=0.07\textwidth]{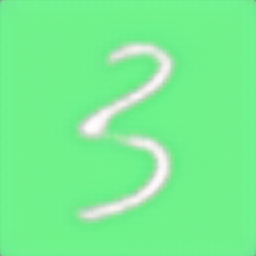} &
    \includegraphics[width=0.07\textwidth]{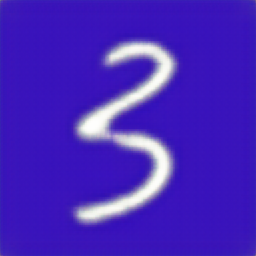} &
    \includegraphics[width=0.07\textwidth]{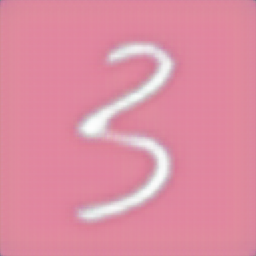} &
    \includegraphics[width=0.07\textwidth]{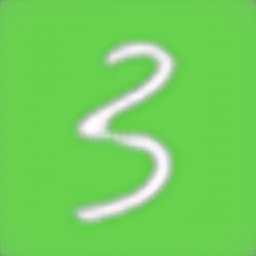} \\
    \includegraphics[width=0.07\textwidth]{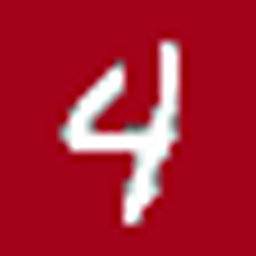} &
    \includegraphics[width=0.07\textwidth]{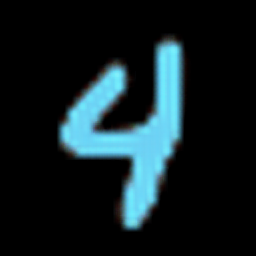} &
    \includegraphics[width=0.07\textwidth]{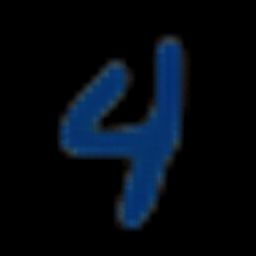} &
    \includegraphics[width=0.07\textwidth]{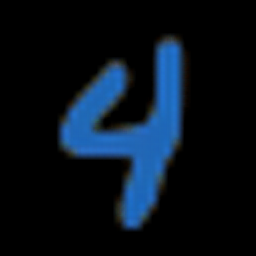} &
    \includegraphics[width=0.07\textwidth]{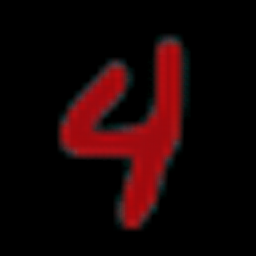} &
    \includegraphics[width=0.07\textwidth]{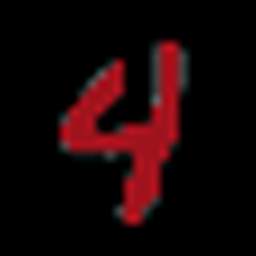} &
    \includegraphics[width=0.07\textwidth]{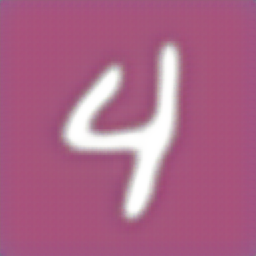} &
    \includegraphics[width=0.07\textwidth]{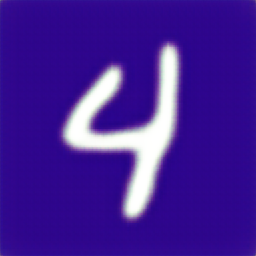} &
    \includegraphics[width=0.07\textwidth]{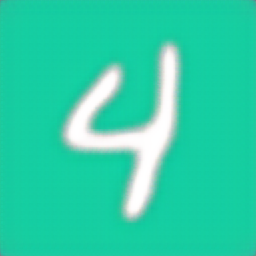} &
    \includegraphics[width=0.07\textwidth]{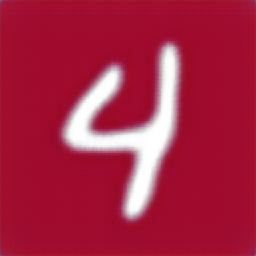} \\
    \includegraphics[width=0.07\textwidth]{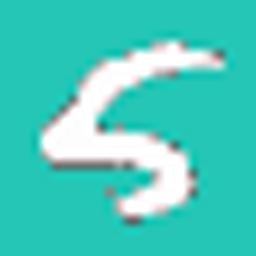} &
    \includegraphics[width=0.07\textwidth]{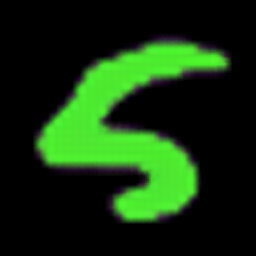} &
    \includegraphics[width=0.07\textwidth]{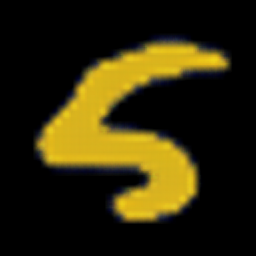} &
    \includegraphics[width=0.07\textwidth]{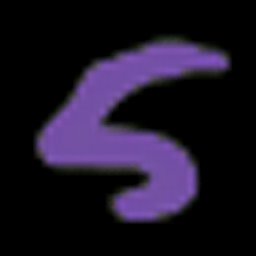} &
    \includegraphics[width=0.07\textwidth]{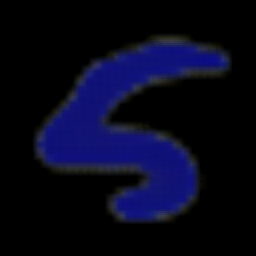} &
    \includegraphics[width=0.07\textwidth]{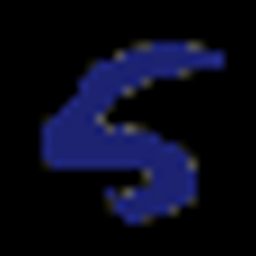} &
    \includegraphics[width=0.07\textwidth]{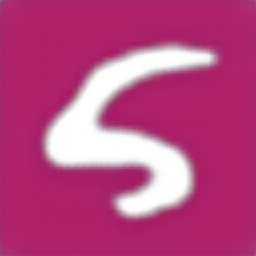} &
    \includegraphics[width=0.07\textwidth]{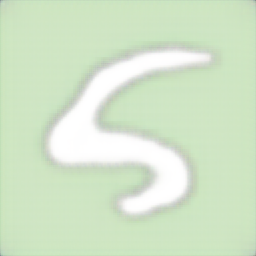} &
    \includegraphics[width=0.07\textwidth]{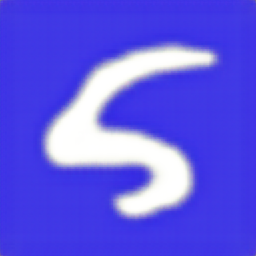} &
    \includegraphics[width=0.07\textwidth]{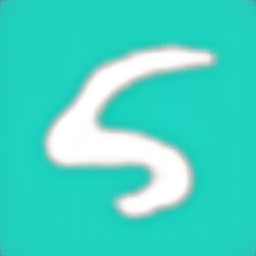} \\
    \includegraphics[width=0.07\textwidth]{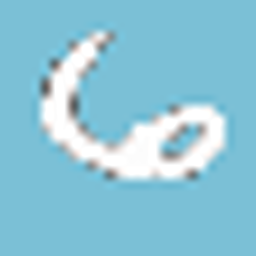} &
    \includegraphics[width=0.07\textwidth]{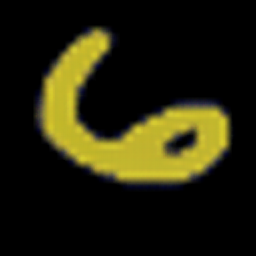} &
    \includegraphics[width=0.07\textwidth]{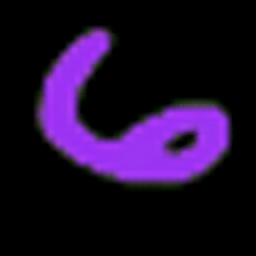} &
    \includegraphics[width=0.07\textwidth]{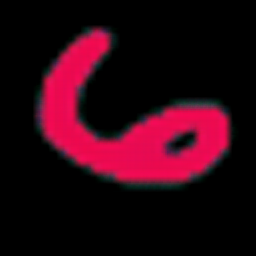} &
    \includegraphics[width=0.07\textwidth]{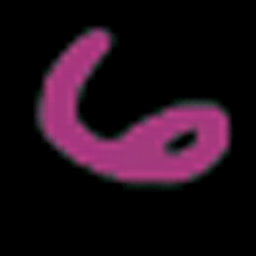} &
    \includegraphics[width=0.07\textwidth]{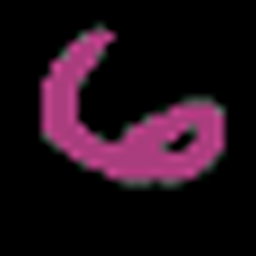} &
    \includegraphics[width=0.07\textwidth]{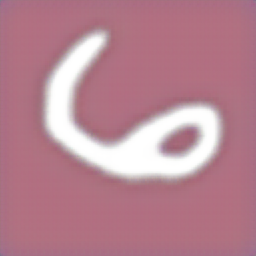} &
    \includegraphics[width=0.07\textwidth]{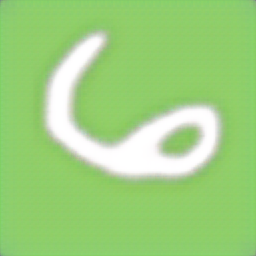} &
    \includegraphics[width=0.07\textwidth]{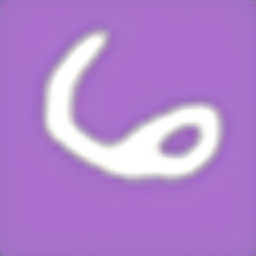} &
    \includegraphics[width=0.07\textwidth]{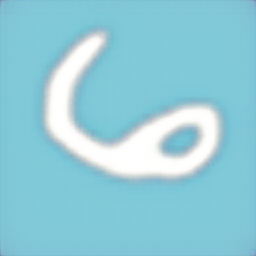} \\
    \includegraphics[width=0.07\textwidth]{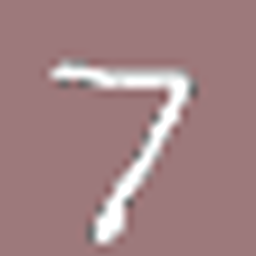} &
    \includegraphics[width=0.07\textwidth]{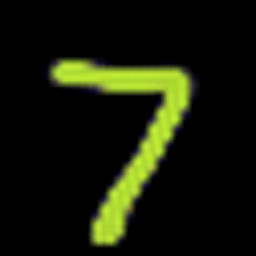} &
    \includegraphics[width=0.07\textwidth]{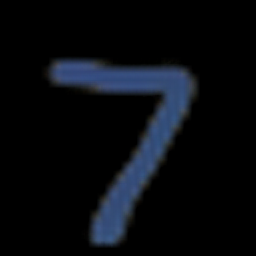} &
    \includegraphics[width=0.07\textwidth]{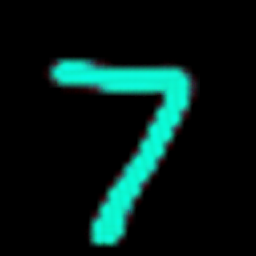} &
    \includegraphics[width=0.07\textwidth]{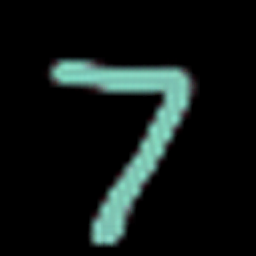} &
    \includegraphics[width=0.07\textwidth]{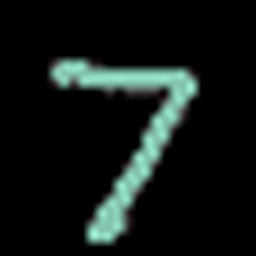} &
    \includegraphics[width=0.07\textwidth]{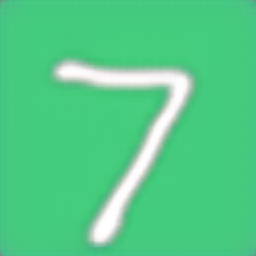} &
    \includegraphics[width=0.07\textwidth]{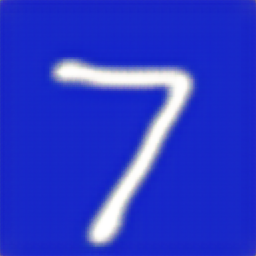} &
    \includegraphics[width=0.07\textwidth]{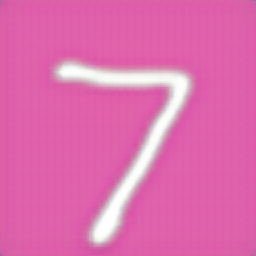} &
    \includegraphics[width=0.07\textwidth]{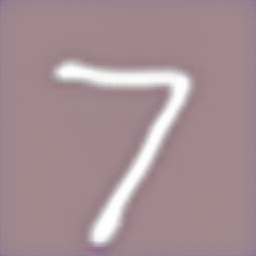} \\
    \includegraphics[width=0.07\textwidth]{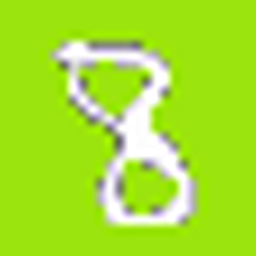} &
    \includegraphics[width=0.07\textwidth]{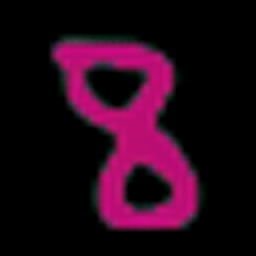} &
    \includegraphics[width=0.07\textwidth]{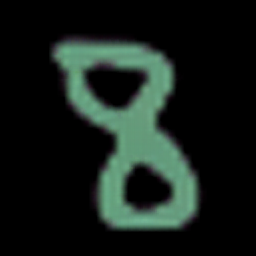} &
    \includegraphics[width=0.07\textwidth]{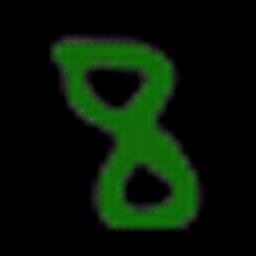} &
    \includegraphics[width=0.07\textwidth]{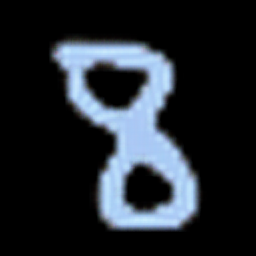} &
    \includegraphics[width=0.07\textwidth]{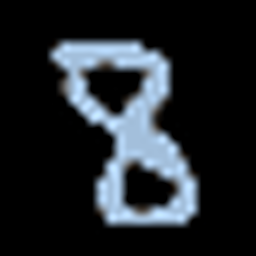} &
    \includegraphics[width=0.07\textwidth]{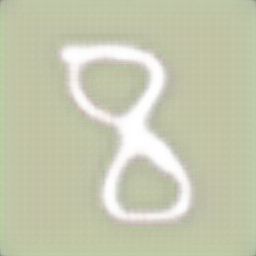} &
    \includegraphics[width=0.07\textwidth]{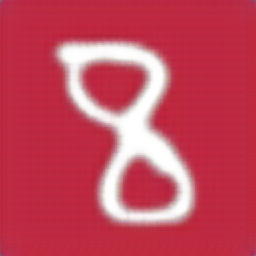} &
    \includegraphics[width=0.07\textwidth]{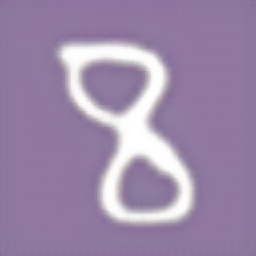} &
    \includegraphics[width=0.07\textwidth]{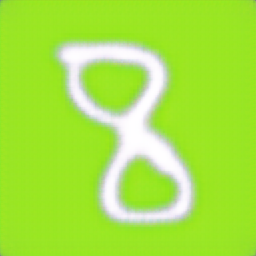} \\
    \includegraphics[width=0.07\textwidth]{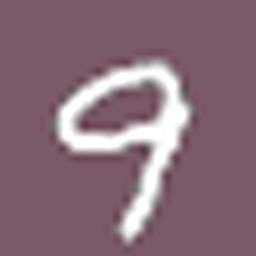} &
    \includegraphics[width=0.07\textwidth]{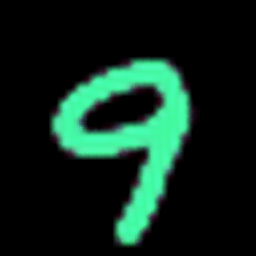} &
    \includegraphics[width=0.07\textwidth]{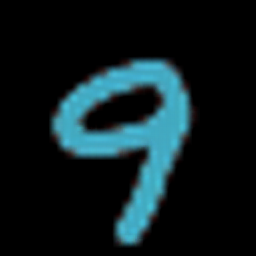} &
    \includegraphics[width=0.07\textwidth]{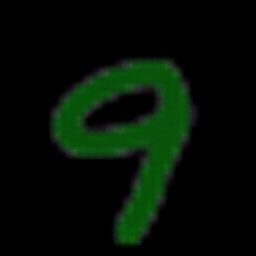} &
    \includegraphics[width=0.07\textwidth]{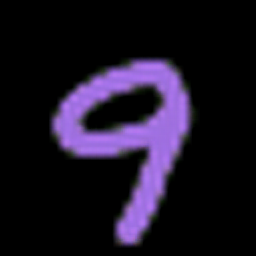} &
    \includegraphics[width=0.07\textwidth]{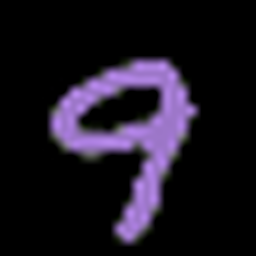} &
    \includegraphics[width=0.07\textwidth]{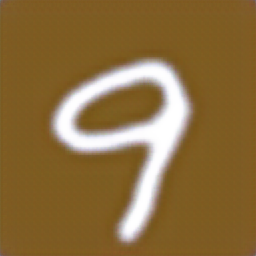} &
    \includegraphics[width=0.07\textwidth]{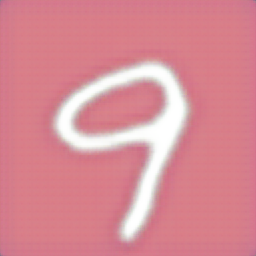} &
    \includegraphics[width=0.07\textwidth]{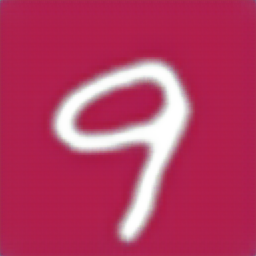} &
    \includegraphics[width=0.07\textwidth]{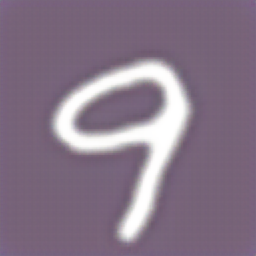} \\
\bottomrule
    \end{tabular}
    \caption{Additional cross-domain translation results in MNIST-CDCB \citep{gonzalez-garcia2018NeurIPS} by IIAE. In MNIST-CDCB, domain-specific factors of variation are color variation in background($X$) and in foreground($Y$) while the common factor is the digit ID.
    }
    \label{tab:I2I_MNIST}
    \end{center}
    \vspace{-1cm}
\end{table}

\begin{table}[H]
    \begin{center}
    \begin{tabular}{ cccccccc }
    \toprule
    \multicolumn{3}{c}{\textbf{ $X$ }} & \multicolumn{3}{c}{\textbf{ $Y$ }} \\
    \cmidrule(lr){1-3}\cmidrule(lr){4-6}
    \textbf{query} & \textbf{reference} & \textbf{Output} &  \textbf{query} & \textbf{reference} & \textbf{Output}\\
    \includegraphics[width=0.07\textwidth]{figs/Mnist2/visual0/00056-inputsX.png} &
    \includegraphics[width=0.07\textwidth]{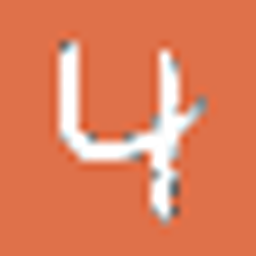} &
    \includegraphics[width=0.07\textwidth]{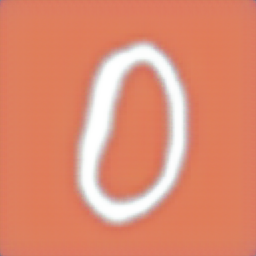} &
    \includegraphics[width=0.07\textwidth]{figs/Mnist2/visual0/00056-inputsY.png} &
    \includegraphics[width=0.07\textwidth]{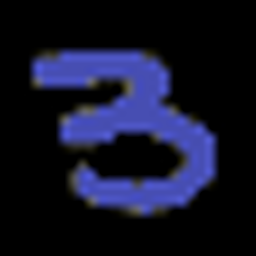} &
    \includegraphics[width=0.07\textwidth]{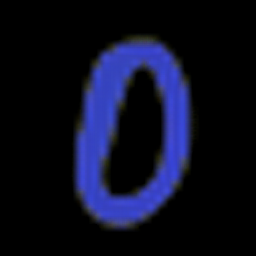} \\
    \includegraphics[width=0.07\textwidth]{figs/Mnist2/visual1/08188-inputsX.png} &
    \includegraphics[width=0.07\textwidth]{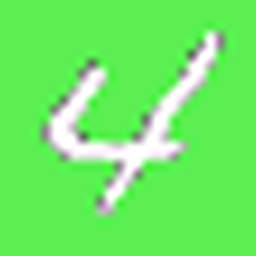} &
    \includegraphics[width=0.07\textwidth]{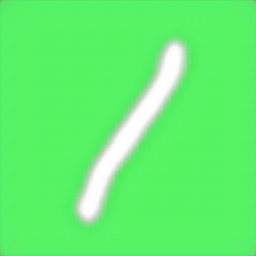} &
    \includegraphics[width=0.07\textwidth]{figs/Mnist2/visual1/08188-inputsY.png} &
    \includegraphics[width=0.07\textwidth]{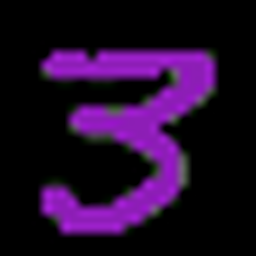} &
    \includegraphics[width=0.07\textwidth]{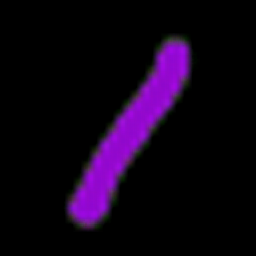} \\
    \includegraphics[width=0.07\textwidth]{figs/Mnist2/visual2/08522-inputsX.png} &
    \includegraphics[width=0.07\textwidth]{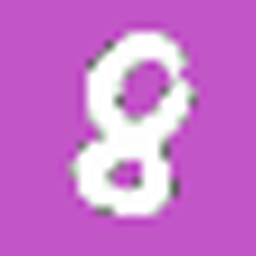} &
    \includegraphics[width=0.07\textwidth]{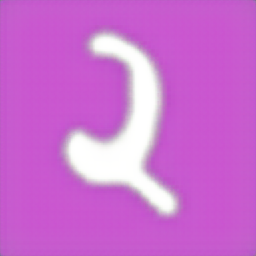} &
    \includegraphics[width=0.07\textwidth]{figs/Mnist2/visual2/08522-inputsY.png} &
    \includegraphics[width=0.07\textwidth]{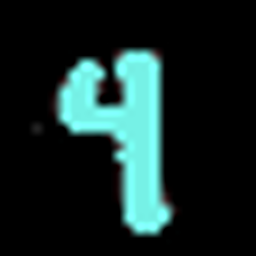} &
    \includegraphics[width=0.07\textwidth]{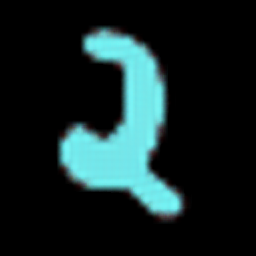} \\
    \includegraphics[width=0.07\textwidth]{figs/Mnist2/visual3/02313-inputsX.png} &
    \includegraphics[width=0.07\textwidth]{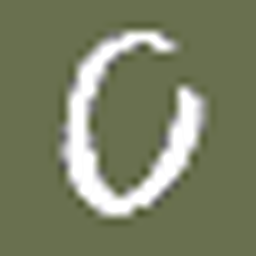} &
    \includegraphics[width=0.07\textwidth]{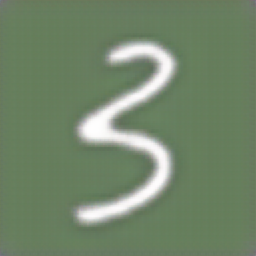} &
    \includegraphics[width=0.07\textwidth]{figs/Mnist2/visual3/02313-inputsY.png} &
    \includegraphics[width=0.07\textwidth]{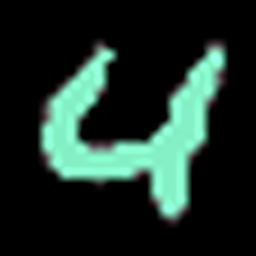} &
    \includegraphics[width=0.07\textwidth]{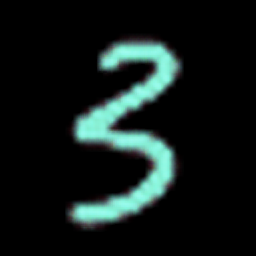} \\
    \includegraphics[width=0.07\textwidth]{figs/Mnist2/visual4/08348-inputsX.png} &
    \includegraphics[width=0.07\textwidth]{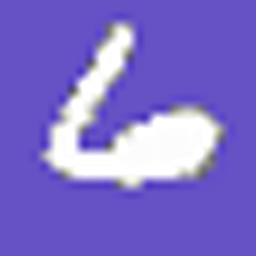} &
    \includegraphics[width=0.07\textwidth]{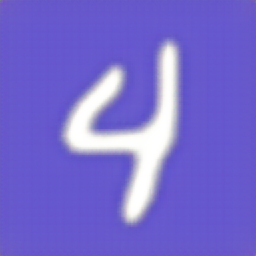} &
    \includegraphics[width=0.07\textwidth]{figs/Mnist2/visual4/08348-inputsY.png} &
    \includegraphics[width=0.07\textwidth]{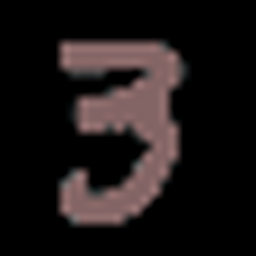} &
    \includegraphics[width=0.07\textwidth]{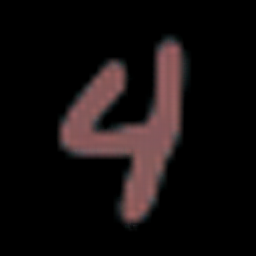} \\
    \includegraphics[width=0.07\textwidth]{figs/Mnist2/visual5/04178-inputsX.png} &
    \includegraphics[width=0.07\textwidth]{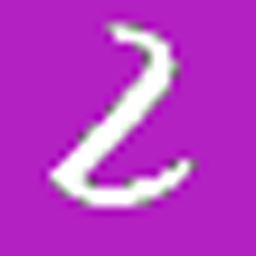} &
    \includegraphics[width=0.07\textwidth]{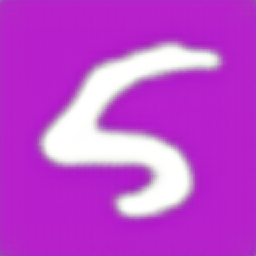} &
    \includegraphics[width=0.07\textwidth]{figs/Mnist2/visual5/04178-inputsY.png} &
    \includegraphics[width=0.07\textwidth]{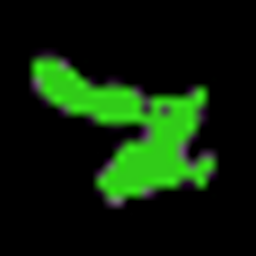} &
    \includegraphics[width=0.07\textwidth]{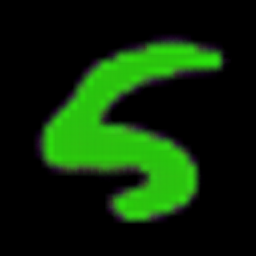} \\
    \includegraphics[width=0.07\textwidth]{figs/Mnist2/visual6/04140-inputsX.png} &
    \includegraphics[width=0.07\textwidth]{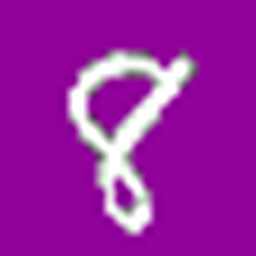} &
    \includegraphics[width=0.07\textwidth]{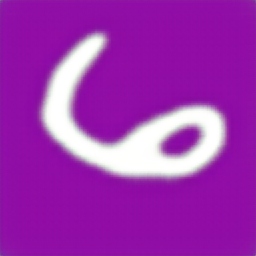} &
    \includegraphics[width=0.07\textwidth]{figs/Mnist2/visual6/04140-inputsY.png} &
    \includegraphics[width=0.07\textwidth]{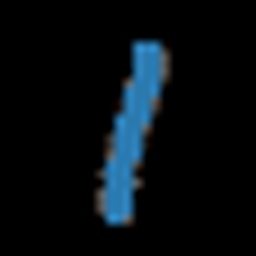} &
    \includegraphics[width=0.07\textwidth]{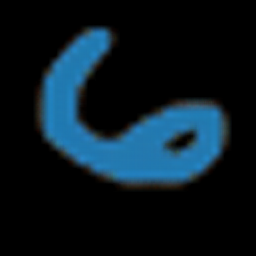} \\
    \includegraphics[width=0.07\textwidth]{figs/Mnist2/visual7/00001-inputsX.png} &
    \includegraphics[width=0.07\textwidth]{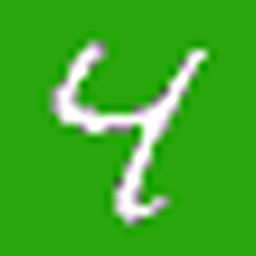} &
    \includegraphics[width=0.07\textwidth]{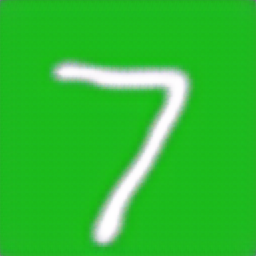} &
    \includegraphics[width=0.07\textwidth]{figs/Mnist2/visual7/00001-inputsY.png} &
    \includegraphics[width=0.07\textwidth]{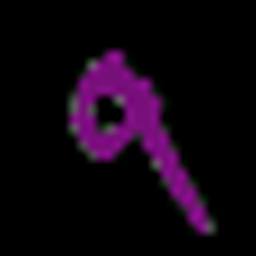} &
    \includegraphics[width=0.07\textwidth]{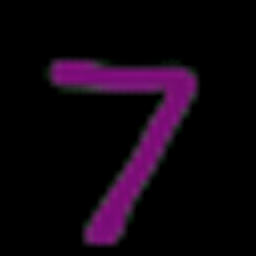} \\
    \includegraphics[width=0.07\textwidth]{figs/Mnist2/visual8/04160-inputsX.png} &
    \includegraphics[width=0.07\textwidth]{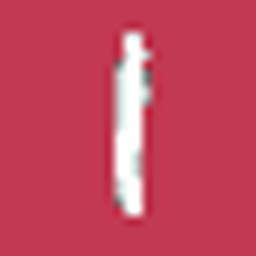} &
    \includegraphics[width=0.07\textwidth]{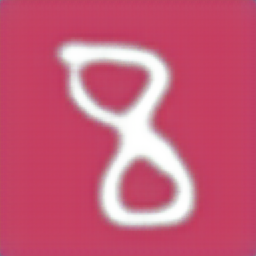} &
    \includegraphics[width=0.07\textwidth]{figs/Mnist2/visual8/04160-inputsY.png} &
    \includegraphics[width=0.07\textwidth]{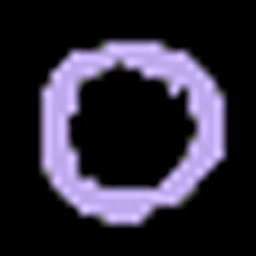} &
    \includegraphics[width=0.07\textwidth]{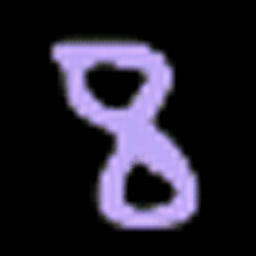} \\
    \includegraphics[width=0.07\textwidth]{figs/Mnist2/visual9/00059-inputsX.png} &
    \includegraphics[width=0.07\textwidth]{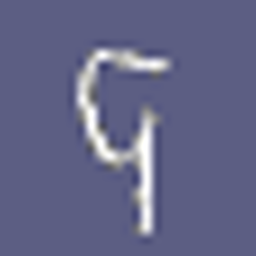} &
    \includegraphics[width=0.07\textwidth]{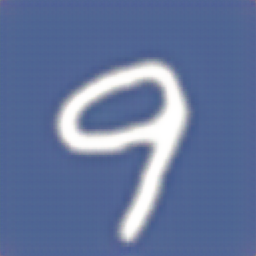} &
    \includegraphics[width=0.07\textwidth]{figs/Mnist2/visual9/00059-inputsY.png} &
    \includegraphics[width=0.07\textwidth]{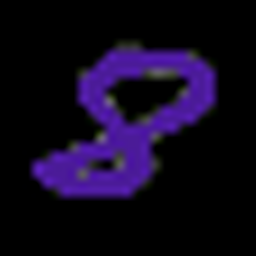} &
    \includegraphics[width=0.07\textwidth]{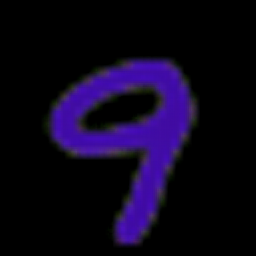} \\
\bottomrule
    \end{tabular}
    \caption{Visual analogies generated by IIAE, synthesizing shared representation from the query and exclusive representations from the reference in each domain.}
    \label{tab:VisAnalogy_MNIST}
    \end{center}
    \vspace{-1cm}
\end{table}

\subsubsection{Cars \citep{reed2015deep}}
In this section, we compare IIAE with CdDN \citep{gonzalez-garcia2018NeurIPS} with Cars dataset. We present additional samples of image translation with IIAE in table \ref{tab:I2I_Cars} and samples from CdDN in table \ref{tab:I2I_Cars_CdDN} with the same input images. Furthermore, we generate visual analogies using IIAE which are presented in table \ref{tab:VisAnalogy_Cars}. To achieve the result of table \ref{tab:I2I_Cars_CdDN} without pretrained model not available, we trained CdDN for Cars dataset (the version with 23 different views) using the code and following the hyperparameter settings released by \citep{gonzalez-garcia2018NeurIPS}. Tables \ref{tab:I2I_Cars} and \ref{tab:I2I_Cars_CdDN} shows that IIAE achieves not only better sample quality but also better disentanglement. In each row of table \ref{tab:I2I_Cars_CdDN}, The content of the given car exposed dependency on the exclusive representation; The details of car such as shape or color varies depending on if exclusive representation is sampled from its prior distribution (the second, third, fourth and seventh, eighth, ninth columns) or extracted from ground-truth pair (the fifth and tenth columns). In table \ref{tab:VisAnalogy_Cars}, we show the query (the first column), 2 references with different orientation (the sencond and fourth columns), and two synthesized images (the third and fifth columns). Queries are sources of shared representation, which is car identity, whereas references are sources of exclusive representations, which are variations in orientation. We present only the analogy of $Y$ domain, since factors of variation only exists in $Y$.

\begin{table}[H]
    \begin{center}
    \begin{tabular}{ cccccccccccc }
    \toprule
    \multicolumn{5}{c}{\textbf{ $X \rightarrow Y$ }} & \multicolumn{5}{c}{\textbf{ $Y \rightarrow X$ }} \\
    \cmidrule(lr){1-5}\cmidrule(lr){6-10}
    \textbf{Input} & \multicolumn{4}{c}{\textbf{ Outputs w/ different $z^{y}$ }} & \textbf{Input} &  \multicolumn{4}{c}{\textbf{ Outputs w/ different $z^{x}$ }} \\
    \cmidrule(lr){1-1}\cmidrule(lr){2-5}\cmidrule(lr){6-6}\cmidrule(lr){7-10}
    x & \multicolumn{3}{c}{$z^{y}_{1},z^{y}_{2},z^{y}_{3} \sim p(z^{y})$} & $\mu^{y}$ & y & \multicolumn{3}{c}{$z^{x}_{1},z^{x}_{2},z^{x}_{3} \sim p(z^{x})$}  & $\mu^{x}$\\
    \cmidrule(lr){2-4}\cmidrule(lr){7-9}
    \includegraphics*[width=0.07\textwidth, viewport=48 64 216 192]{figs/Cars/visual1/159-inputsX.png} &
    \includegraphics*[width=0.07\textwidth, viewport=48 64 216 192]{figs/Cars/visual1/159-outputsYp.png} &
    \includegraphics*[width=0.07\textwidth, viewport=20 64 236 192]{figs/Cars/visual1/159-outputsYpp.png} &
    \includegraphics*[width=0.07\textwidth, viewport=20 64 236 192]{figs/Cars/visual1/159-outputsYppp.png} &
    \includegraphics*[width=0.07\textwidth, viewport=20 64 236 192]{figs/Cars/visual1/159-outputsY_fromX.png} &
    \includegraphics*[width=0.07\textwidth, viewport=20 64 236 192]{figs/Cars/visual1/159-inputsY.png} &
    \includegraphics*[width=0.07\textwidth, viewport=48 64 218 192]{figs/Cars/visual1/159-outputsXp.png} &
    \includegraphics*[width=0.07\textwidth, viewport=48 64 218 192]{figs/Cars/visual1/159-outputsXpp.png} &
    \includegraphics*[width=0.07\textwidth, viewport=48 64 218 192]{figs/Cars/visual1/159-outputsXppp.png} &
    \includegraphics*[width=0.07\textwidth, viewport=48 64 218 192]{figs/Cars/visual1/159-outputsX_fromY.png} \\
    \includegraphics*[width=0.07\textwidth, viewport=48 64 218 192]{figs/Cars/visual2/198-inputsX.png} &
    \includegraphics*[width=0.07\textwidth, viewport=20 64 236 192]{figs/Cars/visual2/198-outputsYp.png} &
    \includegraphics*[width=0.07\textwidth, viewport=32 64 224 192]{figs/Cars/visual2/198-outputsYpp.png} &
    \includegraphics*[width=0.07\textwidth, viewport=32 64 224 192]{figs/Cars/visual2/198-outputsYppp.png} &
    \includegraphics*[width=0.07\textwidth, viewport=32 64 224 192]{figs/Cars/visual2/198-outputsY_fromX.png} &
    \includegraphics*[width=0.07\textwidth, viewport=32 64 224 192]{figs/Cars/visual2/198-inputsY.png} &
    \includegraphics*[width=0.07\textwidth, viewport=48 64 218 192]{figs/Cars/visual2/198-outputsXp.png} &
    \includegraphics*[width=0.07\textwidth, viewport=48 64 218 192]{figs/Cars/visual2/198-outputsXpp.png} &
    \includegraphics*[width=0.07\textwidth, viewport=48 64 218 192]{figs/Cars/visual2/198-outputsXppp.png} &
    \includegraphics*[width=0.07\textwidth, viewport=48 64 218 192]{figs/Cars/visual2/198-outputsX_fromY.png} \\
    \includegraphics*[width=0.07\textwidth, viewport=56 64 200 192]{figs/Cars/visual3/057-inputsX.png} &
    \includegraphics*[width=0.07\textwidth, viewport=16 64 240 192]{figs/Cars/visual3/057-outputsYp.png} &
    \includegraphics*[width=0.07\textwidth, viewport=16 64 240 192]{figs/Cars/visual3/057-outputsYpp.png} &
    \includegraphics*[width=0.07\textwidth, viewport=32 64 224 192]{figs/Cars/visual3/057-outputsYppp.png} &
    \includegraphics*[width=0.07\textwidth, viewport=48 64 218 192]{figs/Cars/visual3/057-outputsY_fromX.png} &
    \includegraphics*[width=0.07\textwidth, viewport=48 64 218 192]{figs/Cars/visual3/057-inputsY.png} &
    \includegraphics*[width=0.07\textwidth, viewport=56 64 200 192]{figs/Cars/visual3/057-outputsXp.png} &
    \includegraphics*[width=0.07\textwidth, viewport=56 64 200 192]{figs/Cars/visual3/057-outputsXpp.png} &
    \includegraphics*[width=0.07\textwidth, viewport=56 64 200 192]{figs/Cars/visual3/057-outputsXppp.png} &
    \includegraphics*[width=0.07\textwidth, viewport=56 64 200 192]{figs/Cars/visual3/057-outputsX_fromY.png} \\
    \includegraphics*[width=0.07\textwidth, viewport=48 64 218 192]{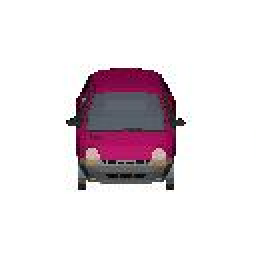} &
    \includegraphics*[width=0.07\textwidth, viewport=16 64 240 192]{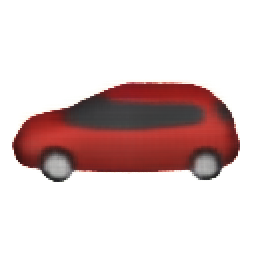} &
    \includegraphics*[width=0.07\textwidth, viewport=16 64 240 192]{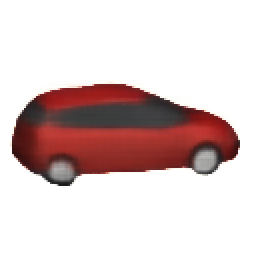} &
    \includegraphics*[width=0.07\textwidth, viewport=32 64 224 192]{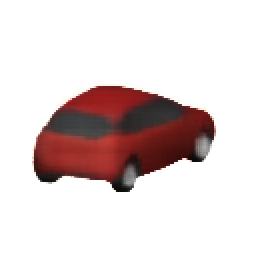} &
    \includegraphics*[width=0.07\textwidth, viewport=20 64 236 192]{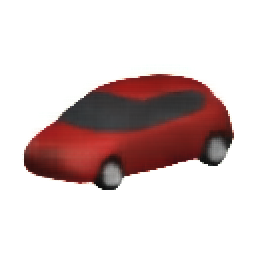} &
    \includegraphics*[width=0.07\textwidth, viewport=20 64 236 192]{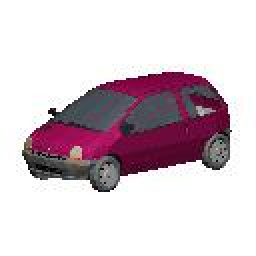} &
    \includegraphics*[width=0.07\textwidth, viewport=48 64 218 192]{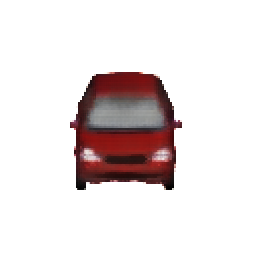} &
    \includegraphics*[width=0.07\textwidth, viewport=48 64 218 192]{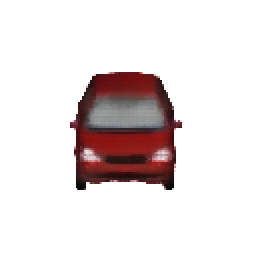} &
    \includegraphics*[width=0.07\textwidth, viewport=48 64 218 192]{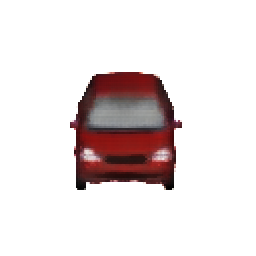} &
    \includegraphics*[width=0.07\textwidth, viewport=48 64 218 192]{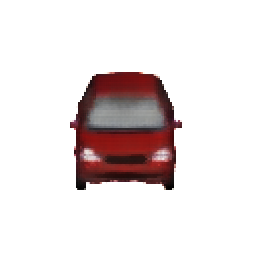} \\
    \includegraphics*[width=0.07\textwidth, viewport=48 64 218 192]{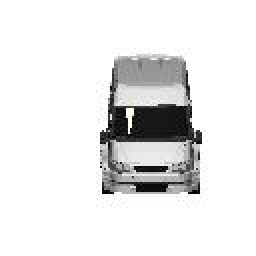} &
    \includegraphics*[width=0.07\textwidth, viewport=16 64 240 192]{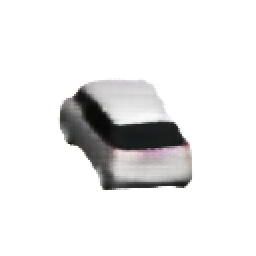} &
    \includegraphics*[width=0.07\textwidth, viewport=16 64 240 192]{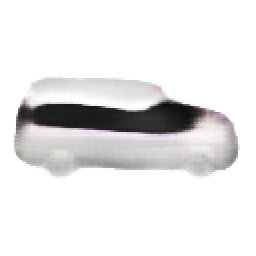} &
    \includegraphics*[width=0.07\textwidth, viewport=32 64 224 196]{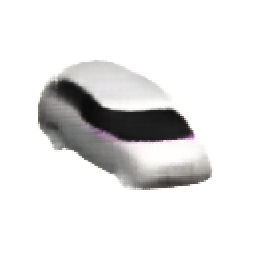} &
    \includegraphics*[width=0.07\textwidth, viewport=20 64 236 192]{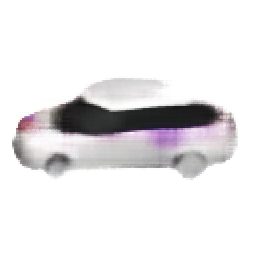} &
    \includegraphics*[width=0.07\textwidth, viewport=20 32 236 224]{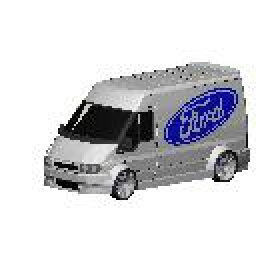} &
    \includegraphics*[width=0.07\textwidth, viewport=48 64 218 192]{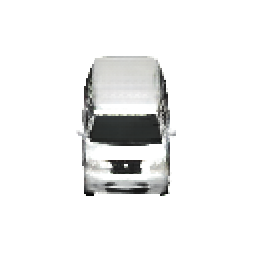} &
    \includegraphics*[width=0.07\textwidth, viewport=48 64 218 192]{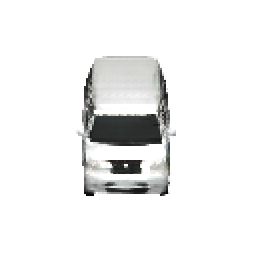} &
    \includegraphics*[width=0.07\textwidth, viewport=48 64 218 192]{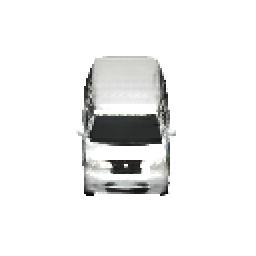} &
    \includegraphics*[width=0.07\textwidth, viewport=48 64 218 192]{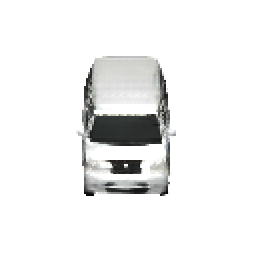} \\
    \includegraphics*[width=0.07\textwidth, viewport=48 64 218 192]{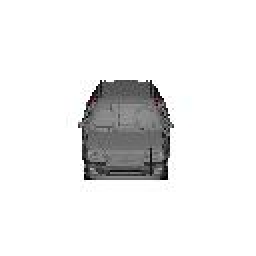} &
    \includegraphics*[width=0.07\textwidth, viewport=16 64 240 192]{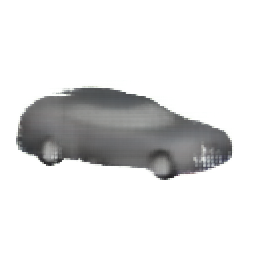} &
    \includegraphics*[width=0.07\textwidth, viewport=28 64 228 192]{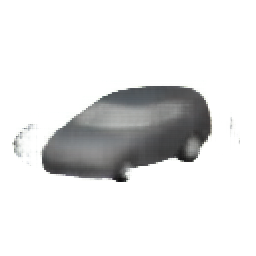} &
    \includegraphics*[width=0.07\textwidth, viewport=20 64 236 192]{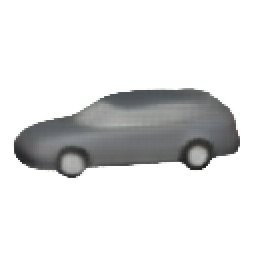} &
    \includegraphics*[width=0.07\textwidth, viewport=56 64 200 192]{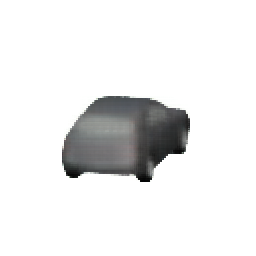} &
    \includegraphics*[width=0.07\textwidth, viewport=56 64 200 192]{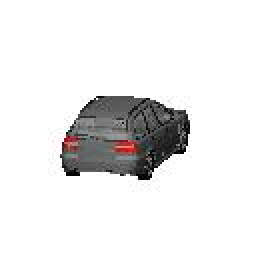} &
    \includegraphics*[width=0.07\textwidth, viewport=48 64 218 192]{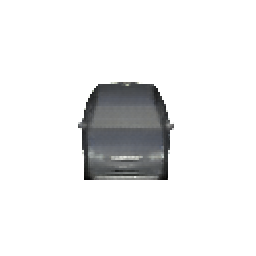} &
    \includegraphics*[width=0.07\textwidth, viewport=48 64 218 192]{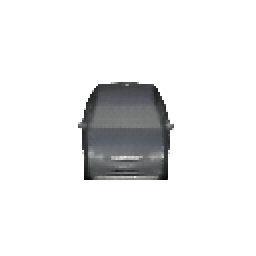} &
    \includegraphics*[width=0.07\textwidth, viewport=48 64 218 192]{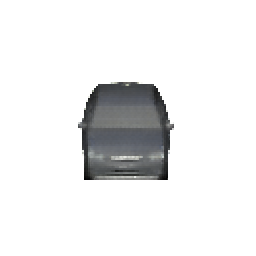} &
    \includegraphics*[width=0.07\textwidth, viewport=48 64 218 192]{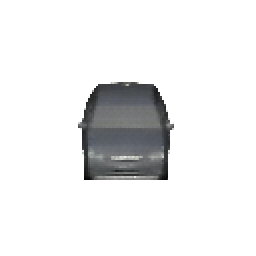} \\
    \includegraphics*[width=0.07\textwidth, viewport=48 64 218 192]{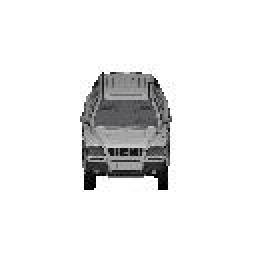} &
    \includegraphics*[width=0.07\textwidth, viewport=32 64 224 192]{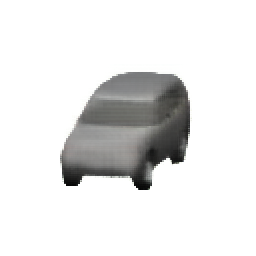} &
    \includegraphics*[width=0.07\textwidth, viewport=28 64 228 192]{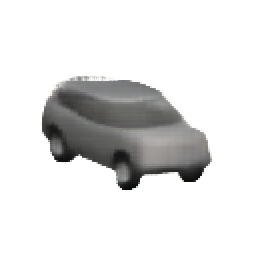} &
    \includegraphics*[width=0.07\textwidth, viewport=20 64 240 192]{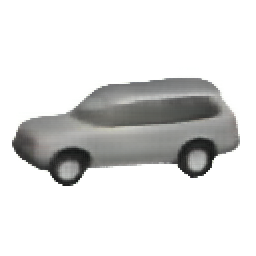} &
    \includegraphics*[width=0.07\textwidth, viewport=40 64 226 192]{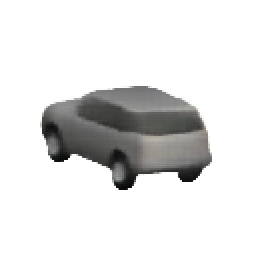} &
    \includegraphics*[width=0.07\textwidth, viewport=40 64 226 192]{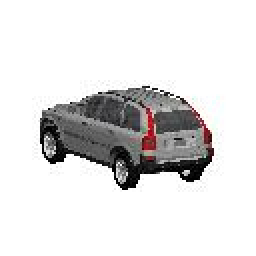} &
    \includegraphics*[width=0.07\textwidth, viewport=48 64 218 192]{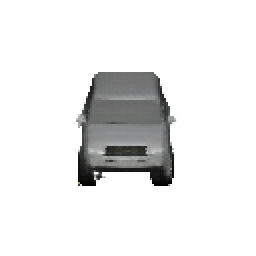} &
    \includegraphics*[width=0.07\textwidth, viewport=48 64 218 192]{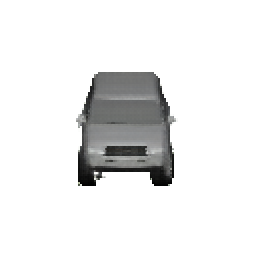} &
    \includegraphics*[width=0.07\textwidth, viewport=48 64 218 192]{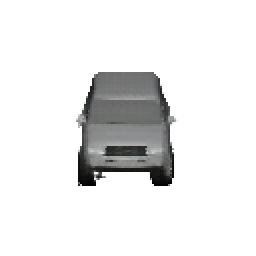} &
    \includegraphics*[width=0.07\textwidth, viewport=48 64 218 192]{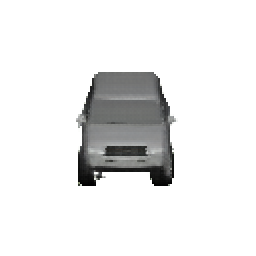} \\    \bottomrule
    \end{tabular}
    \caption{Additional cross-domain translation results in Cars \citep{reed2015deep} generated by IIAE. In Cars, domain-specific factors only exists in $Y$, views in 23 different degrees, while the shared factor is the identity of car.
    }
    \label{tab:I2I_Cars}
    \end{center}
    \vspace{-1cm}
\end{table}

\begin{table}[H]
    \begin{center}
    \begin{tabular}{ cccccccccc }
    \toprule
    \multicolumn{5}{c}{\textbf{ $X \rightarrow Y$ }} & \multicolumn{5}{c}{\textbf{ $Y \rightarrow X$ }} \\
    \cmidrule(lr){1-5}\cmidrule(lr){6-10}
    \textbf{Input} & \multicolumn{4}{c}{\textbf{ Outputs w/ different $z^{y}$ }} & \textbf{Input} &  \multicolumn{4}{c}{\textbf{ Outputs w/ different $z^{x}$ }} \\
    \cmidrule(lr){1-1}\cmidrule(lr){2-5}\cmidrule(lr){6-6}\cmidrule(lr){7-10}
    x & \multicolumn{3}{c}{$z^{y}_{1},z^{y}_{2},z^{y}_{3} \sim p(z^{y})$} & $\mu^{y}$ & y & \multicolumn{3}{c}{$z^{x}_{1},z^{x}_{2},z^{x}_{3} \sim p(z^{x})$}  & $\mu^{x}$\\
    \cmidrule(lr){2-4}\cmidrule(lr){7-9}
    \includegraphics*[width=0.07\textwidth, viewport=48 64 216 192]{figs/Cars/visual1/159-inputsX.png} &
    \includegraphics*[width=0.07\textwidth, viewport=16 64 240 192]{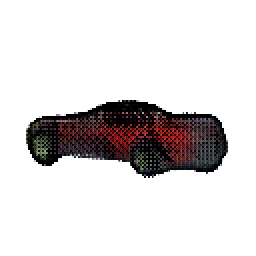} &
    \includegraphics*[width=0.07\textwidth, viewport=32 64 224 192]{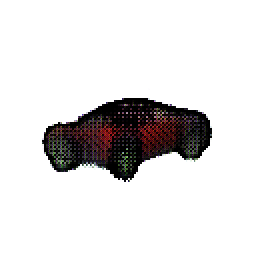} &
    \includegraphics*[width=0.07\textwidth, viewport=16 64 240 192]{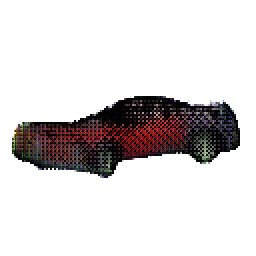} &
    \includegraphics*[width=0.07\textwidth, viewport=20 64 236 192]{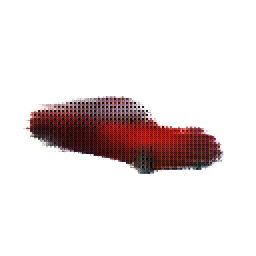} &
    \includegraphics*[width=0.07\textwidth, viewport=20 64 236 192]{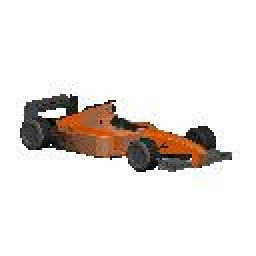} &
    \includegraphics*[width=0.07\textwidth, viewport=48 64 218 192]{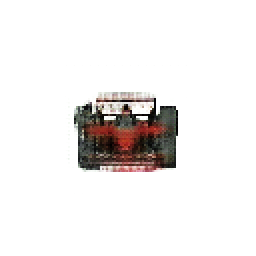} &
    \includegraphics*[width=0.07\textwidth, viewport=48 64 218 192]{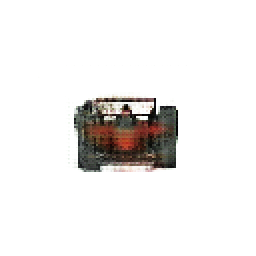} &
    \includegraphics*[width=0.07\textwidth, viewport=48 64 218 192]{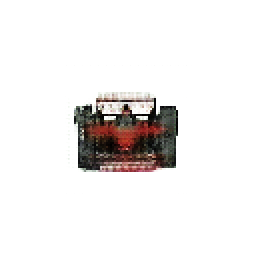} &
    \includegraphics*[width=0.07\textwidth, viewport=48 64 218 192]{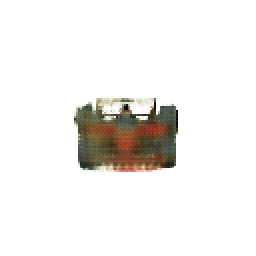} \\
    \includegraphics*[width=0.07\textwidth, viewport=48 64 218 192]{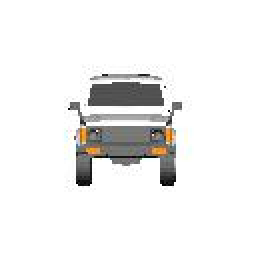} &
    \includegraphics*[width=0.07\textwidth, viewport=20 64 236 192]{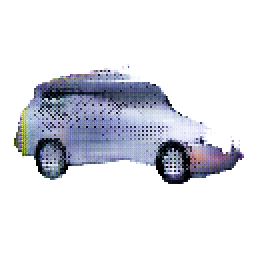} &
    \includegraphics*[width=0.07\textwidth, viewport=32 64 224 192]{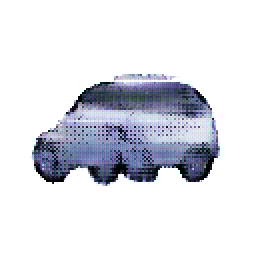} &
    \includegraphics*[width=0.07\textwidth, viewport=32 64 224 192]{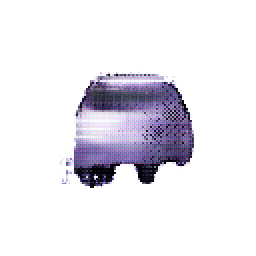} &
    \includegraphics*[width=0.07\textwidth, viewport=32 64 224 192]{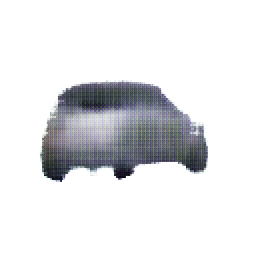} &
    \includegraphics*[width=0.07\textwidth, viewport=32 64 224 192]{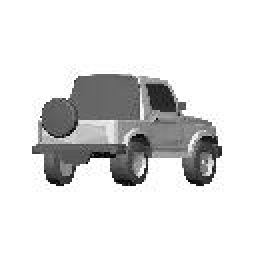} &
    \includegraphics*[width=0.07\textwidth, viewport=48 64 218 192]{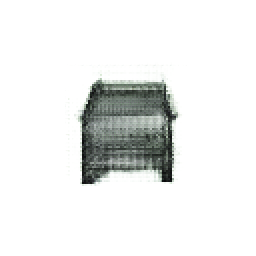} &
    \includegraphics*[width=0.07\textwidth, viewport=48 64 218 192]{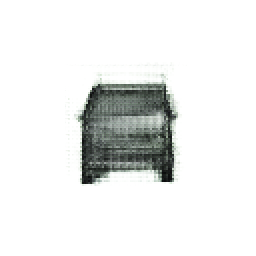} &
    \includegraphics*[width=0.07\textwidth, viewport=48 64 218 192]{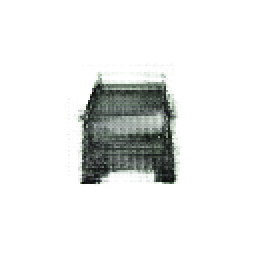} &
    \includegraphics*[width=0.07\textwidth, viewport=48 64 218 192]{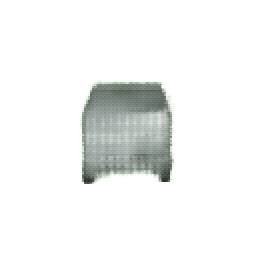} \\
    \includegraphics*[width=0.07\textwidth, viewport=56 64 200 192]{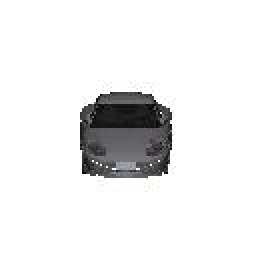} &
    \includegraphics*[width=0.07\textwidth, viewport=16 64 240 192]{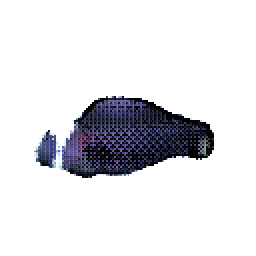} &
    \includegraphics*[width=0.07\textwidth, viewport=16 64 240 192]{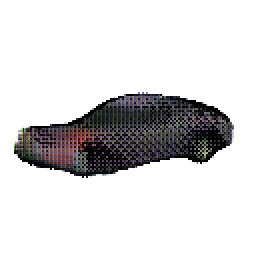} &
    \includegraphics*[width=0.07\textwidth, viewport=16 64 240 192]{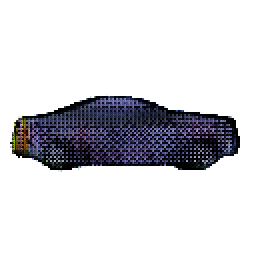} &
    \includegraphics*[width=0.07\textwidth, viewport=32 64 224 192]{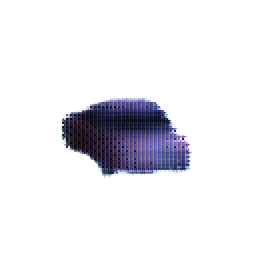} &
    \includegraphics*[width=0.07\textwidth, viewport=48 64 218 192]{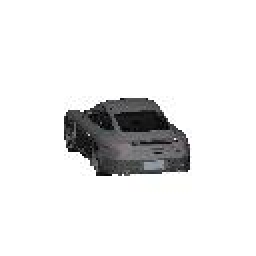} &
    \includegraphics*[width=0.07\textwidth, viewport=56 64 200 192]{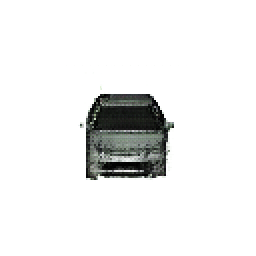} &
    \includegraphics*[width=0.07\textwidth, viewport=56 64 200 192]{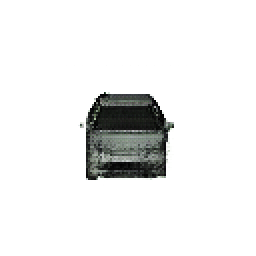} &
    \includegraphics*[width=0.07\textwidth, viewport=56 64 200 192]{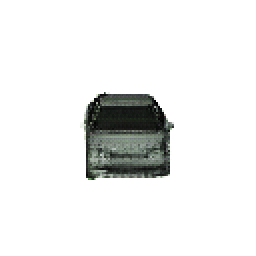} &
    \includegraphics*[width=0.07\textwidth, viewport=56 64 200 192]{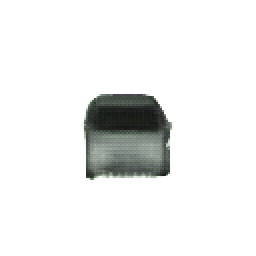} \\
    \includegraphics*[width=0.07\textwidth, viewport=48 64 218 192]{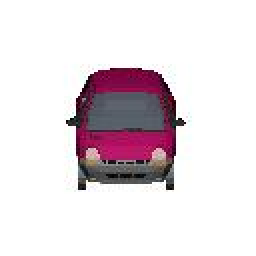} &
    \includegraphics*[width=0.07\textwidth, viewport=16 64 240 192]{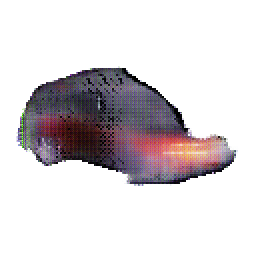} &
    \includegraphics*[width=0.07\textwidth, viewport=16 64 240 192]{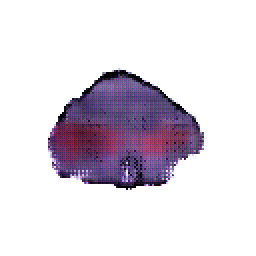} &
    \includegraphics*[width=0.07\textwidth, viewport=16 64 240 192]{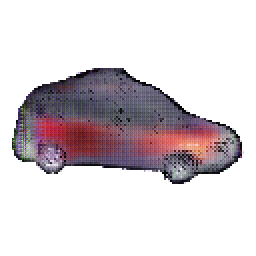} &
    \includegraphics*[width=0.07\textwidth, viewport=20 64 236 192]{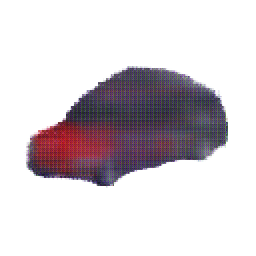} &
    \includegraphics*[width=0.07\textwidth, viewport=20 64 236 192]{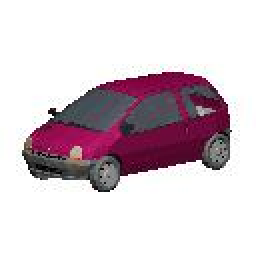} &
    \includegraphics*[width=0.07\textwidth, viewport=48 64 218 192]{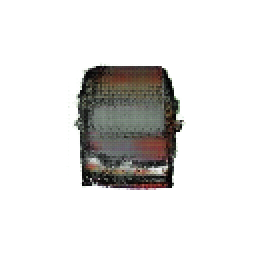} &
    \includegraphics*[width=0.07\textwidth, viewport=48 64 218 192]{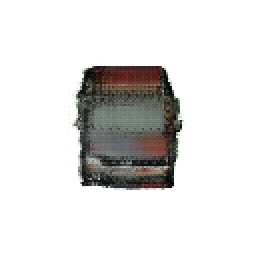} &
    \includegraphics*[width=0.07\textwidth, viewport=48 64 218 192]{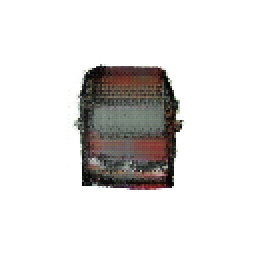} &
    \includegraphics*[width=0.07\textwidth, viewport=48 64 218 192]{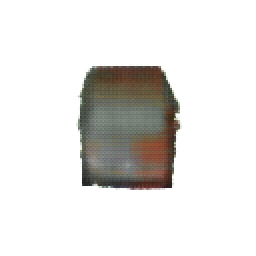} \\
    \includegraphics*[width=0.07\textwidth, viewport=48 64 218 192]{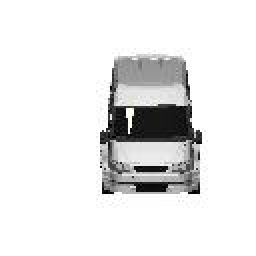} &
    \includegraphics*[width=0.07\textwidth, viewport=16 64 240 192]{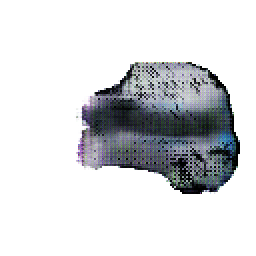} &
    \includegraphics*[width=0.07\textwidth, viewport=16 64 240 192]{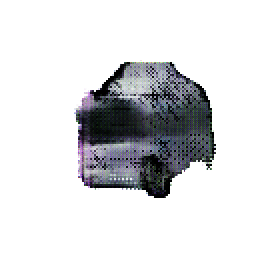} &
    \includegraphics*[width=0.07\textwidth, viewport=16 64 240 192]{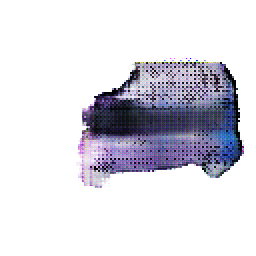} &
    \includegraphics*[width=0.07\textwidth, viewport=20 64 236 192]{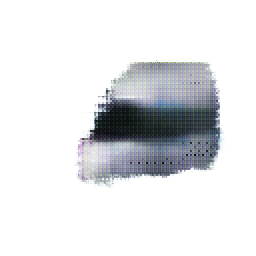} &
    \includegraphics*[width=0.07\textwidth, viewport=20 32 236 224]{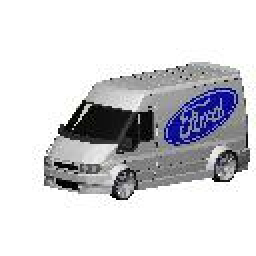} &
    \includegraphics*[width=0.07\textwidth, viewport=48 64 218 192]{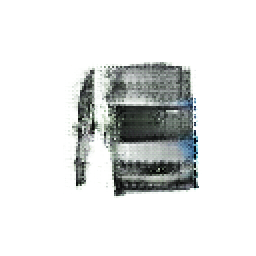} &
    \includegraphics*[width=0.07\textwidth, viewport=48 64 218 192]{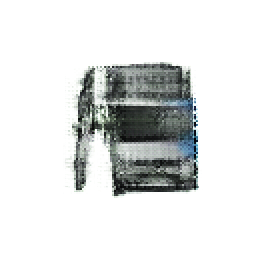} &
    \includegraphics*[width=0.07\textwidth, viewport=48 64 218 192]{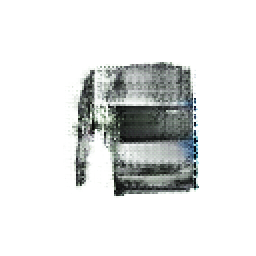} &
    \includegraphics*[width=0.07\textwidth, viewport=48 64 218 192]{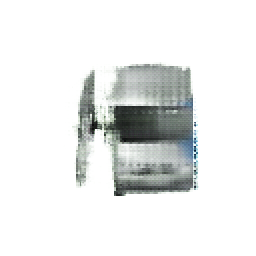} \\
    \includegraphics*[width=0.07\textwidth, viewport=48 64 218 192]{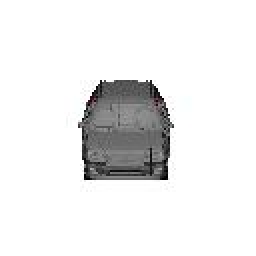} &
    \includegraphics*[width=0.07\textwidth, viewport=16 64 240 192]{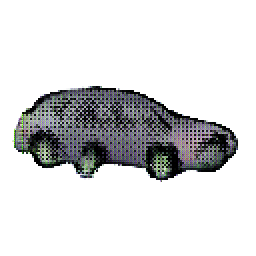} &
    \includegraphics*[width=0.07\textwidth, viewport=28 64 228 192]{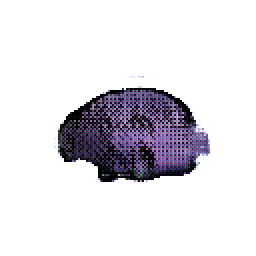} &
    \includegraphics*[width=0.07\textwidth, viewport=28 64 228 192]{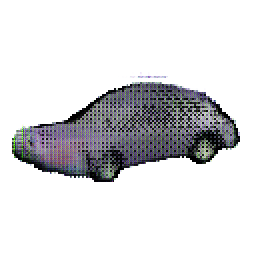} &
    \includegraphics*[width=0.07\textwidth, viewport=56 64 200 192]{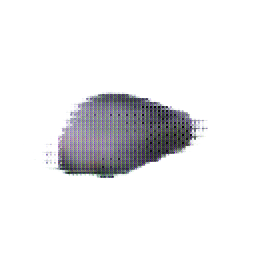} &
    \includegraphics*[width=0.07\textwidth, viewport=56 64 200 192]{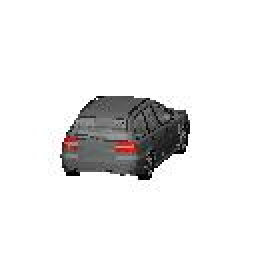} &
    \includegraphics*[width=0.07\textwidth, viewport=48 64 218 192]{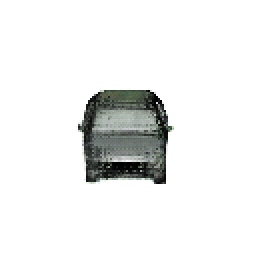} &
    \includegraphics*[width=0.07\textwidth, viewport=48 64 218 192]{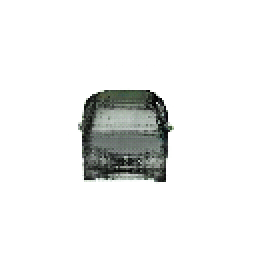} &
    \includegraphics*[width=0.07\textwidth, viewport=48 64 218 192]{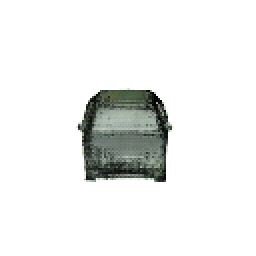} &
    \includegraphics*[width=0.07\textwidth, viewport=48 64 218 192]{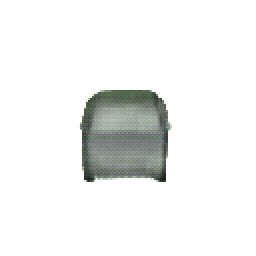} \\
    \includegraphics*[width=0.07\textwidth, viewport=48 64 218 192]{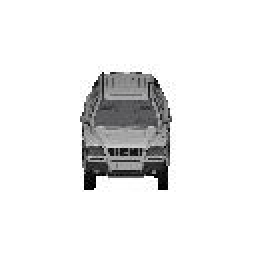} &
    \includegraphics*[width=0.07\textwidth, viewport=32 64 224 192]{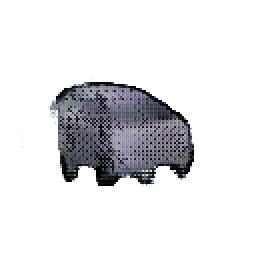} &
    \includegraphics*[width=0.07\textwidth, viewport=16 64 240 192]{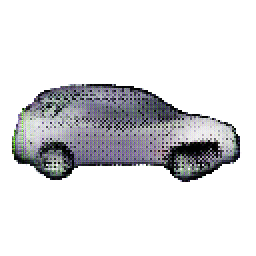} &
    \includegraphics*[width=0.07\textwidth, viewport=16 64 240 192]{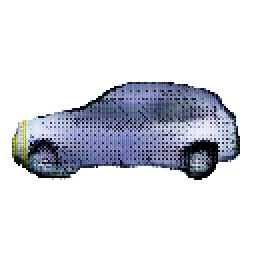} &
    \includegraphics*[width=0.07\textwidth, viewport=40 64 226 192]{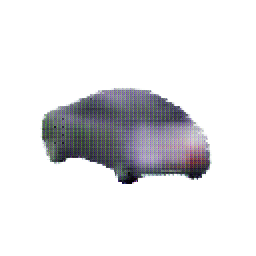} &
    \includegraphics*[width=0.07\textwidth, viewport=40 64 226 192]{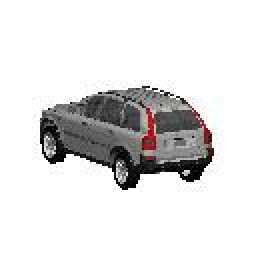} &
    \includegraphics*[width=0.07\textwidth, viewport=48 64 218 192]{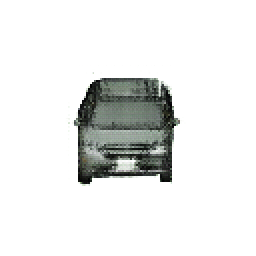} &
    \includegraphics*[width=0.07\textwidth, viewport=48 64 218 192]{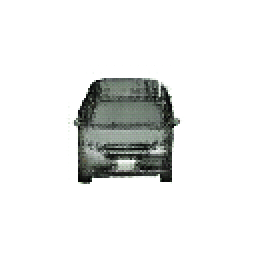} &
    \includegraphics*[width=0.07\textwidth, viewport=48 64 218 192]{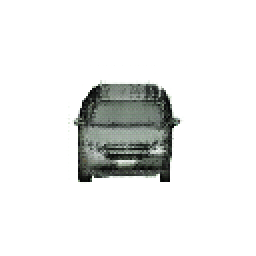} &
    \includegraphics*[width=0.07\textwidth, viewport=48 64 218 192]{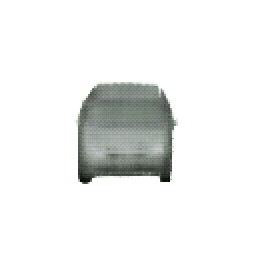} \\    \bottomrule
    \end{tabular}
    \caption{Cross-domain translation results in Cars \citep{reed2015deep} generated by CdDN \citep{gonzalez-garcia2018NeurIPS}.
    }
    \label{tab:I2I_Cars_CdDN}
    \end{center}
    \vspace{-1cm}
\end{table}

\begin{table}[H]
    \begin{center}
    \begin{tabular}{ ccccc }
    \toprule
    \multicolumn{5}{c}{\textbf{ $Y$ }} \\
    \cmidrule(lr){1-5}
    \textbf{query} & \textbf{reference1} & \textbf{Output1} & \textbf{reference2} & \textbf{Output2}\\
    \includegraphics*[width=0.1\textwidth, viewport=16 64 240 192]{figs/Cars/visual1/159-inputsY.png} &
    \includegraphics*[width=0.1\textwidth, viewport=12 64 244 192]{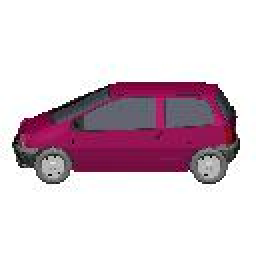} &
    \includegraphics*[width=0.1\textwidth, viewport=16 64 240 192]{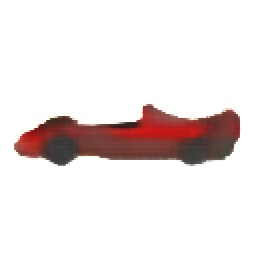} &
    \includegraphics*[width=0.1\textwidth, viewport=32 64 224 192]{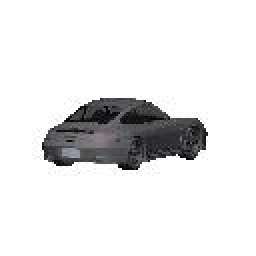} &
    \includegraphics*[width=0.1\textwidth, viewport=32 64 224 192]{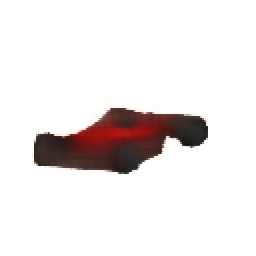} \\
    \includegraphics*[width=0.1\textwidth, viewport=32 64 224 192]{figs/Cars/visual2/198-inputsY.png} &
    \includegraphics*[width=0.1\textwidth, viewport=16 64 248 192]{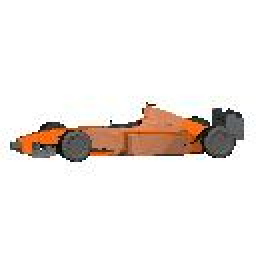} &
    \includegraphics*[width=0.1\textwidth, viewport=16 64 240 192]{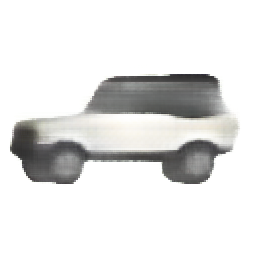} &
    \includegraphics*[width=0.1\textwidth, viewport=16 64 240 192]{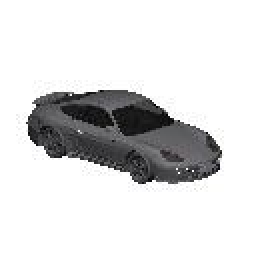} &
    \includegraphics*[width=0.1\textwidth, viewport=16 64 230 192]{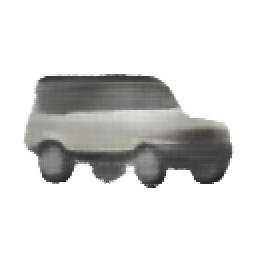} \\
    \includegraphics*[width=0.1\textwidth, viewport=32 64 224 192]{figs/Cars/visual3/057-inputsY.png} &
    \includegraphics*[width=0.1\textwidth, viewport=12 64 244 192]{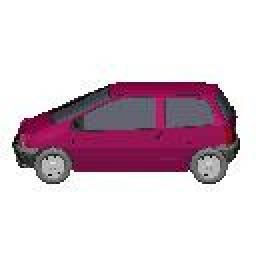} &
    \includegraphics*[width=0.1\textwidth, viewport=16 64 240 192]{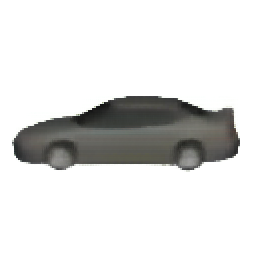} &
    \includegraphics*[width=0.1\textwidth, viewport=24 64 232 192]{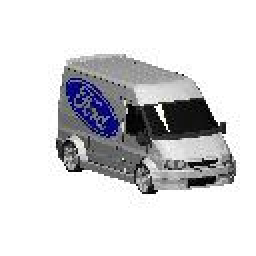} &
    \includegraphics*[width=0.1\textwidth, viewport=32 64 224 192]{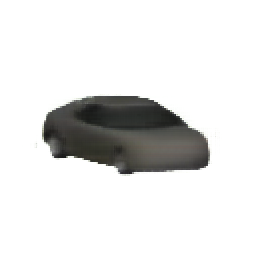} \\
    \includegraphics*[width=0.1\textwidth, viewport=16 64 240 192]{figs/Cars/visual4/417-inputsY.png} &
    \includegraphics*[width=0.1\textwidth, viewport=32 64 224 192]{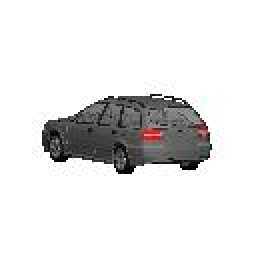} &
    \includegraphics*[width=0.1\textwidth, viewport=32 64 224 192]{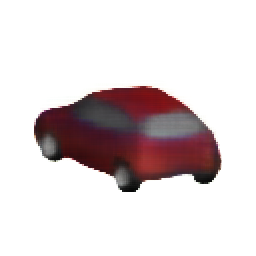} &
    \includegraphics*[width=0.1\textwidth, viewport=32 64 224 192]{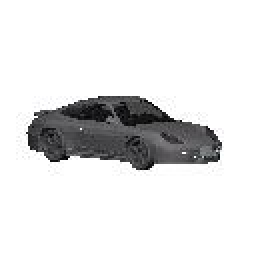} &
    \includegraphics*[width=0.1\textwidth, viewport=24 64 232 192]{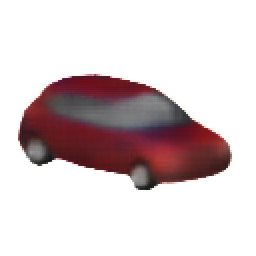} \\
    \includegraphics*[width=0.1\textwidth, viewport=16 48 240 208]{figs/Cars/visual5/325-inputsY.png} &
    \includegraphics*[width=0.1\textwidth, viewport=16 64 240 192]{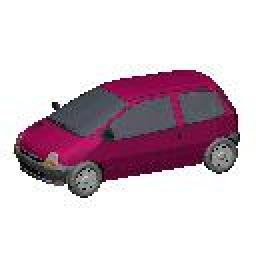} &
    \includegraphics*[width=0.1\textwidth, viewport=16 64 240 192]{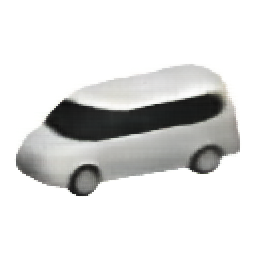} &
    \includegraphics*[width=0.1\textwidth, viewport=40 64 208 192]{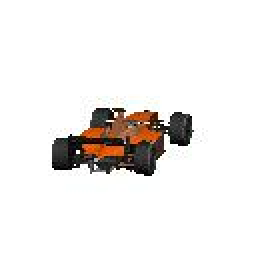} &
    \includegraphics*[width=0.1\textwidth, viewport=40 48 216 208]{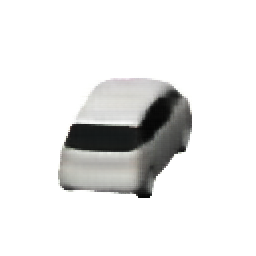} \\
    \includegraphics*[width=0.1\textwidth, viewport=40 64 216 192]{figs/Cars/visual6/519-inputsY.png} &
    \includegraphics*[width=0.1\textwidth, viewport=32 64 224 192]{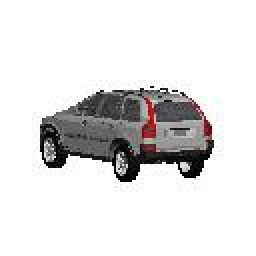} &
    \includegraphics*[width=0.1\textwidth, viewport=40 64 216 192]{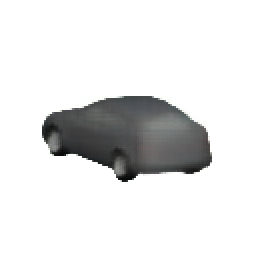} &
    \includegraphics*[width=0.1\textwidth, viewport=10 64 246 192]{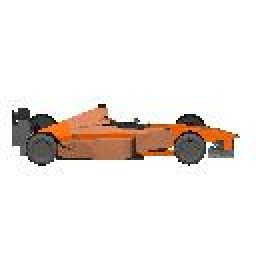} &
    \includegraphics*[width=0.1\textwidth, viewport=12 64 244 192]{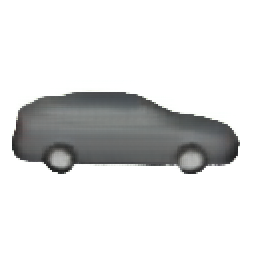} \\
    \includegraphics*[width=0.1\textwidth, viewport=32 64 224 192]{figs/Cars/visual7/608-inputsY.png} &
    \includegraphics*[width=0.1\textwidth, viewport=24 64 232 192]{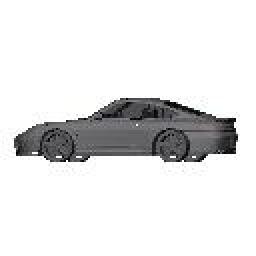} &
    \includegraphics*[width=0.1\textwidth, viewport=16 64 240 192]{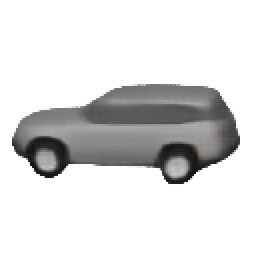} &
    \includegraphics*[width=0.1\textwidth, viewport=8 64 248 192]{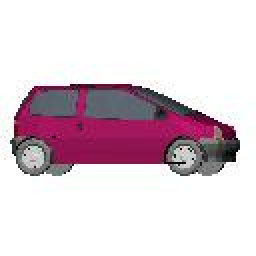} &
    \includegraphics*[width=0.1\textwidth, viewport=16 64 240 192]{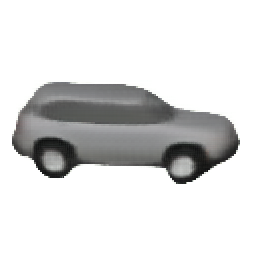} \\
\bottomrule
    \end{tabular}
    \caption{Visual analogies generated by IIAE, synthesizing shared representation from the query and exclusive representations from the reference in $Y$ domain.}
    \label{tab:VisAnalogy_Cars}
    \end{center}
    \vspace{-1cm}
\end{table}
\clearpage
\subsection{Cross-domain retrieval}
\label{appendix:retrieval}
In this section, we visualize the top-3 retrieved images of the cross-domain retrieval task in Facades \citep{10.1007/978-3-642-40602-7_39} and Maps \citep{pix2pix2017} datasets. For each query image, we classify the result as a success only when the ground truth pair of the query is retrieved as the closest one (top-1), failure otherwise. Although IIAE performs close to perfect in this task, there exist a few of failure cases which we present here as well.

\subsubsection{Maps \citep{pix2pix2017}}
\begin{figure}[H]
    \centering
    % \hspace{10mm}
    \begin{subfigure}[b]{0.49\textwidth}
        \captionsetup{justification=raggedright,singlelinecheck=false}
        \caption*{Query(S) \hspace*{0.15cm} GT(M) \hspace*{1.88cm} S $\rightarrow$ M}
        \centering
        \includegraphics*[width=0.18\textwidth, viewport=0 0 600 600]{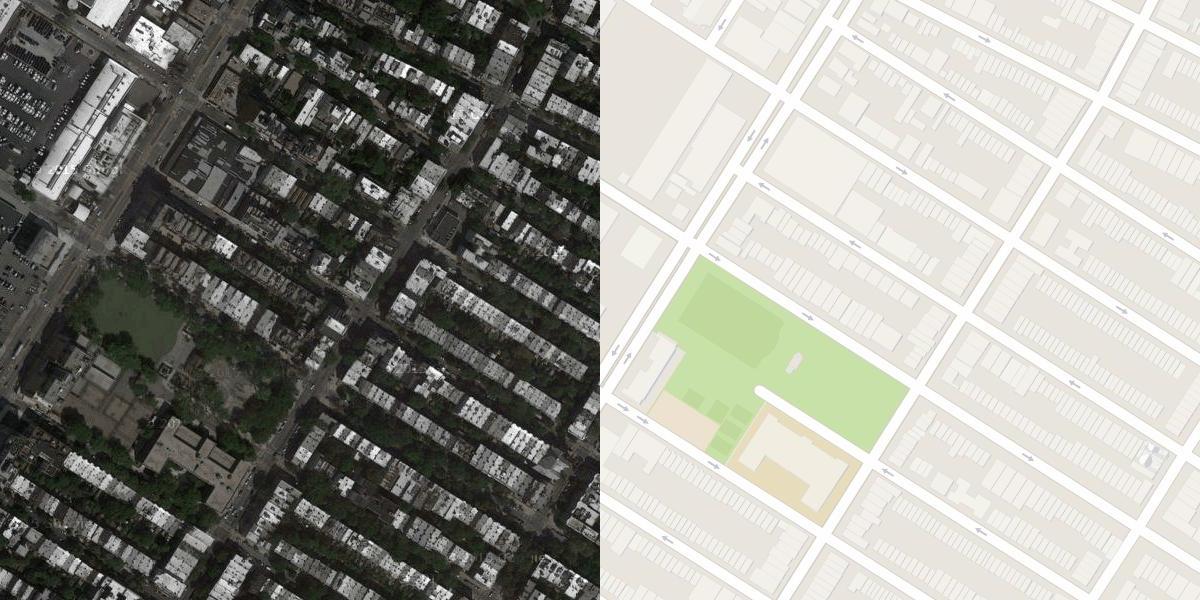}%
        \hspace*{0.02cm}
        \includegraphics*[width=0.18\textwidth, viewport=602 0 1200 600]{figs/Maps/384.jpg}
        \hfill
        \includegraphics*[width=0.18\textwidth, viewport=602 0 1200 600]{figs/Maps/384.jpg}
        \includegraphics*[width=0.18\textwidth, viewport=602 0 1200 600]{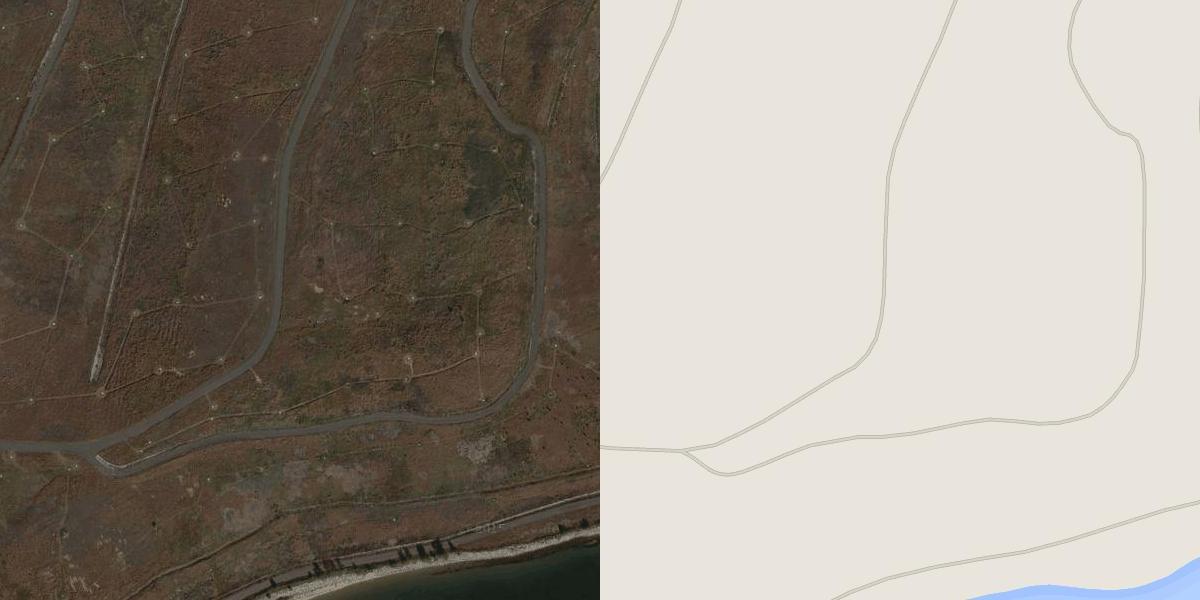}
        \includegraphics*[width=0.18\textwidth, viewport=602 0 1200 600]{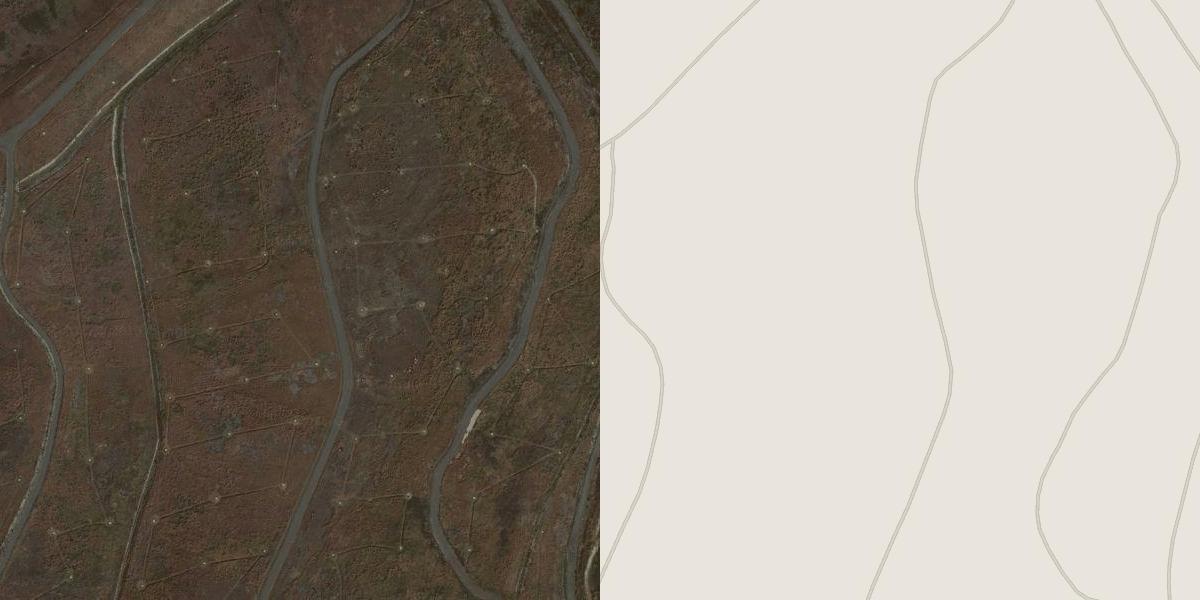}%
        \\
        \includegraphics*[width=0.18\textwidth, viewport=0 0 600 600]{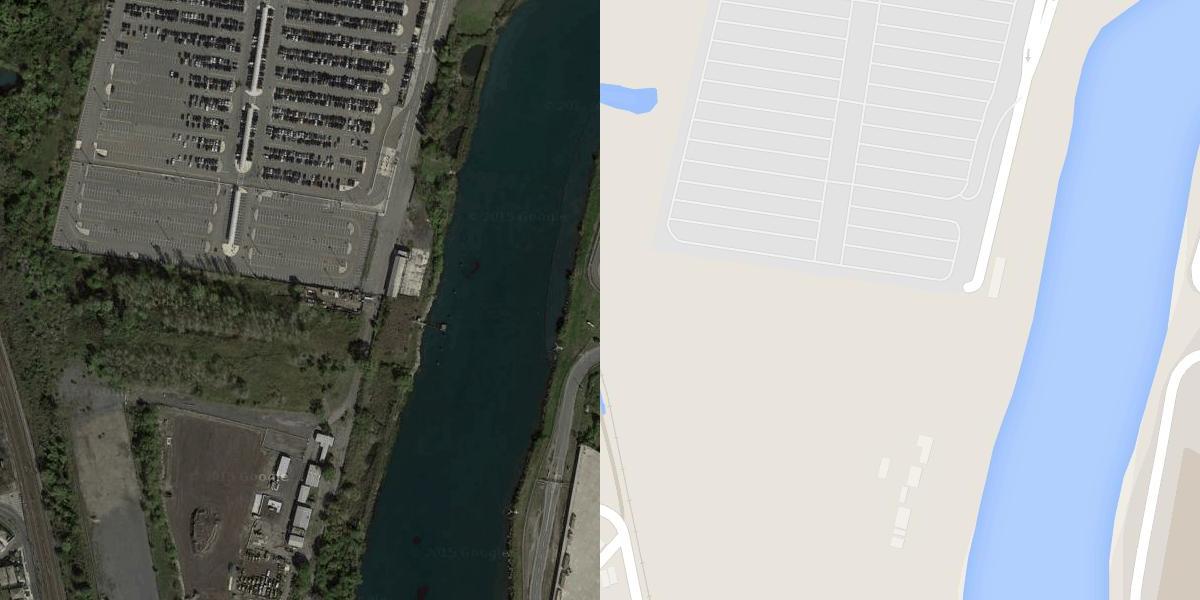}%
        \hspace*{0.02cm}
        \includegraphics*[width=0.18\textwidth, viewport=602 0 1200 600]{figs/Maps/385.jpg}
        \hfill
        \includegraphics*[width=0.18\textwidth, viewport=602 0 1200 600]{figs/Maps/385.jpg}
        \includegraphics*[width=0.18\textwidth, viewport=602 0 1200 600]{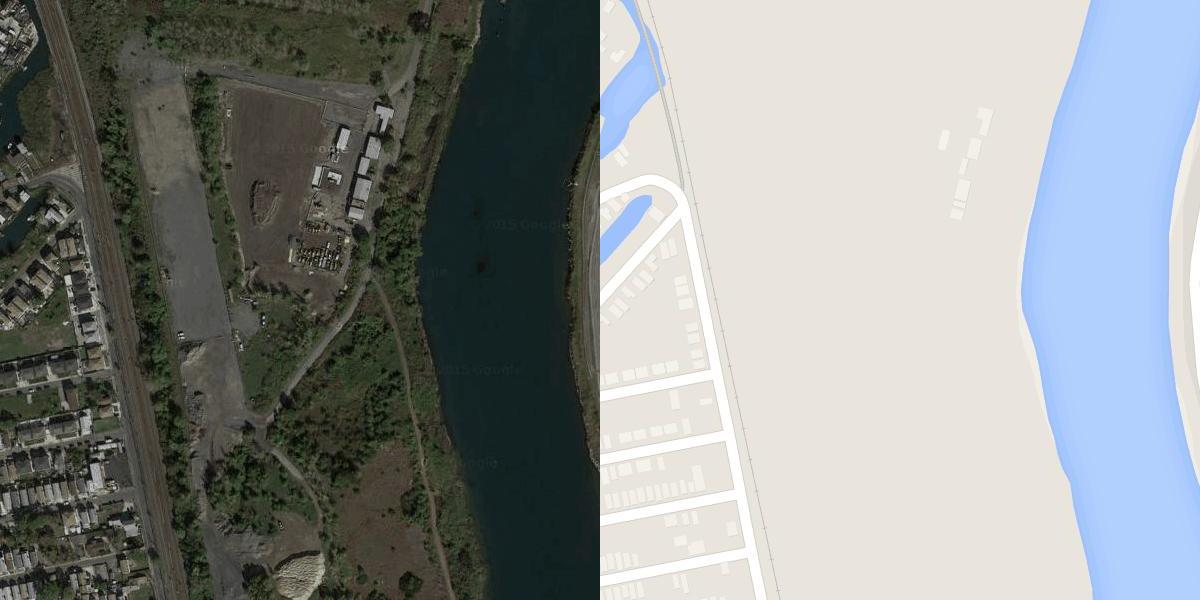}
        \includegraphics*[width=0.18\textwidth, viewport=602 0 1200 600]{figs/Maps/17.jpg}%
        \\
        \includegraphics*[width=0.18\textwidth, viewport=0 0 600 600]{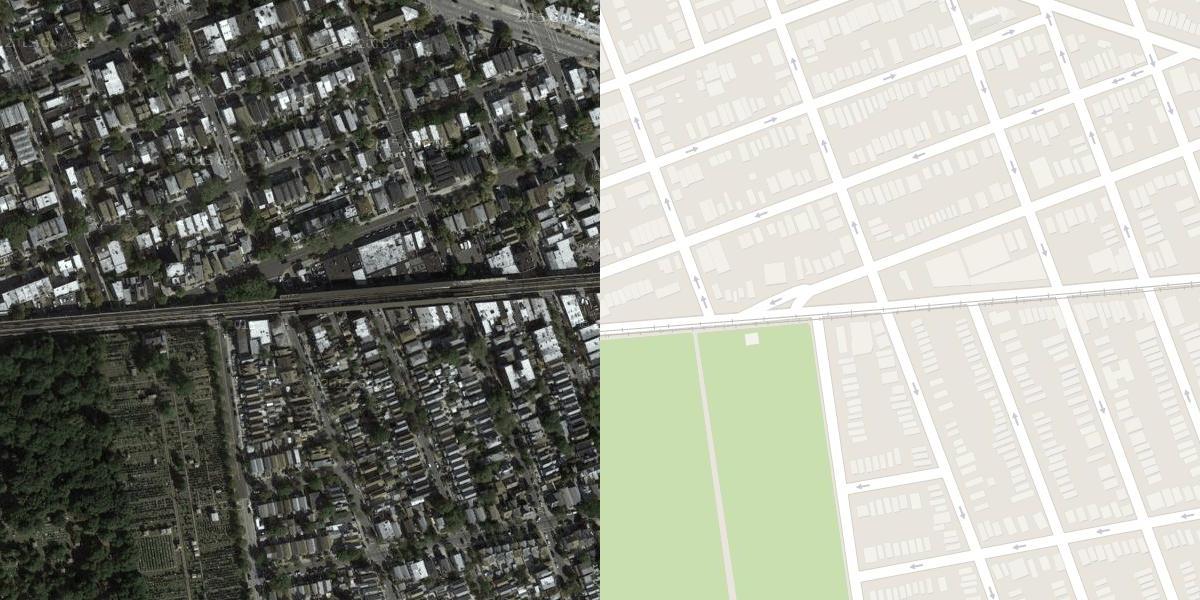}%
        \hspace*{0.02cm}
        \includegraphics*[width=0.18\textwidth, viewport=602 0 1200 600]{figs/Maps/386.jpg}
        \hfill
        \includegraphics*[width=0.18\textwidth, viewport=602 0 1200 600]{figs/Maps/386.jpg}
        \includegraphics*[width=0.18\textwidth, viewport=602 0 1200 600]{figs/Maps/238.jpg}
        \includegraphics*[width=0.18\textwidth, viewport=602 0 1200 600]{figs/Maps/17.jpg}%
        \\
        \includegraphics*[width=0.18\textwidth, viewport=0 0 600 600]{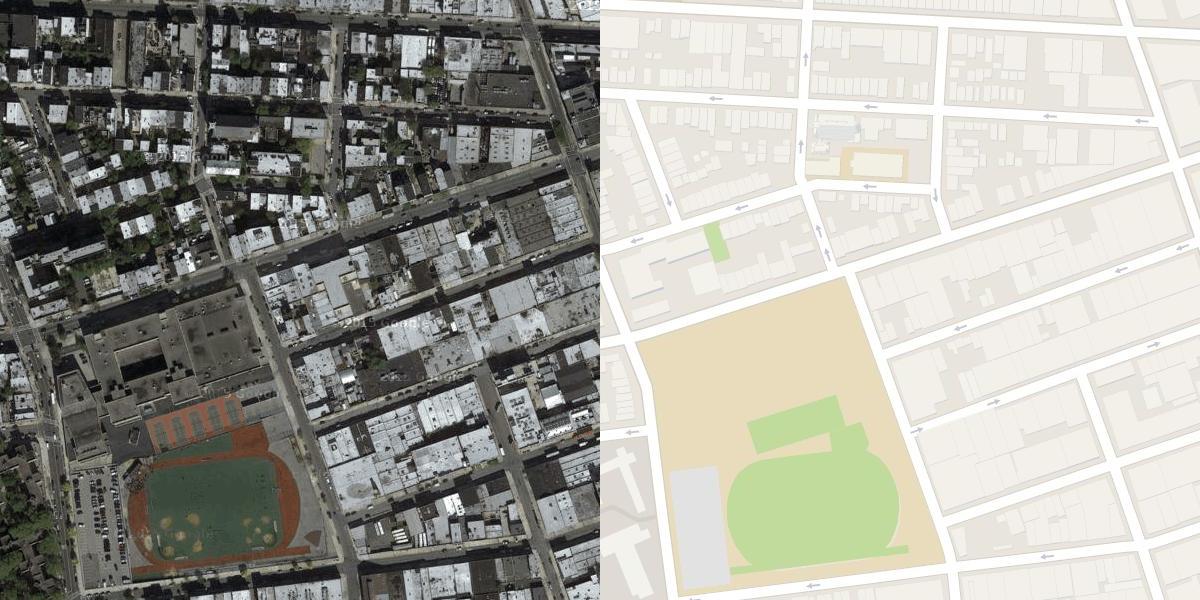}%
        \hspace*{0.02cm}
        \includegraphics*[width=0.18\textwidth, viewport=602 0 1200 600]{figs/Maps/387.jpg}
        \hfill
        \includegraphics*[width=0.18\textwidth, viewport=602 0 1200 600]{figs/Maps/387.jpg}
        \includegraphics*[width=0.18\textwidth, viewport=602 0 1200 600]{figs/Maps/238.jpg}
        \includegraphics*[width=0.18\textwidth, viewport=602 0 1200 600]{figs/Maps/17.jpg}%
        \\
        \includegraphics*[width=0.18\textwidth, viewport=0 0 600 600]{figs/Maps/388.jpg}%
        \hspace*{0.02cm}
        \includegraphics*[width=0.18\textwidth, viewport=602 0 1200 600]{figs/Maps/388.jpg}
        \hfill
        \includegraphics*[width=0.18\textwidth, viewport=602 0 1200 600]{figs/Maps/388.jpg}
        \includegraphics*[width=0.18\textwidth, viewport=602 0 1200 600]{figs/Maps/1065.jpg}
        \includegraphics*[width=0.18\textwidth, viewport=602 0 1200 600]{figs/Maps/1030.jpg}%
        \\
        \includegraphics*[width=0.18\textwidth, viewport=0 0 600 600]{figs/Maps/389.jpg}%
        \hspace*{0.02cm}
        \includegraphics*[width=0.18\textwidth, viewport=602 0 1200 600]{figs/Maps/389.jpg}
        \hfill
        \includegraphics*[width=0.18\textwidth, viewport=602 0 1200 600]{figs/Maps/389.jpg}
        \includegraphics*[width=0.18\textwidth, viewport=602 0 1200 600]{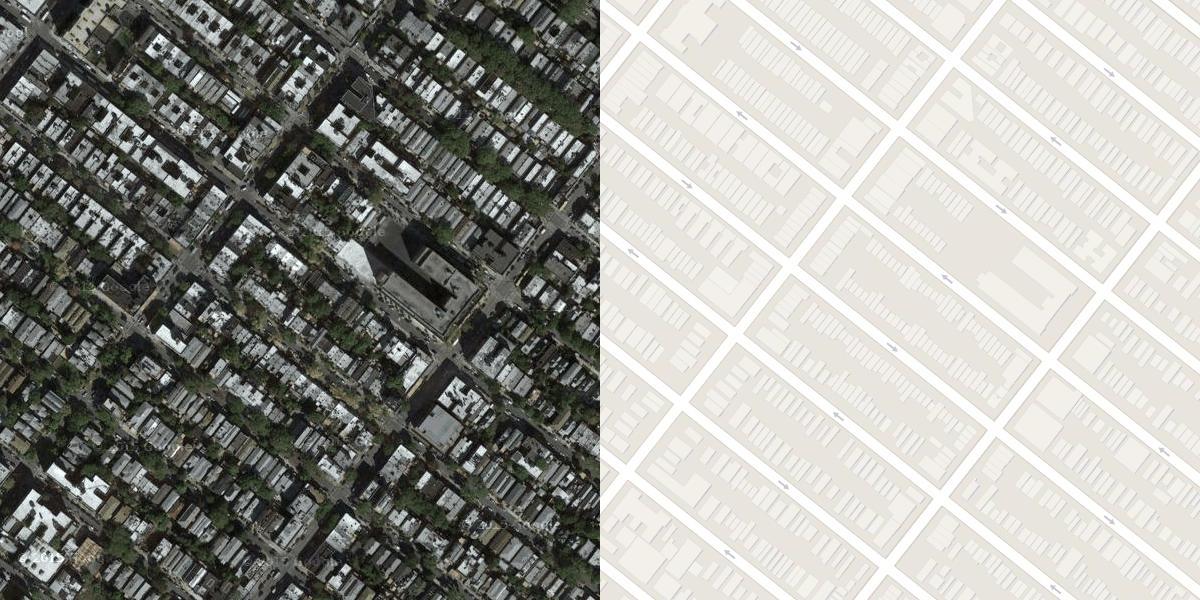}
        \includegraphics*[width=0.18\textwidth, viewport=602 0 1200 600]{figs/Maps/483.jpg}%
        \\
        \includegraphics*[width=0.18\textwidth, viewport=0 0 600 600]{figs/Maps/390.jpg}%
        \hspace*{0.02cm}
        \includegraphics*[width=0.18\textwidth, viewport=602 0 1200 600]{figs/Maps/390.jpg}
        \hfill
        \includegraphics*[width=0.18\textwidth, viewport=602 0 1200 600]{figs/Maps/390.jpg}
        \includegraphics*[width=0.18\textwidth, viewport=602 0 1200 600]{figs/Maps/1010.jpg}
        \includegraphics*[width=0.18\textwidth, viewport=602 0 1200 600]{figs/Maps/238.jpg}%
        \\
        \includegraphics*[width=0.18\textwidth, viewport=0 0 600 600]{figs/Maps/391.jpg}%
        \hspace*{0.02cm}
        \includegraphics*[width=0.18\textwidth, viewport=602 0 1200 600]{figs/Maps/391.jpg}
        \hfill
        \includegraphics*[width=0.18\textwidth, viewport=602 0 1200 600]{figs/Maps/391.jpg}
        \includegraphics*[width=0.18\textwidth, viewport=602 0 1200 600]{figs/Maps/503.jpg}
        \includegraphics*[width=0.18\textwidth, viewport=602 0 1200 600]{figs/Maps/563.jpg}%
        \\
        \includegraphics*[width=0.18\textwidth, viewport=0 0 600 600]{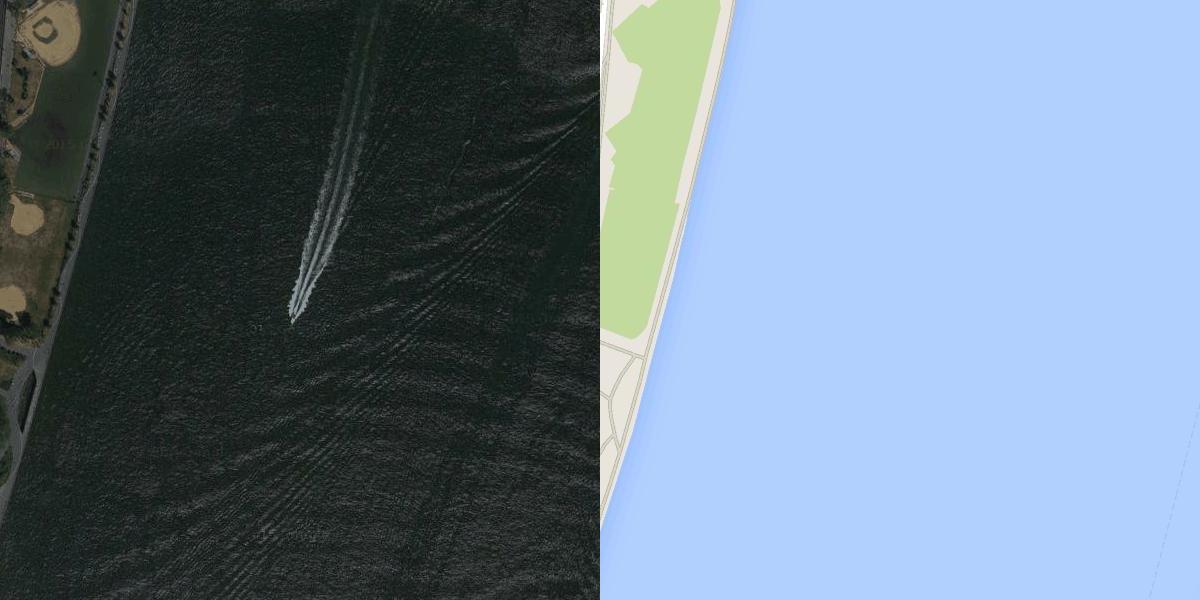}%
        \hspace*{0.02cm}
        \includegraphics*[width=0.18\textwidth, viewport=602 0 1200 600]{figs/Maps/392.jpg}
        \hfill
        \includegraphics*[width=0.18\textwidth, viewport=602 0 1200 600]{figs/Maps/392.jpg}
        \includegraphics*[width=0.18\textwidth, viewport=602 0 1200 600]{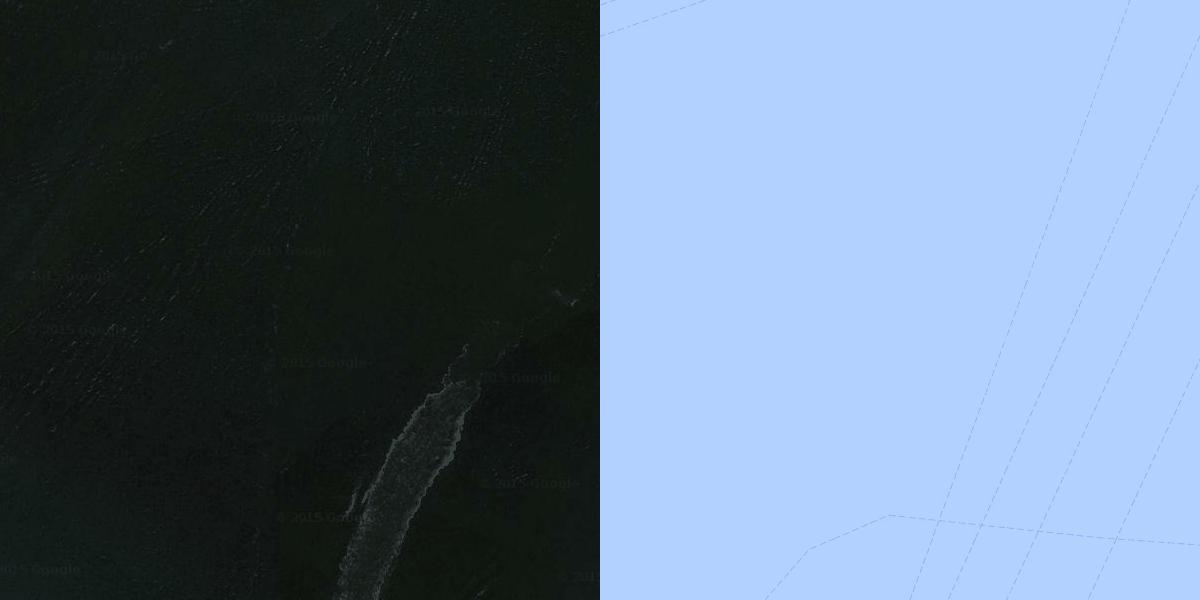}
        \includegraphics*[width=0.18\textwidth, viewport=602 0 1200 600]{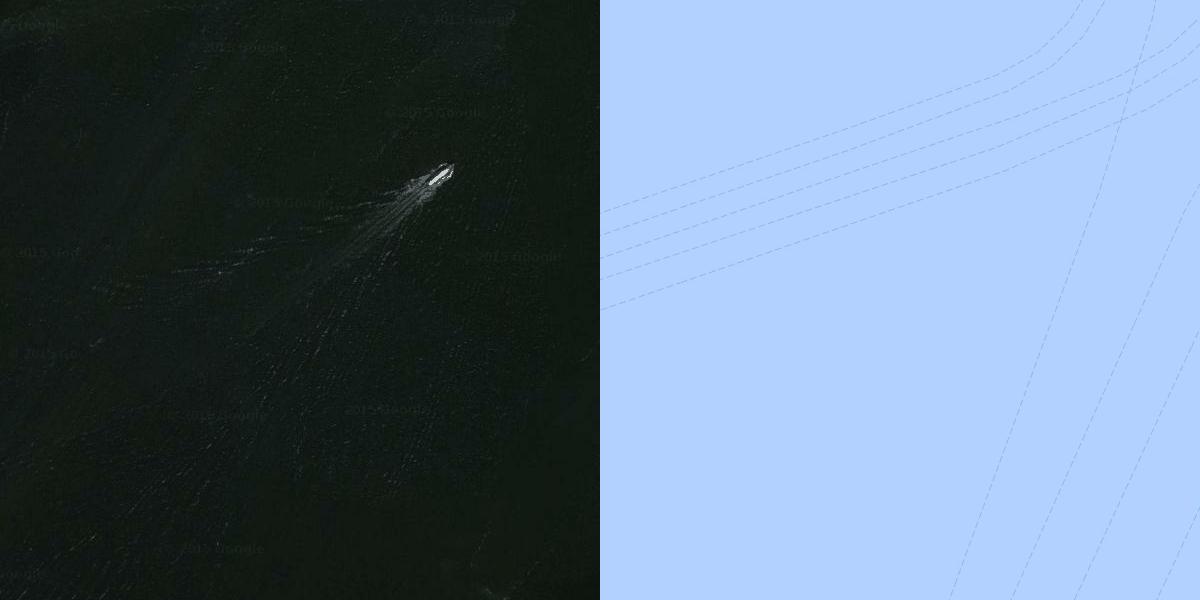}%
        \\
        \includegraphics*[width=0.18\textwidth, viewport=0 0 600 600]{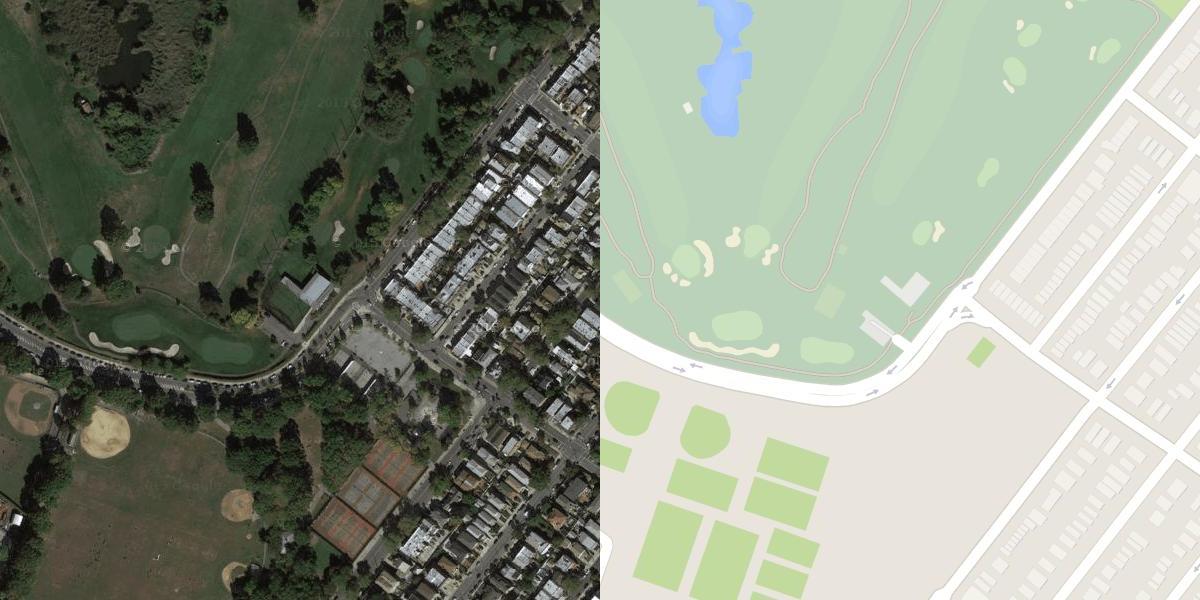}%
        \hspace*{0.02cm}
        \includegraphics*[width=0.18\textwidth, viewport=602 0 1200 600]{figs/Maps/393.jpg}
        \hfill
        \includegraphics*[width=0.18\textwidth, viewport=602 0 1200 600]{figs/Maps/393.jpg}
        \includegraphics*[width=0.18\textwidth, viewport=602 0 1200 600]{figs/Maps/238.jpg}
        \includegraphics*[width=0.18\textwidth, viewport=602 0 1200 600]{figs/Maps/17.jpg}%
    \end{subfigure}
    \vrule
    \hfill
    \begin{subfigure}[b]{0.49\textwidth}
        \captionsetup{justification=raggedright,singlelinecheck=false}
        \caption*{Query(M) \hspace*{0.15cm} GT(S) \hspace*{1.88cm} M $\rightarrow$ S}
        \centering
        \includegraphics*[width=0.18\textwidth, viewport=602 0 1200 600]{figs/Maps/384.jpg}
        \includegraphics*[width=0.18\textwidth, viewport=0 0 600 600]{figs/Maps/384.jpg}%
        \hfill
        \includegraphics*[width=0.18\textwidth, viewport=0 0 600 600]{figs/Maps/384.jpg}
        \includegraphics*[width=0.18\textwidth, viewport=0 0 600 600]{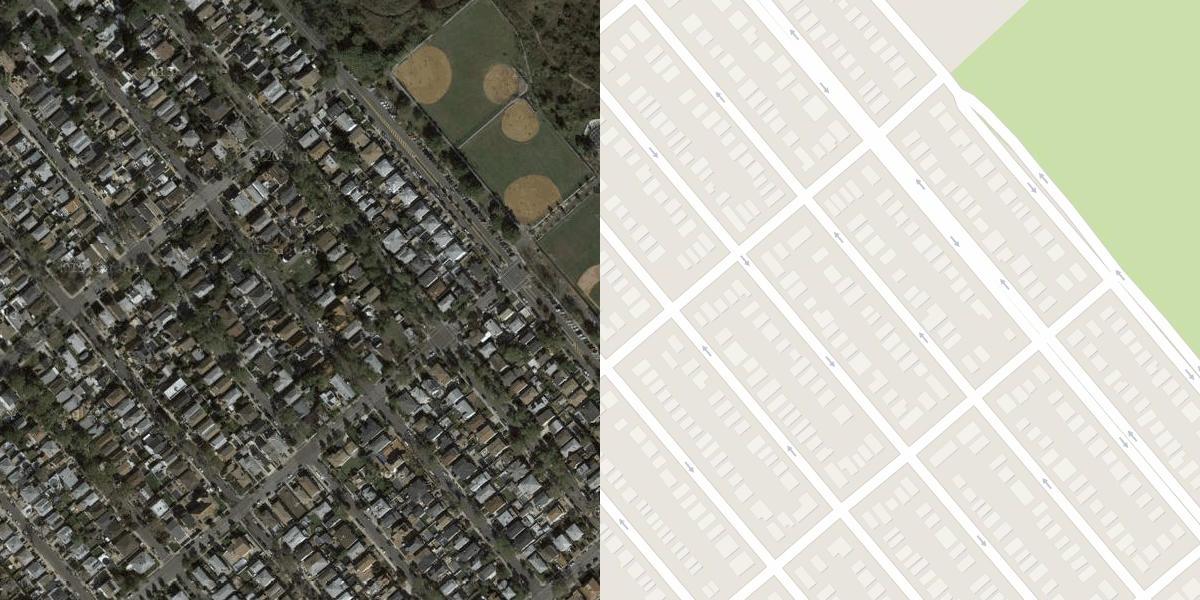}
        \includegraphics*[width=0.18\textwidth, viewport=0 0 600 600]{figs/Maps/238.jpg}%
        \\
        \includegraphics*[width=0.18\textwidth, viewport=602 0 1200 600]{figs/Maps/385.jpg}
        \includegraphics*[width=0.18\textwidth, viewport=0 0 600 600]{figs/Maps/385.jpg}%
        \hfill
        \includegraphics*[width=0.18\textwidth, viewport=0 0 600 600]{figs/Maps/385.jpg}
        \includegraphics*[width=0.18\textwidth, viewport=0 0 600 600]{figs/Maps/238.jpg}
        \includegraphics*[width=0.18\textwidth, viewport=0 0 600 600]{figs/Maps/17.jpg}%
        \\
        \includegraphics*[width=0.18\textwidth, viewport=602 0 1200 600]{figs/Maps/386.jpg}
        \includegraphics*[width=0.18\textwidth, viewport=0 0 600 600]{figs/Maps/386.jpg}%
        \hfill
        \includegraphics*[width=0.18\textwidth, viewport=0 0 600 600]{figs/Maps/386.jpg}
        \includegraphics*[width=0.18\textwidth, viewport=0 0 600 600]{figs/Maps/238.jpg}
        \includegraphics*[width=0.18\textwidth, viewport=0 0 600 600]{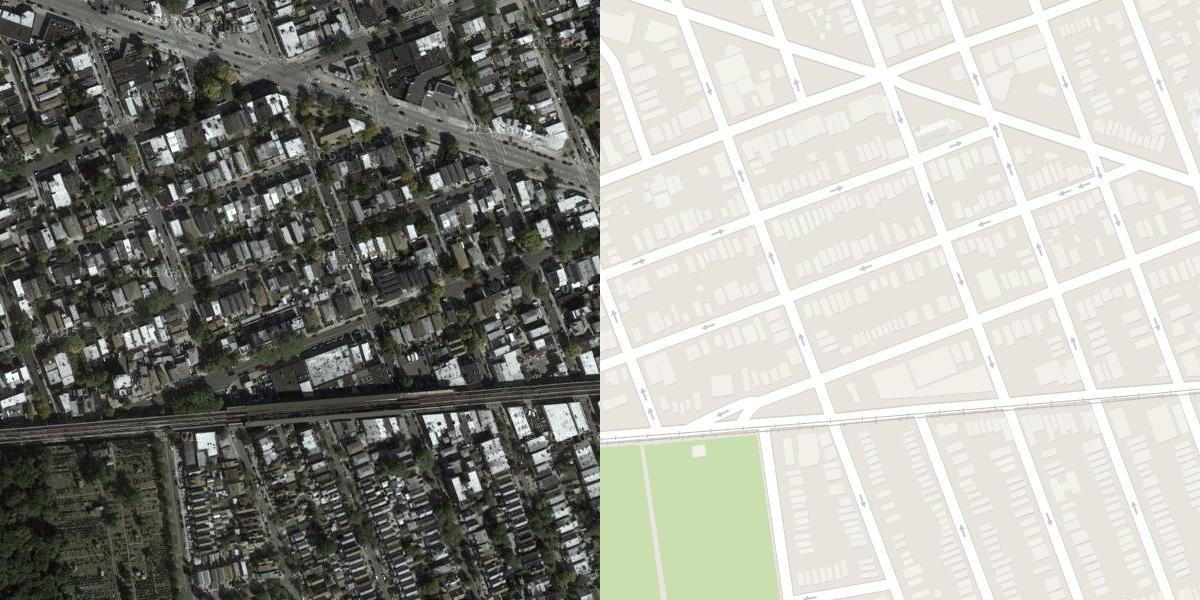}%
        \\
        \includegraphics*[width=0.18\textwidth, viewport=602 0 1200 600]{figs/Maps/387.jpg}
        \includegraphics*[width=0.18\textwidth, viewport=0 0 600 600]{figs/Maps/387.jpg}%
        \hfill
        \includegraphics*[width=0.18\textwidth, viewport=0 0 600 600]{figs/Maps/387.jpg}
        \includegraphics*[width=0.18\textwidth, viewport=0 0 600 600]{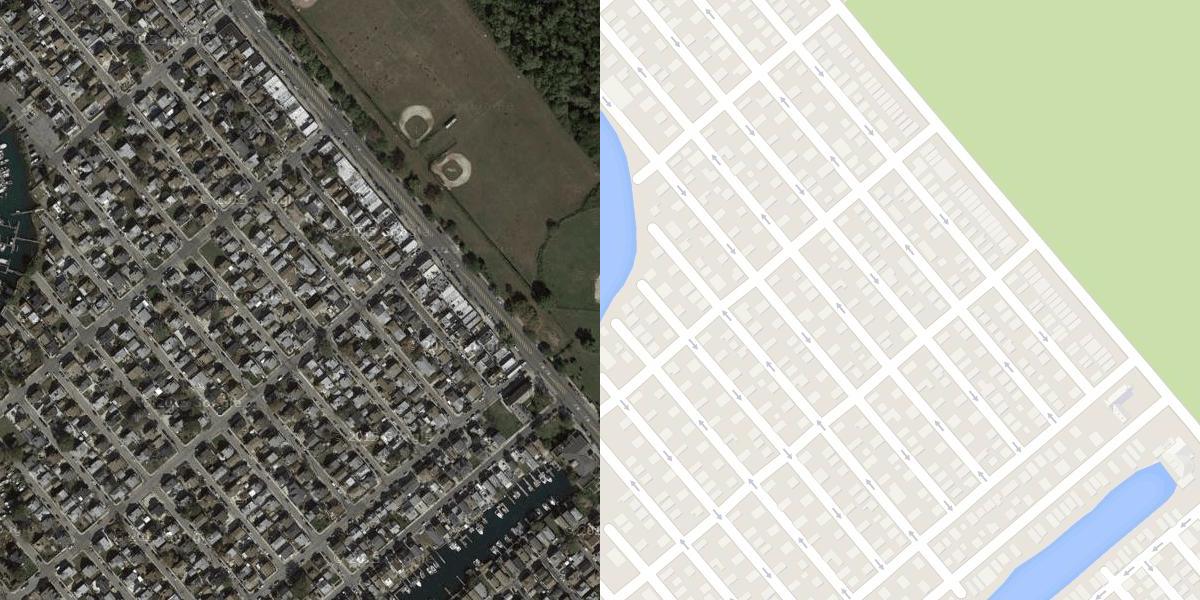}
        \includegraphics*[width=0.18\textwidth, viewport=0 0 600 600]{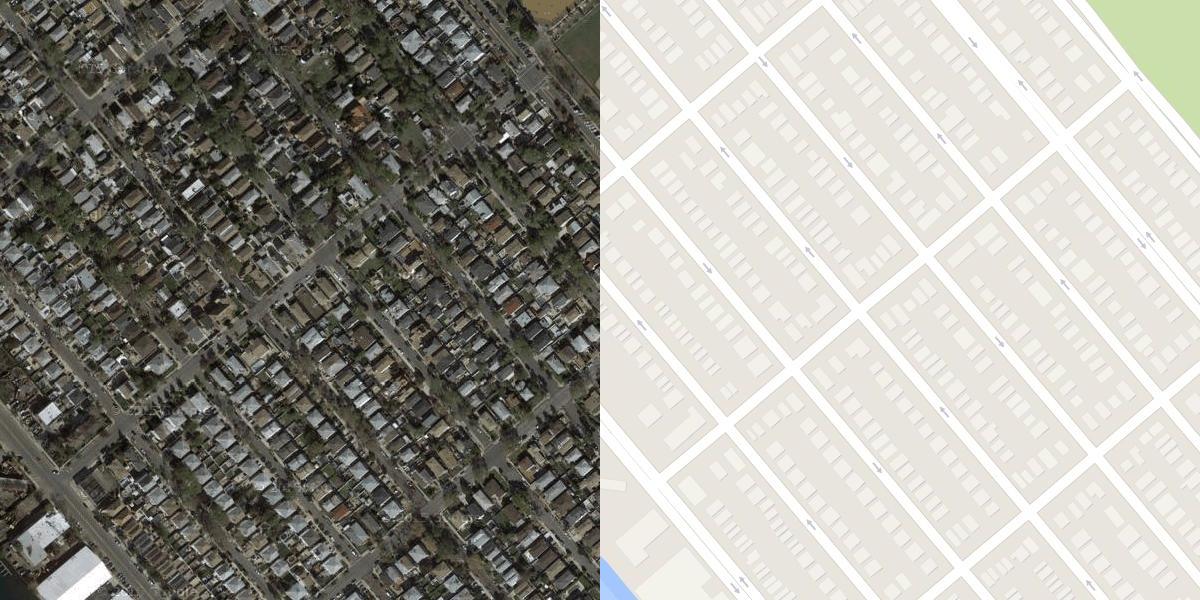}%
        \\
        \includegraphics*[width=0.18\textwidth, viewport=602 0 1200 600]{figs/Maps/388.jpg}
        \includegraphics*[width=0.18\textwidth, viewport=0 0 600 600]{figs/Maps/388.jpg}%
        \hfill
        \includegraphics*[width=0.18\textwidth, viewport=0 0 600 600]{figs/Maps/388.jpg}
        \includegraphics*[width=0.18\textwidth, viewport=0 0 600 600]{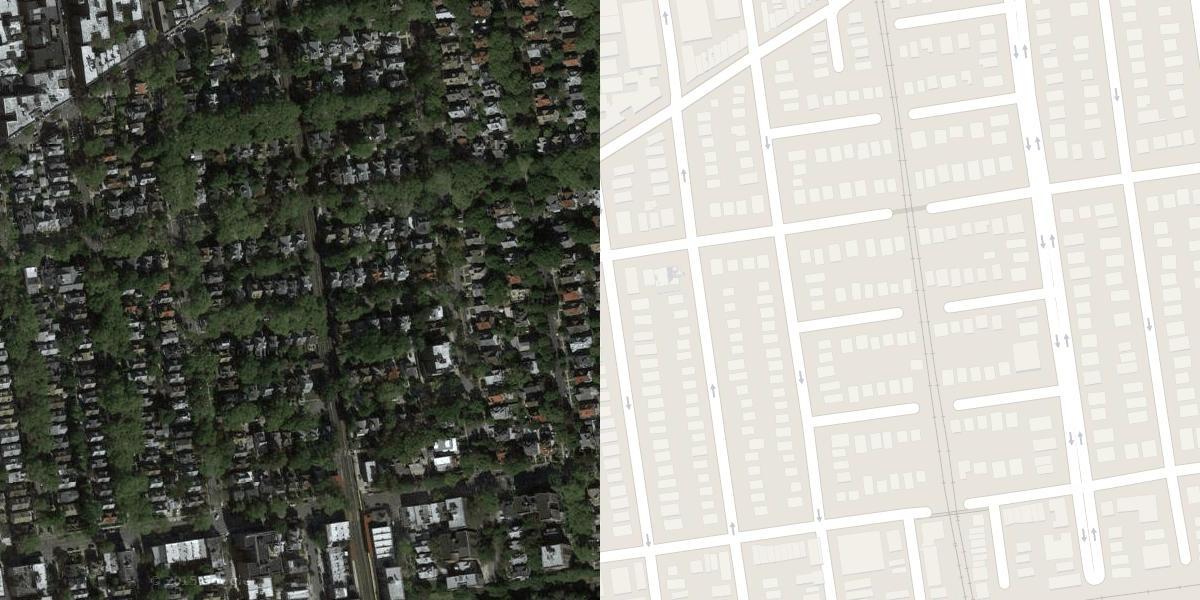}
        \includegraphics*[width=0.18\textwidth, viewport=0 0 600 600]{figs/Maps/1065.jpg}%
        \\
        \includegraphics*[width=0.18\textwidth, viewport=602 0 1200 600]{figs/Maps/389.jpg}
        \includegraphics*[width=0.18\textwidth, viewport=0 0 600 600]{figs/Maps/389.jpg}%
        \hfill
        \includegraphics*[width=0.18\textwidth, viewport=0 0 600 600]{figs/Maps/389.jpg}
        \includegraphics*[width=0.18\textwidth, viewport=0 0 600 600]{figs/Maps/483.jpg}
        \includegraphics*[width=0.18\textwidth, viewport=0 0 600 600]{figs/Maps/183.jpg}%
        \\
        \includegraphics*[width=0.18\textwidth, viewport=602 0 1200 600]{figs/Maps/390.jpg}
        \includegraphics*[width=0.18\textwidth, viewport=0 0 600 600]{figs/Maps/390.jpg}%
        \hfill
        \includegraphics*[width=0.18\textwidth, viewport=0 0 600 600]{figs/Maps/390.jpg}
        \includegraphics*[width=0.18\textwidth, viewport=0 0 600 600]{figs/Maps/1010.jpg}
        \includegraphics*[width=0.18\textwidth, viewport=0 0 600 600]{figs/Maps/975.jpg}%
        \\
        \includegraphics*[width=0.18\textwidth, viewport=602 0 1200 600]{figs/Maps/391.jpg}
        \includegraphics*[width=0.18\textwidth, viewport=0 0 600 600]{figs/Maps/391.jpg}%
        \hfill
        \includegraphics*[width=0.18\textwidth, viewport=0 0 600 600]{figs/Maps/391.jpg}
        \includegraphics*[width=0.18\textwidth, viewport=0 0 600 600]{figs/Maps/503.jpg}
        \includegraphics*[width=0.18\textwidth, viewport=0 0 600 600]{figs/Maps/1083.jpg}%
        \\
        \includegraphics*[width=0.18\textwidth, viewport=602 0 1200 600]{figs/Maps/392.jpg}
        \includegraphics*[width=0.18\textwidth, viewport=0 0 600 600]{figs/Maps/392.jpg}%
        \hfill
        \includegraphics*[width=0.18\textwidth, viewport=0 0 600 600]{figs/Maps/392.jpg}
        \includegraphics*[width=0.18\textwidth, viewport=0 0 600 600]{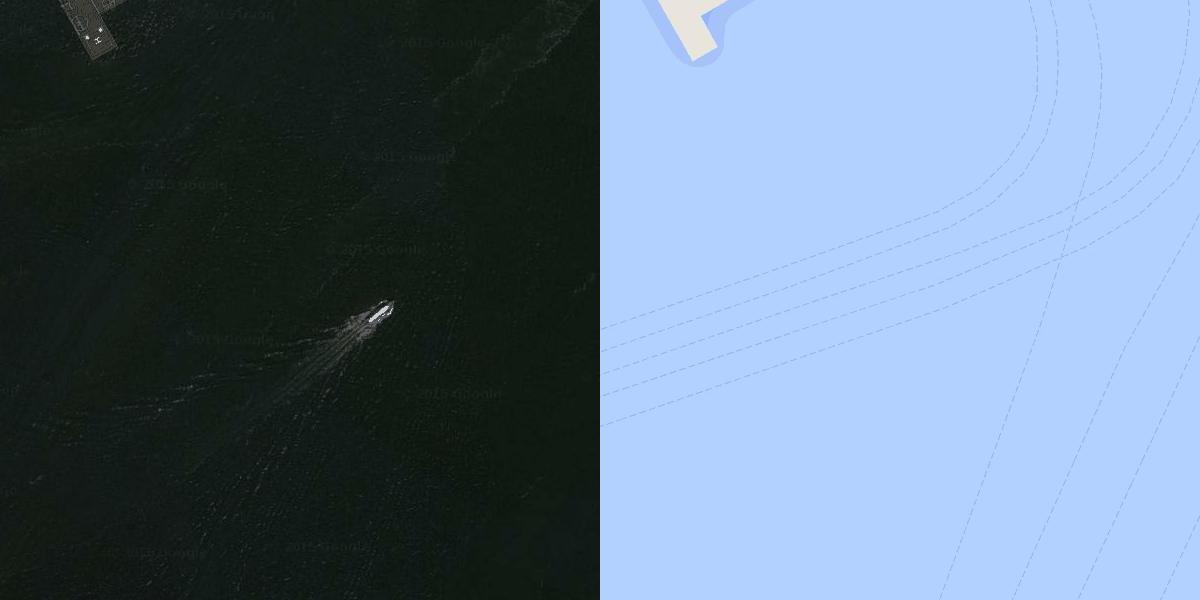}
        \includegraphics*[width=0.18\textwidth, viewport=0 0 600 600]{figs/Maps/538.jpg}%
        \\
        \includegraphics*[width=0.18\textwidth, viewport=602 0 1200 600]{figs/Maps/393.jpg}
        \includegraphics*[width=0.18\textwidth, viewport=0 0 600 600]{figs/Maps/393.jpg}%
        \hfill
        \includegraphics*[width=0.18\textwidth, viewport=0 0 600 600]{figs/Maps/393.jpg}
        \includegraphics*[width=0.18\textwidth, viewport=0 0 600 600]{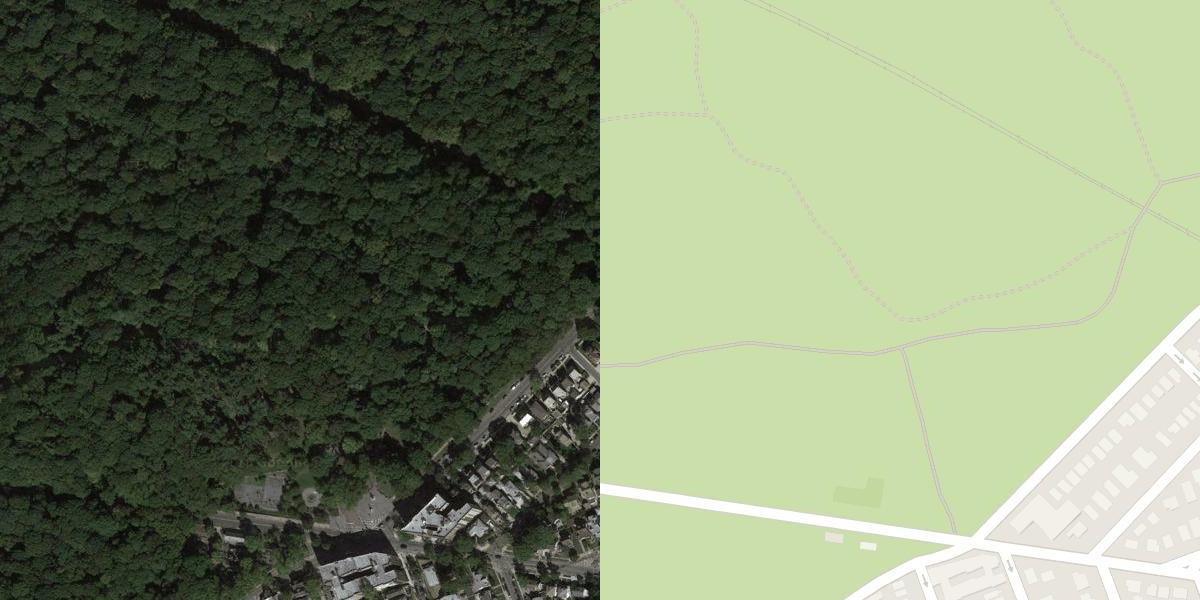}
        \includegraphics*[width=0.18\textwidth, viewport=0 0 600 600]{figs/Maps/238.jpg}%
    \end{subfigure}
    \caption{Successful examples of cross-domain retrieval (Top-3) in Maps using IIAE.}
    \label{fig:SuccessCdMaps}
    \vspace{-0.5cm}
\end{figure}
\begin{figure}[H]
    \centering
    \begin{subfigure}[b]{0.49\textwidth}
        \captionsetup{justification=raggedright,singlelinecheck=false}
        \caption*{Query(S) \hspace*{0.15cm} GT(M) \hspace*{1.88cm} S $\rightarrow$ M}
        \centering
        \includegraphics*[width=0.18\textwidth, viewport=0 0 600 600]{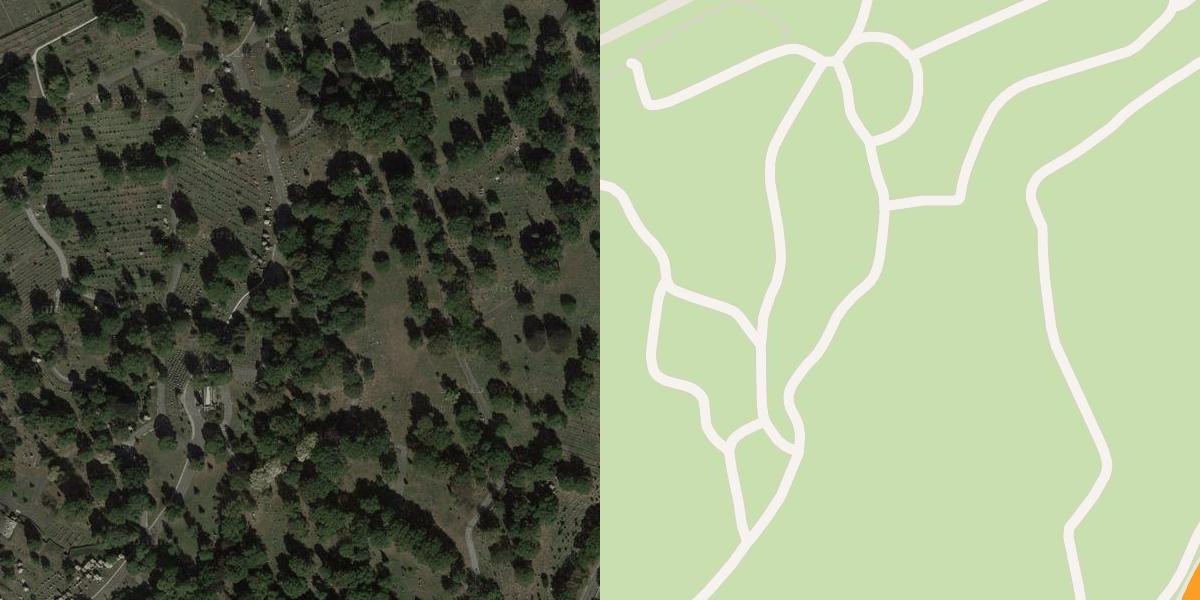}%
        \hspace*{0.02cm}
        \includegraphics*[width=0.18\textwidth, viewport=602 0 1200 600]{figs/Maps/300.jpg}
        \hfill
        \includegraphics*[width=0.18\textwidth, viewport=602 0 1200 600]{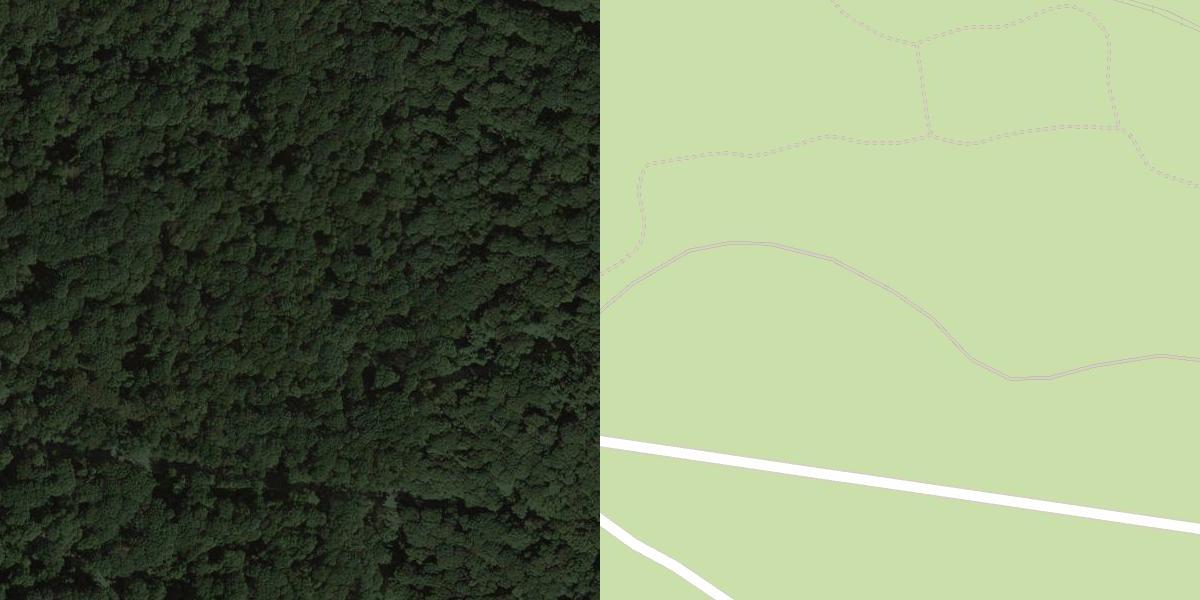}
        \includegraphics*[width=0.18\textwidth, viewport=602 0 1200 600]{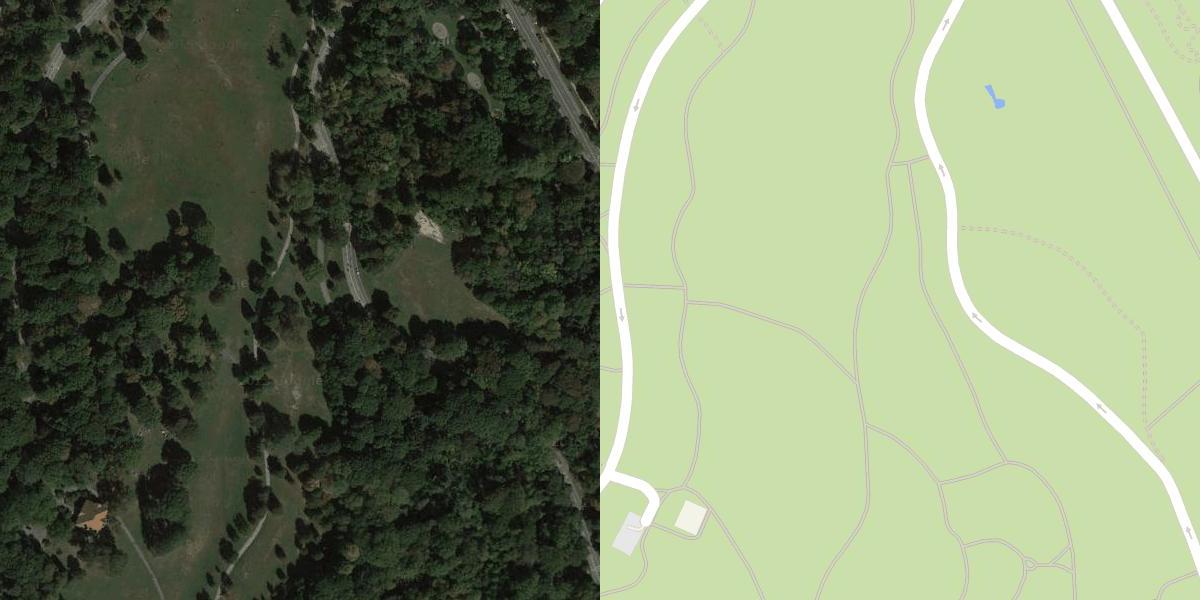}
        \includegraphics*[width=0.18\textwidth, viewport=602 0 1200 600]{figs/Maps/300.jpg}%
        \\
        \includegraphics*[width=0.18\textwidth, viewport=0 0 600 600]{figs/Maps/872.jpg}%
        \hspace*{0.02cm}
        \includegraphics*[width=0.18\textwidth, viewport=602 0 1200 600]{figs/Maps/872.jpg}
        \hfill
        \includegraphics*[width=0.18\textwidth, viewport=602 0 1200 600]{figs/Maps/538.jpg}
        \includegraphics*[width=0.18\textwidth, viewport=602 0 1200 600]{figs/Maps/872.jpg}
        \includegraphics*[width=0.18\textwidth, viewport=602 0 1200 600]{figs/Maps/283.jpg}%
    \end{subfigure}
    \vrule
    \hfill
    \begin{subfigure}[b]{0.49\textwidth}
        \captionsetup{justification=raggedright,singlelinecheck=false}
        \caption*{Query(M) \hspace*{0.15cm} GT(S) \hspace*{1.88cm} M $\rightarrow$ S}
        \centering
        \includegraphics*[width=0.18\textwidth, viewport=602 0 1200 600]{figs/Maps/300.jpg}
        \includegraphics*[width=0.18\textwidth, viewport=0 0 600 600]{figs/Maps/300.jpg}%
        \hfill
        \includegraphics*[width=0.18\textwidth, viewport=0 0 600 600]{figs/Maps/36.jpg}
        \includegraphics*[width=0.18\textwidth, viewport=0 0 600 600]{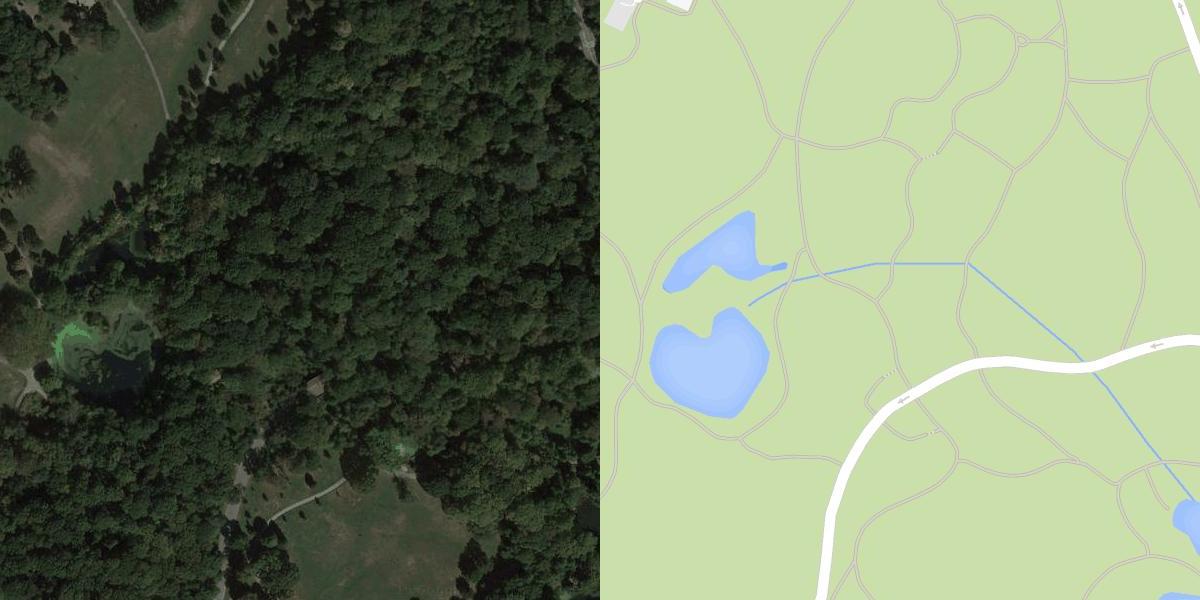}
        \includegraphics*[width=0.18\textwidth, viewport=0 0 600 600]{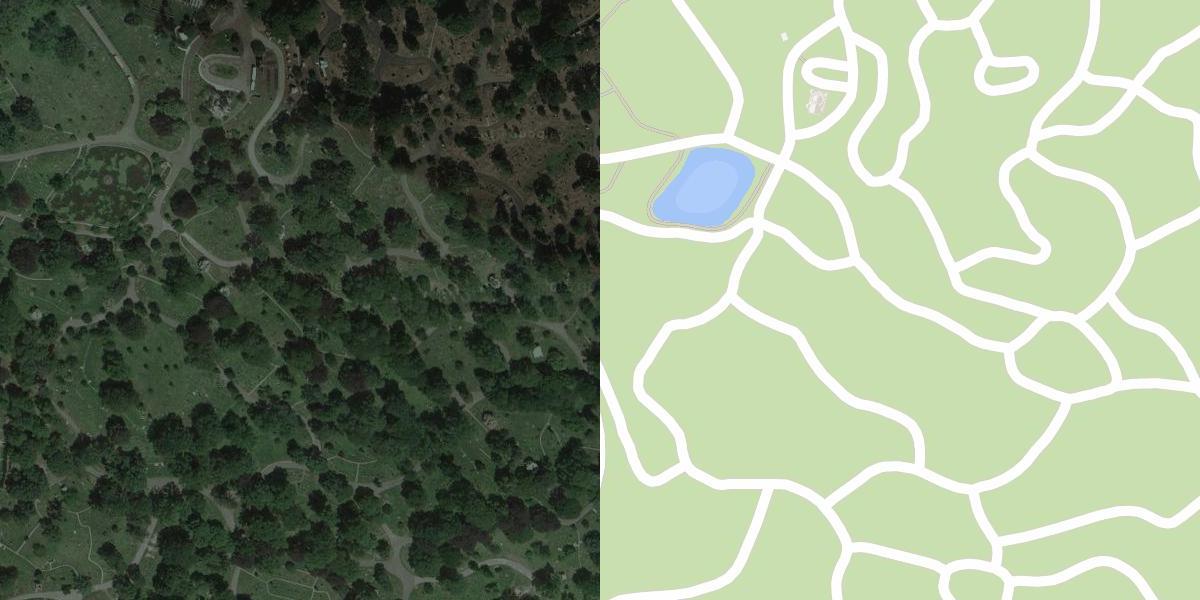}%
        \\
        \includegraphics*[width=0.18\textwidth, viewport=602 0 1200 600]{figs/Maps/872.jpg}
        \includegraphics*[width=0.18\textwidth, viewport=0 0 600 600]{figs/Maps/872.jpg}%
        \hfill
        \includegraphics*[width=0.18\textwidth, viewport=0 0 600 600]{figs/Maps/538.jpg}
        \includegraphics*[width=0.18\textwidth, viewport=0 0 600 600]{figs/Maps/283.jpg}
        \includegraphics*[width=0.18\textwidth, viewport=0 0 600 600]{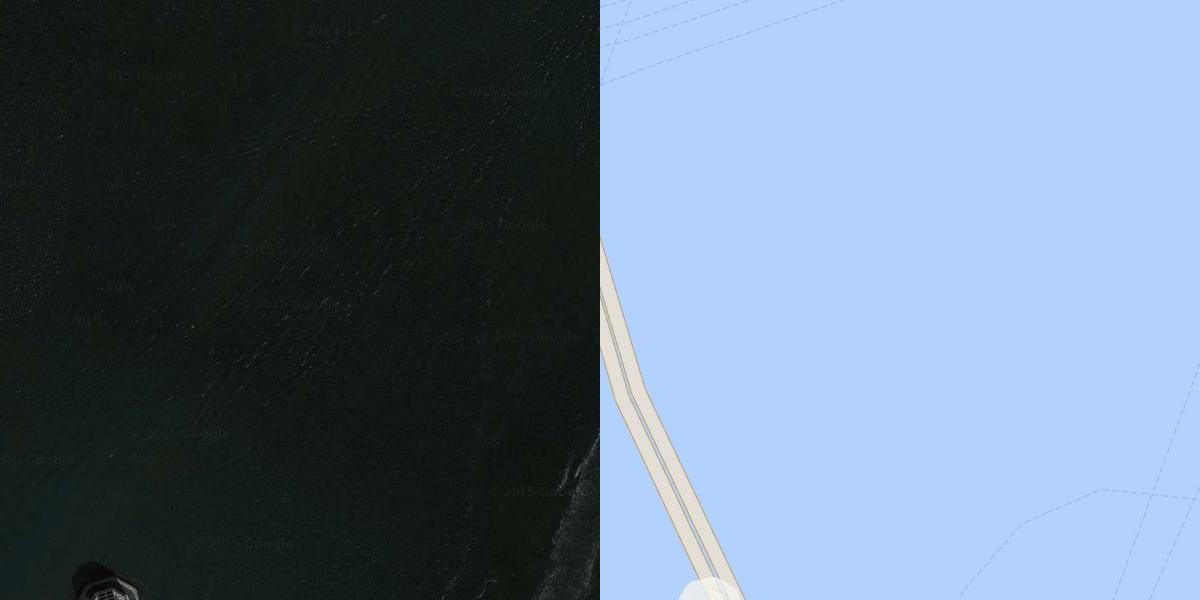}%
    \end{subfigure}
    \caption{Unsuccessful examples of cross-domain retrieval (Top-3) in Maps using IIAE.}
    \label{fig:UnsuccessCdMaps}
    % \vspace{-10em}
\end{figure}

\subsubsection{Facades \citep{10.1007/978-3-642-40602-7_39}}
%\vspace{-0.3cm}
\begin{figure}[H]
    \centering
    % \hspace{10mm}
    \begin{subfigure}[b]{0.49\textwidth}
        \captionsetup{justification=raggedright,singlelinecheck=false}
        \caption*{Query(F) \hspace*{0.16cm} GT(L) \hspace*{2cm} F $\rightarrow$ L}
        \centering
        \includegraphics*[width=0.18\textwidth, viewport=0 0 256 256]{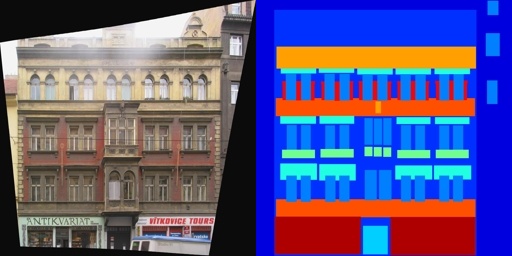}%
        \hspace*{0.02cm}
        \includegraphics*[width=0.18\textwidth, viewport=256 0 512 256]{figs/Facades/2.jpg}
        \hfill
        \includegraphics*[width=0.18\textwidth, viewport=256 0 512 256]{figs/Facades/2.jpg}
        \includegraphics*[width=0.18\textwidth, viewport=256 0 512 256]{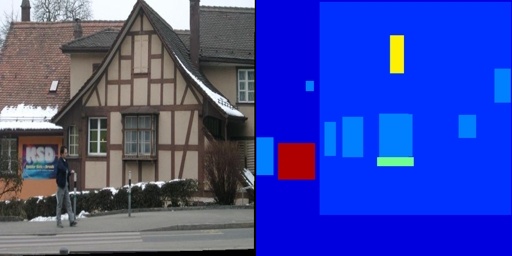}
        \includegraphics*[width=0.18\textwidth, viewport=256 0 512 256]{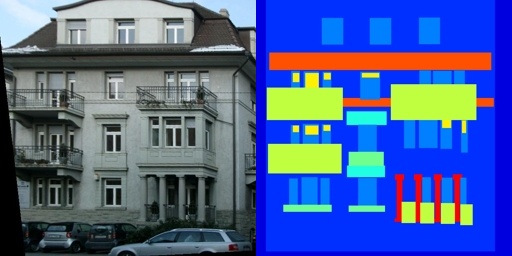}%
        \\
        \includegraphics*[width=0.18\textwidth, viewport=0 0 256 256]{figs/Facades/3.jpg}%
        \hspace*{0.02cm}
        \includegraphics*[width=0.18\textwidth, viewport=256 0 512 256]{figs/Facades/3.jpg}
        \hfill
        \includegraphics*[width=0.18\textwidth, viewport=256 0 512 256]{figs/Facades/3.jpg}
        \includegraphics*[width=0.18\textwidth, viewport=256 0 512 256]{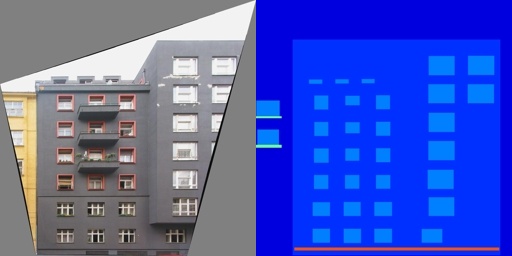}
        \includegraphics*[width=0.18\textwidth, viewport=256 0 512 256]{figs/Facades/66.jpg}%
        \\
        \includegraphics*[width=0.18\textwidth, viewport=0 0 256 256]{figs/Facades/5.jpg}%
        \hspace*{0.02cm}
        \includegraphics*[width=0.18\textwidth, viewport=256 0 512 256]{figs/Facades/5.jpg}
        \hfill
        \includegraphics*[width=0.18\textwidth, viewport=256 0 512 256]{figs/Facades/5.jpg}
        \includegraphics*[width=0.18\textwidth, viewport=256 0 512 256]{figs/Facades/47.jpg}
        \includegraphics*[width=0.18\textwidth, viewport=256 0 512 256]{figs/Facades/62.jpg}%
        \\
        \includegraphics*[width=0.18\textwidth, viewport=0 0 256 256]{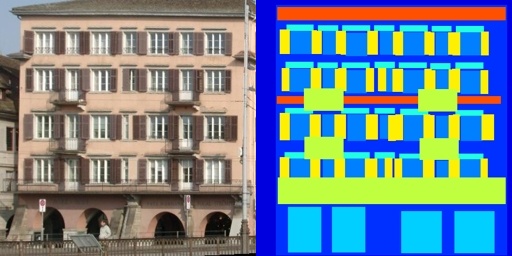}%
        \hspace*{0.02cm}
        \includegraphics*[width=0.18\textwidth, viewport=256 0 512 256]{figs/Facades/6.jpg}
        \hfill
        \includegraphics*[width=0.18\textwidth, viewport=256 0 512 256]{figs/Facades/6.jpg}
        \includegraphics*[width=0.18\textwidth, viewport=256 0 512 256]{figs/Facades/66.jpg}
        \includegraphics*[width=0.18\textwidth, viewport=256 0 512 256]{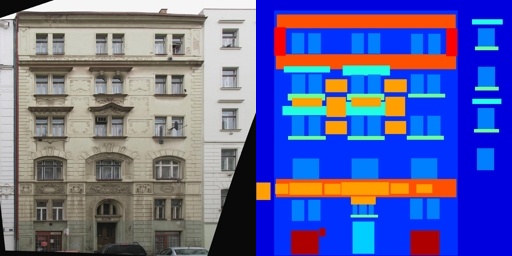}%
        \\
        \includegraphics*[width=0.18\textwidth, viewport=0 0 256 256]{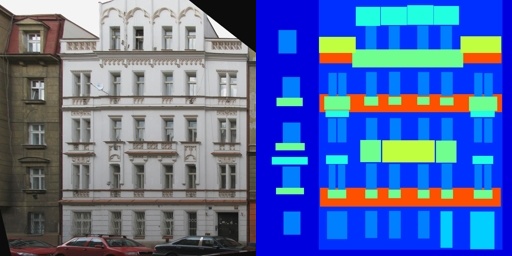}%
        \hspace*{0.02cm}
        \includegraphics*[width=0.18\textwidth, viewport=256 0 512 256]{figs/Facades/7.jpg}
        \hfill
        \includegraphics*[width=0.18\textwidth, viewport=256 0 512 256]{figs/Facades/7.jpg}
        \includegraphics*[width=0.18\textwidth, viewport=256 0 512 256]{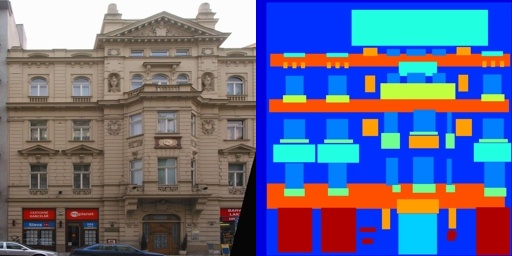}
        \includegraphics*[width=0.18\textwidth, viewport=256 0 512 256]{figs/Facades/43.jpg}%
        \\
        \includegraphics*[width=0.18\textwidth, viewport=0 0 256 256]{figs/Facades/8.jpg}%
        \hspace*{0.02cm}
        \includegraphics*[width=0.18\textwidth, viewport=256 0 512 256]{figs/Facades/8.jpg}
        \hfill
        \includegraphics*[width=0.18\textwidth, viewport=256 0 512 256]{figs/Facades/8.jpg}
        \includegraphics*[width=0.18\textwidth, viewport=256 0 512 256]{figs/Facades/66.jpg}
        \includegraphics*[width=0.18\textwidth, viewport=256 0 512 256]{figs/Facades/21.jpg}%
        \\
        \includegraphics*[width=0.18\textwidth, viewport=0 0 256 256]{figs/Facades/9.jpg}%
        \hspace*{0.02cm}
        \includegraphics*[width=0.18\textwidth, viewport=256 0 512 256]{figs/Facades/9.jpg}
        \hfill
        \includegraphics*[width=0.18\textwidth, viewport=256 0 512 256]{figs/Facades/9.jpg}
        \includegraphics*[width=0.18\textwidth, viewport=256 0 512 256]{figs/Facades/43.jpg}
        \includegraphics*[width=0.18\textwidth, viewport=256 0 512 256]{figs/Facades/21.jpg}%
        \\
        \includegraphics*[width=0.18\textwidth, viewport=0 0 256 256]{figs/Facades/10.jpg}%
        \hspace*{0.02cm}
        \includegraphics*[width=0.18\textwidth, viewport=256 0 512 256]{figs/Facades/10.jpg}
        \hfill
        \includegraphics*[width=0.18\textwidth, viewport=256 0 512 256]{figs/Facades/10.jpg}
        \includegraphics*[width=0.18\textwidth, viewport=256 0 512 256]{figs/Facades/11.jpg}
        \includegraphics*[width=0.18\textwidth, viewport=256 0 512 256]{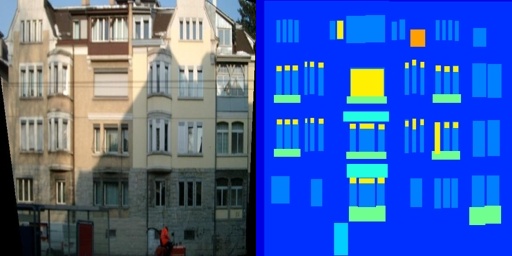}%
        \\
        \includegraphics*[width=0.18\textwidth, viewport=0 0 256 256]{figs/Facades/11.jpg}%
        \hspace*{0.02cm}
        \includegraphics*[width=0.18\textwidth, viewport=256 0 512 256]{figs/Facades/11.jpg}
        \hfill
        \includegraphics*[width=0.18\textwidth, viewport=256 0 512 256]{figs/Facades/11.jpg}
        \includegraphics*[width=0.18\textwidth, viewport=256 0 512 256]{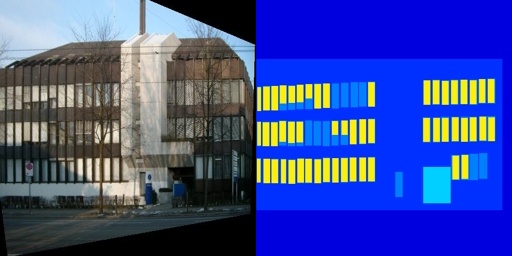}
        \includegraphics*[width=0.18\textwidth, viewport=256 0 512 256]{figs/Facades/14.jpg}%
        \\
        \includegraphics*[width=0.18\textwidth, viewport=0 0 256 256]{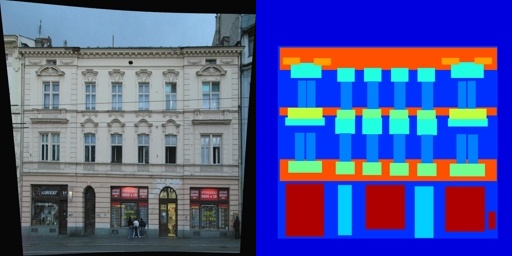}%
        \hspace*{0.02cm}
        \includegraphics*[width=0.18\textwidth, viewport=256 0 512 256]{figs/Facades/12.jpg}
        \hfill
        \includegraphics*[width=0.18\textwidth, viewport=256 0 512 256]{figs/Facades/12.jpg}
        \includegraphics*[width=0.18\textwidth, viewport=256 0 512 256]{figs/Facades/43.jpg}
        \includegraphics*[width=0.18\textwidth, viewport=256 0 512 256]{figs/Facades/47.jpg}%
    \end{subfigure}
    \vrule
    \hfill
    \begin{subfigure}[b]{0.49\textwidth}
        \captionsetup{justification=raggedright,singlelinecheck=false}
        \caption*{Query(L) \hspace*{0.16cm} GT(F) \hspace*{2cm} L $\rightarrow$ F}
        \centering
        \includegraphics*[width=0.18\textwidth, viewport=256 0 512 256]{figs/Facades/2.jpg}
        \includegraphics*[width=0.18\textwidth, viewport=0 0 256 256]{figs/Facades/2.jpg}%
        \hfill
        \includegraphics*[width=0.18\textwidth, viewport=0 0 256 256]{figs/Facades/2.jpg}
        \includegraphics*[width=0.18\textwidth, viewport=0 0 256 256]{figs/Facades/106.jpg}
        \includegraphics*[width=0.18\textwidth, viewport=0 0 256 256]{figs/Facades/9.jpg}%
        \\
        \includegraphics*[width=0.18\textwidth, viewport=256 0 512 256]{figs/Facades/3.jpg}
        \includegraphics*[width=0.18\textwidth, viewport=0 0 256 256]{figs/Facades/3.jpg}%
        \hfill
        \includegraphics*[width=0.18\textwidth, viewport=0 0 256 256]{figs/Facades/3.jpg}
        \includegraphics*[width=0.18\textwidth, viewport=0 0 256 256]{figs/Facades/17.jpg}
        \includegraphics*[width=0.18\textwidth, viewport=0 0 256 256]{figs/Facades/45.jpg}%
        \\
        \includegraphics*[width=0.18\textwidth, viewport=256 0 512 256]{figs/Facades/5.jpg}
        \includegraphics*[width=0.18\textwidth, viewport=0 0 256 256]{figs/Facades/5.jpg}%
        \hfill
        \includegraphics*[width=0.18\textwidth, viewport=0 0 256 256]{figs/Facades/5.jpg}
        \includegraphics*[width=0.18\textwidth, viewport=0 0 256 256]{figs/Facades/12.jpg}
        \includegraphics*[width=0.18\textwidth, viewport=0 0 256 256]{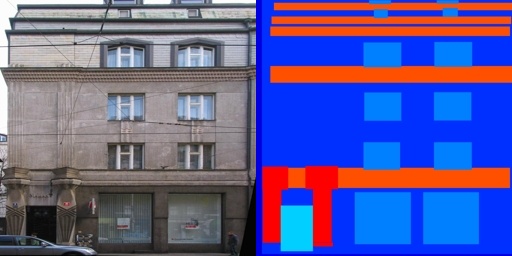}%
        \\
        \includegraphics*[width=0.18\textwidth, viewport=256 0 512 256]{figs/Facades/6.jpg}
        \includegraphics*[width=0.18\textwidth, viewport=0 0 256 256]{figs/Facades/6.jpg}%
        \hfill
        \includegraphics*[width=0.18\textwidth, viewport=0 0 256 256]{figs/Facades/6.jpg}
        \includegraphics*[width=0.18\textwidth, viewport=0 0 256 256]{figs/Facades/66.jpg}
        \includegraphics*[width=0.18\textwidth, viewport=0 0 256 256]{figs/Facades/8.jpg}%
        \\
        \includegraphics*[width=0.18\textwidth, viewport=256 0 512 256]{figs/Facades/7.jpg}
        \includegraphics*[width=0.18\textwidth, viewport=0 0 256 256]{figs/Facades/7.jpg}%
        \hfill
        \includegraphics*[width=0.18\textwidth, viewport=0 0 256 256]{figs/Facades/7.jpg}
        \includegraphics*[width=0.18\textwidth, viewport=0 0 256 256]{figs/Facades/9.jpg}
        \includegraphics*[width=0.18\textwidth, viewport=0 0 256 256]{figs/Facades/66.jpg}%
        \\
        \includegraphics*[width=0.18\textwidth, viewport=256 0 512 256]{figs/Facades/8.jpg}
        \includegraphics*[width=0.18\textwidth, viewport=0 0 256 256]{figs/Facades/8.jpg}%
        \hfill
        \includegraphics*[width=0.18\textwidth, viewport=0 0 256 256]{figs/Facades/8.jpg}
        \includegraphics*[width=0.18\textwidth, viewport=0 0 256 256]{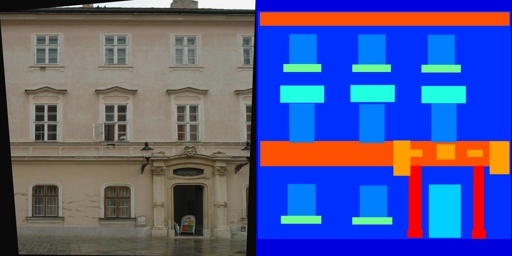}
        \includegraphics*[width=0.18\textwidth, viewport=0 0 256 256]{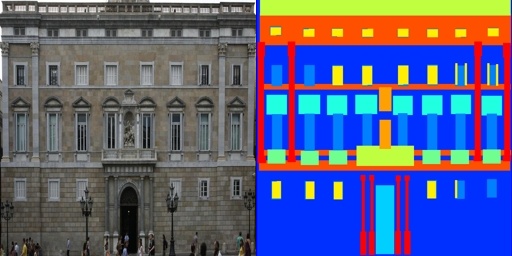}%
        \\
        \includegraphics*[width=0.18\textwidth, viewport=256 0 512 256]{figs/Facades/9.jpg}
        \includegraphics*[width=0.18\textwidth, viewport=0 0 256 256]{figs/Facades/9.jpg}%
        \hfill
        \includegraphics*[width=0.18\textwidth, viewport=0 0 256 256]{figs/Facades/9.jpg}
        \includegraphics*[width=0.18\textwidth, viewport=0 0 256 256]{figs/Facades/7.jpg}
        \includegraphics*[width=0.18\textwidth, viewport=0 0 256 256]{figs/Facades/21.jpg}%
        \\
        \includegraphics*[width=0.18\textwidth, viewport=256 0 512 256]{figs/Facades/10.jpg}
        \includegraphics*[width=0.18\textwidth, viewport=0 0 256 256]{figs/Facades/10.jpg}%
        \hfill
        \includegraphics*[width=0.18\textwidth, viewport=0 0 256 256]{figs/Facades/10.jpg}
        \includegraphics*[width=0.18\textwidth, viewport=0 0 256 256]{figs/Facades/11.jpg}
        \includegraphics*[width=0.18\textwidth, viewport=0 0 256 256]{figs/Facades/45.jpg}%
        \\
        \includegraphics*[width=0.18\textwidth, viewport=256 0 512 256]{figs/Facades/11.jpg}
        \includegraphics*[width=0.18\textwidth, viewport=0 0 256 256]{figs/Facades/11.jpg}%
        \hfill
        \includegraphics*[width=0.18\textwidth, viewport=0 0 256 256]{figs/Facades/11.jpg}
        \includegraphics*[width=0.18\textwidth, viewport=0 0 256 256]{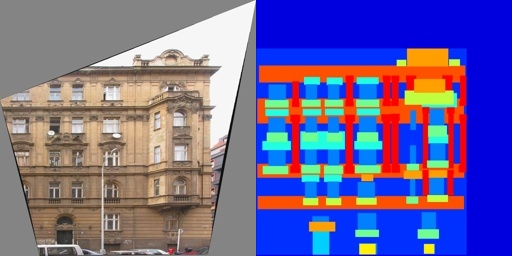}
        \includegraphics*[width=0.18\textwidth, viewport=0 0 256 256]{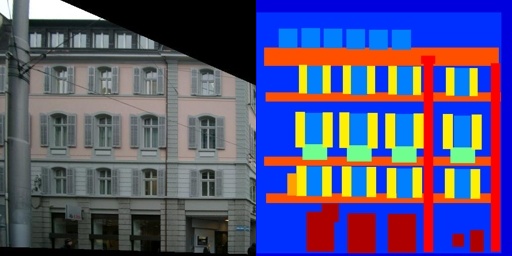}%
        \\
        \includegraphics*[width=0.18\textwidth, viewport=256 0 512 256]{figs/Facades/12.jpg}
        \includegraphics*[width=0.18\textwidth, viewport=0 0 256 256]{figs/Facades/12.jpg}%
        \hfill
        \includegraphics*[width=0.18\textwidth, viewport=0 0 256 256]{figs/Facades/12.jpg}
        \includegraphics*[width=0.18\textwidth, viewport=0 0 256 256]{figs/Facades/5.jpg}
        \includegraphics*[width=0.18\textwidth, viewport=0 0 256 256]{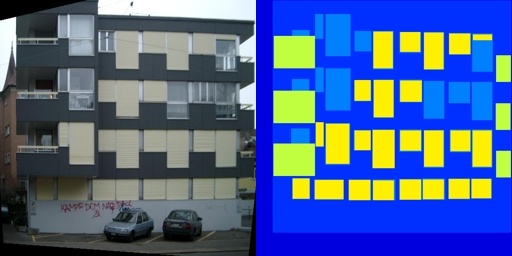}%
        % \label{subfig:GT}
    \end{subfigure}
    \caption{Successful examples of cross-domain retrieval (Top-3) in Facades using IIAE.}
    \label{fig:SuccessCdFacades}
    \vspace{-0.5cm}
\end{figure}

\begin{figure}[H]
    \centering
    \begin{subfigure}[b]{0.49\textwidth}
        \captionsetup{justification=raggedright,singlelinecheck=false}
        \caption*{Query(F) \hspace*{0.16cm} GT(L) \hspace*{2cm} F $\rightarrow$ L}
        \centering
        \includegraphics*[width=0.18\textwidth, viewport=0 0 256 256]{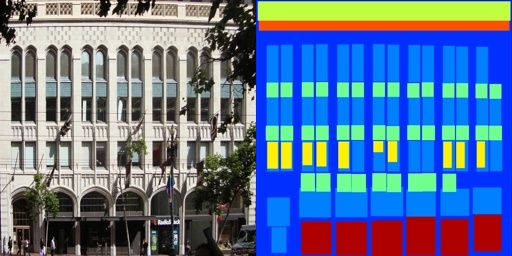}%
        \hspace*{0.02cm}
        \includegraphics*[width=0.18\textwidth, viewport=256 0 512 256]{figs/Facades/4.jpg}
        \hfill
        \includegraphics*[width=0.18\textwidth, viewport=256 0 512 256]{figs/Facades/66.jpg}
        \includegraphics*[width=0.18\textwidth, viewport=256 0 512 256]{figs/Facades/21.jpg}
        \includegraphics*[width=0.18\textwidth, viewport=256 0 512 256]{figs/Facades/14.jpg}%
        \\
        \includegraphics*[width=0.18\textwidth, viewport=0 0 256 256]{figs/Facades/13.jpg}%
        \hspace*{0.02cm}
        \includegraphics*[width=0.18\textwidth, viewport=256 0 512 256]{figs/Facades/13.jpg}
        \hfill
        \includegraphics*[width=0.18\textwidth, viewport=256 0 512 256]{figs/Facades/43.jpg}
        \includegraphics*[width=0.18\textwidth, viewport=256 0 512 256]{figs/Facades/66.jpg}
        \includegraphics*[width=0.18\textwidth, viewport=256 0 512 256]{figs/Facades/14.jpg}%
        \\
        \includegraphics*[width=0.18\textwidth, viewport=0 0 256 256]{figs/Facades/21.jpg}%
        \hspace*{0.02cm}
        \includegraphics*[width=0.18\textwidth, viewport=256 0 512 256]{figs/Facades/21.jpg}
        \hfill
        \includegraphics*[width=0.18\textwidth, viewport=256 0 512 256]{figs/Facades/66.jpg}
        \includegraphics*[width=0.18\textwidth, viewport=256 0 512 256]{figs/Facades/14.jpg}
        \includegraphics*[width=0.18\textwidth, viewport=256 0 512 256]{figs/Facades/13.jpg}%
        \\
        \includegraphics*[width=0.18\textwidth, viewport=0 0 256 256]{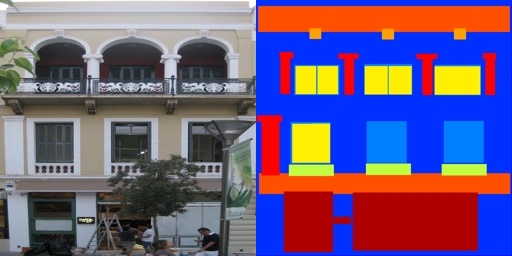}%
        \hspace*{0.02cm}
        \includegraphics*[width=0.18\textwidth, viewport=256 0 512 256]{figs/Facades/81.jpg}
        \hfill
        \includegraphics*[width=0.18\textwidth, viewport=256 0 512 256]{figs/Facades/43.jpg}
        \includegraphics*[width=0.18\textwidth, viewport=256 0 512 256]{figs/Facades/81.jpg}
        \includegraphics*[width=0.18\textwidth, viewport=256 0 512 256]{figs/Facades/14.jpg}%
        % \label{subfig:FL}
    \end{subfigure}
    \vrule
    \hfill
    \begin{subfigure}[b]{0.49\textwidth}
        \captionsetup{justification=raggedright,singlelinecheck=false}
        \caption*{Query(L) \hspace*{0.16cm} GT(F) \hspace*{2cm} L $\rightarrow$ F}
        \centering
        \includegraphics*[width=0.18\textwidth, viewport=256 0 512 256]{figs/Facades/4.jpg}
        \includegraphics*[width=0.18\textwidth, viewport=0 0 256 256]{figs/Facades/4.jpg}%
        \hfill
        \includegraphics*[width=0.18\textwidth, viewport=0 0 256 256]{figs/Facades/4.jpg}
        \includegraphics*[width=0.18\textwidth, viewport=0 0 256 256]{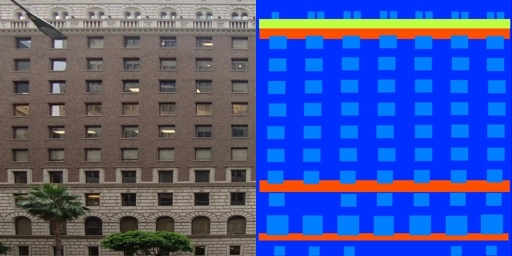}
        \includegraphics*[width=0.18\textwidth, viewport=0 0 256 256]{figs/Facades/21.jpg}%
        \\
        \includegraphics*[width=0.18\textwidth, viewport=256 0 512 256]{figs/Facades/13.jpg}
        \includegraphics*[width=0.18\textwidth, viewport=0 0 256 256]{figs/Facades/13.jpg}%
        \hfill
        \includegraphics*[width=0.18\textwidth, viewport=0 0 256 256]{figs/Facades/43.jpg}
        \includegraphics*[width=0.18\textwidth, viewport=0 0 256 256]{figs/Facades/21.jpg}
        \includegraphics*[width=0.18\textwidth, viewport=0 0 256 256]{figs/Facades/66.jpg}%
        \\
        \includegraphics*[width=0.18\textwidth, viewport=256 0 512 256]{figs/Facades/21.jpg}
        \includegraphics*[width=0.18\textwidth, viewport=0 0 256 256]{figs/Facades/21.jpg}%
        \hfill
        \includegraphics*[width=0.18\textwidth, viewport=0 0 256 256]{figs/Facades/21.jpg}
        \includegraphics*[width=0.18\textwidth, viewport=0 0 256 256]{figs/Facades/66.jpg}
        \includegraphics*[width=0.18\textwidth, viewport=0 0 256 256]{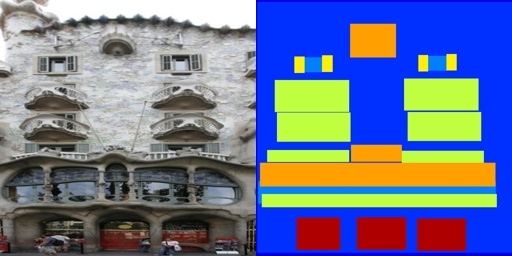}%
        \\
        \includegraphics*[width=0.18\textwidth, viewport=256 0 512 256]{figs/Facades/81.jpg}
        \includegraphics*[width=0.18\textwidth, viewport=0 0 256 256]{figs/Facades/81.jpg}%
        \hfill
        \includegraphics*[width=0.18\textwidth, viewport=0 0 256 256]{figs/Facades/81.jpg}
        \includegraphics*[width=0.18\textwidth, viewport=0 0 256 256]{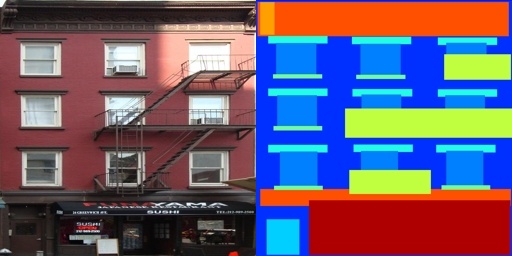}
        \includegraphics*[width=0.18\textwidth, viewport=0 0 256 256]{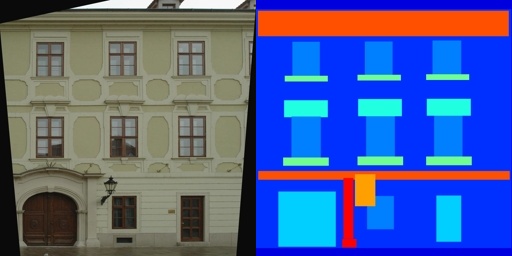}%
        % \label{subfig:LF}
    \end{subfigure}
    \caption{All unsuccessful cases of cross-domain retrieval (Top-3) in Facades using IIAE.}
    \label{fig:FailCdFacades}
\end{figure}

\vspace*{-0.5em}
\subsection{ZS-SBIR}
\label{appendix:ZS-SBIR}
\begin{figure}[H]
    \centering
    \includegraphics*[width=0.085\linewidth, height=0.085\linewidth]{figs/ZS_SBIR/bell/q.jpg}
    \includegraphics*[width=0.085\linewidth, height=0.085\linewidth]{figs/ZS_SBIR/bell/0.jpg}
    \includegraphics*[width=0.085\linewidth, height=0.085\linewidth]{figs/ZS_SBIR/bell/1.jpg}
    \includegraphics*[width=0.085\linewidth, height=0.085\linewidth]{figs/ZS_SBIR/bell/2.jpg}
    \includegraphics*[width=0.085\linewidth, height=0.085\linewidth]{figs/ZS_SBIR/bell/3.jpg}
    \includegraphics*[width=0.085\linewidth, height=0.085\linewidth]{figs/ZS_SBIR/bell/4.jpg}
    \includegraphics*[width=0.085\linewidth, height=0.085\linewidth]{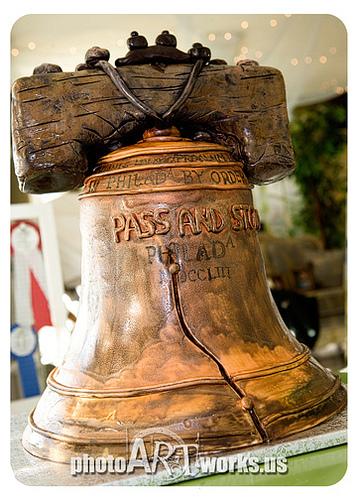}
    \includegraphics*[width=0.085\linewidth, height=0.085\linewidth]{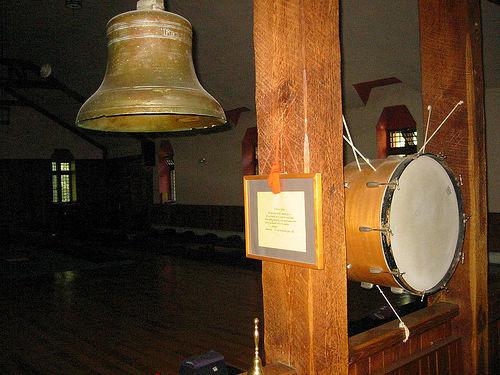}
    \includegraphics*[width=0.085\linewidth, height=0.085\linewidth]{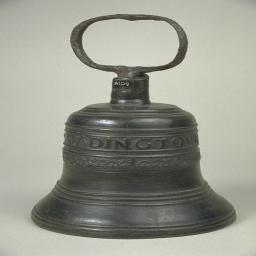}
    \includegraphics*[width=0.085\linewidth, height=0.085\linewidth]{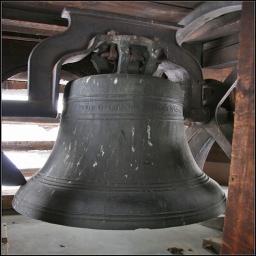}
    \includegraphics*[width=0.085\linewidth, height=0.085\linewidth]{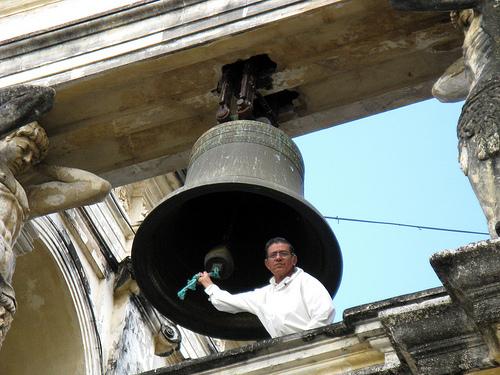}
    \hfill\\
    \vspace{-0.3em}
    \hspace*{0.085\linewidth}
    \includegraphics*[width=0.085\linewidth, height=\fontcharht\font`\B]{figs/gc}
    \includegraphics*[width=0.085\linewidth, height=\fontcharht\font`\B]{figs/gc}
    \includegraphics*[width=0.085\linewidth, height=\fontcharht\font`\B]{figs/gc}
    \includegraphics*[width=0.085\linewidth, height=\fontcharht\font`\B]{figs/gc}
    \includegraphics*[width=0.085\linewidth, height=\fontcharht\font`\B]{figs/gc}
    \includegraphics*[width=0.085\linewidth, height=\fontcharht\font`\B]{figs/gc}
    \includegraphics*[width=0.085\linewidth, height=\fontcharht\font`\B]{figs/gc}
    \includegraphics*[width=0.085\linewidth, height=\fontcharht\font`\B]{figs/gc}
    \includegraphics*[width=0.085\linewidth, height=\fontcharht\font`\B]{figs/gc}
    \includegraphics*[width=0.085\linewidth, height=\fontcharht\font`\B]{figs/gc}
    \hfill\\%
    \vspace{0.3em}
    \includegraphics*[width=0.085\linewidth, height=0.085\linewidth]{figs/ZS_SBIR/cannon/q.jpg}
    \includegraphics*[width=0.085\linewidth, height=0.085\linewidth]{figs/ZS_SBIR/cannon/0.jpg}
    \includegraphics*[width=0.085\linewidth, height=0.085\linewidth]{figs/ZS_SBIR/cannon/1_f.jpg}
    \includegraphics*[width=0.085\linewidth, height=0.085\linewidth]{figs/ZS_SBIR/cannon/2_f.jpg}
    \includegraphics*[width=0.085\linewidth, height=0.085\linewidth]{figs/ZS_SBIR/cannon/3.jpg}
    \includegraphics*[width=0.085\linewidth, height=0.085\linewidth]{figs/ZS_SBIR/cannon/4.jpg}
    \includegraphics*[width=0.085\linewidth, height=0.085\linewidth]{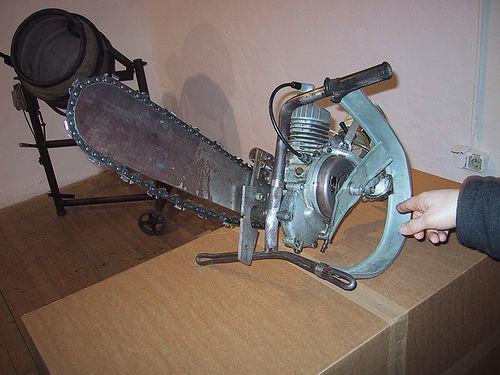}
    \includegraphics*[width=0.085\linewidth, height=0.085\linewidth]{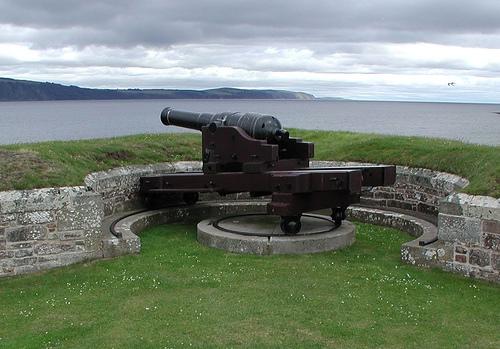}
    \includegraphics*[width=0.085\linewidth, height=0.085\linewidth]{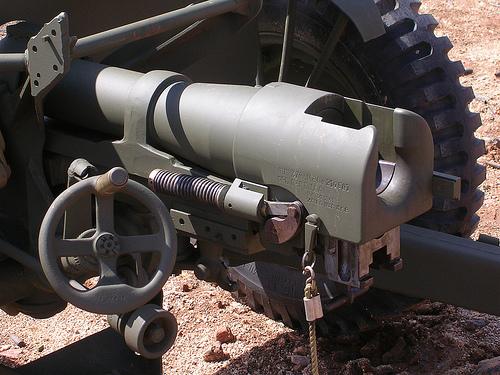}
    \includegraphics*[width=0.085\linewidth, height=0.085\linewidth]{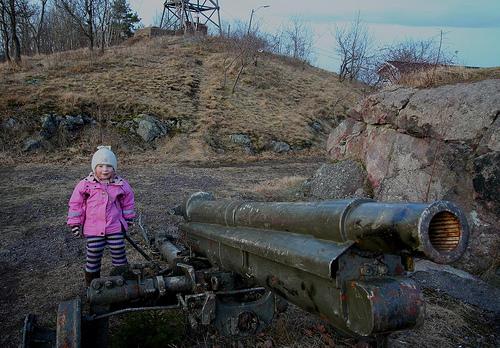}
    \includegraphics*[width=0.085\linewidth, height=0.085\linewidth]{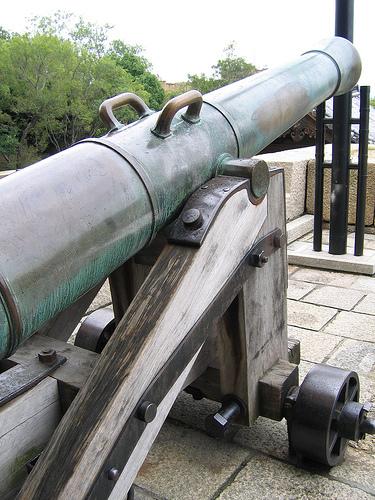}
    \hfill\\
    \vspace{-0.3em}
    \hspace*{0.085\linewidth}
    \includegraphics*[width=0.085\linewidth, height=\fontcharht\font`\B]{figs/gc}
    \includegraphics*[width=0.085\linewidth, height=\fontcharht\font`\B]{figs/rc}
    \includegraphics*[width=0.085\linewidth, height=\fontcharht\font`\B]{figs/rc}
    \includegraphics*[width=0.085\linewidth, height=\fontcharht\font`\B]{figs/gc}
    \includegraphics*[width=0.085\linewidth, height=\fontcharht\font`\B]{figs/gc}
    \includegraphics*[width=0.085\linewidth, height=\fontcharht\font`\B]{figs/rc}
    \includegraphics*[width=0.085\linewidth, height=\fontcharht\font`\B]{figs/gc}
    \includegraphics*[width=0.085\linewidth, height=\fontcharht\font`\B]{figs/gc}
    \includegraphics*[width=0.085\linewidth, height=\fontcharht\font`\B]{figs/gc}
    \includegraphics*[width=0.085\linewidth, height=\fontcharht\font`\B]{figs/gc}
    \hfill\\%
    \vspace{0.3em}
    \includegraphics*[width=0.085\linewidth, height=0.085\linewidth]{figs/ZS_SBIR/door/q.jpg}
    \includegraphics*[width=0.085\linewidth, height=0.085\linewidth]{figs/ZS_SBIR/door/0.jpg}
    \includegraphics*[width=0.085\linewidth, height=0.085\linewidth]{figs/ZS_SBIR/door/1.jpg}
    \includegraphics*[width=0.085\linewidth, height=0.085\linewidth]{figs/ZS_SBIR/door/2_f.jpg}
    \includegraphics*[width=0.085\linewidth, height=0.085\linewidth]{figs/ZS_SBIR/door/3.jpg}
    \includegraphics*[width=0.085\linewidth, height=0.085\linewidth]{figs/ZS_SBIR/door/4.jpg}
    \includegraphics*[width=0.085\linewidth, height=0.085\linewidth]{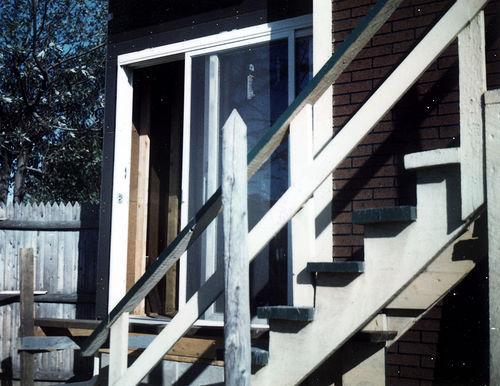}
    \includegraphics*[width=0.085\linewidth, height=0.085\linewidth]{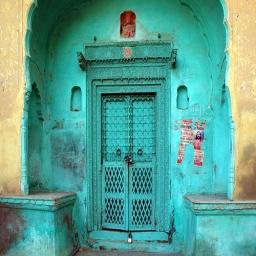}
    \includegraphics*[width=0.085\linewidth, height=0.085\linewidth]{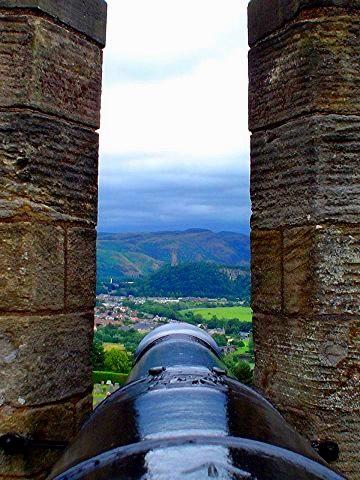}
    \includegraphics*[width=0.085\linewidth, height=0.085\linewidth]{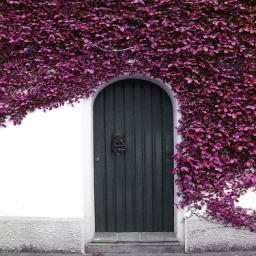}
    \includegraphics*[width=0.085\linewidth, height=0.085\linewidth]{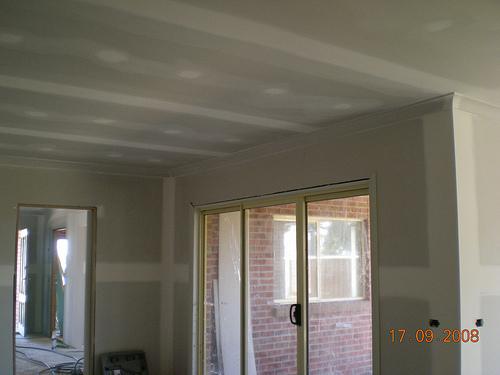}
    \hfill\\
    \vspace{-0.3em}
    \hspace*{0.085\linewidth}
    \includegraphics*[width=0.085\linewidth, height=\fontcharht\font`\B]{figs/gc}
    \includegraphics*[width=0.085\linewidth, height=\fontcharht\font`\B]{figs/gc}
    \includegraphics*[width=0.085\linewidth, height=\fontcharht\font`\B]{figs/rc}
    \includegraphics*[width=0.085\linewidth, height=\fontcharht\font`\B]{figs/gc}
    \includegraphics*[width=0.085\linewidth, height=\fontcharht\font`\B]{figs/gc}
    \includegraphics*[width=0.085\linewidth, height=\fontcharht\font`\B]{figs/gc}
    \includegraphics*[width=0.085\linewidth, height=\fontcharht\font`\B]{figs/gc}
    \includegraphics*[width=0.085\linewidth, height=\fontcharht\font`\B]{figs/rc}
    \includegraphics*[width=0.085\linewidth, height=\fontcharht\font`\B]{figs/gc}
    \includegraphics*[width=0.085\linewidth, height=\fontcharht\font`\B]{figs/gc}
    \hfill\\%
    \vspace{0.3em}
    \includegraphics*[width=0.085\linewidth, height=0.085\linewidth]{figs/ZS_SBIR/eyeglasses/q.jpg}
    \includegraphics*[width=0.085\linewidth, height=0.085\linewidth]{figs/ZS_SBIR/eyeglasses/0.jpg}
    \includegraphics*[width=0.085\linewidth, height=0.085\linewidth]{figs/ZS_SBIR/eyeglasses/1.jpg}
    \includegraphics*[width=0.085\linewidth, height=0.085\linewidth]{figs/ZS_SBIR/eyeglasses/2.jpg}
    \includegraphics*[width=0.085\linewidth, height=0.085\linewidth]{figs/ZS_SBIR/eyeglasses/3.jpg}
    \includegraphics*[width=0.085\linewidth, height=0.085\linewidth]{figs/ZS_SBIR/eyeglasses/4.jpg}
    \includegraphics*[width=0.085\linewidth, height=0.085\linewidth]{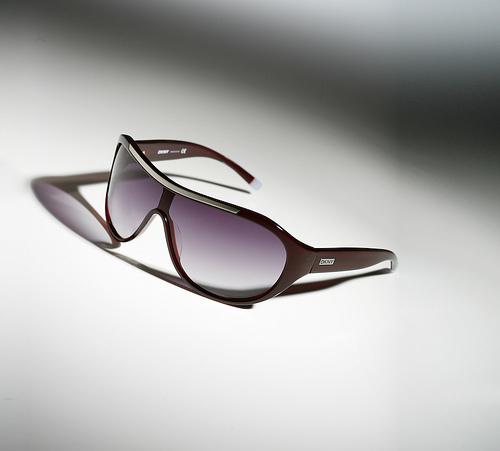}
    \includegraphics*[width=0.085\linewidth, height=0.085\linewidth]{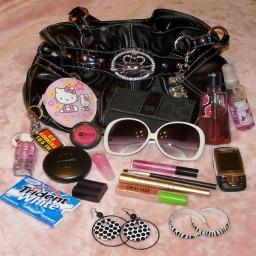}
    \includegraphics*[width=0.085\linewidth, height=0.085\linewidth]{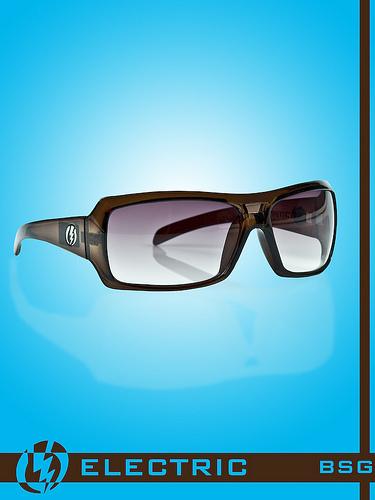}
    \includegraphics*[width=0.085\linewidth, height=0.085\linewidth]{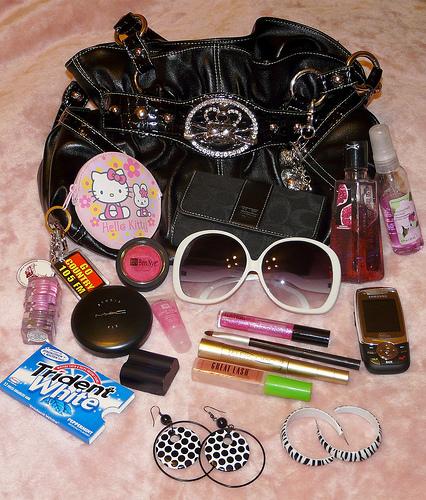}
    \includegraphics*[width=0.085\linewidth, height=0.085\linewidth]{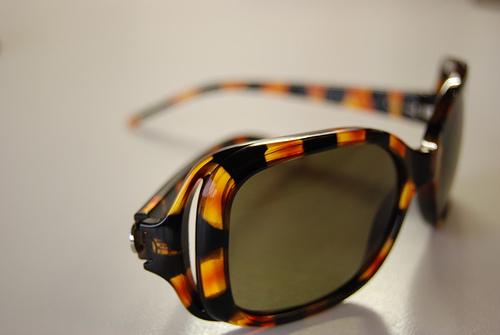}
    \hfill\\
    \vspace{-0.3em}
    \hspace*{0.085\linewidth}
    \includegraphics*[width=0.085\linewidth, height=\fontcharht\font`\B]{figs/gc}
    \includegraphics*[width=0.085\linewidth, height=\fontcharht\font`\B]{figs/gc}
    \includegraphics*[width=0.085\linewidth, height=\fontcharht\font`\B]{figs/gc}
    \includegraphics*[width=0.085\linewidth, height=\fontcharht\font`\B]{figs/gc}
    \includegraphics*[width=0.085\linewidth, height=\fontcharht\font`\B]{figs/gc}
    \includegraphics*[width=0.085\linewidth, height=\fontcharht\font`\B]{figs/gc}
    \includegraphics*[width=0.085\linewidth, height=\fontcharht\font`\B]{figs/gc}
    \includegraphics*[width=0.085\linewidth, height=\fontcharht\font`\B]{figs/gc}
    \includegraphics*[width=0.085\linewidth, height=\fontcharht\font`\B]{figs/gc}
    \includegraphics*[width=0.085\linewidth, height=\fontcharht\font`\B]{figs/gc}
    \hfill\\%
    \vspace{0.3em}
    \includegraphics*[width=0.085\linewidth, height=0.085\linewidth]{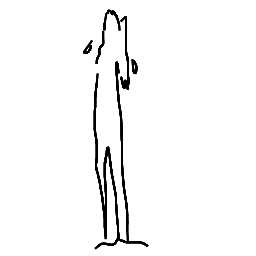}
    \includegraphics*[width=0.085\linewidth, height=0.085\linewidth]{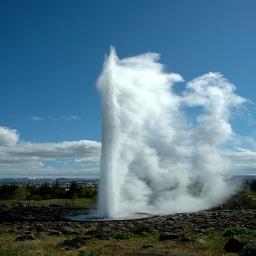}
    \includegraphics*[width=0.085\linewidth, height=0.085\linewidth]{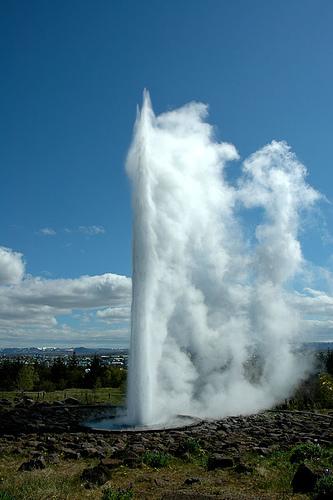}
    \includegraphics*[width=0.085\linewidth, height=0.085\linewidth]{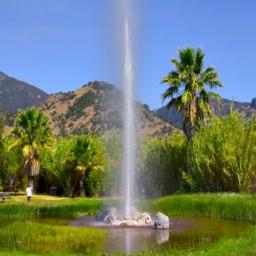}
    \includegraphics*[width=0.085\linewidth, height=0.085\linewidth]{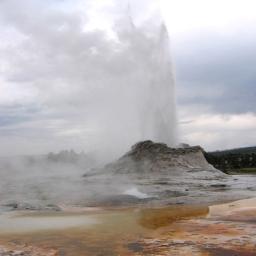}
    \includegraphics*[width=0.085\linewidth, height=0.085\linewidth]{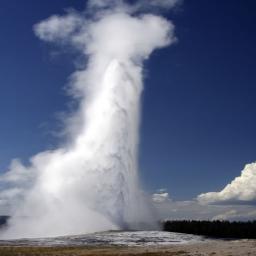}
    \includegraphics*[width=0.085\linewidth, height=0.085\linewidth]{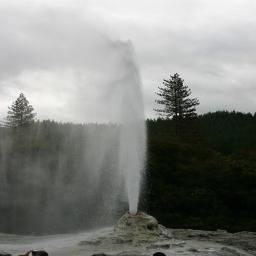}
    \includegraphics*[width=0.085\linewidth, height=0.085\linewidth]{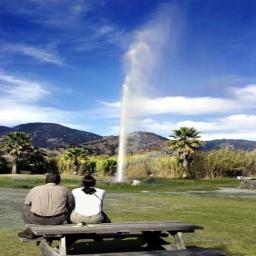}
    \includegraphics*[width=0.085\linewidth, height=0.085\linewidth]{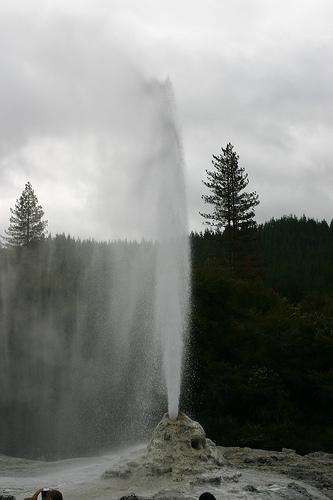}
    \includegraphics*[width=0.085\linewidth, height=0.085\linewidth]{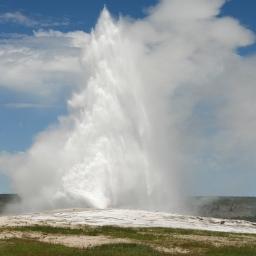}
    \includegraphics*[width=0.085\linewidth, height=0.085\linewidth]{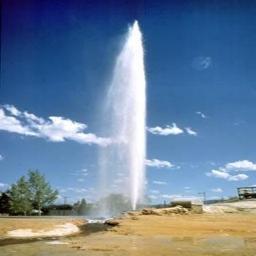}
    \hfill\\
    \vspace{-0.3em}
    \hspace*{0.085\linewidth}
    \includegraphics*[width=0.085\linewidth, height=\fontcharht\font`\B]{figs/gc}
    \includegraphics*[width=0.085\linewidth, height=\fontcharht\font`\B]{figs/gc}
    \includegraphics*[width=0.085\linewidth, height=\fontcharht\font`\B]{figs/gc}
    \includegraphics*[width=0.085\linewidth, height=\fontcharht\font`\B]{figs/gc}
    \includegraphics*[width=0.085\linewidth, height=\fontcharht\font`\B]{figs/gc}
    \includegraphics*[width=0.085\linewidth, height=\fontcharht\font`\B]{figs/gc}
    \includegraphics*[width=0.085\linewidth, height=\fontcharht\font`\B]{figs/gc}
    \includegraphics*[width=0.085\linewidth, height=\fontcharht\font`\B]{figs/gc}
    \includegraphics*[width=0.085\linewidth, height=\fontcharht\font`\B]{figs/gc}
    \includegraphics*[width=0.085\linewidth, height=\fontcharht\font`\B]{figs/gc}
    \hfill\\%
    \vspace{0.3em}
    \includegraphics*[width=0.085\linewidth, height=0.085\linewidth]{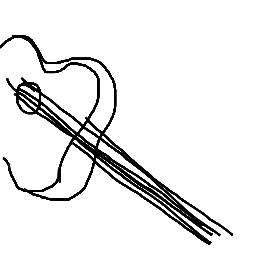}
    \includegraphics*[width=0.085\linewidth, height=0.085\linewidth]{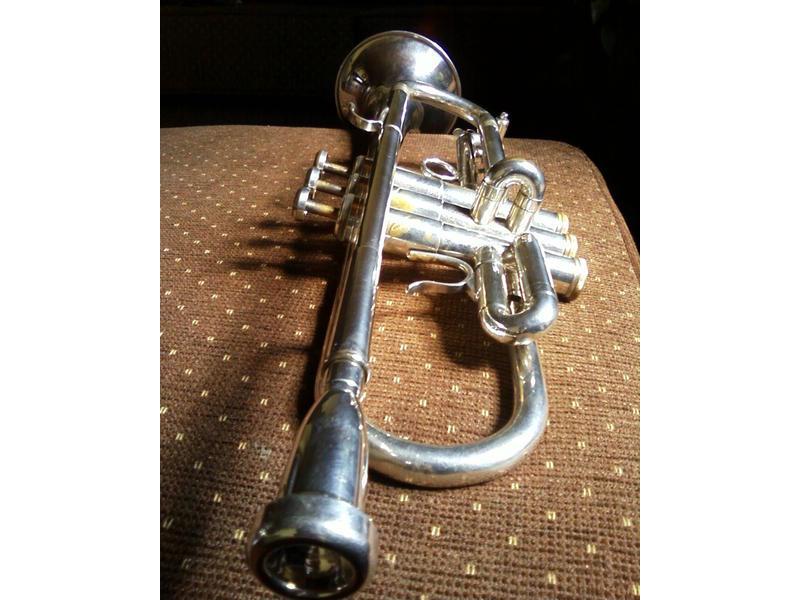}
    \includegraphics*[width=0.085\linewidth, height=0.085\linewidth]{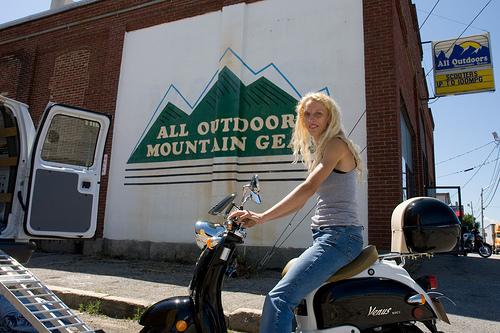}
    \includegraphics*[width=0.085\linewidth, height=0.085\linewidth]{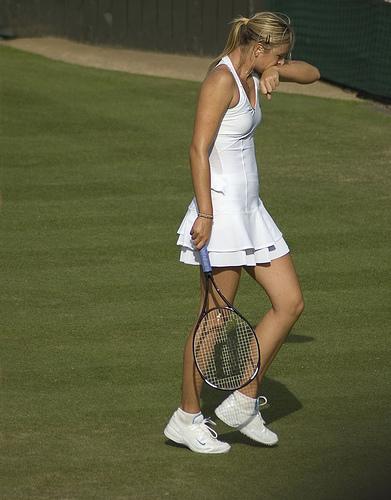}
    \includegraphics*[width=0.085\linewidth, height=0.085\linewidth]{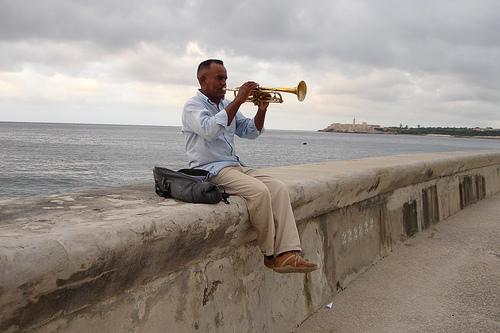}
    \includegraphics*[width=0.085\linewidth, height=0.085\linewidth]{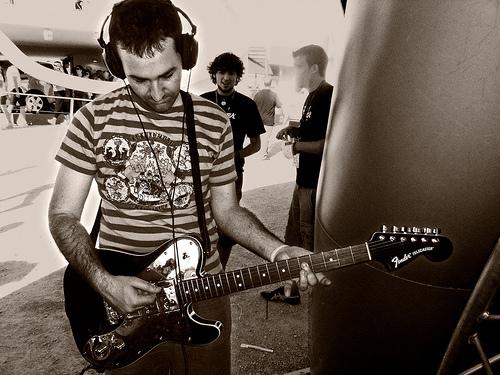}
    \includegraphics*[width=0.085\linewidth, height=0.085\linewidth]{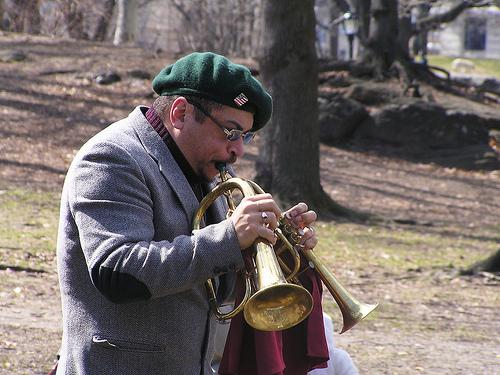}
    \includegraphics*[width=0.085\linewidth, height=0.085\linewidth]{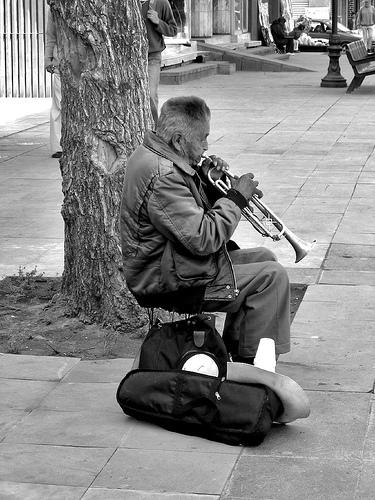}
    \includegraphics*[width=0.085\linewidth, height=0.085\linewidth]{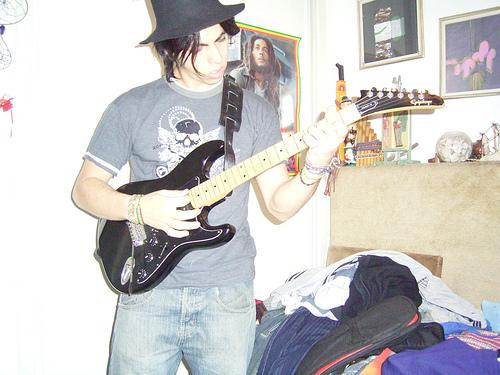}
    \includegraphics*[width=0.085\linewidth, height=0.085\linewidth]{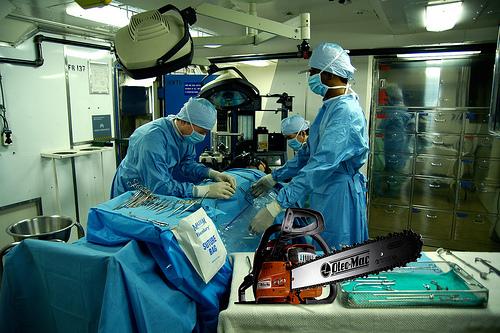}
    \includegraphics*[width=0.085\linewidth, height=0.085\linewidth]{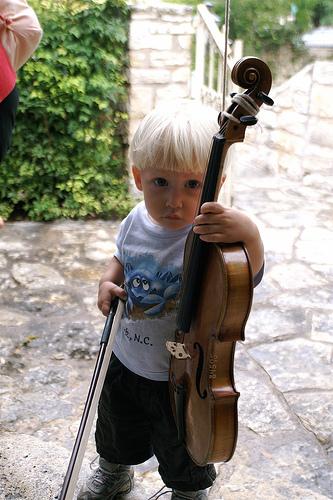}
    \hfill\\
    \vspace{-0.3em}
    \hspace*{0.085\linewidth}
    \includegraphics*[width=0.085\linewidth, height=\fontcharht\font`\B]{figs/rc}
    \includegraphics*[width=0.085\linewidth, height=\fontcharht\font`\B]{figs/rc}
    \includegraphics*[width=0.085\linewidth, height=\fontcharht\font`\B]{figs/rc}
    \includegraphics*[width=0.085\linewidth, height=\fontcharht\font`\B]{figs/rc}
    \includegraphics*[width=0.085\linewidth, height=\fontcharht\font`\B]{figs/gc}
    \includegraphics*[width=0.085\linewidth, height=\fontcharht\font`\B]{figs/rc}
    \includegraphics*[width=0.085\linewidth, height=\fontcharht\font`\B]{figs/rc}
    \includegraphics*[width=0.085\linewidth, height=\fontcharht\font`\B]{figs/gc}
    \includegraphics*[width=0.085\linewidth, height=\fontcharht\font`\B]{figs/rc}
    \includegraphics*[width=0.085\linewidth, height=\fontcharht\font`\B]{figs/rc}
    \hfill\\%
    \vspace{0.3em}
    \includegraphics*[width=0.085\linewidth, height=0.085\linewidth]{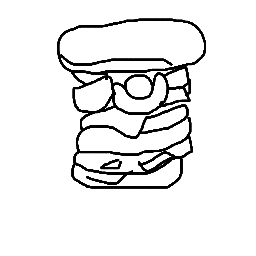}
    \includegraphics*[width=0.085\linewidth, height=0.085\linewidth]{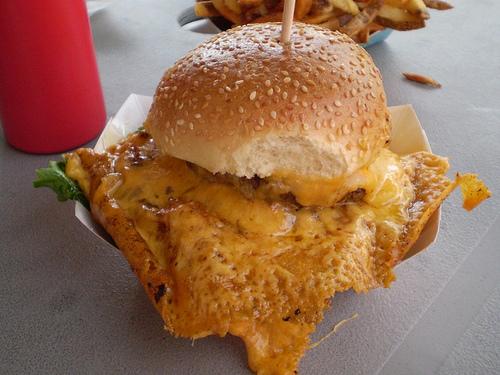}
    \includegraphics*[width=0.085\linewidth, height=0.085\linewidth]{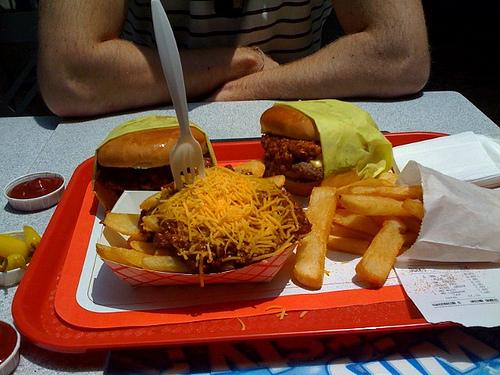}
    \includegraphics*[width=0.085\linewidth, height=0.085\linewidth]{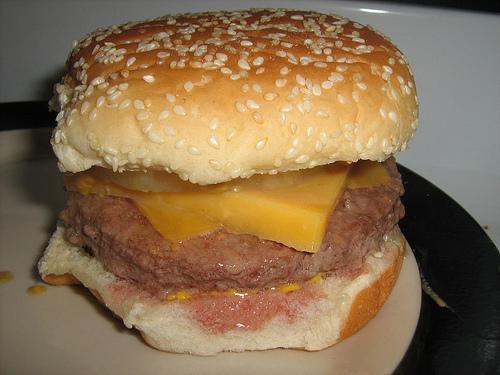}
    \includegraphics*[width=0.085\linewidth, height=0.085\linewidth]{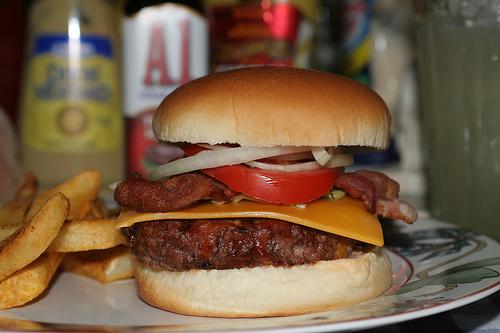}
    \includegraphics*[width=0.085\linewidth, height=0.085\linewidth]{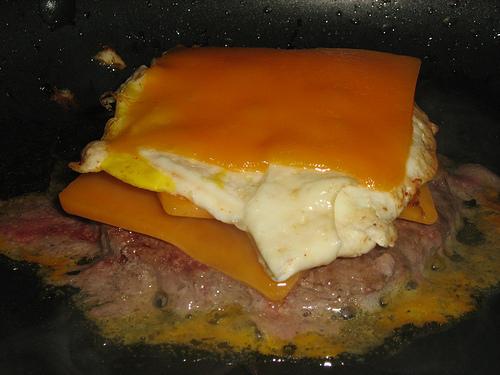}
    \includegraphics*[width=0.085\linewidth, height=0.085\linewidth]{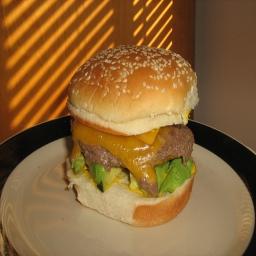}
    \includegraphics*[width=0.085\linewidth, height=0.085\linewidth]{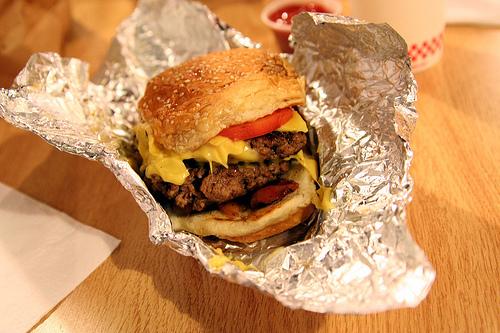}
    \includegraphics*[width=0.085\linewidth, height=0.085\linewidth]{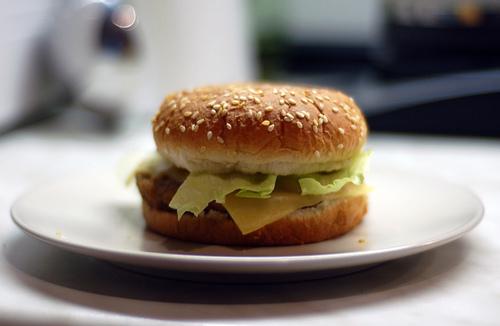}
    \includegraphics*[width=0.085\linewidth, height=0.085\linewidth]{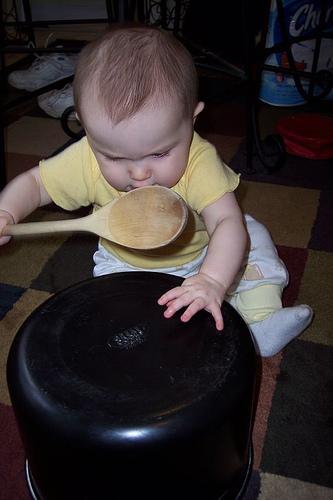}
    \includegraphics*[width=0.085\linewidth, height=0.085\linewidth]{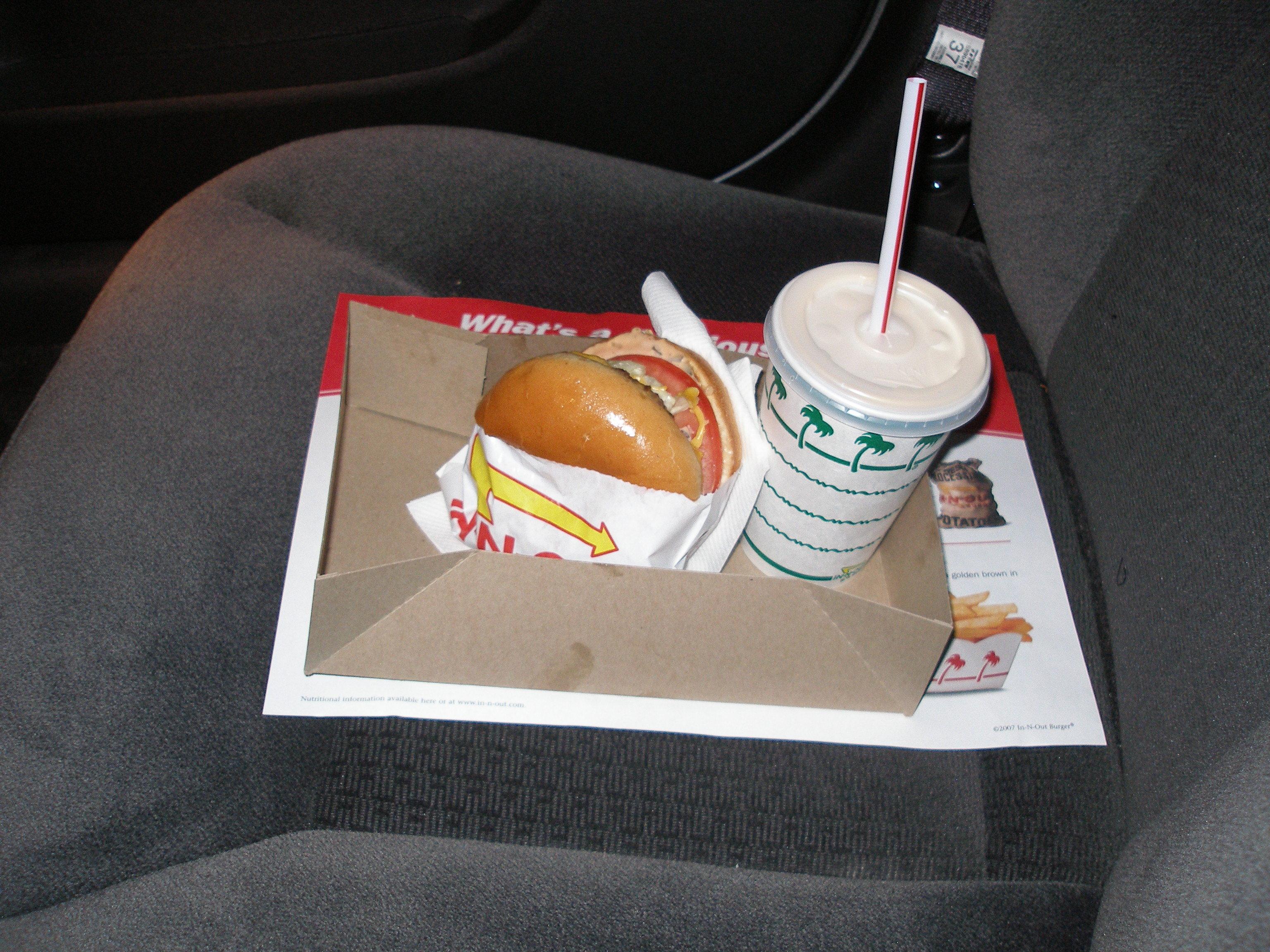}
    \hfill\\
    \vspace{-0.3em}
    \hspace*{0.085\linewidth}
    \includegraphics*[width=0.085\linewidth, height=\fontcharht\font`\B]{figs/gc}
    \includegraphics*[width=0.085\linewidth, height=\fontcharht\font`\B]{figs/gc}
    \includegraphics*[width=0.085\linewidth, height=\fontcharht\font`\B]{figs/gc}
    \includegraphics*[width=0.085\linewidth, height=\fontcharht\font`\B]{figs/gc}
    \includegraphics*[width=0.085\linewidth, height=\fontcharht\font`\B]{figs/gc}
    \includegraphics*[width=0.085\linewidth, height=\fontcharht\font`\B]{figs/gc}
    \includegraphics*[width=0.085\linewidth, height=\fontcharht\font`\B]{figs/gc}
    \includegraphics*[width=0.085\linewidth, height=\fontcharht\font`\B]{figs/gc}
    \includegraphics*[width=0.085\linewidth, height=\fontcharht\font`\B]{figs/rc}
    \includegraphics*[width=0.085\linewidth, height=\fontcharht\font`\B]{figs/gc}
    \hfill\\%
    \vspace{0.3em}
    \includegraphics*[width=0.085\linewidth, height=0.085\linewidth]{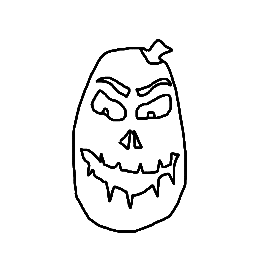}
    \includegraphics*[width=0.085\linewidth, height=0.085\linewidth]{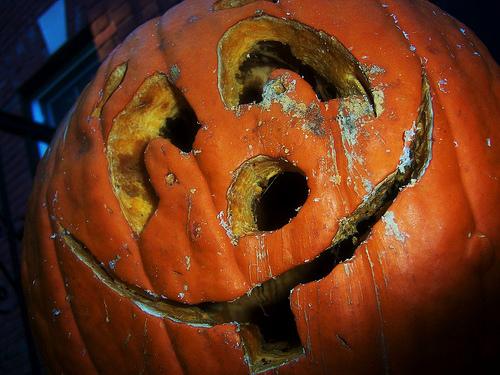}
    \includegraphics*[width=0.085\linewidth, height=0.085\linewidth]{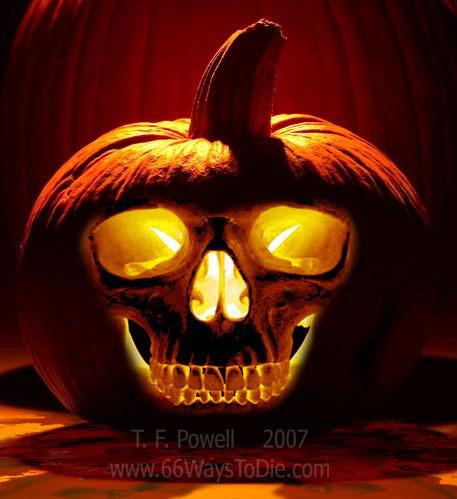}
    \includegraphics*[width=0.085\linewidth, height=0.085\linewidth]{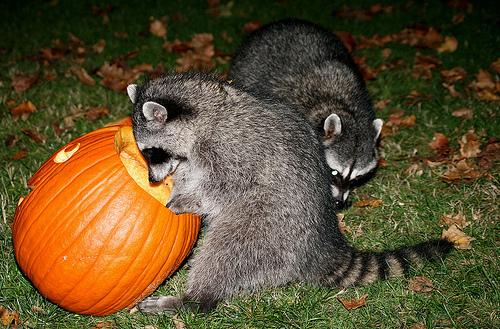}
    \includegraphics*[width=0.085\linewidth, height=0.085\linewidth]{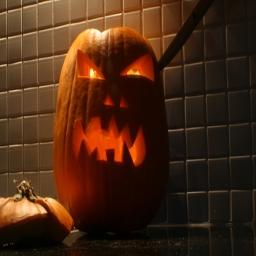}
    \includegraphics*[width=0.085\linewidth, height=0.085\linewidth]{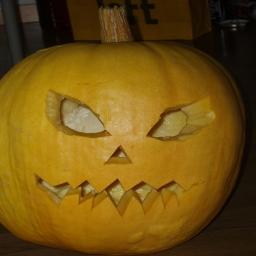}
    \includegraphics*[width=0.085\linewidth, height=0.085\linewidth]{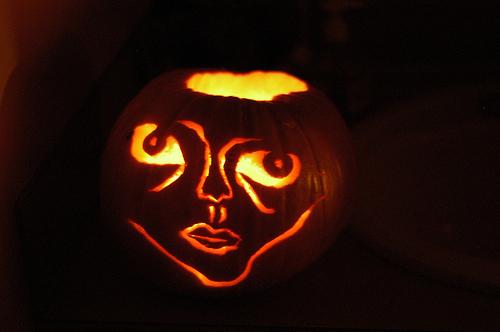}
    \includegraphics*[width=0.085\linewidth, height=0.085\linewidth]{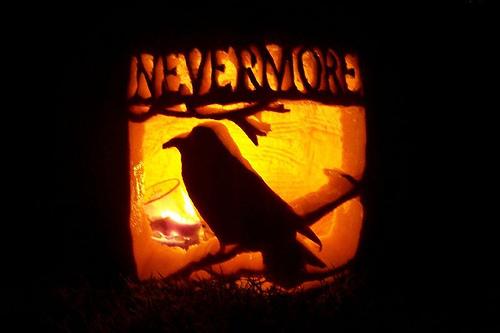}
    \includegraphics*[width=0.085\linewidth, height=0.085\linewidth]{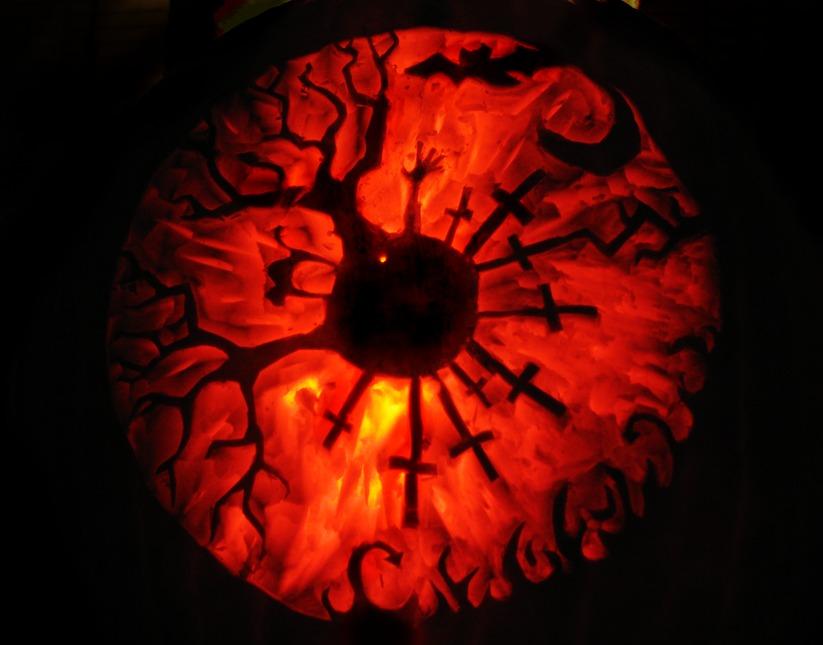}
    \includegraphics*[width=0.085\linewidth, height=0.085\linewidth]{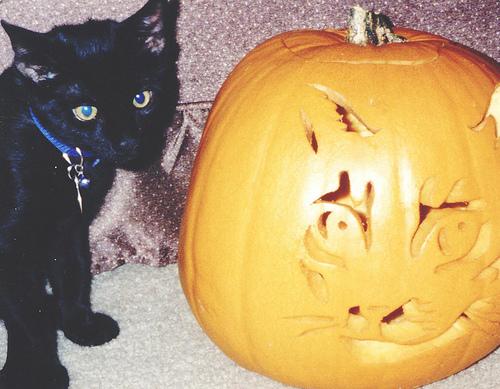}
    \includegraphics*[width=0.085\linewidth, height=0.085\linewidth]{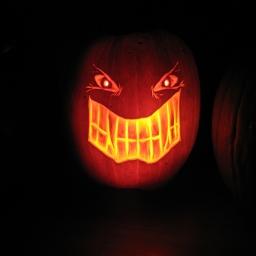}
    \hfill\\
    \vspace{-0.3em}
    \hspace*{0.085\linewidth}
    \includegraphics*[width=0.085\linewidth, height=\fontcharht\font`\B]{figs/gc}
    \includegraphics*[width=0.085\linewidth, height=\fontcharht\font`\B]{figs/gc}
    \includegraphics*[width=0.085\linewidth, height=\fontcharht\font`\B]{figs/gc}
    \includegraphics*[width=0.085\linewidth, height=\fontcharht\font`\B]{figs/gc}
    \includegraphics*[width=0.085\linewidth, height=\fontcharht\font`\B]{figs/gc}
    \includegraphics*[width=0.085\linewidth, height=\fontcharht\font`\B]{figs/gc}
    \includegraphics*[width=0.085\linewidth, height=\fontcharht\font`\B]{figs/gc}
    \includegraphics*[width=0.085\linewidth, height=\fontcharht\font`\B]{figs/gc}
    \includegraphics*[width=0.085\linewidth, height=\fontcharht\font`\B]{figs/gc}
    \includegraphics*[width=0.085\linewidth, height=\fontcharht\font`\B]{figs/gc}
    \hfill\\%
    \vspace{0.3em}
    \includegraphics*[width=0.085\linewidth, height=0.085\linewidth]{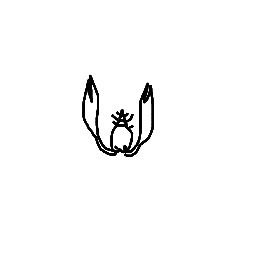}
    \includegraphics*[width=0.085\linewidth, height=0.085\linewidth]{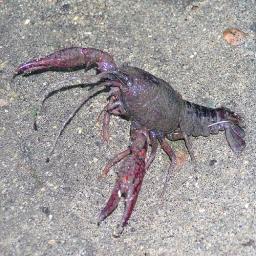}
    \includegraphics*[width=0.085\linewidth, height=0.085\linewidth]{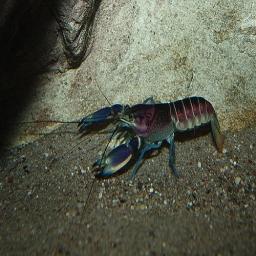}
    \includegraphics*[width=0.085\linewidth, height=0.085\linewidth]{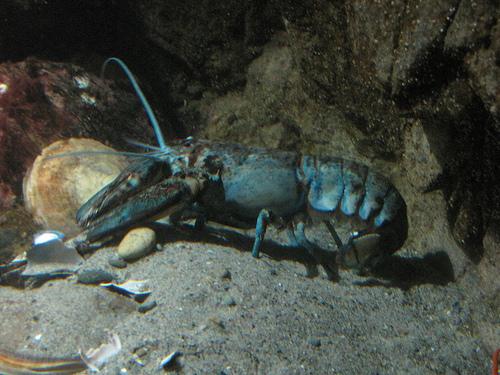}
    \includegraphics*[width=0.085\linewidth, height=0.085\linewidth]{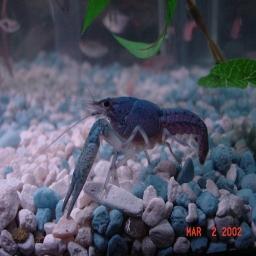}
    \includegraphics*[width=0.085\linewidth, height=0.085\linewidth]{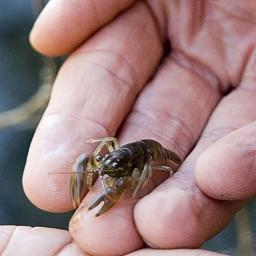}
    \includegraphics*[width=0.085\linewidth, height=0.085\linewidth]{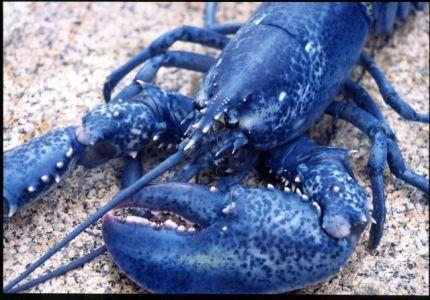}
    \includegraphics*[width=0.085\linewidth, height=0.085\linewidth]{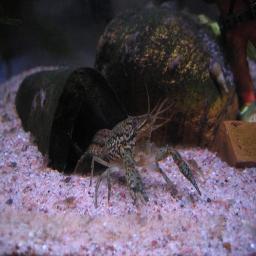}
    \includegraphics*[width=0.085\linewidth, height=0.085\linewidth]{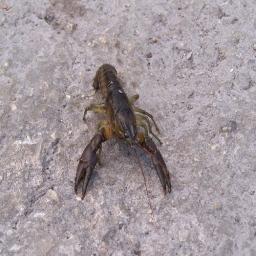}
    \includegraphics*[width=0.085\linewidth, height=0.085\linewidth]{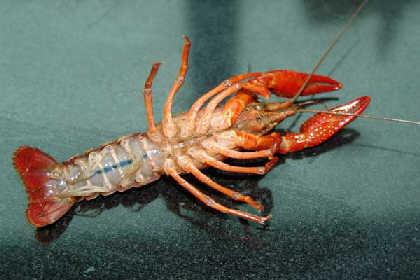}
    \includegraphics*[width=0.085\linewidth, height=0.085\linewidth]{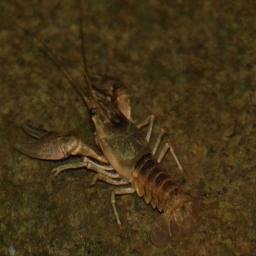}
    \hfill\\
    \vspace{-0.3em}
    \hspace*{0.085\linewidth}
    \includegraphics*[width=0.085\linewidth, height=\fontcharht\font`\B]{figs/gc}
    \includegraphics*[width=0.085\linewidth, height=\fontcharht\font`\B]{figs/gc}
    \includegraphics*[width=0.085\linewidth, height=\fontcharht\font`\B]{figs/gc}
    \includegraphics*[width=0.085\linewidth, height=\fontcharht\font`\B]{figs/gc}
    \includegraphics*[width=0.085\linewidth, height=\fontcharht\font`\B]{figs/gc}
    \includegraphics*[width=0.085\linewidth, height=\fontcharht\font`\B]{figs/gc}
    \includegraphics*[width=0.085\linewidth, height=\fontcharht\font`\B]{figs/gc}
    \includegraphics*[width=0.085\linewidth, height=\fontcharht\font`\B]{figs/gc}
    \includegraphics*[width=0.085\linewidth, height=\fontcharht\font`\B]{figs/gc}
    \includegraphics*[width=0.085\linewidth, height=\fontcharht\font`\B]{figs/gc}
    \hfill\\%
    \vspace{0.3em}
    \includegraphics*[width=0.085\linewidth, height=0.085\linewidth]{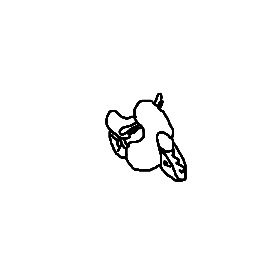}
    \includegraphics*[width=0.085\linewidth, height=0.085\linewidth]{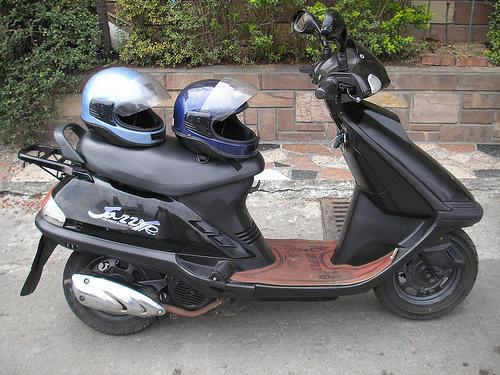}
    \includegraphics*[width=0.085\linewidth, height=0.085\linewidth]{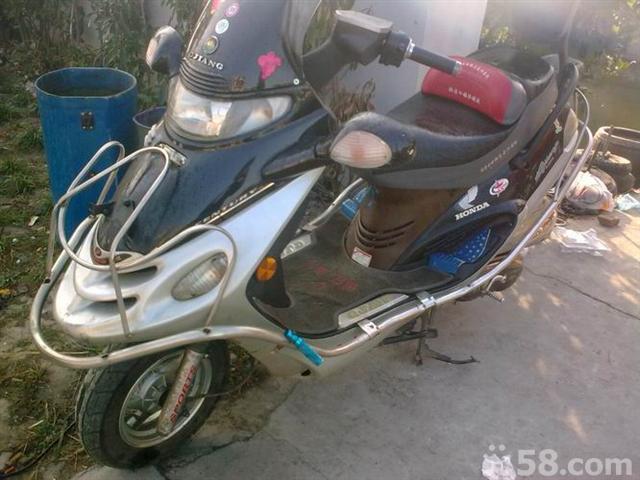}
    \includegraphics*[width=0.085\linewidth, height=0.085\linewidth]{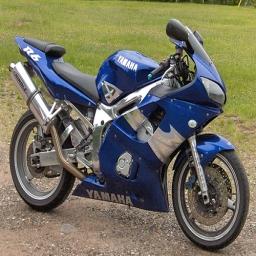}
    \includegraphics*[width=0.085\linewidth, height=0.085\linewidth]{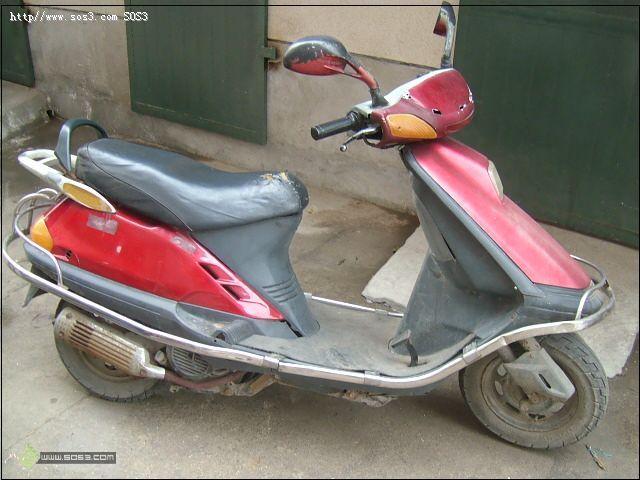}
    \includegraphics*[width=0.085\linewidth, height=0.085\linewidth]{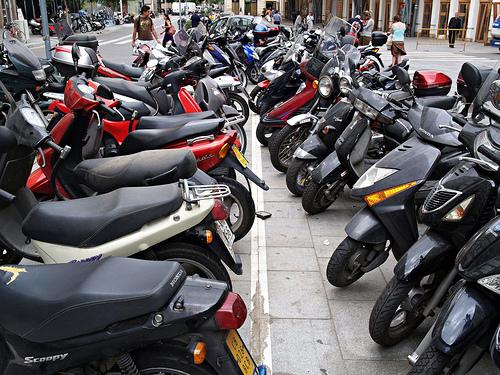}
    \includegraphics*[width=0.085\linewidth, height=0.085\linewidth]{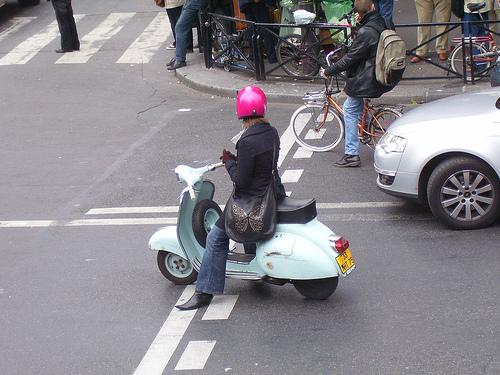}
    \includegraphics*[width=0.085\linewidth, height=0.085\linewidth]{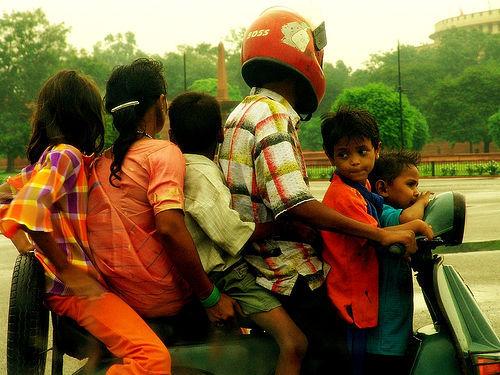}
    \includegraphics*[width=0.085\linewidth, height=0.085\linewidth]{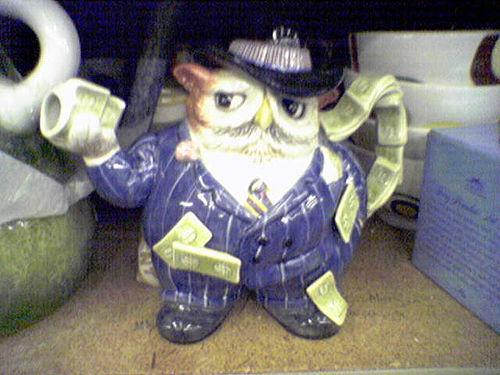}
    \includegraphics*[width=0.085\linewidth, height=0.085\linewidth]{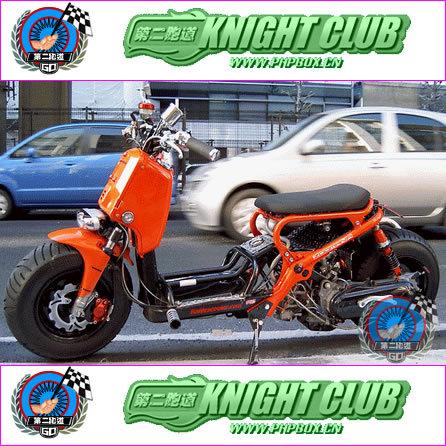}
    \includegraphics*[width=0.085\linewidth, height=0.085\linewidth]{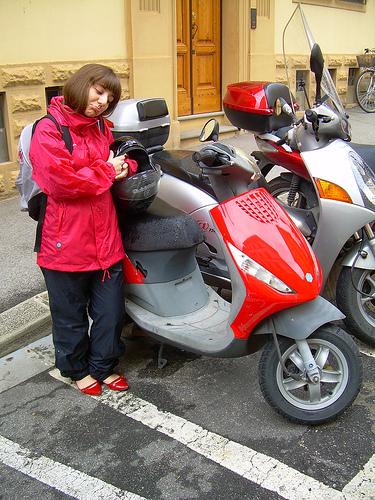}
    \hfill\\
    \vspace{-0.3em}
    \hspace*{0.085\linewidth}
    \includegraphics*[width=0.085\linewidth, height=\fontcharht\font`\B]{figs/gc}
    \includegraphics*[width=0.085\linewidth, height=\fontcharht\font`\B]{figs/gc}
    \includegraphics*[width=0.085\linewidth, height=\fontcharht\font`\B]{figs/gc}
    \includegraphics*[width=0.085\linewidth, height=\fontcharht\font`\B]{figs/gc}
    \includegraphics*[width=0.085\linewidth, height=\fontcharht\font`\B]{figs/gc}
    \includegraphics*[width=0.085\linewidth, height=\fontcharht\font`\B]{figs/gc}
    \includegraphics*[width=0.085\linewidth, height=\fontcharht\font`\B]{figs/gc}
    \includegraphics*[width=0.085\linewidth, height=\fontcharht\font`\B]{figs/rc}
    \includegraphics*[width=0.085\linewidth, height=\fontcharht\font`\B]{figs/gc}
    \includegraphics*[width=0.085\linewidth, height=\fontcharht\font`\B]{figs/gc}
    \hfill\\%
    \vspace{0.3em}
    \includegraphics*[width=0.085\linewidth, height=0.085\linewidth]{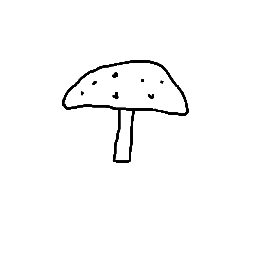}
    \includegraphics*[width=0.085\linewidth, height=0.085\linewidth]{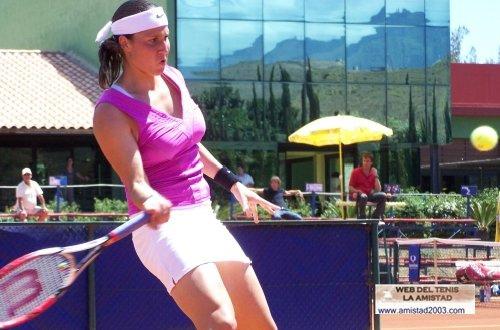}
    \includegraphics*[width=0.085\linewidth, height=0.085\linewidth]{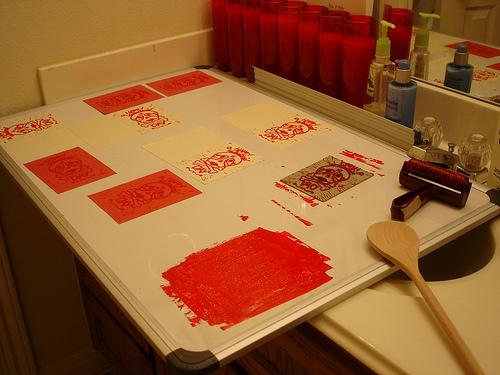}
    \includegraphics*[width=0.085\linewidth, height=0.085\linewidth]{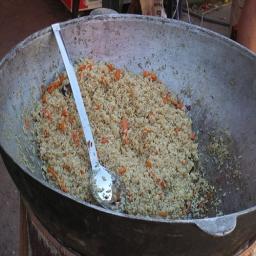}
    \includegraphics*[width=0.085\linewidth, height=0.085\linewidth]{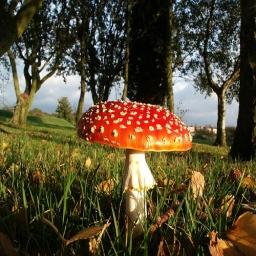}
    \includegraphics*[width=0.085\linewidth, height=0.085\linewidth]{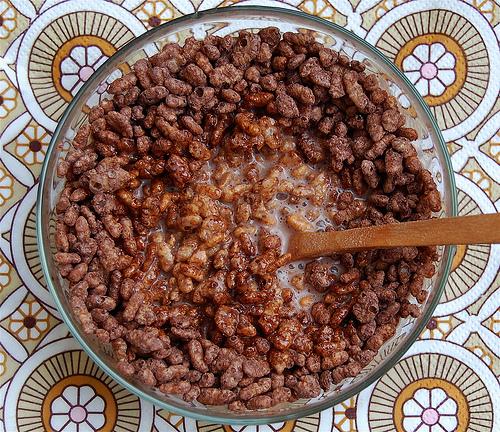}
    \includegraphics*[width=0.085\linewidth, height=0.085\linewidth]{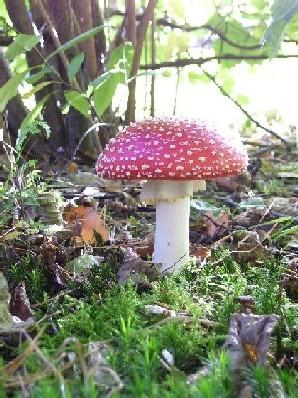}
    \includegraphics*[width=0.085\linewidth, height=0.085\linewidth]{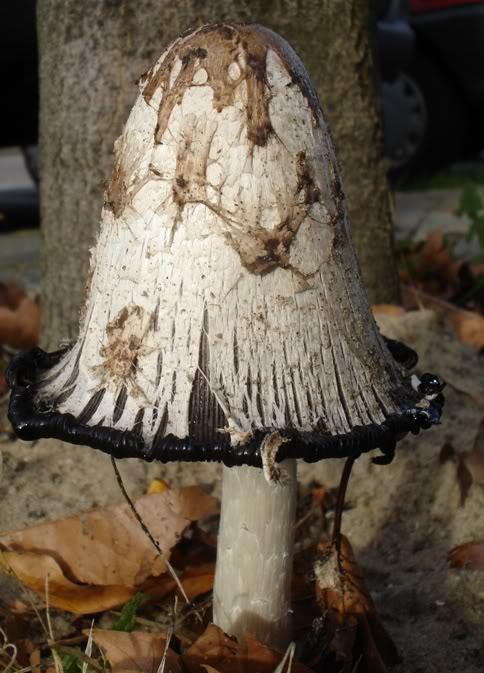}
    \includegraphics*[width=0.085\linewidth, height=0.085\linewidth]{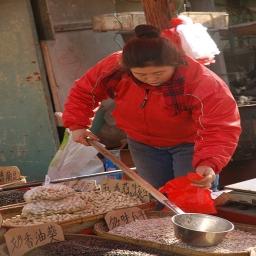}
    \includegraphics*[width=0.085\linewidth, height=0.085\linewidth]{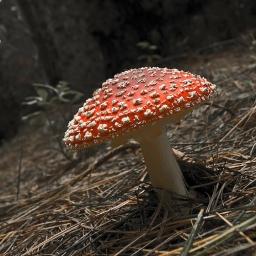}
    \includegraphics*[width=0.085\linewidth, height=0.085\linewidth]{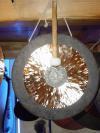}
    \hfill\\
    \vspace{-0.3em}
    \hspace*{0.085\linewidth}
    \includegraphics*[width=0.085\linewidth, height=\fontcharht\font`\B]{figs/rc}
    \includegraphics*[width=0.085\linewidth, height=\fontcharht\font`\B]{figs/rc}
    \includegraphics*[width=0.085\linewidth, height=\fontcharht\font`\B]{figs/rc}
    \includegraphics*[width=0.085\linewidth, height=\fontcharht\font`\B]{figs/gc}
    \includegraphics*[width=0.085\linewidth, height=\fontcharht\font`\B]{figs/rc}
    \includegraphics*[width=0.085\linewidth, height=\fontcharht\font`\B]{figs/gc}
    \includegraphics*[width=0.085\linewidth, height=\fontcharht\font`\B]{figs/gc}
    \includegraphics*[width=0.085\linewidth, height=\fontcharht\font`\B]{figs/rc}
    \includegraphics*[width=0.085\linewidth, height=\fontcharht\font`\B]{figs/gc}
    \includegraphics*[width=0.085\linewidth, height=\fontcharht\font`\B]{figs/rc}
    \hfill\\%
    \vspace{0.3em}
    \includegraphics*[width=0.085\linewidth, height=0.085\linewidth]{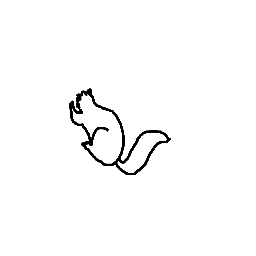}
    \includegraphics*[width=0.085\linewidth, height=0.085\linewidth]{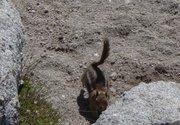}
    \includegraphics*[width=0.085\linewidth, height=0.085\linewidth]{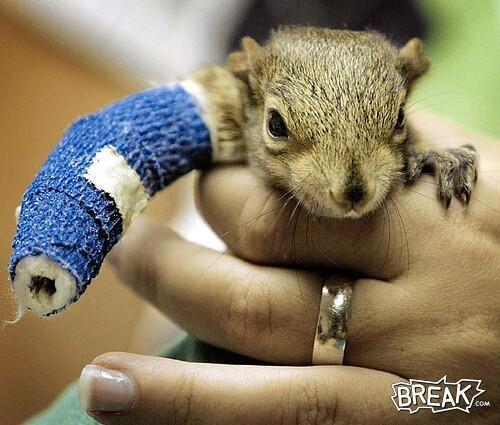}
    \includegraphics*[width=0.085\linewidth, height=0.085\linewidth]{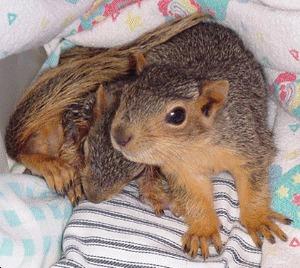}
    \includegraphics*[width=0.085\linewidth, height=0.085\linewidth]{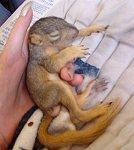}
    \includegraphics*[width=0.085\linewidth, height=0.085\linewidth]{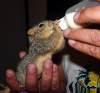}
    \includegraphics*[width=0.085\linewidth, height=0.085\linewidth]{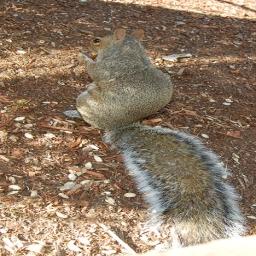}
    \includegraphics*[width=0.085\linewidth, height=0.085\linewidth]{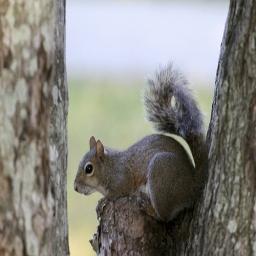}
    \includegraphics*[width=0.085\linewidth, height=0.085\linewidth]{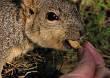}
    \includegraphics*[width=0.085\linewidth, height=0.085\linewidth]{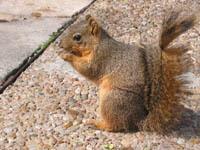}
    \includegraphics*[width=0.085\linewidth, height=0.085\linewidth]{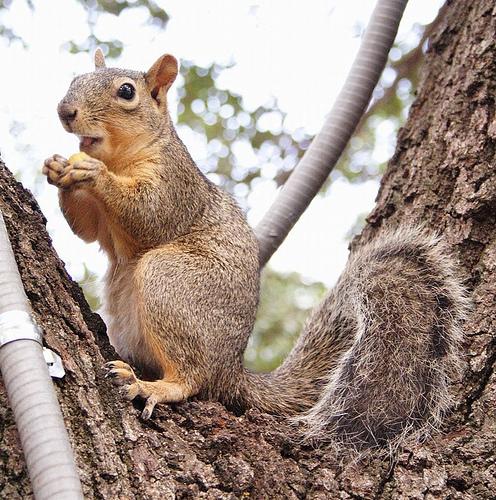}
    \hfill\\
    \vspace{-0.3em}
    \hspace*{0.085\linewidth}
    \includegraphics*[width=0.085\linewidth, height=\fontcharht\font`\B]{figs/gc}
    \includegraphics*[width=0.085\linewidth, height=\fontcharht\font`\B]{figs/gc}
    \includegraphics*[width=0.085\linewidth, height=\fontcharht\font`\B]{figs/gc}
    \includegraphics*[width=0.085\linewidth, height=\fontcharht\font`\B]{figs/gc}
    \includegraphics*[width=0.085\linewidth, height=\fontcharht\font`\B]{figs/gc}
    \includegraphics*[width=0.085\linewidth, height=\fontcharht\font`\B]{figs/gc}
    \includegraphics*[width=0.085\linewidth, height=\fontcharht\font`\B]{figs/gc}
    \includegraphics*[width=0.085\linewidth, height=\fontcharht\font`\B]{figs/gc}
    \includegraphics*[width=0.085\linewidth, height=\fontcharht\font`\B]{figs/gc}
    \includegraphics*[width=0.085\linewidth, height=\fontcharht\font`\B]{figs/gc}
    \hfill\\%
    \vspace{0.3em}
    \includegraphics*[width=0.085\linewidth, height=0.085\linewidth]{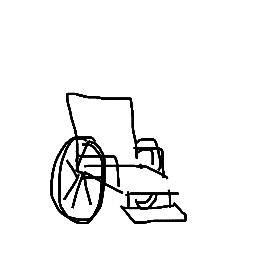}
    \includegraphics*[width=0.085\linewidth, height=0.085\linewidth]{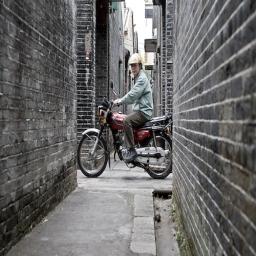}
    \includegraphics*[width=0.085\linewidth, height=0.085\linewidth]{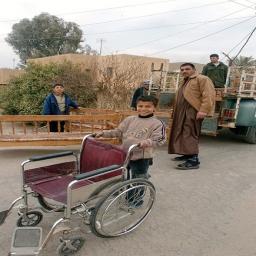}
    \includegraphics*[width=0.085\linewidth, height=0.085\linewidth]{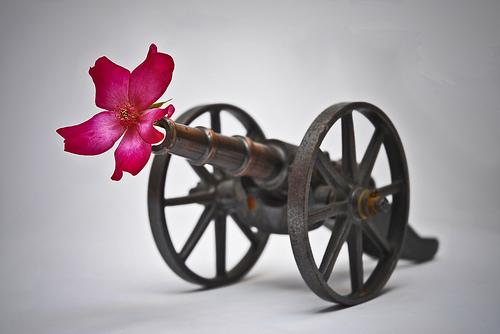}
    \includegraphics*[width=0.085\linewidth, height=0.085\linewidth]{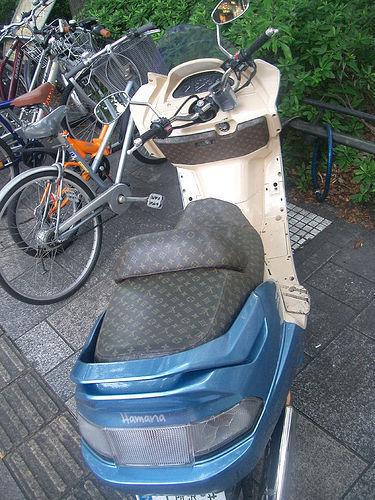}
    \includegraphics*[width=0.085\linewidth, height=0.085\linewidth]{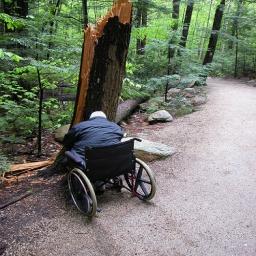}
    \includegraphics*[width=0.085\linewidth, height=0.085\linewidth]{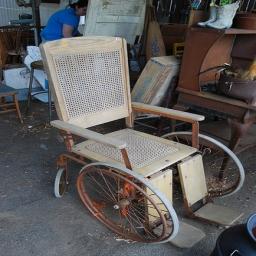}
    \includegraphics*[width=0.085\linewidth, height=0.085\linewidth]{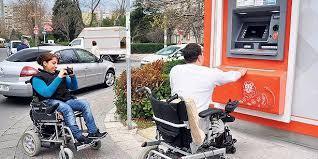}
    \includegraphics*[width=0.085\linewidth, height=0.085\linewidth]{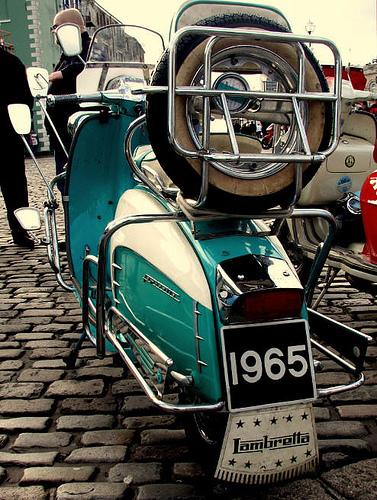}
    \includegraphics*[width=0.085\linewidth, height=0.085\linewidth]{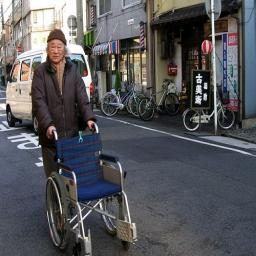}
    \includegraphics*[width=0.085\linewidth, height=0.085\linewidth]{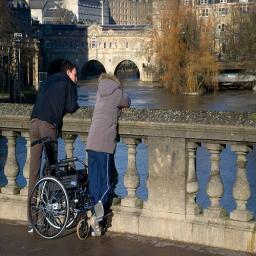}
    \hfill\\
    \vspace{-0.3em}
    \hspace*{0.085\linewidth}
    \includegraphics*[width=0.085\linewidth, height=\fontcharht\font`\B]{figs/rc}
    \includegraphics*[width=0.085\linewidth, height=\fontcharht\font`\B]{figs/gc}
    \includegraphics*[width=0.085\linewidth, height=\fontcharht\font`\B]{figs/rc}
    \includegraphics*[width=0.085\linewidth, height=\fontcharht\font`\B]{figs/rc}
    \includegraphics*[width=0.085\linewidth, height=\fontcharht\font`\B]{figs/gc}
    \includegraphics*[width=0.085\linewidth, height=\fontcharht\font`\B]{figs/gc}
    \includegraphics*[width=0.085\linewidth, height=\fontcharht\font`\B]{figs/gc}
    \includegraphics*[width=0.085\linewidth, height=\fontcharht\font`\B]{figs/rc}
    \includegraphics*[width=0.085\linewidth, height=\fontcharht\font`\B]{figs/gc}
    \includegraphics*[width=0.085\linewidth, height=\fontcharht\font`\B]{figs/gc}
    \hfill\\%
    \caption{Top-10 ZS-SBIR samples from IIAE on the Sketchy Extended dataset. Sketches in the leftmost columns are queries and rest are retrieved candidates (Top-1 to 10 from the left to right). Green checkmark indicates correct retrieval, whereas red crossmark indicates wrong retrieval.}
    % \label{subfig:GT}
\end{figure}

\section{Implementation details}
\label{appendix:implementation}
Here we describe the network architectures of our implementation.
For any dataset, every convolutional layer or fully connected layer in encoders is followed by batch normalization (BN) and LeakyReLU with slope 0.2, except the last layers of distribution encoders $q(z^x|x)$, $q(z^y|y)$, $r(z^s|x)$, $r(z^s|y)$, and $q(z^s|x,y)$.
The output of those last layers are means and log variances. Note that feature extractors (FE) of $r(z^s|x)$ and $r(z^s|y)$ are shared with $q(z^{s}|x,y)$.
Our implementation is publicly available.\footnote{\url{https://github.com/gr8joo/IIAE}}
\subsection{Network Architecture for MNIST-CDCB \citep{gonzalez-garcia2018NeurIPS}, Cars \citep{reed2015deep}, Maps \citep{pix2pix2017}, and Facades\citep{10.1007/978-3-642-40602-7_39}}
\begin{table}[H]
\begin{tabular}{ccc}
\hline
\multicolumn{1}{|c|}{Encoder} & \multicolumn{1}{c|}{$q(z^x|x)$ or $q(z^y|y)$}              & \multicolumn{1}{c|}{FE}                \\ \hline
\multicolumn{1}{|c|}{Input}   & \multicolumn{1}{c|}{256 x 256 x 3 image}                   & \multicolumn{1}{c|}{256 x 256 x 3 image}                   \\ \hline
\multicolumn{1}{|c|}{Layer1}  & \multicolumn{1}{c|}{4x4 Conv. w/ stride 2 and 32 filters}  & \multicolumn{1}{c|}{4x4 Conv. w/ stride 2 and 32 filters}  \\ \hline
\multicolumn{1}{|c|}{Layer2}  & \multicolumn{1}{c|}{4x4 Conv. w/ stride 2 and 64 filters}  & \multicolumn{1}{c|}{4x4 Conv. w/ stride 2 and 64 filters}  \\ \hline
\multicolumn{1}{|c|}{Layer3}  & \multicolumn{1}{c|}{4x4 Conv. w/ stride 2 and 128 filters} & \multicolumn{1}{c|}{4x4 Conv. w/ stride 2 and 128 filters} \\ \hline
\multicolumn{1}{|c|}{Layer4}  & \multicolumn{1}{c|}{4x4 Conv. w/ stride 2 and 256 filters} & \multicolumn{1}{c|}{4x4 Conv. w/ stride 2 and 256 filters} \\ \hline
\multicolumn{1}{|c|}{Layer5}  & \multicolumn{1}{c|}{FC. 16}                                & \multicolumn{1}{c|}{-}                                     \\ \hline
                              &                                                            &                                                            \\ \hline
\multicolumn{1}{|c|}{Encoder} & \multicolumn{1}{c|}{$r(z^s|x)$ or $r(z^s|y)$}              & \multicolumn{1}{c|}{$q(z^s|x,y)$}                          \\ \hline
\multicolumn{1}{|c|}{Input}   & \multicolumn{1}{c|}{FE($x$) or FE($y$)}                & \multicolumn{1}{c|}{[FE($x$) ; FE($y$)]}               \\ \hline
\multicolumn{1}{|c|}{Layer1}  & \multicolumn{1}{c|}{4x4 Conv. w/ 256 filters}              & \multicolumn{1}{c|}{4x4 Conv. w/ 256 filters}              \\ \hline
\multicolumn{1}{|c|}{Layer2}  & \multicolumn{1}{c|}{FC. 256}                               & \multicolumn{1}{c|}{FC. 256}                               \\ \hline
\multicolumn{1}{|c|}{Layer3}  & \multicolumn{1}{c|}{FC. 256}                               & \multicolumn{1}{c|}{FC. 256}                               \\ \hline
\end{tabular}
\end{table}
\begin{table}[H]
\begin{tabular}{|c|c|}
\hline
Decoder & $p(x|z^x,z^s)$ or $p(y|z^y,z^s)$                                      \\ \hline
Input   & $\left[z^x;z^s\right]$ or $\left[z^y;z^s\right]$                      \\ \hline
Layer1  & FC. 262,144, BN, Dropout(0.5), ReLU                            \\ \hline
Layer2  & 4x4 Deconv. w/ stride 1/2 and 512 filters, BN, Dropout(0.5), ReLU\\ \hline
Layer3  & 4x4 Deconv. w/ stride 1/2 and 256 filters, BN, Dropout(0.5), ReLU\\ \hline
Layer4  & 4x4 Deconv. w/ stride 1/2 and 128 filters, BN, ReLU            \\ \hline
Layer5  & 4x4 Deconv. w/ stride 1/2 and 64 filters, BN, ReLU                 \\ \hline
Layer6  & 4x4 Deconv. w/ stride 1/2 and 3 filters and Tanh activation               \\ \hline
\end{tabular}
\end{table}
Note that the last two fully connected layers in shared representation encoders ($r(z^s|x)$, $r(z^s|y)$, and $q(z^s|x,y)$) and the first fully connected layer in decoders are only applied to Cars \citep{reed2015deep} dataset.

\subsection{Network architecture for Sketchy Extended \citep{sangkloy2016sketchy, liu2017deep} (ZS-SBIR)}
\begin{table}[H]
\begin{tabular}{cccccc}
\cline{1-3}
\multicolumn{1}{|c|}{Encoder} & \multicolumn{1}{c|}{$q(z^x|x)$ or $q(z^y|y)$} & \multicolumn{1}{c|}{FE}       &                       &                              &                                                       \\ \cline{1-3}
\multicolumn{1}{|c|}{Input}   & \multicolumn{1}{c|}{512 image feature}        & \multicolumn{1}{c|}{512 image feature}       &                       &                              &                                                       \\ \cline{1-3}
\multicolumn{1}{|c|}{Layer1}  & \multicolumn{1}{c|}{FC. 512}                  & \multicolumn{1}{c|}{FC. 512}                 &                       &                              &                                                       \\ \cline{1-3}
\multicolumn{1}{|c|}{Layer2}  & \multicolumn{1}{c|}{FC. 256}                  & \multicolumn{1}{c|}{-}                       &                       &                              &                                                       \\ \cline{1-3}
\multicolumn{1}{|c|}{Layer5}  & \multicolumn{1}{c|}{FC. 128}                  & \multicolumn{1}{c|}{-}                       &                       &                              &                                                       \\ \cline{1-3}
                              &                                               &                                              &                       &                              &                                                       \\ \cline{1-3} \cline{5-6} 
\multicolumn{1}{|c|}{Encoder} & \multicolumn{1}{c|}{$r(z^s|x)$ or $r(z^s|y)$} & \multicolumn{1}{c|}{$q(z^s|x,y)$}            & \multicolumn{1}{c|}{} & \multicolumn{1}{c|}{Decoder} & \multicolumn{1}{c|}{$p(x|z^x,z^s)$ or $p(y|z^y,z^s)$} \\ \cline{1-3} \cline{5-6} 
\multicolumn{1}{|c|}{Input}        & \multicolumn{1}{c|}{FE($x$) or FE($y$)}   & \multicolumn{1}{c|}{[FE($x$) ; FE($y$)]} & \multicolumn{1}{c|}{} & \multicolumn{1}{c|}{Input}   & \multicolumn{1}{c|}{$\left[z^x;z^s\right]$ or $\left[z^y;z^s\right]$}                                 \\ \cline{1-3} \cline{5-6} 
\multicolumn{1}{|c|}{Layer1}  & \multicolumn{1}{c|}{FC. 256}                  & \multicolumn{1}{c|}{FC. 512}                 & \multicolumn{1}{c|}{} & \multicolumn{1}{c|}{Layer1}  & \multicolumn{1}{c|}{FC. 128}                                 \\ \cline{1-3} \cline{5-6} 
\multicolumn{1}{|c|}{Layer2}  & \multicolumn{1}{c|}{FC. 128}                  & \multicolumn{1}{c|}{FC. 128}                 & \multicolumn{1}{c|}{} & \multicolumn{1}{c|}{Layer2}  & \multicolumn{1}{c|}{FC. 512}                                 \\ \cline{1-3} \cline{5-6} 
\end{tabular}
\end{table}
\subsection{Hyperparameters}
\begin{table}[H]
\begin{tabular}{|c|c|c|c|c|c|}
\hline
Hyper-                & \multicolumn{5}{c|}{Datasets}                          \\ \cline{2-6} 
parameters            & MNIST-CDCB & Cars & Facades & Maps  & Sketchy Extended \\ \hline
Learning rate         & \multicolumn{5}{c|}{0.0002}                            \\ \hline
Lambda                & 5          & 50   & 1,000    & 50    & 2                \\ \hline
Reconstruction weight & \multicolumn{3}{c|}{1,000}   & 20,000 & 10               \\ \hline
% Epochs                & 25         & 500  & 1,100    & 600   & 10                \\ \hline
\end{tabular}
\end{table}

\end{document}